\documentclass[b5paper,10pt]{book}

\usepackage{phdthesis}
\usepackage[
  paperwidth=17cm
  ,paperheight=24cm
  ,inner=2.6cm
  ,outer=2.3cm
  ,top=2.3cm
  ,bottom=2.3cm
]{geometry}

\usepackage{graphicx,times}
\usepackage[hyphens]{url}
\usepackage{tabularx,amsmath}
\usepackage{multirow}
\usepackage{booktabs}
\usepackage{amssymb}
\usepackage[comma,compress,sectionbib]{natbib}
\usepackage{wrapfig}
\usepackage{enumerate}
\usepackage{supertabular}
\usepackage{multicol}
\usepackage{setspace}
\usepackage{import}
\usepackage{graphicx}
\usepackage{url}
\usepackage{times}
\usepackage{microtype}
\usepackage{tabularx,amsmath}
\usepackage{amssymb}
\usepackage{color}
\usepackage[usenames,dvipsnames,table]{xcolor}
\usepackage[normalem]{ulem}
\usepackage[english]{babel}
\usepackage{statex}
\usepackage{algorithm}
\usepackage{paralist}
\usepackage{flushend}
\usepackage{enumerate}
\usepackage{enumitem}
\usepackage{setspace}
\usepackage{textcomp}
\usepackage{listings}
\usepackage{keyval}
\usepackage{rotating}
\usepackage{acronym}
\usepackage{adjustbox}
\usepackage{latexsym}
\usepackage{epigraph}

\usepackage{lipsum}
\usepackage[noend]{algpseudocode}

\usepackage{bm}
\usepackage{dblfloatfix}    % To enable figures at the bottom of page
\usepackage{capt-of}
\usepackage{mathabx}
\usepackage [autostyle, english = american]{csquotes}
\usepackage{tabularx}
\usepackage{subcaption}
\usepackage{float}
\usepackage{xcolor}
\usepackage{tikz}
\usepackage{pgfplots, pgfplotstable}
\usetikzlibrary{patterns}

\usepackage[labelfont=bf]{caption}

\usepackage{tcolorbox,varwidth}
\usepackage{pifont}% http://ctan.org/pkg/pifont
\usepackage{array}
\usepackage{stackengine}

\usepackage{collcell}
\usepackage{pgf}

\usepackage{arydshln}

\usepackage{verbatim}
\MakeOuterQuote{"}
\mathchardef\mhyphen="2D

\newcommand\given[1][]{\:#1\vert\:}
\newcommand{\code}[1]{\texttt{#1}}
\DeclareMathOperator*{\argmax}{arg\,max}
\DeclareMathOperator*{\softmax}{softmax}
\DeclareMathOperator*{\topk}{topK}
\DeclareMathOperator*{\concat}{concat}
\DeclareMathOperator*{\gammaa}{Gamma}
\DeclareMathOperator*{\attention}{Attention}
\DeclareMathOperator*{\countop}{count}
\DeclareMathOperator*{\unigram}{unigram}
\DeclareMathOperator*{\gram}{-gram}
\DeclareMathOperator*{\bleu}{BLEU}
\DeclareMathOperator*{\bp}{BP}
\DeclareMathOperator*{\similarity}{Sim}
\newcommand{\vt}[1]{\bm{\mathbf{#1}}}

\newcommand{\gap}{\vspace{2mm}}

\renewcommand{\arraystretch}{1.2} %row spacing of all tables is increased here

\definecolor{gr}{HTML}{A6CEE3}
\definecolor{blu}{HTML}{1C9099}

\definecolor{mygray}{gray}{0.6}
\definecolor{tablegray}{gray}{0.9}

\usepackage[backref=page]{hyperref}
\hypersetup{%
pdfborder = {0 0 0},
pdftitle = {Understanding and Enhancing Context for MT},
pdfsubject = {PhD Thesis},
pdfkeywords = {machine translation, context, neural models, NLP},
pdfauthor = {Marzieh Fadaee}
}
\usepackage{bookmark}

\def\blankpage{%
      \clearpage%
      \thispagestyle{empty}%
      \addtocounter{page}{-1}%
      \null%
      \clearpage}

\renewcommand*{\backref}[1]{} 
\renewcommand*{\backrefalt}[4]{%
\ifcase #1
\or (Cited on page~#2.)  %
\else 
(Cited on pages~#2.)  
\fi
}

\definecolor{mygray}{gray}{0.4}

\newcolumntype{R}[2]{%
    >{\adjustbox{angle=#1,lap=\width-(#2)}\bgroup}%
    l%
    <{\egroup}%
}
\newcommand{\cmark}{\ding{51}}%
\newcommand{\xmark}{\ding{55}}%

\newcommand*{\MinNumber}{0}%
\newcommand*{\MaxNumber}{1}%
\newcommand{\ApplyGradient}[1]{%
  \pgfmathsetmacro{\PercentColor}{30.0*(#1-\MinNumber)/(\MaxNumber-\MinNumber)}%
  \edef\x{\noexpand\cellcolor{black!\PercentColor}}\x\textcolor{black}{#1}%
}
\newcolumntype{S}{>{\collectcell\ApplyGradient}{c}<{\endcollectcell}}

\newcommand{\grade}[1]{%
  \par
  \vspace*{-\parskip}% Correct insertion of paragraph skip due to \par
  {\raggedleft #1\par}}
  
\begin{document}

% Start the front matter
% Latex takes care of Roman page numbering ( I, II, III, IV, etc) for front matter
% Front matter, acknowledgements, and ToC
\frontmatter
% !TEX root = thesis-main.tex
% This is the first page, consisting of your title and name.

% Title goes here
{\pagestyle{empty}
\newcommand{\printtitle}{%
{\Huge\bf Understanding and Enhancing \\ the Use of Context \\ for Machine Translation \\[0.8cm]
}}

% Some markup followed by your name
\begin{titlepage}
\par\vskip 2cm
\begin{center}
\printtitle
\vfill
{\LARGE\bf Marzieh Fadaee}
\vskip 2cm
\end{center}
\end{titlepage}

% Skip a page to start on a right page again.
\mbox{}\newpage
\setcounter{page}{1}

% This is the title page (titelblad)
% You also need to send it to the Bureau Pedel
% Make sure you change:
%   the date of your defense
%   your name
%   your place of birth (including country, but ONLY if you were born outside the Netherlands)
\clearpage
\par\vskip 2cm
\begin{center}
\printtitle
\par\vspace {4cm}
{\large \sc Academisch Proefschrift}
\par\vspace {1cm}
{\large ter verkrijging van de graad van doctor aan de \\
Universiteit van Amsterdam\\
op gezag van de Rector Magnificus\\
prof.\ dr.\ ir.\ K.I.J. Maex\\
ten overstaan van een door het College voor Promoties ingestelde \\
commissie, in het openbaar te verdedigen in \\
de Agnietenkapel\\
%de Aula der Universiteit\\
op dinsdag 10 november 2020, te 12.00 uur \\ }
\par\vspace {1cm} {\large door}
\par \vspace {1cm}
{\Large Marzieh Fadaee}
\par\vspace {1cm}
{\large geboren te Berlijn} %MAKE VERY SURE THIS MATCHES YOUR PASSPORT, THE BEADLE WILL CHECK THIS! 
\end{center}

% The page following the titelblad. This usiallu contains:
%   committee members
%   SIKS logo + text
%   sponsors/ project numbers
%   ISBN
%   copyrights, cover design, printer
\clearpage
\noindent%
\textbf{Promotiecommissie} \\\\
\begin{tabular}{@{}l l l}
Promotor: \\
& dr. C. Monz & Universiteit van Amsterdam \\  %promotor
Co-promotor: \\
& dr. A. Bisazza & Rijksuniversiteit Groningen \\
Overige leden: \\
& dr. A. Birch & University of Edinburgh \\
 & prof. dr. A.P.J. van den Bosch & Radboud Universiteit Nijmegen \\
 & prof. dr. P.T. Groth & Universiteit van Amsterdam \\
  & prof. dr. E. Kanoulas & Universiteit van Amsterdam \\
 & dr. I. Markov & Universiteit van Amsterdam \\
\end{tabular}

\bigskip\noindent%
Faculteit der Natuurwetenschappen, Wiskunde en Informatica\\

\vfill

% Sponsors and projects
\noindent
%The research was supported by SomeOne under project number XXX-YYY.  \\
\bigskip

% Copyrights
\noindent
Copyright \copyright~2020 Marzieh Fadaee, Amsterdam, The Netherlands\\
Cover by Mostafa Dehghani, Painting "Series VIII: Picture of the Starting Point" by Hilma af Klint \\
Printed by Ipskamp printing, Amsterdam\\
ISBN: 978-94-6421-059-0\\
\clearpage
}

{\pagestyle{empty}

\vfill
{
\begin{center}
\noindent
{
\vskip 5cm
\em To those who see the world through words.} \\ \vskip .5cm
\end{center}
}

\clearpage
} %print version
\blankpage

{\pagestyle{empty}

{
\begin{center}

\noindent
\textbf{Acknowledgements} \\ \vskip .5cm
\end{center}
}

\begin{quote}
{``In the good mystery there is nothing wasted, no sentence, no word that is not significant. And even if it is not significant, it has the potential to be so - which amounts to the same thing.''  \footnote{Paul Auster, The New York Trilogy}}
\end{quote}
\noindent 
\textit{No word that is not significant}; this is the main discussion of this book on the importance of context as well as an accurate description of how I (and this book) came to be. 
%rewrite ^
When I started my PhD the world was a different place. Then again, that can be said about every minute that goes by. What remains a constant is the impact of people around us, in the beginning of things and in the end.
It has been an agonizingly fun ride! And I am very fortunate to have many amazing people around me that made it possible. 

%christof
First, Christof.
I came to Amsterdam with a promise and it turned out to be quite an experience. 
I owe my deepest gratitude to Christof for trusting me and taking me on this journey. 
Over the years he taught me how to research new areas, how to break down complicated questions to get to the heart of the problem, and how to never be happy with an answer and always push for more.
I specially would like to thank him for his attentiveness when it counted.
I enjoyed the walks, the talks, and making him exasperated with my wordplays. 

%arianna
I would like to thank Arianna for her guidance and for being a friend. 
Her critical thinking and questioning the things that I took for granted taught me how to explore novel corners of research and ask new questions. 
I will always be grateful to have her by my side, whether it was during discussions with Christof or the disappointment that was the Star Wars sequels.

%maarten
Next, I would like to express my gratitude to Maarten for building a great research group at ILPS. 
From the first day of joining ILPS, I felt part of a team. I enjoyed the fun impromptu discussions, the group activities, and everything in between.
Now that I've been away for some time, I see that what we had at ILPS was something truly special and Maarten is a big reason for that. 
Thanks for the guidance and for encouraging me to push my boundaries. 

%committee
My sincere gratitude to Alexandra, Antal, Paul, Evangelos, and Ilya who were generous with their time to read this thesis and accepted to be part of my defense committee. 

%mt
I would like to thank the MT members: Marlies, Ke, Katya, and Praveen, my brothers and sisters in arms. 
We shared laughs and tears, successes and failures, valuable and worthless discussions. And we all came out of it alive! 
What I got out of this PhD was more than I expected and I am glad to have met these people along the way. 
Thanks for making my PhD experience a fun and enjoyable one. 

%ilps
I am grateful for many current and former members at ILPS for our times together. From showing the ropes and sharing their wisdom, to asking intriguing questions, to fun sushi parties. 
Thanks Alexey, Amir, Anna, Artem, Boris, Chang, Chuan, Daan, Damien, Dan, David, David, Dilek, Eva, Evgeny, Fei, Harrie, Hendra, Hendrike, Hinda, Ivan, Jiahuan, Jie, Julia, Julien, Kaspar, Maarten, Maurits, Mostafa, Pengjie, Richard, Ridho, Rolf, Shangsong, Shaojie, Spyretta, Svitlana, Tobias, Trond, Vera, Xiaohui, Xiaojuan, Xinyi, Yangjun, Yifan, Zhaochun, and Ziming.

I particularly like to thank Anne, Ana, Bob, Christophe, Evangelos, Ilya, Isaac, Manos, and Tom, who are responsible for some of my good memories.
Special thanks to Petra, who knows the answer to every question and is a delight to be around. 
%
%cover
I would also like to thank Mostafa for designing the cover and putting up with all of my typography suggestions.
%et.
Cristina, thanks for being a ray of sunshine full of support and insight. We will always have Toronto.

%paranymphs
Next, I want to thank my wonderful paranymphs: Maartje, my office-mate and friend who makes an excellent arrabiata pasta, and Nikos, my partner in crime, who makes every moment the opposite of boring. 
Thanks for agreeing to be by my side on my defense day.

%ebay
I spent a summer in Aachen doing an internship at eBay which was an invaluable practice. Thanks, Shahram and Jose for the guidance and the interesting discussions, and Daniel, Leonard, Michael, Nicola, and Sivan for making it a nice experience. 
Parnia thanks for the talks, the hikes, and the Ash Reshtehs. 

%nl friends
Moving to Amsterdam meant a fresh start and a new beginning. I was lucky to meet many amazing people that help me weather the gray days and savor the sunny ones. I am particularly thankful to Ali, Ali, Ameneh, Amin, Amirhosein, Arash, Aylar, Azad, Azadeh, Behrouz, Danial, Fahimeh, Fatemeh, Hester, Hoda, Hoda, Hojat, Hoora, Irene, Iris, Jasmijn, Jeyran, Mahdi, Mahdi, Mahdieh, Maria, Maryam, Marzieh, Marzieh, Masoud, Masoumeh, Mohammad, Mohammad Amin, Mohammad Hosein, Mohsen, Narges, Naser, Nasim, Nasrin, Parisa, Parisa, Parisa, Saba, Samira, Samira, Sara, Shayan, Shima, Vahid, Vivianne, Wilker, Zahra, and Zoheir.   
A particular thanks to a special group of people with whom I had many imaginative conversations: Ava, Behsa, Borna, NikAyeen, Pouya, Radin, Saba, Samin, Sarina, Sepehr, Sina, and Zeinab. They made me an optimist about the future.

%algon
My Algonquin Round Table, whom I am grateful to have in my life:
Abbas, Ali, Arian, Atieh, Elaheh, Faezeh, Farzad, Ghazaleh, Louise, Mandana, Maryam, Mehdi, Safoora, Sara, Sara, Siavash. 
I met each of these people at different points in my life and something about each of them got stuck with me. Geographical determinism might be working against us now, but I cherish every time we got a chance to meet somewhere in this world. 

%bros
My brothers Mohammad and Mohsen, thanks for doing what brothers do best: supporting me in times of need and humbling me the rest of the time. Fatemeh and Mahsa, I am thankful for the kind words and the support.

%parents
I am eternally grateful to my amazing parents. 
Baba, you showed me every day that there is always something new worth learning. 
Your dedication to diving blindly into new areas, languages, and technologies is always inspiring.
You taught me to never stop fighting for what I want. 
Maman, you are the strongest woman I know and I appreciate everything you have done for me.
Every step I took has been a little bit easier because of the hard decisions you had to make. 
I followed my dreams because of you.

%hamid
Lastly, Hamid. 
You are the kindest, smartest, most driven person I know. 
And yes I am still not a fan of using superlatives. 
It is truly earned in your case though.
This would not have been possible without you. 
You were the best sounding board for everything that did (and did not) end up in this book.
I am lucky to have your love, your support, and your wisdom by my side in this journey. 
\\
\\
\grade{Marzieh}
\grade{September 2020}

\vfill

\begin{quote}
{``Everything becomes essence; the center of the book shifts with each event that propels it forward. The center, then, is everywhere, and no circumference can be drawn until the book has come to its end.'' \footnote{Paul Auster, The New York Trilogy}}
\end{quote}

\clearpage
}

\blankpage
 %print version
\tableofcontents

% Define some commonly used acronyms. 
%\acrodef{IR}{information retrieval}

% Start the main matter
% Here, Latex will use Arabic page numbering (1, 2, 3, etc.)
% Add all your chapters:
%  1. introduction
%  2. background/ related work
%  3. experimental methodology
%  4,5,X. technical chapters
%  X+1. conclusion
\mainmatter
% !TEX root = thesis-main.tex

\begin{savequote}[45mm]
--- Il n'y a pas de hors-texte.
\qauthor{Jacques Derrida}
\end{savequote}

\chapter{Introduction}

\label{chapter:introduction}

%%%%%%%%%%%%%%%%%%%%%%%%%%%%%%%%%%%%%%%%%%%%%%%%%%%%%%
%  OPENING
%%%%%%%%%%%%%%%%%%%%%%%%%%%%%%%%%%%%%%%%%%%%%%%%%%%%%%

% NN learning from data
Neural networks learn patterns from data to solve complex problems.
To understand and infer meaning in language, neural models have to learn complicated nuances.

% detour to nuances of language
Discovering distinctive linguistic phenomena from data is not an easy task. 
For instance, lexical ambiguity is a fundamental feature of language which is challenging to learn \citep{small2013lexical}. 
Even more prominently, inferring the meaning of rare and unseen lexical units is difficult with neural networks \citep{koehn2017six}. 
For instance, \citet{rios-etal-2018-word} provide an example where an English-German translation model translates the sentence ``[$\ldots$] \textit{Hedge-Fund- \textbf{Anlagen} nicht zwangsl\"{a}ufig risikoreicher sind als traditionelle  \textbf{Anlagen}}'' to 
``[$\ldots$] \textit{hedge fund \underline{assets} are not necessarily more risky than traditional \underline{plants}}''. Here, the ambiguous word \textit{`Anlagen'} is first translated correctly to \textit{`assets'}, but then incorrectly to \textit{`plants'} in the second occurrence.

To understand many of these phenomena, a model has to learn from a few instances and be able to generalize well to unseen cases.
%
% connect using context to learn nuances of language
%
Natural language speakers typically learn the meanings of words by the \textit{context} in which they are used.
 \citet{miller-1985-dictionaries} states that:
\begin{quote}
``\emph{When subtle semantic distinctions are at issue, it is customary to remark that a satisfactory language understanding system will have to know a great deal more than the linguistic values of words.}''
\end{quote}

\noindent Sentence and document-level context provide the possibility to go beyond lexical instances and study words in a broader context.
Neural models use a sizable amount of data that often consists of contextual instances to learn patterns.
% back to learning from data and the necessity of {data-driven} learning 
However, the learning process is hindered when training data is scarce for a task \citep{45801,edunov-etal-2018-understanding}.
Even with sufficient data, learning patterns for the long tail of the lexical distribution is challenging \citep{NIPS2017_7278}. 
To address these problems, one approach is to augment the training data \citep{sennrich-haddow-birch:2016:P16-11}.
Many strategies for data augmentation focus on increasing the amount of data to assist the learning process of data-driven neural models.
While simply increasing the size of data is helpful, it is not entirely clear \textit{where} the improvements come from and \textit{how} neural models benefit from the additional context with augmentation. 

%%%%%%%%%%%%%%%%%%%%%%%%%%%%%%%%%%%%%%%%%%%%%%%%%%%%%%
%  FIRST PART OF THE TITLE: "ENHANCING THE USE OF CONTEXT [...]"
%%%%%%%%%%%%%%%%%%%%%%%%%%%%%%%%%%%%%%%%%%%%%%%%%%%%%%
% What is important about context and why we want to understand and enhance it?
Arguably, it is important to understand the impact of new contexts to design augmentation models that exploit these contexts.
This includes understanding what constitutes a beneficial context, and how to enhance the use of context in neural models.
In this thesis, we focus on understanding certain potentials of contexts in a neural model, and design augmentation models to benefit from them. 
%We are particularly interested in the use of context in understanding meaning in a language.  

%%%%%%%%%%%%%%%%%%%%%%%%%%%%%%%%%%%%%%%%%%%%%%%%%%%%%%
%  SECOND PART OF THE TITLE: "[...] FOR MACHINE TRANSLATION"
%%%%%%%%%%%%%%%%%%%%%%%%%%%%%%%%%%%%%%%%%%%%%%%%%%%%%%
% why machine translation?
We focus on machine translation as a prominent instance of the more general language understanding problems. 
In order to translate from a source language to a target language, a neural model has to understand the meaning of constituents in the provided context and generate constituents with the same meanings in the target language.
This task accentuates the value of capturing nuances of language and the necessity of generalization from few observations \citep{DBLP:journals/corr/abs-2004-02181}.
Additionally, the lack of large amounts of labeled data is even more pronounced in machine translation in the form of bilingual corpora.
This signifies the need for efficient and informed data augmentation models.

%%%%%%%%%%%%%%%%%%%%%%%%%%%%%%%%%%%%%%%%%%%%%%%%%%%%%%
%  CLOSING THE INTRO SECTION IF THE INTRODUCTION CHAPTER!
%%%%%%%%%%%%%%%%%%%%%%%%%%%%%%%%%%%%%%%%%%%%%%%%%%%%%%
\medskip

% focus on _where_ and _why_
The main problem we study in this thesis is what neural machine translation models (NMT) learn from data, and how we can devise more focused contexts to enhance this learning.
We believe that looking more in-depth into the role of context and the impact of data on learning models is essential to advance the Natural Language Processing (NLP) field. 
Understanding the importance of data in the learning process and how neural network models interact, utilize, and benefit from data can help develop more {accurate} NLP systems. 
Moreover, it helps highlight vulnerabilities and volatilities of current neural networks and provides insights into designing more  robust models.

%% The research questions and sub questions
% !TEX root = thesis-main.tex

\section{Research outline and questions}
\label{section:introduction:rqs}

This thesis explores the role of context in language understanding and in particular, machine translation using recent advances in deep learning.
We develop novel models and learning algorithms to examine the abilities of neural networks in learning from data. 
Specifically, we are interested in the importance of contextual cues in translating words and various ways we can use data to advance translation systems further. 

Before investigating the role of context in the bilingual setting of machine translation, we ask ourselves a more general question about the impact of context in monolingual settings.
As a preliminary investigation into this question, we look into ambiguous words where, by definition, context is the prominent factor in understanding word meaning. 
We study how document-level contexts as topics aid in distinguishing different meanings of a word. 

Next, in more detail, we focus on the influence of context in the bilingual setting of machine translation. 
While recent advances in neural networks have been very successful in translation, the significance of different aspects of data is still largely unexplored.
We investigate how the translation models exploit context to learn and transfer meaning and show that manipulating data improves translation quality.
In particular, our proposed models examine how different and diverse contexts resolve various obstacles of translation.

Lastly, we address the shortfalls of relying only on the observed context to learn word meaning and focus on particularly interesting cases. 
Neural networks optimize the learning process on the available data. 
We examine under which conditions the observed context in the training data is not enough for meaning inference and capturing various linguistic phenomena. 
Moreover, we raise questions about the learning abilities of current translation models and where they fail to capture the available information in data. 
With contextual modifications, we identify an underlying generalization problem in state-of-the-art translation models.

Concretely, we set out to answer the following research questions in this thesis:

\acrodef{rq:topic}[\ref{rq:topic}]{\textit{Can document-level topic distribution help infer the meaning of a word?}}
\acrodef{rq:topic1}[\ref{rq:topic1}]{\textit{ To what extent can distributions over word senses be approximated by distributions over topics of documents without assuming these concepts to be identical?}}  % [embedding]}}
\acrodef{rq:topic2}[\ref{rq:topic2}]{\textit{ How can we exploit document-level topics to distinguish between different meanings of a word and learn the corresponding representations?}}%  [embedding]}}
\acrodef{rq:topic3}[\ref{rq:topic3}]{\textit{ What are the advantages of using document-level topics in learning multiple representations per word? }}% [embedding]}}

\acrodef{rq:tdabt}[\ref{rq:tdabt}]{\textit{How is the translation quality of a word influenced by the availability of diverse contexts?}}

\acrodef{rq:tda1}[\ref{rq:tda1}]{\textit{How can we successfully augment the training data with diverse contexts for rare words?}}%   [augmentation]}}
\acrodef{rq:tda2}[\ref{rq:tda2}]{\textit{Do rare words benefit from augmentation via paraphrasing during test time?}}% [augmentation]}}
\acrodef{rq:bt1}[\ref{rq:bt1}]{\textit{ Do signals from the NMT model help identify low-confidence words that could benefit from additional context?  }}%  [backtrans]}}
\acrodef{rq:bt2}[\ref{rq:bt2}]{\textit{ How can we successfully apply data selection of monolingual data to diversify the contexts of low-confidence words? }}% [backtrans]}}

\acrodef{rq:vol}[\ref{rq:vol}]{\textit{To what extent are neural translation models vulnerable as a result of relying on the observed context in the training data to infer meaning? }}
\acrodef{rq:id1}[\ref{rq:id1}]{\textit{What are the challenges of idiom translation with neural models? }} %   [idiom]}}
\acrodef{rq:id2}[\ref{rq:id2}]{\textit{How is the translation quality of NMT influenced by idiomatic expressions? }} %  [idiom]}}
\acrodef{rq:vol1}[\ref{rq:vol1}]{\textit{ How can contextual modifications during testing reveal a lack of robustness of translation models and affect the translation quality?}}%    [volatility]}}
\acrodef{rq:vol2}[\ref{rq:vol2}]{\textit{ To what extent is a lack of robustness an indicator of a generalization problem in neural machine translation models?  }}% [volatility]}}

\begin{enumerate}[label=\textbf{Research Question \arabic*:},ref={RQ\arabic*},wide = 0pt,resume]
\setlength\itemsep{1em}
\item \acl{rq:topic} \label{rq:topic}

In this research question, we investigate whether using document-level context, as opposed to sentence-level only, has an impact on learning word representations. Word representations are abstract feature vectors that capture word meanings. To produce good representations, the learning model must capture various linguistic phenomena such as the ambiguity of the language.  
Notably, we tackle the problem of representing ambiguous words by defining multiple representations per word and using implicit topics of documents to distinguish between different meanings of a word. 
We divide this research question into three sub-questions and address them in Chapter~\ref{chapter:research-01}:

\begin{enumerate}[label=\textbf{RQ1.\arabic*},wide = 0pt, leftmargin=2em]
\setlength\itemsep{1em}
\item \acl{rq:topic1}  \label{rq:topic1}

\medskip

Modeling document topics is commonly used in different ways to address the challenging task of word sense disambiguation \citep{boyd-graber-etal-2007-topic,li-etal-2010-topic,ChaplotS18}. 
However, the topic of a document does not directly correspond to the senses of the words in that document.
We investigate whether a document topic distribution is an informative signal to help distinguish between different senses of a word and how we can leverage this information to learn word representations. 
Next, we ask:

\item \acl{rq:topic2}  \label{rq:topic2}

\medskip

To answer this question, we estimate document-topic distributions using unsupervised topic modeling techniques. 
We observe that the produced distribution over topics is different for different senses of an ambiguous word. 
We propose three variants of the Skipgram word embedding model \citep{mikolov2013efficient} to integrate topic distributions and learn multiple representations per word. 

\item \acl{rq:topic3}  \label{rq:topic3}

\medskip

To further evaluate our models, we analyze the linguistic phenomena captured by topic-sensitive word representations. 
Namely, we show that different senses of a word are separated into different representations. 
We observe that the additional context of a document topic is most beneficial when the task is more complex.
We find that these representations achieve improvements over the baselines for word similarity and lexical substitution tasks. 

\end{enumerate}

Having examined the effectiveness of learning word representations using auxiliary contextual information, we then investigate how the diversity of the context affects language understanding and transfer of meaning between two languages. Concretely we ask:

\item \acl{rq:tdabt}  \label{rq:tdabt}

In this research question, we choose machine translation as the task of interest.
We investigate this question by diversifying the local context for different words and propose various data augmentation techniques with the new contexts. 
Additionally, we explore the influence of these synthetic contexts on translation quality. 
We divide this research question into four sub-questions and discuss them in Chapters~\ref{chapter:research-02} and~\ref{chapter:research-03} of this thesis. 

\begin{enumerate}[label=\textbf{RQ2.\arabic*},wide = 0pt, leftmargin=2em]
\setlength\itemsep{1em}
\item \acl{rq:tda1} \label{rq:tda1}

\medskip

In this question, we are interested in translation of rare words in low-resource settings where the available data is scarce for one or both languages. 
The success of neural networks is partly due to their ability to learn from vast amounts of data efficiently. 
These models suffer significantly when sufficient data is not available \citep{ngo-etal-2019-overcoming}.
Subsequently, even with adequate data, neural machine translation models have difficulty learning the meaning of rare words existing in the source language \citep{koehn2017six}. 
Additionally, they are also not successful in generating rare words in the target language \citep{luong2014addressing}.  
To answer this question, in Chapter~\ref{chapter:research-02}, we propose a data augmentation technique that targets rare words and substitute them in new sentences with novel contexts. 
Leveraging a monolingual corpus, which is available in much larger quantities in comparison to a bilingual corpus, we create new contexts for rare words in the training data.
We investigate how additional data can improve the learning and the generation of rare words. 
In Chapter~\ref{chapter:research-02}, we show that by increasing the diversity of the contexts of rare words, we can achieve significant improvements in translation quality.

\item \acl{rq:tda2} \label{rq:tda2}

\medskip

Diversifying context is only valid when both source and target sentences are modified, i.e., at training time when the model has access to the sentence pairs.
During inference, only the source sentence is available and we use the reference sentence solely for evaluation. 
As a consequence, any changes to the source sentence have to be meaning-preserving so that we do not modify the reference translations.
We propose a data augmentation technique at test time, focusing on \textit{paraphrasing} rare and unknown words in the source sentence. 
In contrast to our previous approach where the goal was to diversify the context of rare words in the training data, here we substitute rare words with more common synonyms.
In Chapter~\ref{chapter:research-02}, we show that with paraphrasing rare words at test time, we gain improvements in translation quality.

\item \acl{rq:bt1} \label{rq:bt1}

\medskip

In the previous research questions, we identify rare words as words that can benefit from additional contexts. 
While the translation quality of these words improves with our proposed data augmentation technique, these are not the only words that suffer due to inadequacies in the training data. 
In Chapter~\ref{chapter:research-03}, we expand our investigation in this direction. 
Rather than using features like frequency in the training data, we look into the \textit{model} itself and where it struggles.
We detect the words for which the model has low confidence during translation. 
We examine various approaches to identify these low-confidence words as signaled by the model and augment the training data accordingly. 
Hence, we ask:

\item \acl{rq:bt2} \label{rq:bt2}

\medskip

To generate new contexts and augment the training data, we propose targeted back-translation. 
Back-translation leverages monolingual data in the target language and a trained translation model to translate randomly selected sentences into the source language \citep{sennrich-haddow-birch:2016:P16-11}. 
The automatically generated bilingual data, although noisy, is added to the training data and the translation model is trained on the augmented data.  
In Chapter~\ref{chapter:research-03}, we identify words that can benefit from diverse context.
We show that by back-translating sentences containing low-confidence words, we achieve improvements over the baselines. 

\end{enumerate}

Having demonstrated the advantages of using contextual cues in various forms to improve word representation learning and translation modeling, we come to the final research question of this thesis.
Here, we investigate the shortcomings of relying on the observed context. 
Concretely we ask:

\item \acl{rq:vol} \label{rq:vol}

While the success of neural networks in NLP is indisputable, it is well worth to ask whether neural networks have hidden vulnerabilities. 
In this research question, we also choose machine translation as the task of interest.
We are interested in vulnerabilities of the translation models that can be exposed by looking into the data. 
In particular, we divide this research question into four sub-questions and discuss them in Chapters~\ref{chapter:research-04} and~\ref{chapter:research-05} of this thesis:

\begin{enumerate}[label=\textbf{RQ3.\arabic*},wide = 0pt, leftmargin=2em]
\setlength\itemsep{1em}
\item \acl{rq:id1} \label{rq:id1}

\medskip

Neural translation models struggle in handling idiosyncratic linguistic patterns. 
One of these patterns are idioms, which are semantic lexical units whose meaning is not merely a function of the meaning of its constituent parts.
In Chapter~\ref{chapter:research-04}, we look into idiomatic expressions in particular and why the translation of such phrases is a challenge. 
Furthermore, we automatically label parallel training and test data for idiomatic expressions using a bilingual dictionary of idioms.
We assess whether the sentential context is enough for inferring idiomatic meanings and show that it is indeed not the case.

Next, we ask:

\item \acl{rq:id2} \label{rq:id2}

\medskip

There is no explicit indicator in the data to signal whether a phrase should be translated literally or idiomatically in any given context. 
Researchers have shown that neural models can benefit from side constraints in data in various cases. For instance, 
\citet{sennrich-etal-2016-controlling} note that adding side constraints as unique tokens at the end of the source text help the model translate to the desired level of politeness. 
In Chapter~\ref{chapter:research-04}, we investigate whether a similar technique is useful for the translation of sentences containing idiomatic expressions. 

\medskip

\noindent Finally, we look into other vulnerabilities of neural models which can be highlighted by contextual cues. 
Our next research question focuses on other cases where NMT models fail to generate a correct translation. 
To investigate this question, we first examined how to expose this shortcoming in translation models by asking:  

\item \acl{rq:vol1} \label{rq:vol1}

\medskip

In Chapter~\ref{chapter:research-05}, we ask how receptive the translation models are to manipulations of data. While previous works have investigated the performance of neural models when encountering noise in the form of adversarial instances \citep{goodfellow6572explaining,D18-1050,DBLP:journals/corr/abs-1711-02173}, we are interested in unexpected performance when the data is \textit{not} noisy.

Next, we investigate the robustness of neural translation models by asking:

\item \acl{rq:vol2} \label{rq:vol2}

\medskip

We propose an approach to generate contextual modifications in the test data, yielding semantically and syntactically correct sentences.
Our new test data sheds light on volatile behaviour in current state-of-the-art translation models. 
In Chapter~\ref{chapter:research-05}, we show that identifying this volatility is already achievable with extremely minor modifications.
Our findings highlight unexpected but recurring patterns of errors and possible problems of generalization in neural translation models.

\end{enumerate}
\end{enumerate}

%% Lists the main contributions of the thesis
% !TEX root = thesis-main.tex

\section{Main contributions}
\label{section:introduction:contributions}

Here we summarize the main algorithmic and empirical contributions of this thesis to the field of natural language processing and in particular machine translation, as well as the constructed resources.

\subsection{Algorithmic contributions}

We develop novel learning algorithms and neural network models for investigating the influence of context in learning capacities of models. 

\begin{enumerate}
\item We present a framework for learning multiple embeddings per word using topical context.  With three variants of our model, we employ topical context in various ways and learn distinctions between different senses of the words (Chapter~\ref{chapter:research-01}).
\item We introduce a data augmentation technique for generating new contexts for rare words in machine translation. 
Leveraging monolingual data, we propose a neural language model that given a sentence, suggests rare words to substitute into the given context. 
This new method can be applied to any low-resource language pair as long as there are monolingual data available in both languages (Chapter~\ref{chapter:research-02}).
\item We introduce a novel method to identify difficult words, where the neural translation model has low prediction confidence.
Leveraging this information, we improve upon an existing augmentation technique by replacing its random selection with targeted selection and specifically provide new contexts for low-confidence words (Chapter~\ref{chapter:research-03}).
\item We propose a procedure to (i) automatically detect idiomatic expressions in sentences using a dictionary of idioms, and (ii) automatically annotate the bilingual data with the corresponding idioms (Chapter~\ref{chapter:research-04}).
\item We introduce an effective technique to shed light on the lack of robustness of neural translation models. 
Our approach generates variants of the same sentences that differ slightly and are semantically and syntactically correct. 
We investigate the behaviour of the neural model in translating these variants by proposing metrics to identify volatile performance (Chapter~\ref{chapter:research-05}).
\end{enumerate}

\subsection{Empirical contributions} 

We evaluate our proposed models on large scale data sets as well as controled experiments to validate our hypotheses. We provide empirical results for each research question asked in this thesis. 
More specifically:

\begin{enumerate}
\item We compare how different approaches of incorporating topical context affect the resulting representations. 
We assess the topic-sensitive word representations on word similarity and lexical substitution tasks and perform a qualitative analysis between different representations of a word (Chapter~\ref{chapter:research-01}).
\item We evaluate the effectiveness of our first data augmentation approach in machine translation for two language directions: English$\rightarrow$German and German$\rightarrow$English.
We simulate a low-resource setting by only using a subset of the available training data, while simultaneously being able to compute the upper bound of performance in case more data is available.
Our approach successfully mitigates the problem of rare word translation, where sufficient bilingual training data is not available. 
We perform an analysis of the confidence of the translation model for both generating and translating rare words (Chapter~\ref{chapter:research-02}).
\item We evaluate our second proposed data augmentation approach in machine translation for two language directions: English$\rightarrow$German and German$\rightarrow$English.
We study the effects of previous data augmentation techniques on confidence and the learning capacity of the translation model. 
We compare various ways of identifying low-confidence words and show that targeted data augmentation using these words improves translation quality.
We demonstrate that with diversifying contexts of difficult words, the confidence of the model in predicting these words and consequently the translation quality improve (Chapter~\ref{chapter:research-03}).
\item We conduct an empirical evaluation of translation models facing sentences that include an idiomatic expression. 
Using annotated training and test data, we demonstrate how the current neural translation models struggle with translating idioms.
We show that even when we annotate them in the training data, translating these expressions is a challenge and the translation models require much broader knowledge to learn them (Chapter~\ref{chapter:research-04}).
\item  We show that fluctuations in translations of extremely similar sentences are more prominent than expected. These findings can be used to develop more robust models (Chapter~\ref{chapter:research-05}).
\end{enumerate}

\subsection{Resource contributions}

We release the resources of the proposed models in this thesis including source codes and annotated data. More specifically:

\begin{enumerate}
\item Chapter~\ref{chapter:research-01}: We released the code for the proposed models where we use document topics to learn word representations.
\item Chapter~\ref{chapter:research-02}: We released the code for targeted data augmentation of parallel corpora using language models.
\item Chapter~\ref{chapter:research-04}: We released the annotations of idiomatic phrases in training, development, and test data. The bilingual corpora can be used for translation of English$\rightarrow$German and German$\rightarrow$English.
\item Chapter~\ref{chapter:research-05}: We released a data set which contains multiple variants for each sentence pair in the standard WMT English$\leftrightarrow$German test data.
We annotate the translations of these variants and label different types of errors. Additionally, we release the code for generating sentence variations of bilingual corpora for a more in-depth evaluation of translation quality.
\end{enumerate}

%% Overview of the thesis; what is described in which chapter
% !TEX root = thesis-main.tex

\section{Thesis overview}
\label{section:introduction:overview}

After this introductory chapter, the remainder of this thesis consists of a background chapter (Chapter~\ref{chapter:background}), five research chapters (Chapters~\ref{chapter:research-01}-\ref{chapter:research-05}), and a concluding chapter (Chapter~\ref{chapter:conclusions}). 
Below we present a high-level overview of the main content of each of these chapters.

\paragraph{Chapter~\ref{chapter:background}: Background} provides an introduction to the neural machine translation (NMT) paradigm used in this thesis. We briefly review the core models, the training and test data required, and the learning and optimization strategies we employ. We also discuss different representation learning approaches. Additionally, we describe the basic experimental settings for our systems. 
Finally, we provide an overview of evaluation metrics used in this thesis.  

\paragraph{Chapter~\ref{chapter:research-01}: Representation learning using documental context } introduces the concept of learning multiple representations per word to capture lexical ambiguity in a language. 
We first investigate the influence of document topics on distinguishing different meanings of a word, then propose various models to integrate topical information in representation learning, and finally analyze the performance of these contextual representations and compare them to single representations. 
Our findings in in this chapter provide answers to \textbf{\ref{rq:topic}}.

\paragraph{Chapter~\ref{chapter:research-02}: Data augmentation for rare words} focuses on the impact of additional context in influencing translation quality of rare words. 
Notably, we use language models to substitute rare words in existing bilingual contexts. 
We augment the translation model with the newly generated data and as a result, improve both the generation frequency and the translation quality of rare words.
Our results in this chapter provide answers to \ref{rq:tda1} and \ref{rq:tda2}.

\paragraph{Chapter~\ref{chapter:research-03}: Data augmentation based on model failure} examines the influence of augmenting data with diverse context for difficult words on translation models. 
We first inspect the learning process of state-of-the-art translation models and identify where they are not confident in their predictions. 
After further analyzing the words that translation models have difficulties in learning, we introduce an augmentation approach to target these words.
We improve upon an existing data augmentation approach by devising new contexts for low-confidence words.
Our results in this chapter provide an answer to \ref{rq:bt1}~and \ref{rq:bt2}.

\paragraph{Chapter~\ref{chapter:research-04}: Analyzing idiomatic expressions} investigates translation errors prevalent in current models. 
First, we identify multiword expressions that are syntactically or semantically idiosyncratic and challenging to translate. 
Next, we create a parallel corpus consisting of sentence pairs with idiomatic expressions.
For this study, we introduce new error analysis measures to evaluate the translation quality of these expressions individually.
We provide empirical answers to \ref{rq:id1} and \ref{rq:id2} in this chapter.

\paragraph{Chapter~\ref{chapter:research-05}: Analyzing volatility} investigates the robustness of state-of-the-art translation models to variants in source sentences. 
We propose an effective technique to generate modifications in test sentences while avoiding the introduction of semantic or syntactic noise.
Investigating the translation outputs of different models on the modified test corpus reveals the extent of volatility that exists in translation models.
We perform an analysis of robustness of our models to answer \ref{rq:vol1} and \ref{rq:vol2}.

%Finally, we summarize findings from all research chapters in the concluding chapter:

\paragraph{Chapter~\ref{chapter:conclusions}: Conclusion} concludes this thesis by revisiting the research questions and their corresponding answers. 
We also reflect on future research directions and on what the community can learn from the findings in this thesis.

%% Describes the papers from which the chapters  originate
% !TEX root = thesis-main.tex

\section{Origins}
\label{section:introduction:origins}

The research presented in Chapters~\ref{chapter:research-01}-\ref{chapter:research-05} of this thesis is based on a number of peer-reviewed publications. 
Below, we indicate the origins of each chapter.

\paragraph{Chapter~\ref{chapter:research-01}} is based on Marzieh Fadaee and Arianna Bisazza and Christof Monz,
``Learning Topic-Sensitive Word Representations'',
\textit{In Proceedings of the 55th Annual Meeting of the Association for Computational Linguistics (ACL)},
%pages 441--447,
%Vancouver, Canada,
%July 2017.
\citep{fadaee-etal-2017-learning}.
Fadaee designed and carried out the experiments. All authors contributed to the discussion and text.

\paragraph{Chapter~\ref{chapter:research-02}} is based on Marzieh Fadaee and Arianna Bisazza and Christof Monz,
``Data Augmentation for Low-Resource Neural Machine Translation'',
\textit{In Proceedings of the 55th Annual Meeting of the Association for Computational Linguistics (ACL)},
%pages 567--573,
%Vancouver, Canada,
%July 2017.
 \citep{fadaee-bisazza-monz:2017:Short2}.
Fadaee designed the methods, performed the experiments and wrote most of the text.
Bisazza and Monz contributed to the discussion and editing.

\paragraph{Chapter~\ref{chapter:research-03}} is based on Marzieh Fadaee and Christof Monz,
``Back-Translation Sampling by Targeting Difficult Words in Neural Machine Translation'',
\textit{In Proceedings of the 2018 Conference on Empirical Methods in Natural Language Processing (EMNLP)},
%pages 436--446,
%Brussels, Belgium,
%October 2018.
\citep{fadaee-monz-2018-back}.
Fadaee designed the methods, performed the experiments and wrote most of the text.
Monz contributed to the discussion and editing.

\paragraph{Chapter~\ref{chapter:research-04}} is based on Marzieh Fadaee and Arianna Bisazza and Christof Monz,
``Examining the Tip of the Iceberg: A Data Set for Idiom Translation'',
\textit{In Proceedings of the Ninth International Conference on Language Resources and Evaluation (LREC)},
%pages 925--929,
%Miyazaki, Japan,
%May 2018.
\citep{L18-1148}.
Fadaee designed the methods, performed the experiments, and wrote the text.
Bisazza and Monz contributed to the discussion and editing.

\paragraph{Chapter~\ref{chapter:research-05}} is based on Marzieh Fadaee and Christof Monz,
``The Unreasonable Volatility of Neural Machine Translation'', 
\textit{In Proceedings of the 4th Workshop on Neural Generation and Translation (WNGT)},
%Seattle, Washington, USA.
%July 2020.
\citep{fadaee_new}.
Fadaee designed the methods, performed the experiments, and wrote the text.
Monz contributed to the discussion and editing.

% !TEX root = thesis-main.tex

\chapter{Background}
\label{chapter:background} \label{nmtsec}

Neural machine translation (NMT) is an end-to-end learning approach to machine translation that is based on neural networks.
In contrast to traditional translation systems such as phrase-based machine translation (PBMT) \citep{koehn-etal-2003-statistical}, all components of the neural translation model are trained jointly to maximize translation performance.
In this chapter, we discuss the NMT paradigm and the properties of building a translation model.

The training data, in the format of parallel data, is a fundamental part of building NMT models.
We first explain the training data used in the NMT paradigm in Section~\ref{databg}, followed by an overview of data preparation and building the translation vocabulary in Section~\ref{bgvocab}.
Next, we discuss different word representation models in Section~\ref{bgemb}. 
In the following sections, we review the two main NMT frameworks used in this thesis: recurrent neural networks (Section~\ref{RNN}) and the transformer model (Section~\ref{TRNN}). 
Both models are classes of artificial neural networks and use large amounts of parallel data to learn a translation model. 
Finally, in Section~\ref{bgexp}, we describe the evaluation approaches used in the later chapters of this thesis. 

\section{Parallel and monolingual corpora} \label{databg}

Neural models, and specifically neural translation models, rely heavily on training data. 
The primary training data for learning translation models are parallel corpora, which are aligned texts in two or more languages. 
These corpora are often paired at the sentence-level, ideally providing an exact translation of every sentence in the source and target language. 

The quality of the translation system depends on the quality and the size of the training data. 
Acquiring good-quality parallel corpora requires manual translation by professional translators and as a result is expensive.
Examples of available parallel corpora gathered by experts in the domain include Europarl
\citep{koehn2005europarl}, which is the proceedings of the European Parliament published on the web, and JRC-Acquis 
\citep{steinberger2006jrc}, which is the total body of the European Union law applicable in the EU Member States.
\citet{callisonburch-EtAl:2007:WMT} gathered News Commentary corpora which consist of political and economic commentary crawled from the web site Project Syndicate.
This data is extracted every year for the translation task of the WMT conference \citep{barrault-EtAl:2019:WMT}. 

Monolingual data, in comparison, are available in abundance for many languages. 
Phrase-based machine translation models use monolingual corpora in the target language \citep{koehn-etal-2003-statistical, brants-etal-2007-large, koehn-etal-2007-moses} to improve the fluency of the generated translation \citep{lembersky-etal-2011-language}. 
Monolingual parallel corpora of aligned complex-simple sentences are also used with pharse-based \citep{wubben-etal-2012-sentence,kajiwara-komachi-2016-building} and neural \citep{zhang-lapata-2017-sentence} models to learn to simplify text.
The monolingual News Crawl corpus from WMT and many available corpora in the Linguistic Data Consortium (LDC)\footnote{\url{https://www.ldc.upenn.edu/}} are examples of commonly utilized data in machine translation.

Vanilla NMT models typically do not use any monolingual data in their training. 
In Chapters~\ref{chapter:research-02} and~\ref{chapter:research-03} of this thesis, we address this matter by focusing on the use of monolingual data for NMT.
Recently there have been studies that propose various approaches for incorporating information from monolingual data in the models \citep{domhan-hieber-2017-using,burlot-yvon-2018-using,currey-heafield-2019-zero}. 
\citet{currey2017copied} created a parallel corpus from monolingual data in the target language by copying it so that each source sentence is identical to its corresponding target sentence. With this simple technique, they observe improvements on relatively low-resource language pairs.
Another category of approaches is to translate sentences from monolingual data and augment the bitext with the resulting pseudo parallel corpora. 
This category of approaches is discussed in the following section.

\subsection{Back-Translation in machine translation} \label{bgbtref}

In this section, we introduce the conventional method of generating synthetic data, namely back-translation and its effectiveness in PBMT and NMT.
Back-translation uses an intermediate MT model, trained on parallel data, to translate target monolingual data into the source language.
The result of back-translation is a parallel corpus where the source side is synthetic MT output while the target is actual text written by humans.

This technique is not bounded to neural networks, and prior to NMT models, it has been used in combination with PBMT. 
\citet{Schwenk2008InvestigationsOL} proposes to translate large amounts of monolingual data with a PBMT system and use those as additional training data. They observe that this lightly-supervised training achieves improvements in translation quality.
\citet{Rapp:2009:BSA:1667583.1667625} introduces the back-translation score as an alternative mean for the evaluation of PBMT models.
He trains a translation model in both directions and evaluates the quality of the model by translating the target sentences back to the source language. 
The score is therefore computed by comparing the back-translated sentence to the original source sentence. 
As part of their experiments, \citet{tiedemann-etal-2016-phrase} note that back-translating sentences from monolingual news data and augmenting the parallel training data improves the translation quality of a PBMT system. 
In these experiments, the models have to be re-tuned from scratch with the additional synthetic data.

In the framework of NMT, \citet{sennrich-haddow-birch:2016:P16-11} show that back-translating sentences from monolingual data improves the performance of NMT models. 
This approach of augmenting the training data has since become common practice in training NMT models. 
\citet{pham2017karlsruhe} experimented with using domain adaptation methods to select monolingual data for back-translation based on the cross-entropy between monolingual data and the in-domain corpus \citep{axelrod2015class}, 
but did not find any improvements over random sampling as initially proposed by \citet{sennrich-haddow-birch:2016:P16-11}.

\citet{edunov-etal-2018-understanding} investigate back-translation in NMT at a large scale by adding hundreds of millions of back-translated sentences to the bitext.
They study different methods for generating synthetic sentences and show that synthetic data based on sampling and noised beam search provides a stronger training signal than using pure beam.
They observe that the generated corpora tend to stray away from the distribution of natural data.
\citet{brants-etal-2007-large} suggest a distributed language model infrastructure, which allows direct integration into the hypothesis-search algorithm.  
They observe that translation quality continues to improve with increasing language model size.
\citet{Ueffing2007} use an iterative procedure that translates the monolingual source language data in each iteration and then re-trains the phrased-based translation model.
They conclude that when bilingual training data are scarce, a PBMT system could be trained on a small amount of data and then iteratively improved by adding reliable translations of monolingual data to the training data. 

\citet{NIPS2016_6469} observe that any machine translation task has a dual task, for instance, English$\rightarrow$French translation (primal) versus French$\rightarrow$English translation (dual). 
They propose an approach based on reinforcement learning, where two agents, representing the primal and dual task, teach each other. 
The agents leverage monolingual data by translating it forward to the other language and then translate backward to the original language.
\citet{gulcehre2017integrating} propose two methods, shallow and deep fusion, for integrating a neural language model into an NMT system.
They observe improvements by combining the scores of a neural language model trained on target monolingual data with an NMT system.

\section{Translation vocabulary} \label{bgvocab}

In translation models, the vocabulary of the source and target language is defined as what the model is exposed to during training. 
Word-level translation models are unable to translate or generate unseen words at inference.
The number of words in the vocabulary can be remarkably large and training models on large vocabularies is computationally expensive.  

An early practice was to limit the vocabulary to the $K$ most frequent words, where $K$ is often in the range of 30k \citep{DBLP:journals/corr/BahdanauCB14} to 80k \citep{sutskever2014sequence}.
The tail of the vocabulary not included in this shortlist is mapped to a special token \texttt{[unk]} representing an unknown or out-of-vocabulary word.
This method results in neural models that can be trained and tested within a reasonable amount of time, however, as a consequence of this simplification, the translation quality of the model suffers. 
Specifically, the performance decreases significantly when the translation of a source sentence requires many unknown words \citep{cho2014properties}.

To address this issue, \citet{jean-etal-2015-using} proposed an approximate training algorithm that can use a very large target vocabulary (vocabularies of 500,000 source and target words). 
They show that decoding the target sentence by sampling only a small subset of the whole vocabulary achieves competitive results without sacrificing too much speed.
\citet{luong2014addressing} proposed a copy mechanism that aligns the OOV words on both the source and the target side by learning to copy indices.
\citet{sennrich-haddow-birch:2016:P16-12} analyzed NMT models that work with subword units and observed that the majority of tokens are potentially translatable through smaller units. 
They modify Byte Pair Encoding (BPE) \citep{10.5555/177910.177914} to segment words into subword units, where each of which should be frequently observed in the corpus. 
While some segmentations correspond to correct morphemes, for many words that is not the case. 
For instance, the word \textit{`quixotism'} would be segmented into \textit{`quixot + -ism'} and the word \textit{`sceptical'} would be segmented into \textit{`scep + -tic + -al'}.
This approach is very effective in generalization and is able to generate words not seen during training using these subword units.

In this thesis, we segment words during preprocessing using the BPE technique in all translation experiments unless stated otherwise.
We refer to the subword units throughout the chapters as \textit{tokens}. 

\section{Word representations} \label{bgemb} 

The first step in using neural models for text is to map the words in the vocabulary to \textit{dense vectors} of real numbers.
These vectors are somewhat similar to \textit{sparse vectors} used in distributional semantics, where they represent meaning by capturing similarities between lexical units based on their distributional properties \citep{baroni-etal-2014-dont,baroni-lenci-2010-distributional}.
The context of the lexical unit is commonly used for the computation of dense and sparse embeddings.
The intuition is that since similar words appear in similar contexts, they end up with similar embeddings \citep{firth1957synopsis}.
Computation of dense vectors is often a by-product of solving a natural language processing task such as language modeling or translation. 

Word representations can be categorized into two groups \citep{Wang2019UsingDE}: \textit{static embeddings} where a fixed vector is learned for each word in the vocabulary, and \textit{dynamic embeddings} where vectors are dynamically calculated for each sentence. 
In the next sections, we discuss different approaches in each category.

\subsection{Static embeddings}  \label{bgembstatic} 

\begin{figure}
\centering
\begin{subfigure}[t]{.45\textwidth}
  \centering
\includegraphics[width=0.8\linewidth]{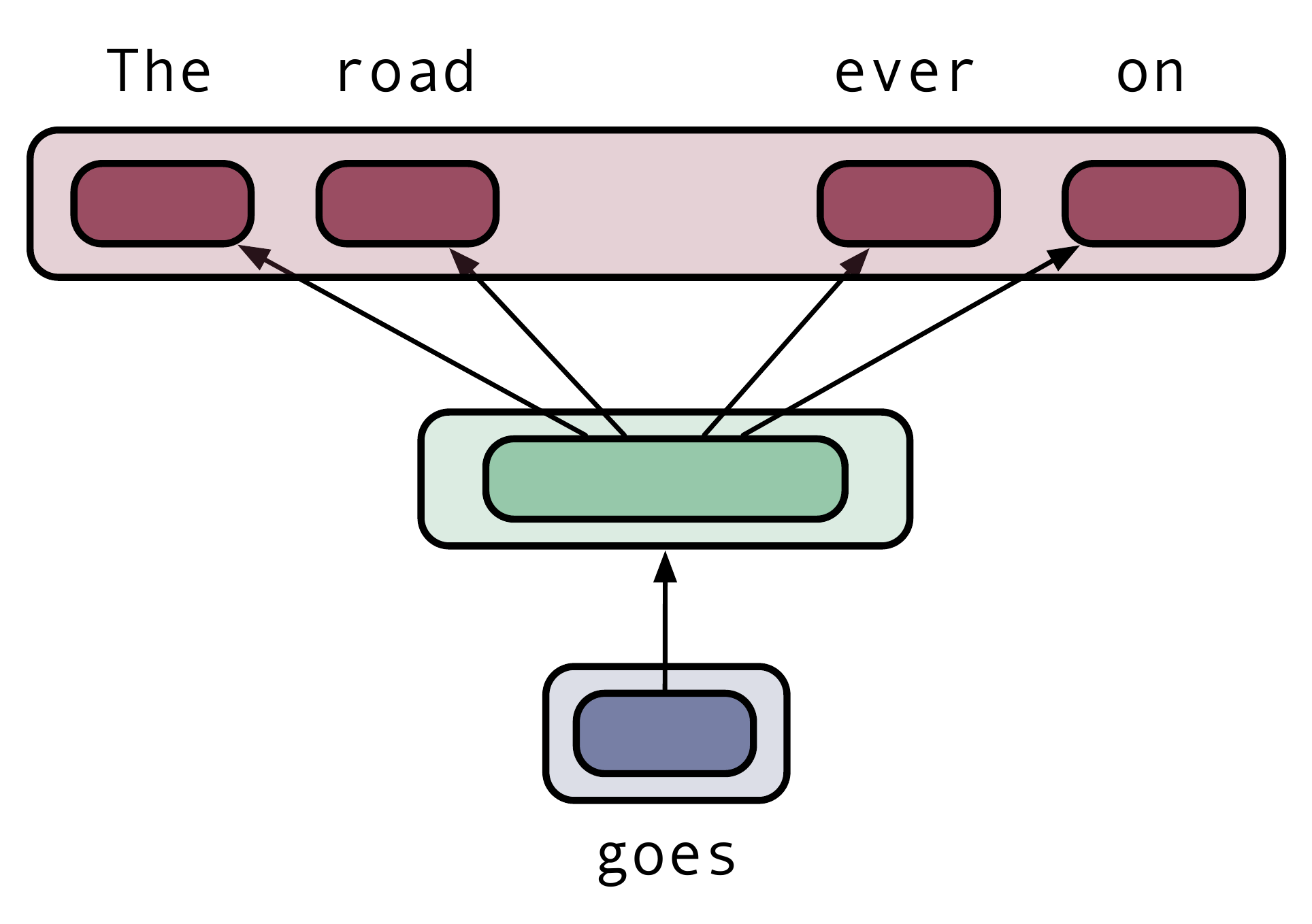}
\caption{Skipgram model}
\label{bgsgfig}
\end{subfigure}%
\begin{subfigure}[t]{.45\textwidth}
  \centering
\includegraphics[width=.8\linewidth]{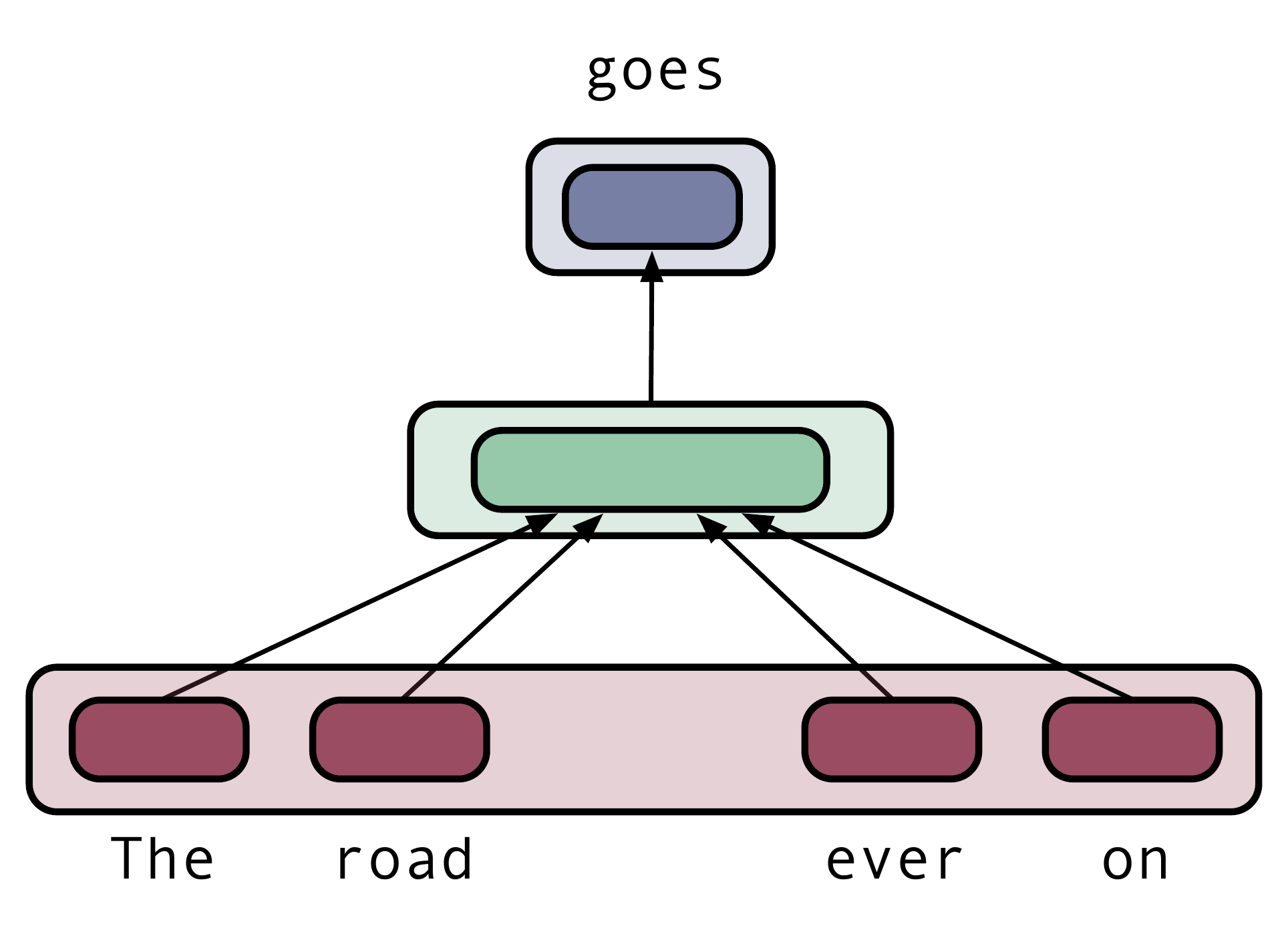}
  \caption{CBOW model}
  \label{bgcbowfig}
\end{subfigure}
\caption{Representation learning architectures proposed by \citet{mikolov2013efficient}. The CBOW model predicts the current word given the context, and the Skipgram model predicts the surrounding words given the current word}
\label{bgsgcbowfig}
\end{figure}

Traditional word embedding techniques learn a global and static word embedding for every word in the vocabulary. 
\citet{mikolov2013efficient} proposed two models for learning word representations: continuous Skipgram and continuous bag-of-words (CBOW).
Both models use a feed-forward neural network architecture with the objective of modeling language, illustrated in Figure~\ref{bgsgcbowfig}.
This architecture does not include any non-linearity. 
The CBOW model has a projection layer which is shared between all words. This layer averages the input vectors.
Next, using a classifier, the model predicts the word $w_t$ given the context of words surrounding $w_t$ in a fixed sized window: $[w_{t-c}, \ldots, w_{t+c}]$. The objective of the CBOW model is to maximize the following average log probability:
\begin{align}
\frac{1}{T} \sum_{t=1}^T \log p(w_{t} \mid w_{t-c}, \ldots, w_{t-1}, w_{t+1}, \ldots, w_{t+c})
\end{align}

\noindent where $T$ is the length of the sequence of training words and $c$ is the context window size.
The Skipgram model is similar to CBOW, but instead of predicting $w_t$, the model predicts the words within a fixed range surrounding $w_t$.
The objective of the Skipgram model is to maximize the following average log probability:
\begin{align}
\frac{1}{T} \sum_{t=1}^T \sum_{\substack{-c \le j \le c \\ j \ne 0}} \log p(w_{t+j} \mid w_{t})
\end{align}

\noindent where $T$ is the length of the sequence of training words.
Note that in both CBOW and Skipgram models, the context window includes both the past and the future.

\citet{pennington2014glove} combined count-based matrix factorization and context-based Skipgram model together.
The intuition is that meaning of words can be captured by the ratios of co-occurrence probabilities.
They proposed a weighted least squares model that trains on global word co-occurrence counts. 
They showed that the vector space learned from this model captures meaningful vector space substructures.
While some syntactic and semantic features in language are captured by these word embeddings \citep{mikolov-etal-2013-linguistic}, the dimensions are often not interpretable. 

These models utilize surrounding words as context.
However, word representations can capture different phenomena if the definition of context is different. 
\citet{levy2014dependency} proposed to use dependency-based contexts, extracted from dependency parse-trees. 
They observed that these embeddings are less topical and exhibit more functional similarity than the original Skipgram embeddings.

Static embeddings for the most part learn a static matrix of embeddings for each word type and ignore capturing some nuances of language such as ambiguity. 
In Chapter~\ref{chapter:research-01}, we address this issue by exploring document topics and learning multiple embeddings per word type to capture polysemy. 

\subsection{Dynamic embeddings} 

Static models generate out-of-context embeddings for word types and are simple and efficient to train and use. 
However, learning meaningful word representations has recently been elevated beyond this paradigm. 
Rather than learning static representations for word types, these models learn \textit{dynamic} vectors for word instances in context using language modeling objectives. 
We denote this kind of embeddings as dynamic because instead of a static matrix of embeddings, they are obtained through the hidden states of a language model given the context.

\citet{peters-etal-2018-deep} proposed to use a bidirectional recurrent neural network to extract context-dependent representations.
The learning objective is to predict the next word in a sequence, given the previous context words. 
\citet{devlin-etal-2019-bert} use a transformer architecture and define two new objectives for training: \textit{masked language modeling}, and \textit{next sentence prediction}.
During masked language modeling, they mask a randomly selected word in a sentence, and the model has to predict that word given the context.
The second objective gets two input sentences and predicts whether the second sentence is indeed the next sentence.
The contextualized word embeddings are successful at downstream NLP tasks such as question answering and textual entailment \citep{zhang2019semanticsaware,garg2019tanda,Lan2020ALBERT:,mandar2020}.

While word vectors in neural translation models can be initialized with these static or dynamic word representations, they are often initialized randomly \citep{wu2016google}. 
One reason can be that with large-scale parallel data, these initial word representations will be forgotten during the training of the NMT model. 
\citet{qi-etal-2018-pre} showed that for low-resource language pairs and when languages are more similar, pre-trained embeddings can be effective.
\citet{lewis2019bart} recently proposed a denoising autoencoder model named BART for pretraining sequence-to-sequence models. 
They corrupt text with an arbitrary noising function and learn a model to reconstruct the original text.
BART is effective when fine-tuned for text generation and translation but also works well for comprehension tasks.
The research presented in this thesis mostly predates the work mentioned in this section.
We use \textit{static} embeddings in Chapter~\ref{chapter:research-01} where we investigate the effect of having more than one representation per word type. 
As for the chapters on machine translation, we consider the most widespread setup where embeddings are initialized randomly before training on the parallel data.

\section{Recurrent translation models} \label{RNN}

In this section, we discuss a category of neural models that are effective in modeling languages. 
Earlier developments of neural models addressing language modeling tasks incorporated the temporal structure of the language in the structure of the network \citep{Jacquemin1994ATC,Schmidhuber:HabilitationThesis}.  
A recurrent neural network (RNN) is an example of these sequential models \citep{10.5555/65669.104451}.
RNNs are powerful models that achieve state-of-the-art results in a variety of tasks such as question answering \citep{garg2019tanda}, reading comprehension \citep{Zhang2020RetrospectiveRF}, image semantic segmentation \citep{yuan2019objectcontextual}, and speech recognition \citep{8049322}. 

RNN models are capable of modeling sequences of text with various length, while selectively passing on information between different time steps in the sequence. 
A long short-term memory (LSTM) network \citep{hochreiter1997long} is an RNN structure that uses special LSTM units in addition to standard ones. 
These special units include a memory cell that can maintain information for long sequences.
LSTM models address the \textit{vanishing gradient problem} in the earlier RNN architecture where the weights and biases of the hidden layers are not updated effectively because the gradient decreases exponentially \citep{10.1142/S0218488598000094}. 

\begin{figure}[htb!]
\centering
\includegraphics[width=0.95\linewidth]{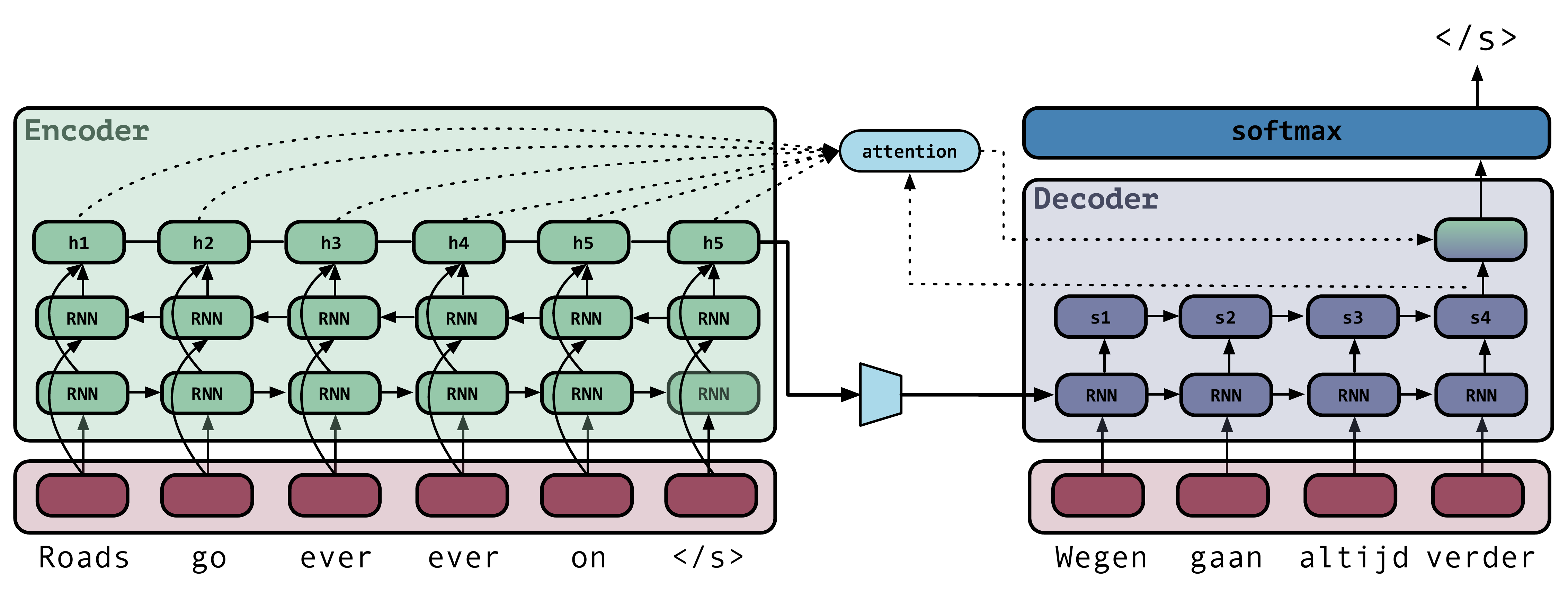}
\caption{An illustration of an RNN encoder-decoder with attention.}
\label{bgRNNfig}
\end{figure}

\citet{sutskever2014sequence} and \citet{cho2014properties} were among the firsts to employ RNNs to build an end-to-end machine translation model.
\citet{DBLP:journals/corr/BahdanauCB14} and \citet{luong:2015:EMNLP} introduced an \textit{attention mechanism} that achieved performance on par with traditional statistical models. 
In the following sections, we describe the RNN architecture with attention used in the NMT experiments in this thesis.

\subsection{Model architecture}

Neural machine translation models fall under a sequence-to-sequence framework where an encoder builds up a representation of the source sentence and a decoder 
generates the target translation. 
Both the encoder and the decoder can be recurrent neural models.
Figure~\ref{bgRNNfig} illustrates this architecture which we will discuss in detail in this section. 
In order to train an NMT system, two sequences of tokens, $X =  \big[ x_1, \ldots, x_n \big] $ and $Y =  \big[ y_1, \ldots, y_m \big] $, are given in the source and target language, respectively.
As discussed in Section~\ref{bgemb}, the input tokens are mapped to an embedding space.
As a result, the source sequence is the input to the encoder as vectors: $\big[ \vt{x}_1, \ldots, \vt{x}_n \big]$.

The encoder then encodes the input sequence into hidden states, where at time step $t$ the hidden state is a function of the current input vector and the previous hidden state:
\begin{align}
\vt{h}_t = f(\vt{x}_t, \vt{h}_{t-1})
\end{align}

Function $f$ adds non-linearities to the transformation of the input sequence to the output of the encoder.
With a bidirectional architecture, two RNNs are run on the input sequence: one in forward and one in backward direction. The hidden state at time $t$ is created by concatenating the forward and backward hidden states
at each point in time, the input has access to the information on both sides. 
Note that the forward and backward hidden states are concatenated to create the top hidden states of the encoder, $\vt{h}_t$ as follows:
\begin{align}
\vt{h}_t = \big[ \overrightarrow{\vt{h}_t}^\intercal; \overleftarrow{\vt{h}_t}^\intercal \big]^\intercal,      t = 1, \ldots, n
\end{align}

%%%%%%%%%%%%%%%%%%

The decoder then generates the target translation one word at a time starting with the last hidden state of the encoder and the representation for the start-of-sentence symbol \mbox{\texttt{\textless s\textgreater}}.
Each decoder hidden state $\vt{s}_t$ is computed as:
\begin{align}
\vt{s}_t = g(\vt{s}_{t-1}, \vt{y}_{t-1}, \vt{c}_t)
\end{align}

\noindent where $g$ is a transformation function that outputs a vocabulary-sized vector and $\vt{y}_{t-1}$ is the representation of the previously predicted token.
$\vt{c}_i$ is the context vector for output at position $i$ and is defined as:
\begin{align}
\vt{c}_i = \sum_{j=1}^{n} \alpha_{ij} \vt{h}_j
\end{align}

This context vector is recomputed at each time step.
$\alpha_{ij}$ is the attention weight and it is computed for all source words at each time step $i$.
We will discuss different approaches to computing attention weights in Section~\ref{BGlstmATT}.

Next, the decoder predicts each target token $y_t$ by computing the conditional probability:
\begin{align}
p(y_t \mid y_{<t}, X) = \softmax \,(\vt{s}_t)
\end{align}

This conditional probability is computed over the vocabulary of the target language which is fixed during training and testing. 
For token $y_t$, the conditional probability $p(y_t \mid y_{<t}, X)$ during training quantifies the difficulty of predicting that token in the context $y_{<t}$.
The prediction loss of token $y_t$ is the negative log-likelihood of this probability.
During training on a parallel corpus $\mathbb{D}$, the cross-entropy objective function is defined as:
\begin{align}
\mathcal{L} = \sum_{(X,Y) \in \mathbb{D}} \sum_{i=1}^{m} - \log p(y_i \mid y_{<i}, X)
\end{align}

The objective of this function is to improve the model's estimation of predicting target words given the source sentence and the target context. 
The model is trained end-to-end by minimizing the negative log-likelihood of the target words using stochastic gradient descent.

NMT systems often benefit from multiple layers of stacked RNNs during training \citep{wu2016google}. 
By increasing the number of parameters, the learning capability of the model also increases \citep{britz-etal-2017-massive}.
\citet{belinkov-etal-2017-neural} show that different layers in the encoder capture different linguistic features, namely that higher layers capture semantics while lower layers tend to capture syntax.
Encoding the input sequence in both directions also provides advantages \citep{luong:2015:EMNLP,DBLP:journals/corr/BahdanauCB14}. The backward layer in a recurrent model learns more about the semantics of words, whereas the forward layer encodes more of the local context \citep{ghader-monz-2019-intrinsic}.

\subsection{Inference} \label{bgrnninference}

During inference, a trained model is given a source sentence and it generates the target translation word by word using a left-to-right beam search technique \citep{jelinek98}
This procedure was already adopted by pre-neural translation methods such as phrase-based translation models \citep{koehn-etal-2003-statistical}. 
Generation of target words stops when a special end-of-sentence symbol \mbox{\texttt{\textless /s\textgreater}} is generated. 
At each step, the model computes a probability distribution over all words in the target language and chooses the most likely word:
\begin{align}
\hat{Y} = \underset{Y}\argmax \; p({Y} \mid {X})
\end{align}

\citet{sutskever2014sequence} showed that increasing the beam size beyond 2 does not improve the predictions significantly and even with a beam size of 1, the model performs well. 
With a large enough beam size, the best translation performance can be reached with the drawback of efficiency \citep{freitag-al-onaizan-2017-beam}.
It is common practice to set beam size to around 5 to 10 \citep{wu2016google,edunov-etal-2018-understanding}.
Beam search decoding, even though effective, still suffers from \textit{exposure bias}.
Exposure bias results from the mismatch between how the models are trained and how they are used at inference \citep{wiseman-rush-2016-sequence,DBLP:journals/corr/RanzatoCAZ15}.
During training, the model is guided by the ground-truth target translation. 
However, at inference, target translations are not available and the model has to rely on its own predictions which can be wrong.
\citet{collobert2019a} proposed a fully differentiable beam search decoder that can be used during training and eliminates this bias.

\subsection{Attention mechanism} \label{BGlstmATT}

One of the shortcomings of the discussed models is that the translation quality decreases considerably as sentences become longer \citep{koehn2017six}.
One reason is that the source sentence is encoded into one \textit{fixed length} vector and this vector is expected to be a complete and static representation of the source sentence.
To address this problem, several works focus on learning a context vector with connections to the source sentence \citep{Graves2014NeuralTM,DBLP:journals/corr/BahdanauCB14,luong:2015:EMNLP}. 
This context vector regulates the alignment between the source and the target sentences and is a sum of the hidden states of the input, weighted by alignment scores.
At each time step $t$, the model computes a variable-length alignment weight vector based on the current target state and all source inputs. 
Table~\ref{backgroundattentions} summarizes different approaches for computing alignment scores.

\begin{table}
\centering
\small
\begin{tabularx}{0.93\linewidth}{lll}
 \toprule
\textbf{Name} & \textbf{Proposed by} & \textbf{Alignment score}  \\ \midrule
Additive & \citet{DBLP:journals/corr/BahdanauCB14} & $\text{score}(\vt{s}_i, \vt{h}_t) = \vt{v}^\intercal_a \tanh (\vt{W}_a[\vt{s}_i, \vt{h}_t])$ \\
 Location-base	 & \citet{luong:2015:EMNLP} & $\alpha_{n, t} = \softmax\,(\vt{W}_a\vt{s}_i) $ \\
General	 &  \citet{luong:2015:EMNLP} & $ \text{score}(\vt{s}_i, \vt{h}_t) = \vt{s}^\intercal_i  \vt{W}_a  \vt{h}_t $ \\
Dot-product	 &  \citet{luong:2015:EMNLP} & $\text{score}(\vt{s}_i, \vt{h}_t) = \vt{s}^\intercal_i \vt{h}_t  $ \\
 Scaled dot-product & \citet{vaswani2017attention} & $\text{score}(\vt{s}_i, \vt{h}_t) = \frac{\vt{s}^\intercal_i \vt{h}_t }{\sqrt{n}} $ \\
\bottomrule
\end{tabularx}
\caption{Different alignment scores in the literature used for creating the context vector. }
\label{backgroundattentions}
\end{table}

It is worth noting that while attention matches traditional word alignment at times, it often captures relations beyond that between the source and target sentence \citep{ghader-monz-2017-attention,koehn2017six}.

\section{Fully attention-based translation models} \label{TRNN}

In the previous section, we discussed attention mechanisms where the model selectively attends to the source sequence to make predictions. 
Self-attention is a type of attention mechanism that connects different positions \textit{within} a sequence to compute a representation.
The Transformer model proposed by \citet{vaswani2017attention} is a sequence-to-sequence model that relies solely on attention to encode the input and generate the output sequence. 
One of the main advantages of this architecture is that it can be trained with massive parallelization because it bypasses the recurrent dependency that exists in RNN models.  
The transformer model has been shown to perform quite well in bilingual and multilingual settings \citep{lakew-etal-2018-comparison} and has become the most common choice to implement NMT models \citep{edunov-etal-2018-understanding,aharoni-etal-2019-massively}.
In this section, we describe this architecture in more detail.

\subsection{Model architecture}

\begin{figure}
\centering
\includegraphics[width=0.95\linewidth]{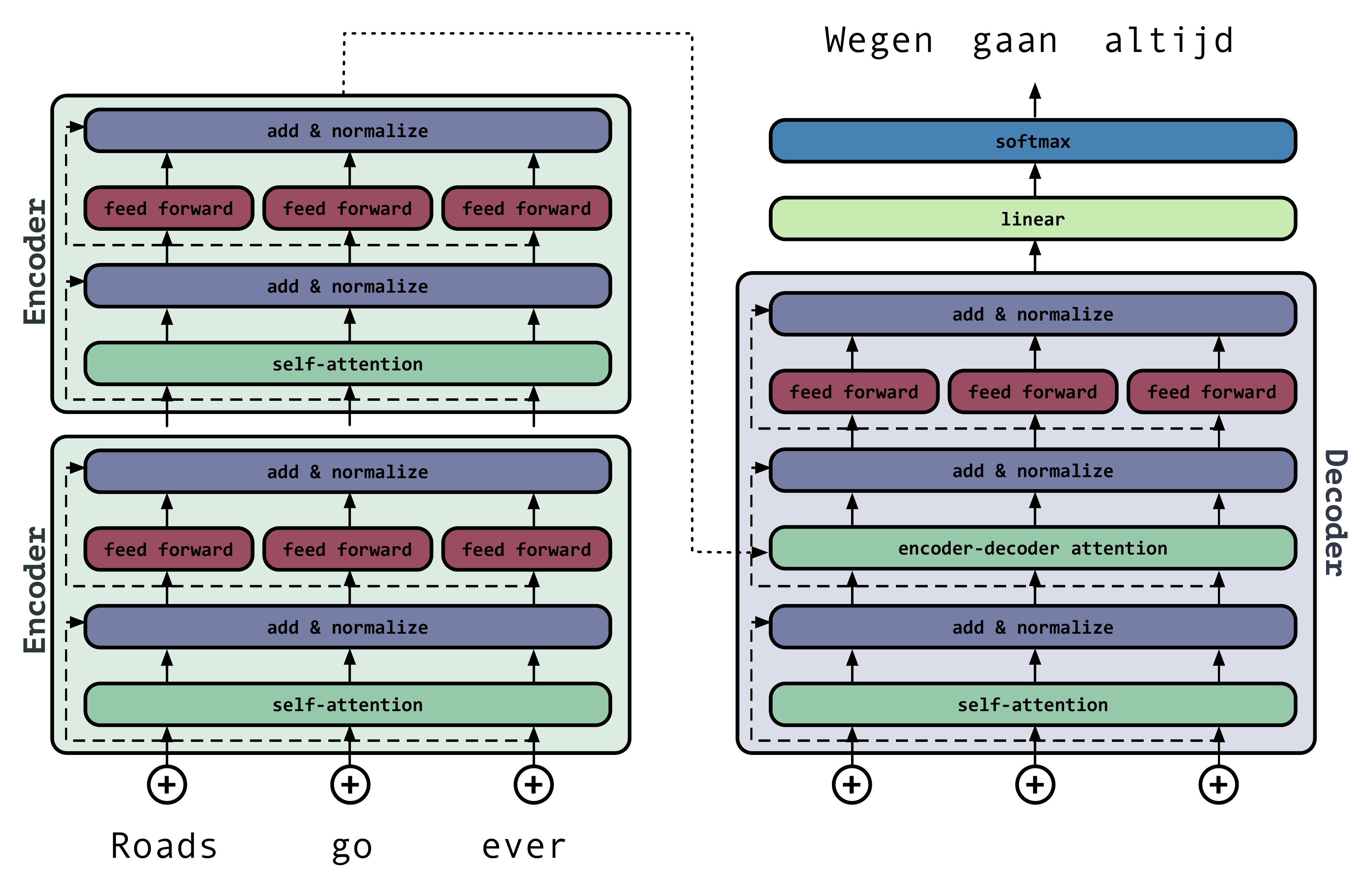}
\caption{An illustration of a Transformer model introduced in \citet{vaswani2017attention}.}
\label{bgTRNfig}
\end{figure}

\begin{figure}
\centering
\includegraphics[width=0.5\linewidth]{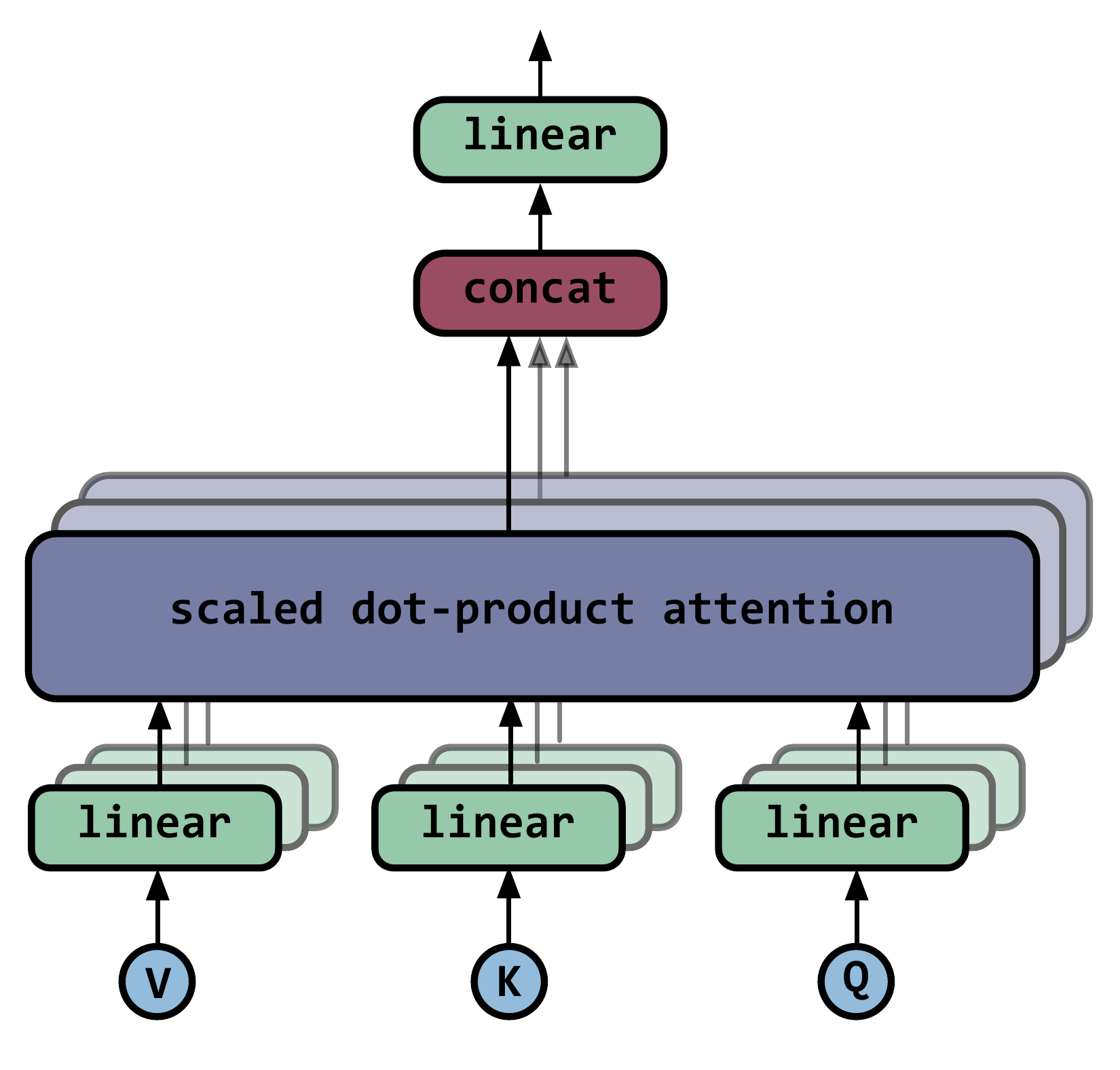}
\caption{An illustration of self-attention in the Transformer model based on \citet{vaswani2017attention}.}
\label{bgTRNattfig}
\end{figure}

The transformer model is an encoder-decoder architecture without a sequential structure.
The encoder is given an input sequence of tokens $X = \big[ x_1, \ldots, x_n \big] $, and encodes it as a continuous representation $\vt{X} = \big[\vt{x}_1, \ldots, \vt{x}_n \big] $ based on the attention. 
The decoder then generates the output sequence $Y = \big[ y_1, \ldots, y_m \big] $ token by token, given the representation $\vt{X}$ and the previously generated token. 
Figure~\ref{bgTRNfig} illustrates this architecture which we will discuss in detail below. 

For every token $x_i$ in the input sequence, we first create a query $\vt{q}_i$, a key $\vt{k}_i$, and a value $\vt{v}_i$ vector.
Self-attention then uses \textit{scaled dot product attention} (last row in Table~\ref{backgroundattentions}) to compute the attention score of token $x_i$ against other words in the input sequence.
This attention has a scaling factor where $n$ is the dimension of the source hidden states.
To calculate representation $\vt{x}_i$, a softmax layer is then used to normalize the self-attention scores and multiplies it with $\vt{v}_i$.
In practice, the attention is computed on matrices of inputs ($\vt{Q}, \vt{K}, \vt{V}$) as follows:
\begin{align}
\attention (\vt{Q}, \vt{K}, \vt{V}) = \softmax (\frac{\vt{K}^\intercal \vt{Q} }{\sqrt{d_k}})\vt{V}
\end{align}

\noindent where $d_k$ is the embedding dimension of the key vectors which scales the dot product. The encoder is a stacking of identical layers each consisting of a \textit{multi-head self-attention} layer and a point-wise fully connected feed-forward network.
$\vt{Q}$, $\vt{K}$, and $\vt{V}$ matrices are split up into multiple heads and the multi-head attention mechanism computes the attention in parallel. 
Each token in the sequence goes through the encoder independently. 
During encoding, there are dependencies between the paths of different tokens in the self-attention layer, but the feed-forward layer of each token does not have any dependencies.
As a result words in the sequence can be processed in parallel. 
The independent attention outputs are then concatenated and linearly projected as follows: %transformed into expected dimensions.
\begin{align}
\text{Multi-Head Attention} (\vt{Q}, \vt{K}, \vt{V}) &= \concat (\text{head}_1, \ldots \text{head}_h) \vt{W}^o 
\end{align}
\noindent
where
\begin{align}
\text{head}_i = \attention(\vt{Q}\vt{W}_i^Q, \vt{K}\vt{W}_i^K, \vt{V}\vt{W}_i^V) 
\end{align}
\noindent
where $\vt{W}_i^Q, \vt{W}_i^K, \vt{W}_i^V$ are weight matrices that map the input representations to the query, key, and value matrices. $\vt{W}^o$ is the linear transformation that generates the output. 
All weight matrices are learned during training of the model.
Figure ~\ref{bgTRNattfig} illustrates this component. 
Similarly to the encoder, the decoder consists of a stack of identical layers, as well as a third sub-layer, which computes multi-head attention over the output of the encoder stack.
The self-attention layers in the decoder work slightly differently from the ones in the encoder. 
The computation of attention is \textit{masked} before the softmax step to prevent looking to the future of the sequence during training. 

Finally, there is a fully connected neural network that transforms the output of the stack of decoders into the target vocabulary vector.
The softmax function turns the scores into probabilities and the word with the highest probability is generated (greedy decoding). 
Alternatively, decoding can be done using the beam search technique similar to the RNN models discussed in Section~\ref{bgrnninference}.

\subsection{Residual connections} 

Another effective detail of the transformer architecture is the inclusion of residual connections \citep{He2016DeepRL} to facilitate optimization.
Residual connections connect the output of one layer with the input of an earlier layer.
Every self-attention and feed-forward neural network in the encoder and the decoder stack has a residual connection around it and a normalization layer \citep{Ba2016LayerN}.
This shortcut connection is particularly effective in training very deep architectures and mitigates the vanishing gradient problem. 

\subsection{Positional Encoding}

As discussed earlier, the transformer model does not have a recurrent structure and can be trained with a high degree of parallelization. 
However, languages are structured sequentially and it is necessary to encode some form of word order in the sequence \citep{tran-etal-2018-importance}. 
To address this shortcoming, the transformer adds a positional encoding vector to every word in the input sequence. 
These embeddings model the position of each word, or the relative distance between different words in the input.

\citet{vaswani2017attention} proposed sine and cosine functions of different frequencies to compute positional encodings:
\begin{align}
\text{positional encoding}_{(i, \delta)} = \begin{cases}
    \sin (\frac{i}{10000^{2\delta'/d}}) & \text{if $\delta = 2\delta'$}\\
        \\[5pt]
    \cos (\frac{i}{10000^{2\delta'/d}}) & \text{if $\delta = 2\delta' + 1$}
  \end{cases}
\end{align}

\noindent where $i$ is the position and $\delta = 1, \ldots, d$ is the dimension.
They also experimented with learned positional embeddings similar to \citet{pmlr-v70-gehring17a}, by assigning each input token with a learned vector that encodes its absolute position, and observed similar results to the sinusoidal version.

\section{Translation evaluation} \label{bgexp}

We evaluate all translation experiments in this thesis using the BiLingual Evaluation Understudy metric, better known as BLEU \citep{Papineni2001}.
This metric assesses the closeness of the generated translation to a human reference translation.
It includes a \textit{brevity penalty} (BP) to avoid preferring shorter translations. 
The BLEU score for $n$-grams up to length $N$ is defined as:
\begin{align}
\bleu {_n} = \bp. \exp \bigg( \sum_{n=1}^N w_n \log p_n \bigg)
\end{align}

\noindent 
where $w_n$ is a weight assigned to the size of $n$-gram (often set uniformly to $1/N$). $p_n$ is computed as:
\begin{align} 
p_n &= \frac{ \sum\limits_{{c} \in \{candidates\}} \sum\limits_{{n\!\gram} \in {c}} \countop_{clip}({n\!\gram}) }{  \sum\limits_{{c'} \in \{candidates\}}^{} \sum\limits_{{n\!\gram }' \in {c'}} \countop({n\!\gram}')  }
\end{align}
\noindent
where:
\begin{align} 
\countop{_{clip}}(x) &= \min(\countop(x), \textit{max\_ref\_count})(x)) 
\end{align}
\noindent
Here, $candidates$ are translation candidates, and \textit{max\_ref\_count} is the largest count observed in the reference for that word. 
Scores are calculated over sentence pairs in the test set and the average BLEU is reported for the entire test set.  
Unless stated otherwise, in this thesis we compute case-sensitive BLEU up to and including $n$-grams of length 4. 

We also use other evaluation metrics, namely METEOR and Translation Error Rate (TER), in some chapters of this thesis.
METEOR is another metric to automatically evaluate translation quality \citep{banerjee-lavie-2005-meteor,denkowski-lavie-2011-meteor,denkowski:lavie:meteor-wmt:2014}.
Similar to BLEU, this metric compares the translation output with a reference translation, however, it addresses some of the deficiencies of the BLEU metric.
This is done by aligning the two sentences not only based on the exact match, but also on matching synonyms and paraphrases. 
METEOR has to be fine-tuned to achieve maximum correlation with human judgments \citep{agarwal-lavie-2008-meteor}.
TER is an easy-to-explain metric to compare translation output and manually created reference translation \citep{Snover06astudy}. 
It measures the number of edits required to change a translation output into one of the references.
A higher score of TER is a sign of more post-editing effort and it may not always correlate with translation quality.

While all these metrics attempt to measure translation quality, they assume inexact models of permissible variations in translation and may not capture the precise quality of a system \citep{callison-burch-etal-2006-evaluating}.
However, they allow for systematic evaluation of incremental changes to a single system and are very inexpensive to perform. 
We specifically choose BLEU because it is the most common metric and it makes it possible to compare various systems.
In Chapters~\ref{chapter:research-04} and~\ref{chapter:research-05} of this thesis, we explore cases where the translation quality of NMT models are affected, but individual automatic metrics do not reflect this change in quality.

% !TEX root = thesis-main.tex

\chapter{Topic-Sensitive Word Representations}
\label{chapter:research-01}

\section{Introduction and research questions}

Word representations in the form of dense vectors, or word embeddings, capture semantic and syntactic information \citep{mikolov2013efficient,pennington2014glove} and are widely used in many NLP tasks such as sentiment analysis \citep{tang2014learning,yu-etal-2017-refining}, identifying multiword expressions \citep{salehi-etal-2015-word,gharbieh2016word}, and translation \citep{zou2013bilingual, artetxe-etal-2018-unsupervised}.
These representation models are based on the assumption that the meaning of a word can be inferred from its textual context \citep{firth1957synopsis}. 

Currently, there are two categories of approaches to learning word representations (discussed in Section~\ref{bgemb}): \textit{static embeddings} where a fixed vector is learned for each word in the vocabulary, and \textit{dynamic embeddings} where vectors are dynamically calculated for each sentence. 
Dynamic embeddings such as ELMo \citep{peters-etal-2018-deep} and BERT \citep{devlin-etal-2019-bert} store the learned weights of the network, and use that to get word representations by computing them on the fly for a given context.
As a result, these models can capture context-dependent characteristics of the language such as polysemy: in natural language, words usually have more than one meaning (or sense).

%While these models perform extremely well on several NLP tasks, they are computationally expensive to acquire and use. 
%Static embeddings, on the other hand, are cost-effective and instantly accessible for nearly all downstream tasks.
Like dynamic embeddings, the research presented in this chapter aims at overcoming the inability of static embeddings to capture polysemy.
However, it predates dynamic embeddings.
Before the advent of contextualized embedding approaches, most static representation models learned \textit{one} fixed-length representation per word. 
However, this approach and how it is evaluated has some shortcomings.

Firstly, many tasks can benefit from using multiple representations per word to capture polysemy \citep{Bengio:2003,reisinger-mooney:2010:NAACLHLT}.
Many intricate distinctions of word senses are lost when we use one embedding vector to capture multiple meanings.
Additionally, this simplification of natural language unintentionally leads to more simplistic evaluation tasks.  
Most studies on static word embeddings used word similarity task to assess the accuracy of the static word representations where word pairs are ranked based on how similar or related they are.
However, most of the word similarity benchmarks present words out of context. The word pair `\textit{bank}' and `\textit{reef}' can have very different similarity scores depending on the context of the word \textit{bank}.
Finally, there is no clear and quantifiable definition of \textit{similarity} and \textit{relatedness} when comparing two words, and as a result, different benchmarks have different interpretations \citep{faruqui2016problems}. 

In this chapter, we seek to address these shortcomings and propose an approach for learning multiple static word representations per word.
We aim to understand the role of a particular kind of context, namely document topics, for learning these representations.
We analyze to what extent learning multiple topic-sensitive embeddings per word captures polysemy, which we believe is a necessary step towards further understanding the impact of context. 
Concretely, we ask:

\paragraph{Research Question 1:} \acl{rq:topic} 

\medskip

 \noindent We first look at the importance of document-level context and how it can help to separate different meanings of the word.
We study the integration of this topical information in learning word representation and evaluate the embeddings on a contextual task. 
Concretely, we ask: 
 
\begin{enumerate}[label=\textbf{RQ1.\arabic* },wide = 0pt, leftmargin=2em]
\setlength\itemsep{1em}
 \setcounter{enumi}{0}
\item \acl{rq:topic1}

\medskip

\noindent We introduce a model that uses a nonparametric Bayesian model, namely Hierarchical Dirichlet Process (HDP), to learn multiple topic-sensitive representations per word. 
\citet{yao2011nonparametric} showed that HDP is effective in learning topics yielding state-of-the-art performance for sense induction. This approach learns the granularity of senses from the data and does not require heuristic parameter setting. 
The authors assumed that topics and senses are entirely interchangeable, and so they trained individual models per word.
However, this assumption makes it difficult to scale to large data. 
In our approach, we do not hold the same assumption, which enables us to use HDP to model topics effectively using large unannotated training data. 
We aim to \textit{approximate} the word senses with topics and further use this additional signal for training the embeddings of each topic-word pair separately. 

\item \acl{rq:topic2}

\medskip

\noindent We propose three unsupervised, language-independent approaches to approximate senses with topics and learn multiple topic-sensitive embeddings per word.
Our first model uses a hard topic labeling approach to learn representations. The second model jointly learns topic-labeled and generic representations for each word in order to share statistical information between different meanings of a particular word. The third model uses topic distributions for each word following the notion that meanings of words are not mutually exclusive in a given context.
We show that in the lexical substitution ranking task \citep{mccarthy2007semeval} 
our models outperform two competitive baselines and perform comparably to the best-performing methods despite the fact that---unlike those methods---our approach does not use any syntactic information.

\item \acl{rq:topic3}

\medskip

\noindent The process of learning topics and topic-sensitive representations is applied to the same corpus ensuring compatibility between the granularity of topics and diversity of meanings of word embeddings. 
By learning the granularity of topics from the corpus we do not use any external knowledge sources.
As a result, this approach can be used for low-resource languages with no manually curated knowledge sources. 
Additionally, this approach broadens the contextual signals for learning more accurate representations when sentence-level context is not sufficient. 
We evaluate our representations on the contextual word similarity task and the lexical substitution task, both of which showcase the importance of learning multiple embeddings per word. 

\end{enumerate}

\paragraph{Organization.} This chapter is organized as follows: In Section~\ref{toprel}, we provide an overview of existing work on word sense disambiguation and static sense representations literature. 
Next, in Section~\ref{sect:model}, we introduce our representation models.
We present experimental details and study different tasks to evaluate the representations in Section~\ref{embevaluationsec}.
In Section~\ref{embindepth}, we present a more in-depth analysis of the resulting representations. 
Finally, we discuss the conclusions and implications of this work in Section~\ref{embconc}. 

\section{Related work}\label{toprel}

In this section, we discuss previous works that focus on learning multiple static word embeddings per word to capture polysemy, as well as related work in the area of word sense disambiguation. 

\subsection{Word sense disambiguation}
Word sense disambiguation is the problem of determining which \textit{sense} of a word is activated by the use of the word in a particular context \citep{ide-veronis-1998-introduction}. 
%To study different meanings of a word, we need to identify which sense of a word is used in a given context. 

There have been several attempts to build repositories for word senses \citep{miller1995wordnet,navigli2010babelnet}, but this is laborious and therefore limited to few languages. Moreover, defining a universal set of word senses is challenging as polysemous words can exist at many levels of granularity \citep{journals/lre/Kilgarriff97,Navigli2012}. 
For this reason, earlier work focuses on unsupervised sense induction, often following a Bayesian framework \citep{brody-lapata:2009:EACL,lau2014learning}. 

\citet{yao2011nonparametric} show that a nonparametric Bayesian model, such as the Hierarchical Dirichlet Process (HDP), is effective in learning topics and can yield state-of-the-art results in sense induction. 
The advantage of nonparametric methods is that they learn the granularity of topics from the data and do not require to fix the number of senses per word a priori.
By using HDP for sense induction, they assume that topics and senses are interchangeable and train a topic model for each target word with a sampled number of context instances.
%However, there are some concerns in using it for building sense repositories. 
However, inference with HDP does not scale to large corpus sizes due to the complexities of the model \citep{Jordan2011THEEO,6802355}.

\subsection{Static sense representations} 

We discussed in Section~\ref{bgembstatic} that the most commonly used approaches learn exactly one embedding per word \citep{mikolov2013efficient,pennington2014glove}.
However, even before the advent of dynamic embeddings, several studies have focused on learning multiple embeddings per word due to the ambiguous nature of language \citep{qiu-tu-yu:2016:EMNLP2016}.
\citet{huang2012improving} cluster word contexts and use the average embedding of each cluster as word sense embeddings, which yields improvements on a word similarity task. 
\citet{neelakantan2014efficient} propose two approaches, both based on clustering word contexts: In the first, they fix the number of senses manually, and in the second, they use an ad-hoc greedy procedure that allocates a new representation to a word if existing representations explain the context below a certain threshold.

\citet{li-jurafsky:2015:EMNLP} used a Chinese Restaurant Process (CRP) model to distinguish between senses of words and train vectors for senses, where the number of senses is not fixed. 
They change the Skipgram model \citep{mikolov2013efficient} to perform sense induction and sense embedding updates simultaneously. 
They use two heuristic approaches for assigning senses in a context: `greedy' which assigns the locally optimum sense label to each word, and `expectation' which computes the expected value for a word in a given context with probabilities for each possible sense. 
%and observe improvements in several NLP tasks.
%Like \citet{li-jurafsky:2015:EMNLP}, we employ a Bayesian nonparametric model, however, while they focus on learning senses for words individually, we learn the topic distributions and estimate senses from them. The HDP model has the advantage of sharing the statistical strength of the topic components across words and allows dependencies between documents to be modeled effectively.

\section{Topic-Sensitive representations} \label{sect:model}

In this section, we introduce our approach to learn topic-sensitive word representations based on the Skipgram model proposed by \citet{mikolov2013efficient}.
We previously mentioned that inference with HDP does not scale to large corpus sizes \citep{Jordan2011THEEO,6802355}.
Here, we describe our proposed models to learn topics from a corpus using HDP \citep{teh2006hierarchical,lau2014learning} in a way that is applicable to large corpora.

The main advantage of this model compared to non-hierarchical methods like the Chinese Restaurant Process (CRP) is that 
each document in the corpus is modeled using a mixture model with %mixture components 
topics shared between all documents \citep{citeulike:635668,brody-lapata:2009:EACL}. 
HDP yields two sets of distributions that we use in our methods: (i) distributions over topics for words in the vocabulary, and (ii) distributions over topics for documents in the corpus. 

  \begin{figure}    
  \centering 
  \includegraphics[scale=0.4]{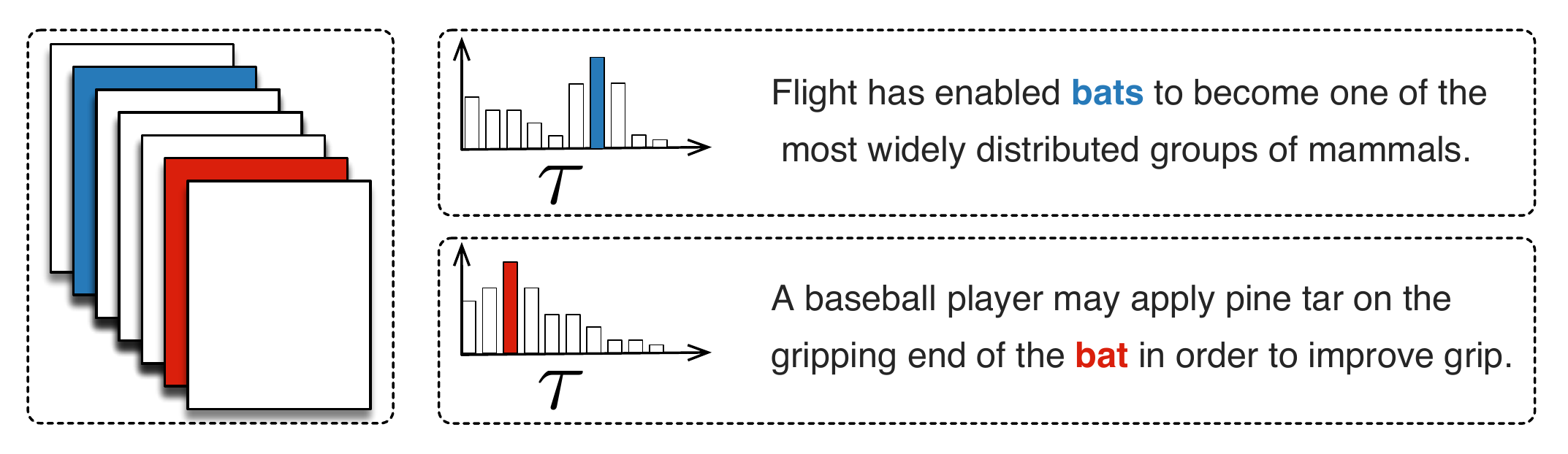} 
  \caption{An example of sentences including the word `bat' from two documents with different topic distribution.}
    \label{modelsfig0}
\end{figure}

Similarly to \citet{neelakantan2014efficient}, we use neighboring words %of a particular target word 
to detect the meaning of the context, however,
we also use the two HDP distributions. 
By doing so, we take advantage of the topic of the document beyond the scope of the neighboring words, which is helpful when the immediate context of the target word is not sufficiently informative. 
We modify the Skipgram model \citep{mikolov2013efficient} to obtain multiple topic-sensitive representations per word type using topic distributions.

Additionally, the context vectors of a word type with multiple topics are shared in our model.
This is especially beneficial for infrequent words, where a rare sense of the word can use the contextual information of other senses of the word.
%In addition, In our models the contexts of the identical target words with different topics are shared which 
%reduces the sparsity problem and provides a better representation estimation for rare words.
%In addition, \citet{neelakantan2014efficient} cluster context windows
%unlike \citet{neelakantan2014efficient}, we train for every sense of a word individually. and do not cluster the context windows
%This %In our models the contexts of the identical target words with different topics are shared which 
%reduces the sparsity problem and provides a better representation estimation for rare words.
We assume that meanings of words can be determined by their contextual information and use the distribution over topics to differentiate between occurrences of a word in different contexts, i.e., documents with different topics (see example in Figure~\ref{modelsfig0}).
We propose three different approaches illustrated in Figures~\ref{modelsfigH} and \ref{modelsfigS}:
two methods with hard topic labeling of words and one with soft labeling.
In the following sections, we discuss each of these model variants in detail.
  
\subsection{Hard topic-labeled representations}
\label{sect:hardReps}

\begin{figure*}[ht!]
\centering
\begin{subfigure}[t]{0.285\textwidth}
\includegraphics[width=\linewidth]{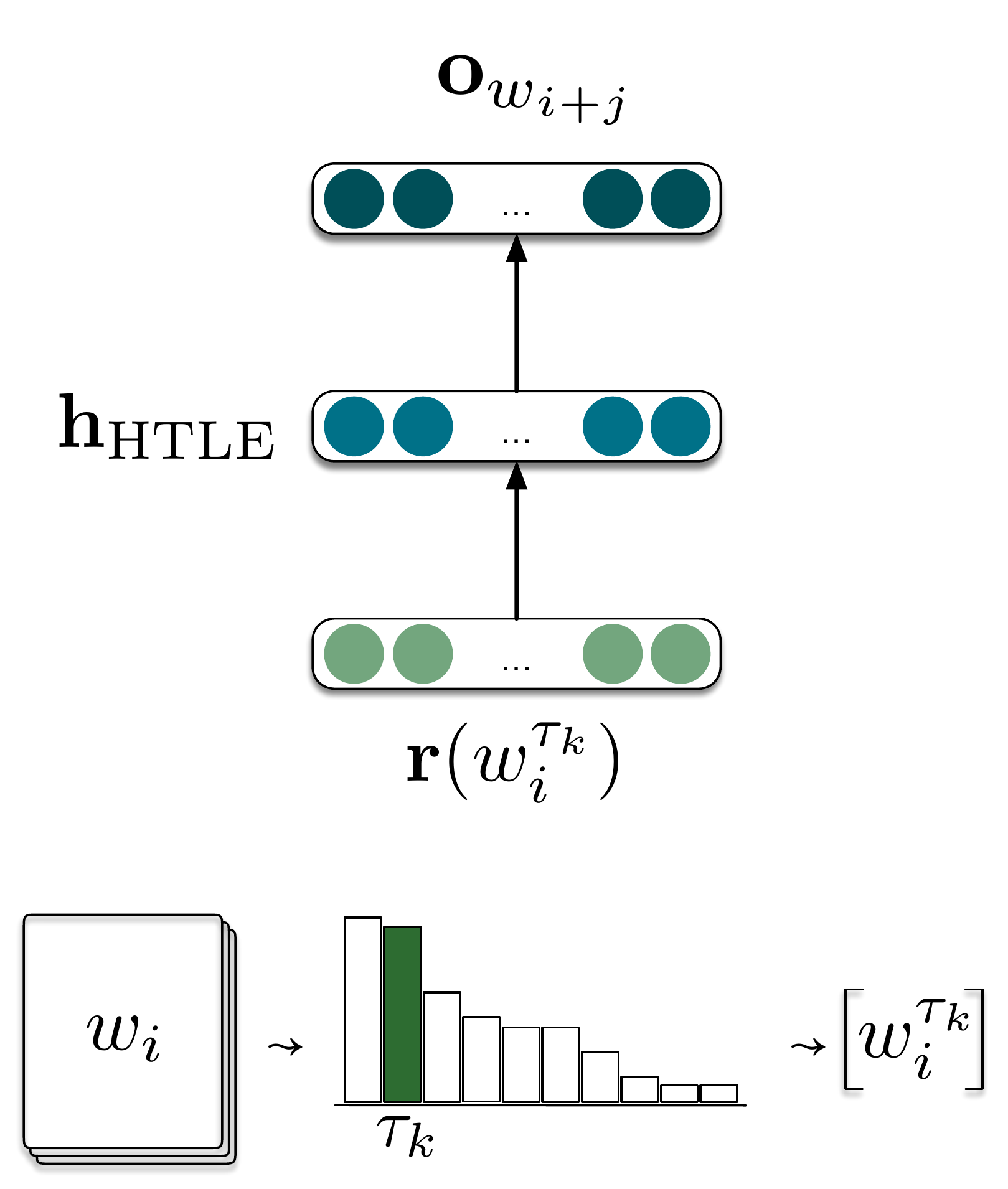}
\caption{\textbf{HTLE}: Hard-topic labeled representations model}
\end{subfigure}
\hspace{0.15\textwidth}
\begin{subfigure}[t]{0.315\textwidth}
\includegraphics[width=\linewidth]{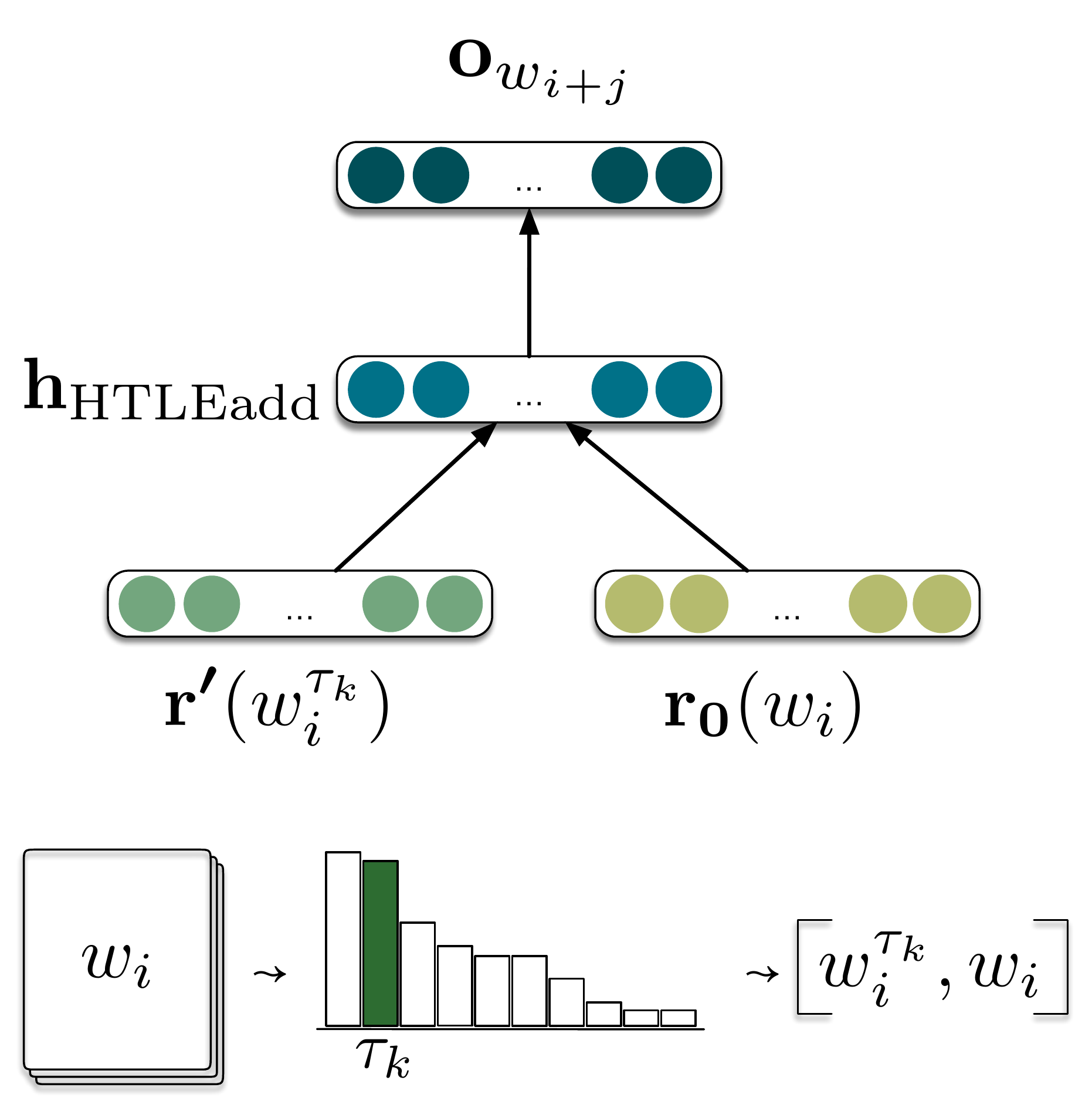}
\caption{\textbf{HTLEadd}: Hard topic-labeled representations plus generic word representations model}
\end{subfigure}
\caption{Illustrations of two proposed models with hard-labeling topics in this chapter. \label{modelsfigH}}
\end{figure*}

In the hard-labeling approach, we assign exactly one topic to each word based on sampling from the topic distribution.
We use the trained HDP model to label every word in the training data with the chosen topic ID.

Our first model variant (Figure~\ref{modelsfigH} (a)) considers each word-topic pair as a separate vocabulary entry.
To reduce sparsity on the context side and share the word-level information between similar contexts, we use topic-sensitive representations for target words (input to the Skipgram network) and standard, i.e., unlabeled, word representations for context words (output to the Skipgram network). Note that this results in different input and output vocabularies.
The training objective is then to maximize the log-likelihood of context words $w_{i+j}$ given the target word-topic pair $w_{i}^{\tau}$:
\begin{align}\small
\mathcal{L}_{HardT\textrm{-}SG} = \frac{1}{I} \sum_{i=1}^I \sum_{\substack{-c \le j \le c \\ j \ne 0}}{\log p(w_{i+j} \mid w^{\tau}_{i})}
\end{align}
\noindent%
where $I$ is the number of %running 
words in the training corpus, $c$ is the context %window 
size and $\tau$ is the topic assigned to $w_i$ by HDP sampling.
$\vt{o}_{w_{i+j}}$ is the context (i.e., output) representation for the word $w_i$.
Note that $w_i$ is an occurrence of word $w$ in context $[i-c, i+c]$.

The embedding of a word in context $\vt{h}({w_i})$ is obtained by simply extracting the row of the input lookup table (\vt{r}) corresponding to the HDP-labeled word-topic pair:
\begin{equation}\label{eq:HTLE}
\vt{h}_\textrm{HTLE}({w_i}) = \vt{r}(w_i^{\tau})
\end{equation}

A possible shortcoming of the HTLE model is that the representations are trained separately and information is not shared between different topic-sensitive representations of the same word. 
To address this issue, we introduce a model variant that learns multiple topic-sensitive word representations and generic word representations simultaneously (Figure~\ref{modelsfigH} (b)). 
In this variant (HTLEadd), the target word embedding is obtained by adding the word-topic pair representation ($\vt{r'}$) to the generic representation of the corresponding word ($\vt{r_0}$):
\begin{equation}\label{eq:HTLEadd}
\vt{h}_\textrm{HTLEadd}({w_i}) = \vt{r'}(w_i^\tau) + \vt{r_0}(w_i)
\end{equation}

This representation captures both the generic and the contextual meaning of the word.

\begin{figure*}[hbt!]
\centering
\includegraphics[scale=0.2]{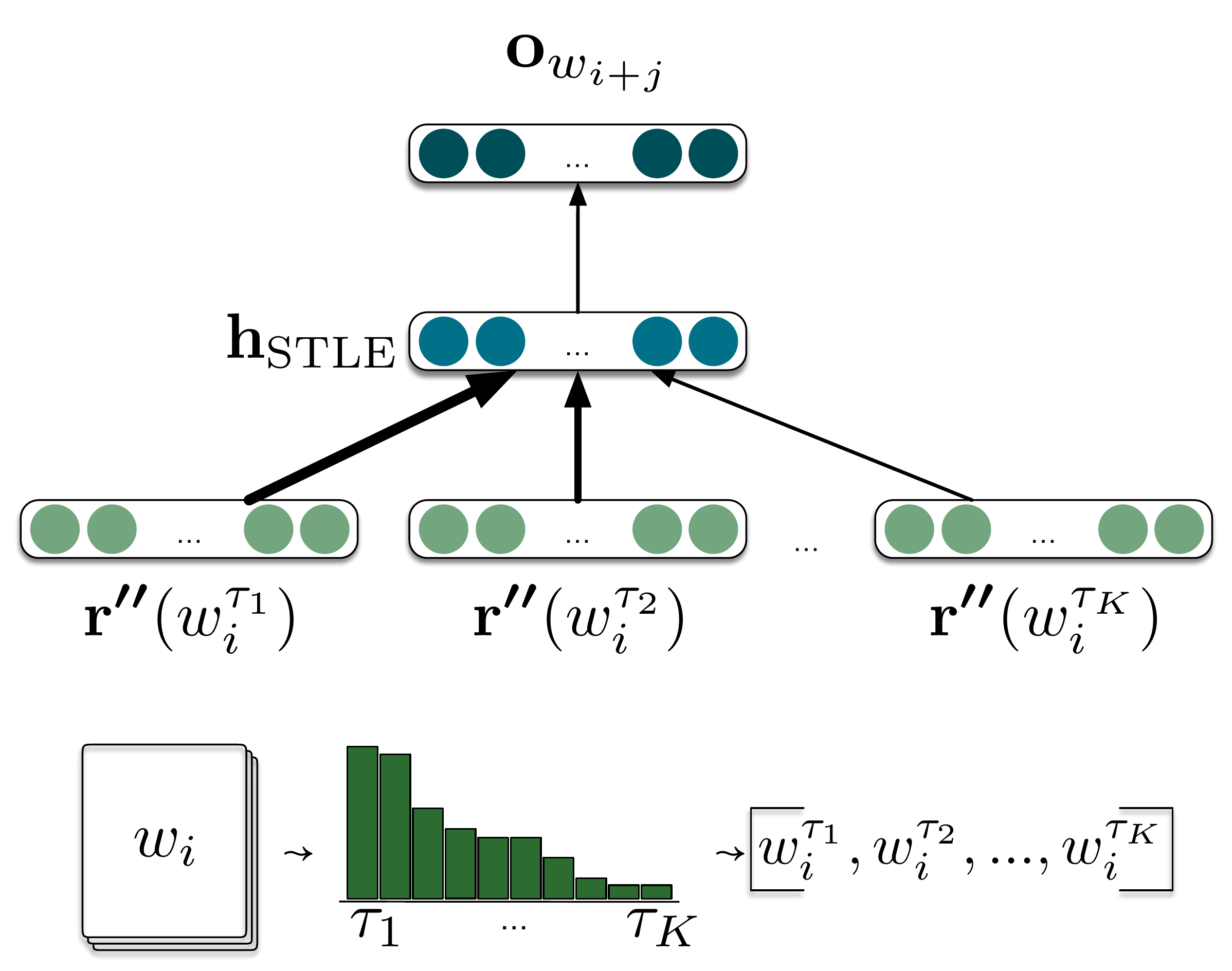}
\caption{Illustration of our soft topic-labeled representation model (STLE). \label{modelsfigS}}
\end{figure*}

\subsection{Soft topic-labeled representations}
\label{sect:softReps}

 \begin{table} \setlength{\tabcolsep}{0.2em}
\centering\small
\caption{Nearest neighbors of three examples in different representation spaces using cosine similarity. \textbf{word2vec} and \textbf{GloVe} are pre-trained embeddings from \protect\citep{mikolov2013efficient} and \protect\citep{pennington2014glove}, respectively. \textbf{SGE} is the Skipgram baseline and \textbf{HTLE} is our topic-sensitive Skipgram (cf. Equation~(\ref{eq:HTLE})), both trained on Wikipedia. $\tau_k$ stands for HDP-inferred topic $k$.\label{embs}}
\begin{tabular}{@{\extracolsep{4pt}} c|l|l|l|ll @{}}
\toprule
  \multicolumn{3}{c}{\bf Pre-trained}   & \multicolumn{3}{c}{\bf  Trained on Wikipedia}  \\ \cline{2-3} \cline{4-6} 
& \bf word2vec & \bf Glove & \bf SGE & \bf HTLE: $\tau_1$ & \bf HTLE: $\tau_2$ \\ \toprule
\multirow{10}{*}{\rotatebox[origin=c]{90}{\textbf{bat}}}  &bats&bats&uroderma&ball&vespertilionidae \\
&batting&batting&magnirostrum&pitchout&heran\\
&Pinch\_hitter\_Bray\textellipsis&Bat&sorenseni&batter &hipposideros\\
&batsman&catcher&miniopterus&toss-for&sorenseni\\
&batted&fielder&promops&umpire &luctus\\
&Hawaiian\_hoary&hitter&luctus&batting &coxi\\
&Lelands.com\textellipsis&outfield&micronycteris&bowes &kerivoula\\
&yelled\_Cheater&hitting&hipposideros&straightened &natterer\\
&wicketkeeper\_Andr\textellipsis&batted&chaerephon&fielder &nyctophilus\\
&lefthanded\_batter&catchers&pteronotus&flies &artibeus\\
%&Batting&balls&bats&ball-a&carollia\\
%&skipper\_Nicky\_Boje&slugger&gallagheri&yardley &pteronotus\\
%&batsmen&hitters&saccolaimus&hitting &lucifugus\\
%&baseman\_mitt&Bats&carollia&base &nasutus\\
%&uppercutting&baseball&ball-a&plate &macleayii\\
%&fielder&batsman&emballonura&batsmen &nosed \\
\midrule
\multirow{10}{*}{\rotatebox[origin=c]{90}{\textbf{jaguar}}}  & jaguars&jaguars&electramotive&ford &wiedii \\
&Macho\_B&xk8&vk66de&bmw &	puma \\
&panther&xj6&viper&chevrolet &	margay\\
&lynx&xjs&id66&honda &	tapirus\\
&rhino&panther&xj666&porsche &	jaguarundi\\
&lizard&xkr&roadster&multimatic&	yagouaroundi\\
&tapir&xj8&saleen&monza &	vison\\
&tiger&mercedes&siata&nissan &	concolor\\
&leopard&Jaguar&enetered&xj&	tajacu\\
& Florida\_panther	& porsche	& chevrolet	& dodge &		tayassu \\
\midrule
\multirow{10}{*}{\rotatebox[origin=c]{90}{\textbf{appeal}}}  &appeals&appeals&court&court&sfa  \\
&appealing&appealed&appeals&case&steadfast\\
&appealed&appealing&appealed&appeals&lackadaisical\\
&Appeal&Appeal&carmody&appealed&assertions\\
& rehearing&court&upheld&decision&lack\\
& apeal&decision&verdict&proceedings&symbolize\\
& Appealing&conviction&jaruvan&disapproves&fans\\
& ceasing\_hostilities\textellipsis &plea&affirmed&ruling&attempt\\
& ruling&sought&appealable&upholding&unenthusiastic\\
& Appeals&dismiss&battin&carmody&cancellation \\
\bottomrule                  
\end{tabular}
\end{table}

The model variants above rely on the hard labels resulting from HDP sampling.
As a soft alternative to this, we can directly %incorporate into our Skipgram model 
include the topic distributions estimated by HDP for each document, see Figure~\ref{modelsfigS}. 
Since the topics are not clearly separated, every identified topic of a word can contribute to the learning process proportional to its value.
Specifically, for each update, we use the topic distribution to compute a weighted sum over the word-topic representations ($\vt{r''}$):
\begin{equation}\label{eq:STLE}
\vt{h}_\textrm{STLE}(w_i) = \sum_{k=1}^{T}  \ p(\tau_k \mid d_i) \ \vt{r''}(w_i^{\tau_k})
\end{equation}
%\noindent% 
where $T$ is the total number of topics, $d_i$ the document containing $w_i$, and $p(\tau_k \mid d_i)$ the probability assigned to topic $\tau_k$ by HDP in document $d_i$.
The training objective for this model is:
\begin{align}\small
\mathcal{L}_{SoftT\textrm{-}SG} = \frac{1}{I} \sum_{i=1}^I \sum_{\substack{-c \le j \le c \\ j \ne 0}}{\log p(w_{i+j} \mid w_i, \tau)}
\end{align}

\noindent%
where $\tau$ is the topic of document $d_i$ learned by HDP. 
The STLE model has the advantage of directly applying the distribution over topics in the Skipgram model. 
Also, for each instance, we update all topic representations of a given word with non-zero probabilities, which has the potential to reduce the sparsity problem.

\subsection{Embeddings for polysemous words}

The representations obtained from our models are expected to capture the meaning of a word in different topics.
We now examine whether these representations can distinguish between different word senses.
Table~\ref{embs} provides examples of nearest neighbors. 
For comparison, we include our own baseline, i.e., embeddings learned with Skipgram on our corpus, as well as Word2Vec \citep{mikolov2013distributed} 
and GloVe embeddings \citep{pennington2014glove} 
pre-trained on large data.

\begin{table}
\centering
\small
\caption{Statistics of the degree of polysemy in Wordnet and HTLE. \label{wordnetinfo}} %out of 46255 overlap between wn and ours 
\begin{tabularx}{0.65\textwidth}{lcc}
\toprule
& \textbf{Wordnet} & \textbf{HTLE} \\
\midrule
 Degree of polysemy & 2.08 & 4.79 \\
  Single [sense/representation] words &  26,755 & 21,490 \\
\bottomrule
\end{tabularx}
\end{table}

In the first example, the word \textit{bat} has two different meanings: animal or sport device. We can see that the nearest neighbors of the baseline and pre-trained word representations either center around one primary, i.e., most frequent, meaning of the word, or looks like a mixture of different meanings. 
The topic-sensitive representations, on the other hand, correctly distinguish between the two different meanings. A similar pattern is observed for the word \textit{jaguar} and its two meanings: car or animal.
The last example, \textit{appeal}, illustrates a case where topic-sensitive embeddings are not clearly detecting different meanings of the word, despite having some correct words in the lists. %(\textit{unenthusiastic}, \textit{fans})
%there are also errors present. 
Here, the meaning \textit{attract} does not seem to be captured by any embedding set. 

These observations suggest that topic-sensitive representations capture different word senses to some extent.
To quantify the degree of polysemy in our embeddings, we compare it to Wordnet \citep{miller1995wordnet}. 
Wordnet is a manually curated lexical database of English that interlinks different senses of the words through conceptual-semantic and lexical relations.
As a result, words that are found near one another in the network are semantically disambiguated. 
Table~\ref{wordnetinfo} shows statistics for Wordnet and our proposed embedding method.
We observe that the \textit{degree of polysemy} of our embeddings is more than double that of Wordnet.
Note that while in our models we do not explicitly specify the desired number of senses per word type, the hyperparameters have an impact on it: $\gamma$ manages the variability of the global sense distribution and $\alpha$ manages the variability of each word type's selection of senses. These hyperparameters are discussed in Section~\ref{embexpsetup}.

Moreover, we observe that not every topic-sensitive word representation corresponds to a distinct and unique sense. 
In our experiments, we see that at times, multiple embeddings capture the same sense of the word.
However, they are also in close proximity in the embedding space and end up being very similar.
To provide a systematic validation of our approach, we now investigate whether topic-sensitive representations can improve tasks where polysemy is a known issue.

\section{Evaluation} \label{embevaluationsec}

In this section, we present the setup for our experiments and empirically evaluate our approach on the context-aware word similarity and lexical substitution tasks.

\subsection{Experimental setup} \label{embexpsetup}

%We use the English Wikipedia corpus containing 4.8M documents %and 1B tokens 
All word representations are learned on the English Wikipedia corpus containing 4.8M documents (approximately 1 billion tokens).
Preprocessing of the training data includes lowercasing, removing stop words, and removing words occurring less than 100 times.
The topics are learned on a 100K-document subset of this corpus using the HDP implementation of \citet{teh2006hierarchical}.
HDP has two hyperparameters, $\gamma$  and $\alpha$, which control the variability of the global topic distribution and each word's choice of topics, respectively. We do not tune these parameters and following the literature, %set these values to 
we put gamma priors $\gammaa(1, 1)$ and $\gammaa(1, 0.1)$ on hyperparameters $\gamma$ and $\alpha$ respectively. These parameters encourage skewed topic distributions which are typically observed in natural languages \citep{Gale1992}. 
Once the topics have been learned, we run HDP on the whole corpus to obtain the word-topic labeling (Section~\ref{sect:hardReps}) and the document-level topic distributions (Section~\ref{sect:softReps}).
We train each model variant with window size $c=10$ and different embedding sizes (100, 300, 600) with random initialization. 
All model variants in this chapter are trained on the same training data with the same settings, following suggestions by \citet{mikolov2013efficient} and \citet{levy2015improving}. 

\begin{table}[htb!]
\centering
\small
\caption{Word similarity benchmarks for intrinsic evaluation of word representations. \label{wstab}}
\begin{tabular}{lcl}%@{}}
\toprule
\textbf{Data set} & \textbf{Word pairs} & \textbf{Reference} \\
\midrule
RG & 65 & \citet{Rubenstein:1965:CCS:365628.365657} \\ %
%MC & 30 & \citet{miller1991contextual} \\ %
WS353 & 353 &  \citet{Finkelstein:2001:PSC:371920.372094} \\ %
%YP-130 & 130 &  \citet{Yang06verbsimilarity} \\  
MTurk287 & 287 &  \citet{Radinsky:2011:WTC:1963405.1963455} \\ %
%MTurk771 & 771 &  \citet{Halawi:2012:LLW:2339530.2339751} \\
MEN & 3000 &  \citet{bruni-etal-2012-distributional} \\ %
RW & 2034 &  \citet{luong-etal-2013-better} \\ %
%Verb & 144 &  \citet{baker-etal-2014-unsupervised} \\
SimLex & 999 &  \citet{hill-etal-2015-simlex} \\ %
\bottomrule
\end{tabular}
\end{table}

\subsection{Word similarity task}

The most popular intrinsic evaluation of static word representations is the word similarity task. 
In this task, a list of pairs of words with their similarity scores judged by human annotators is provided. 
The goal is to measure how well the word vector representations capture the notion of word similarity by ranking the word pairs according to their similarity scores. 
Table~\ref{wstab} provides a list of benchmarks with the number of word pairs in each data set that we use for evaluation. 

Similar to most previous approaches \citep{Radinsky:2011:WTC:1963405.1963455,hassan2011semantic,yih-qazvinian-2012-measuring}, we use Spearman's $\rho$ (rank correlation coefficient) to assess the monotonic relationship between the model's ranking of word pairs and the gold standard's ranking. 

Our models learn multiple embeddings per word, but these benchmarks do not include any context to help distinguish between the vectors.
Therefore we employ several techniques for selecting and combining the representation vectors:
\begin{itemize}
\itemsep0em 
\item \textbf{Max}: computes the pairwise similarity between the nearest topic-sensitive embeddings of the word pair.
\item \textbf{Mean}: computes the pairwise similarity between the means of all topic-sensitive embeddings of each word. 
\item \textbf{wMean}: computes the pairwise similarity between the weighted means of all topic-sensitive embeddings of each word. The weights are defined according to the frequency of each topic.
\end{itemize}

Table~\ref{wstabres} provides the results for the word similarity experiments. 
We observe slight improvements in different settings, but there is no clear indication that one model performs best across all data sets. 
It is also clear that each data set has a different level of difficulty, and because of the differences in quality of the word pairs and the definition of \textit{similarity} for the annotators, they are not analogous. 

\begin{table}[tbh!]  \setlength{\tabcolsep}{0.4em} 
\centering
\small
\caption{Spearman's rank correlation performance on word similarity tasks. All vectors are 100-dimensional. \label{wstabres}}
\begin{tabular} {llccccccc} 
\toprule
\multicolumn{1}{c} {} &
  \multicolumn{1}{c} {} &
  \multicolumn{1}{c} {\textbf{RG}} &
  \multicolumn{1}{c} {\textbf{WS353-rel}} &
  \multicolumn{1}{c} {\textbf{WS353-sim}} &
  \multicolumn{1}{c} {\textbf{MTurk287}} &
  \multicolumn{1}{c} {\textbf{MEN}} &
  \multicolumn{1}{c} {\textbf{RW}} &
  \multicolumn{1}{c} {\textbf{SimLex}} \\
\midrule
  \multicolumn{2}{l}{SGE} & 0.77  & 0.44 & 0.69 & 0.66  & \bf 0.71 & \bf 0.38 & 0.29 \\ \midrule
\multirow{3}{*}{\rotatebox[origin=c]{90}{HTLE}} & Max & 0.68 & 0.20 & 0.46 & 0.42 & 0.51 & 0.15 & 0.20 \\
 & Mean & 0.65 & 0.29 & 0.61 & 0.62 & 0.57 & 0.35 & 0.22 \\
 & wMean & 0.48 & 0.32 & 0.40 & 0.55 & 0.50 & 0.08 & 0.10 \\ \midrule
\multirow{3}{*}{\rotatebox[origin=c]{90}{HTLEadd}}  & Max &\bf  0.81 &  0.30 & 0.57 & 0.56  & 0.63 & 0.24 & 0.22  \\
 & Mean & 0.77 & 0.42 & 0.67 & \bf 0.68 & 0.69 & 0.36 & 0.28 \\
 & wMean & 0.60 & 0.36 & 0.48 & 0.63 & 0.57 & 0.13 & 0.14 \\ \midrule
\multirow{3}{*}{\rotatebox[origin=c]{90}{STLE}}  &Max & 0.74  & 0.43 & 0.69 &0.67 & 0.69 & 0.20 & 0.30 \\
 & Mean & 0.71 & \bf 0.45 & \bf 0.69 & 0.67 &  0.67 & 0.23 & \bf 0.30 \\
 & wMean & 0.65 & 0.43 & 0.64 & 0.65 & 0.68 & 0.16 & 0.24 \\
\bottomrule
\end{tabular}
\end{table}

One of the main concerns of using these benchmarks, in general, is that the notion of word similarity is subjective and there is no clear division between similarity and relatedness \citep{faruqui2016problems,torabi-asr-etal-2018-querying}.
As a result, some data sets penalize representation models that consider two related words as `not similar', while others do not.
For instance, in MEN \citep{bruni-etal-2012-distributional}, the guidelines did not distinguish between similarity and relatedness and gave examples of both similarity (e.g., \textit{"car-automobile"}), and relatedness (e.g., \textit{"wheels-car"}) as valid options to the annotators. 
The instructions for the SimLex data set \citep{hill-etal-2015-simlex}, however, included guidelines for the annotators with examples of related pairs (e.g., \textit{"car-tyre"}) that are \textit{not} to be labeled similar.

\citet{faruqui2016problems} evaluated several issues of the word similarity task.
These include low correlation with extrinsic evaluation, no consideration of polysemy, absence of statistical significance, and semantic versus task-specific embeddings. 
To specifically address the lack of context to identify polysemous words, \citet{huang2012improving} proposed the Stanford contextual word similarity data set (SCWS). In the following subsection, we evaluate our embeddings using this data set.

\subsection{Context-Aware word similarity task}

As mentioned before, there are multiple test sets available for intrinsic evaluation of embeddings, but in almost all of them word pairs are considered out of context. 
To evaluate our static embeddings intrinsically, we use the SCWS data set \citep{huang2012improving}. 
To the best of our knowledge, this was the only word similarity data set considering word context at the time our models were developed.
Note that more recently, instead of intrinsic evaluations, the performance of dynamic contextual embeddings is typically evaluated on downstream NLP tasks \citep{peters-etal-2018-deep,devlin-etal-2019-bert}. 

The SCWS data set contains word pairs and their respective contexts with average human ratings indicating the similarity of the target words.
Table~\ref{iwdtexample} presents examples of word pairs and their contexts in SCWS. 
\begin{table}[tbh!]
\caption{Examples from SCWS data set. Each example includes the word pair (identical or non-identical), their corresponding contexts, and the average human score between 0 and 10 to indicate the similarity. \label{iwdtexample}} %Subscripts are the topic IDs assigned to the word in the given context.   
\centering
\small
\renewcommand{\tabcolsep}{3pt}
\begin{tabularx}{\textwidth}{lX}
\toprule
\textbf{Word pair} & \textit{bitter,  bitter}  \\
  \hdashline
$\text{context}_1$ & It has an aromatic, warm and slightly \underline{bitter} taste. \\
$\text{context}_2$ & AK - a very common beer name in the 1800s - was often referred to as a "mild \underline{bitter} beer" interpreting "mild" as "unaged". \\
  \hdashline
Human score & 6.0 \\
\midrule
\textbf{Word pair} & \textit{ bitter,   resentful}       \\
  \hdashline
$\text{context}_1$ &  Named for the tattoos they decorated themselves with and \underline{bitter} enemies of encroaching Roman legions, the Picts fired Howard's imagination and crystallized in him a love for barbarians and outsiders from civilization who lived lives of great hardship and struggle but also great freedom and verve. \\
$\text{context}_2$ &   Legge-Bourke had been hired by Prince Charles as a young companion for his sons while they were in his care, and Diana was extremely \underline{resentful} of Legge-Bourke and her relationship with the young princes.\\
  \hdashline
Human score & 9.0 \\
\midrule
\textbf{Word pair} & \textit{bitter,     taste}  \\
  \hdashline
$\text{context}_1$ & This practice began during the Prohibition as a means of covering the \underline{bitter} taste. \\
$\text{context}_2$ & Once it has decayed, it leaves no \underline{taste} or odor in drinking water. \\
  \hdashline
 Human score  & 7.0 \\
  %\midrule
%\textbf{Word pair} & \textit{bitter, ale}  \\
  %\hdashline
%$\text{context}_1$ &  The designation of such beers as "\underline{bitter}" or "mild" has tended to change with fashion. \\
%$\text{context}_2$ &  When the beer has fermented, it is packaged either into casks for cask \underline{ale} or kegs, aluminium cans, or bottles for other sorts of beer. \\
  %\hdashline
% Human score & 8.0 \\
 \bottomrule
\end{tabularx}
\end{table}
To evaluate our models on SCWS, we run HDP on the data treating each word's context as a separate document.
We compute the similarity of each word pair as follows:
\begin{align}\small
\text{Sim}(w_1, w_2) &= \cos(\vt{h}(w_1), \vt{h}(w_2)) 
\end{align}
where $\vt{h}(w_i)$ refers to any of the topic-sensitive representations defined in Section~\ref{sect:model}. 
Note that $w_1$ and $w_2$ can refer to the same word.

%%%%%%%%%%
We compare our models to various baselines: The Skipgram model (SGE), the context-aware Skipgram model (SGE + context), and the best-performing multi-sense embeddings model per word type (MSSG) \citep{neelakantan2014efficient}. 
The context-aware Skipgram baseline (SGE + context) computes the average pairwise cosine similarity between a target word in each context with every word in the opposing context.  

For MSSG we use the best performing similarity measure (avgSimC) as proposed by \citet{neelakantan2014efficient}:
\begin{align}\small \label{mssgavgsim}
\text{avgSimC}(w_1, w_2)  = \sum_{j=1}^{K} \sum_{i=1}^{K} P(w_1, c_1, i)P(w_2, c_2, j) d(\vt{v}(w_1, i), \vt{v}(w_2, j))
\end{align}

\noindent where $P(w, c, k)$ is the probability that $w$ takes the $k$-th sense given context $c$.
$\vt{v}(w, k)$ is the embedding for word $w$ with assigned sense $k$.
$d(\vt{v}(w_1, i'), \vt{v}(w_2, j'))$ is the similarity measure between the given embeddings $\vt{v}(w_1, i')$ and $\vt{v}(w_2, j')$.
avgSimC measures the similarity between each pair of senses by how well each sense fits the context at hand.
%%%%%%%%%%

Table~\ref{scws} provides the Spearman's correlation scores for different models against the human ranking. 
We see that with dimensions 100 and 300, two of our proposed models obtain slight improvements over the baseline. 
However, for higher dimensions (embedding size 600), the MSSG model is the best performing system. 

\begin{table}[tbh!]
\centering
\small
\caption{Spearman's rank correlation performance for the Word Similarity task on SCWS \citep{huang2012improving}.\label{scws}}
\begin{tabular}{lccc}%@{}}
\toprule
                            & \multicolumn{3}{c}{\textbf{Dimension}} \\ \cmidrule(r){2-4}
      \textbf{Model}        & 100         & 300         & 600         \\ \midrule
SGE + context  \citep{mikolov2013efficient}  &     0.59        &    0.59         &      0.62       \\
MSSG \citep{neelakantan2014efficient} & 0.60 & \bf 0.61  & \textbf{0.64} \\\cmidrule(r){1-4}
HTLE          &     \textbf{0.63}        &       0.56      &       0.55      \\
HTLEadd              &         0.61    &      \textbf{0.61}       &      0.58       \\
STLE                      &     0.59        &      0.58       &       0.55      \\ \bottomrule
\end{tabular}
\end{table}

The main advantage of having multiple embeddings per word for different meanings is in comparing pairs of identical word with ambiguous meanings. 
With multiple representations per word, we can have a better estimation of similarities for identical words, given that we detect different senses correctly. 
Spearman's rank correlation between the two systems is based on the average differences between the two ranks of each observation. 
To further understand the performance of our models, we look into the rank differences for the two types of word pairs in SCWS (identical and non-identical) and the two types of embeddings (assigning the same topic or not).
The former explicitly evaluates the performance of our models on identical words, and the latter evaluates the impact of the topic-labeling step.
This results in four categories for comparison.
%: identical words and non-identical words with the same topic assignment or not.

   \begin{figure}[htb!]
\centering
\includegraphics[scale=0.4]{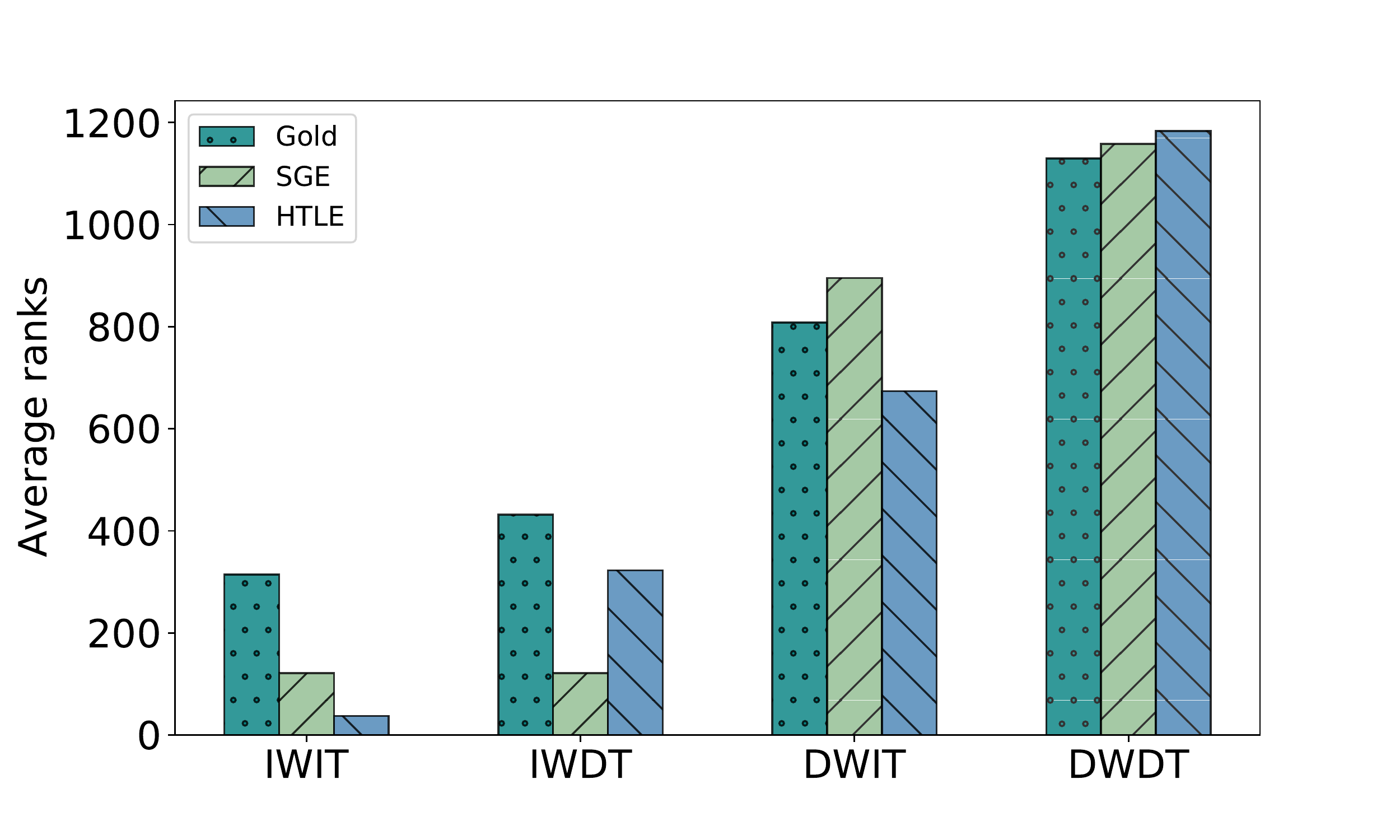} 
  \captionof{figure}{Average absolute rank of the baseline embeddings and the HTLE embeddings in four categories, where being closer to the gold rankings is better. The categories are marked with a combination of these labels:
  %Average ranks of different embedding spaces in four categories of word and topic. 
  \textbf{I}: identical, \textbf{D}: different. \textbf{W}: word, \textbf{T}: topic. %Lower values are better.
  For instance \texttt{IWDT} is the category of word pairs where \textit{\textbf{i}dentical \textbf{w}ords have \textbf{d}ifferent \textbf{t}opics}. Examples of the categories are presented in Table~\ref{iwdtexample}.}
  \label{scwsbar}
   \end{figure}
   
Since the difficulty of each category is different, we expect different performances from the models.
Figure~\ref{scwsbar} shows the average absolute rank of the baseline embeddings and the HTLE embeddings with 300 dimensions in these four categories. 
The gold rank, which is the ranking of all word pairs in the data set according to human judgments, is also shown for each category.
The best-performing model is the one that is the closest to the gold ranking.
Note that the gold rank naturally increases when words are different in comparison to when they are the same. 
%This displays the distribution of the test data where the number of identical words that are also semantically similar is higher. 
   
One can see that for identical words, labeled with different topics, \texttt{IWDT}, the rank assigned by topic-sensitive embeddings is much closer to the gold ranking than the one produced by the baseline. 
The average rank is also still higher when considering both categories of identical words: \texttt{IWDT} and \texttt{IWIT}.
This indicates that the estimation of the similarity scores of identical words is notably more accurate in our model. 
However, for non-identical word pairs (\texttt{DWIT} and \texttt{DWDT}), the rank difference is higher for topic-sensitive embeddings and since the evaluation set consists of mostly non-identical word pairs, the correlation with gold ranking decreases in total in comparison with baseline word embeddings. 

\subsection{Lexical substitution task}

Continuing our evaluation of word representations, in this section, we explore the lexical substitution task. 
This task requires one to identify the best replacements for a word in a sentential context. The replacements should be both semantically compatible with the word, and syntactically correct in the context. For example:

\vspace{1mm}

\begin{center}\small
\setlength{\tabcolsep}{20pt}
\begin{tabular}{l l}
\textbf{sentence} & \textbf{substitutions} \\ \hline
The sun was \underline{bright}. & \textit{luminous, colorful}  \\
He was \underline{bright} and independent.  & \textit{intelligent, clever, smart} \\
\end{tabular}
\end{center}

\vspace{1mm}

The presence of many polysemous target words makes this task more suitable for evaluating sense embeddings. %Instead of using an open vocabulary, 
Following \citet{melamud2015simple}, we pool substitution candidates 
from different instances and rank them by the number of annotators that selected them for a given context. 
We use two evaluation sets: LS-SE07 \citep{mccarthy2007semeval}, and LS-CIC \citep{kremer2014substitutes}.
The Concept-in-Context (LS-CIC) set for the lexical substitution task is a large-scale corpus constructed by crowdsourcing \citep{kremer2014substitutes} and contains a more extensive set of words. 
The main difference between LS-CIC and LS-SE07 is that the former was constructed as a large-scale "all-words" corpus, while LS-SE07 mostly includes ambiguous words. 

Unlike previous work \citep{DBLP:conf/emnlp/SzarvasBH13,kremer2014substitutes,melamud2015simple}, we do not use any syntactic information in our models, % during training or testing, 
motivated by the fact that high-quality parsers are not available for most languages.
% The main reason is that for most languages parsers are not available, hence we only focus on discovering semantic relations. 
The evaluation is performed by computing the Generalized Average Precision (GAP) score \citep{kishida2005property}. %Below we discuss the evaluation pipeline and analyze the performance of the models. 
Given a gold standard of size $R$ of ranked candidates, the GAP score is defined as:
\begin{equation}
\small
\textrm{GAP} = \frac{\sum_{i=1}^n I(x_i) p_i}{R'}    \quad \quad \quad         R' = \sum_{i=1}^R I(y_i)\overline{y}_i
 \end{equation}
 
\noindent where $x_i$ is a binary variable symbolizing whether the $i$-th candidate as ranked by the model is in the gold standard and $n$ is the number of candidates to be ranked. 
$I(x_i)$ is one if $x_i$ is larger than zero, and otherwise, it is zero. 
$\overline{y}_i$ is the average weight of the ideal ranked list in the gold standard.

In order to rank substitution candidates, we compute the similarity between the target word and each candidate similar to \citet{melamud2015simple} but adapt it to include topic distributions as well as context words for word embeddings. 
%As for the word similarity task, 
We run HDP on the evaluation set % differently from the SCWS task, we cannot have topic information 
and compute the similarity between target word $w_t$ and each substitution $w_s$ using two different inference methods in line with how we incorporate topics during training.
We define the first method, Sampled \textit{(Smp)}, as:
\begin{equation}
\small
\cos(\vt{h}(w_s^\tau), \vt{h}(w_t^{\tau'})) +  \frac{\sum_{c}  \cos(\vt{h}(w_s^{\tau}), \vt{o}(w_c)) }{C},  
 \end{equation}
 
\noindent 
and define the second method, Expected \textit{(Exp)}, as:
\begin{equation}
\small
\begin{split}
\sum_{\tau, \tau'}  \ p(\tau) \ p(\tau') \cos(\vt{h}(w_s^\tau), \vt{h}(w_t^{\tau'})) +  \frac{\sum_{\tau,c}  \cos(\vt{h}(w_s^{\tau}), \vt{o}(w_c)) \ p(\tau) }{C},  
\end{split}
\end{equation}

\noindent where $p(\tau)$ and $p(\tau')$ are the topic probabilities, $\vt{h}(w_s^\tau)$ and $\vt{h}(w_t^{\tau'})$ are the representations for substitution word $s$ with topic $\tau$ and target word $t$ with topic $\tau'$ respectively (see Section~\ref{sect:model}),
$w_c$ are context words of $w_t$ taken from a sliding window of the same size as the embeddings,
$\vt{o}(w_c)$ is the context (i.e., output) representation of $w_c$, and $C$ is the total number of context words.
Note that these two methods are consistent with how we train HTLE and STLE.
\begin{table}[bht!]
\centering
\small
\setlength{\tabcolsep}{8pt}
\caption{\label{embs_tables00}GAP scores on LS-SE07 and LS-CIC sets. For \textsc{SGE + context} we use the \textit{context} embeddings to disambiguate the substitutions. Improvements over the best baseline (MSSG) are marked $^\blacktriangle$ at $p<.01$ and $^\vartriangle$ at $p < .05$.} 
\begin{tabular}{@{\extracolsep{4pt}}lcllllll@{}}
\toprule
                           && \multicolumn{3}{c}{\textbf{LS-SE07}}  & \multicolumn{3}{c}{\textbf{LS-CIC}} \\ 
                           \midrule
                                                      && \multicolumn{3}{c}{Dimension} &  \multicolumn{3}{c}{Dimension}   \\ 
                                                      \cmidrule{3-5}  \cmidrule{6-8}
              \textbf{Model}      & \textbf{Infer.}       & 100         & 300         & 600       & 100         & 300         & 600     \\ \midrule
SGE  &   \multirow{3}{*}{n/a}  & 36.2          &      40.5       &      41.1   &      30.4     &       32.1      &       32.3       \\
SGE + context  &  &      36.6       &       40.9      &     41.6   &  32.8      &    36.1         &      36.8         \\
MSSG & & 37.8 & 41.1 & 42.9 & 33.9 & 37.8 &  39.1 \\
\midrule
HTLE    & \multirow{3}{*}{Smp} & 39.8$^\blacktriangle$   & 42.5$^\blacktriangle$  &   43.0$^\blacktriangle$  & 32.1 & 32.7 &  33.0   \\
HTLEadd  &   & 39.4$^\vartriangle$   & 41.3$^\blacktriangle$  &   41.8  & 30.4 & 31.5 &    31.7   \\
STLE     &   & 35.2  & 36.7 & 39.0    & 32.9 & 32.3 &  33.9  \\ 
\midrule
HTLE      &  \multirow{3}{*}{Exp}   &     \textbf{40.3}$^\blacktriangle$       &      \textbf{42.8}$^\blacktriangle$       &      \textbf{43.4}$^\blacktriangle$     &     36.6$^\blacktriangle$       &      \textbf{40.9}$^\blacktriangle$        &      \textbf{41.3}$^\blacktriangle$      \\
HTLEadd  &                &       39.9$^\blacktriangle$     &       41.8$^\blacktriangle$       &      42.2       &        35.5$^\vartriangle$     &  37.9$^\vartriangle$          &      38.6      \\
STLE           &     &       38.7$^\vartriangle$      &     41.0         &      41.0       &      \textbf{36.8}$^\blacktriangle$      &  36.8   &     37.1      \\ \bottomrule
\end{tabular}
\end{table}
The \textit{Smp} method, similar to HTLE, uses the HDP model to assign topics to word occurrences during testing. The \textit{Exp} method, similar to STLE, uses the HDP model to learn the probability distribution of topics of the context sentence and uses the entire distribution to compute the similarity. 
Both of these inference methods can be used with either model.

\medskip

For the context-aware Skipgram baseline (SGE + context), we compute the similarity as follows:
\begin{align}
\similarity(w_s, w_t) = \cos(\vt{h}(w_s), \vt{h}(w_t)) +\frac{\sum_{c} \cos(\vt{h}(w_s), \vt{o}(w_c))}{C} 
\end{align}
This computation uses the similarity between the candidate word and all words in the context, as well as the similarity between target and candidate words.
We also report results for the baseline Skipgram model (SGE) without using the provided context as well as the MSSG model. 
The MSSG baseline uses the best performing similarity measure (Equation~\ref{mssgavgsim}) as proposed by \citet{neelakantan2014efficient} for a context-aware comparison.

Table~\ref{embs_tables00} shows the GAP scores of our models and the baselines. We use the nonparametric rank-based Mann-Whitney-Wilcoxon test 
\citep{sprent2016applied} to check for statistically significant differences between runs. % to check for statistically significant differences } 
We observe that all models using multiple embeddings per word perform better than SGE. 
Our proposed models outperform both SGE and MSSG in both evaluation sets, with more pronounced improvements in the LS-CIC data set. 
Note that we do not require any syntactic information and only focus on the semantic aspect of the task. 
We further observe that our \textit{Exp} method is more robust and performs better for all embedding sizes.  
Moreover, we can see a decrease in GAP for the model variant HTLEadd compared to HTLE. By including a generic representation for each word, different topic representations drift close to each other and obtain a more general meaning of the word as well as the topic-specific meaning. Such representations are not beneficial for this task. %which requires distinguishing between different meanings of a word.

Table~\ref{nvaa} shows the GAP scores broken down by the main word classes: noun, verb, adjective, and adverb. With 100 dimensions, our best model (HTLE) yields improvements across all POS tags, with the largest improvements for adverbs and smallest improvements for adjectives. 
\bgroup
\def\arraystretch{1.25}
\begin{table}[htb!]
\centering
\small
\setlength{\tabcolsep}{.6em}
\caption{GAP scores on the candidate ranking task on LS-SE07 for different part-of-speech categories. \label{nvaa}}
\begin{tabular}{c|l|cccc}
\toprule
& \textbf{Model} &  \textbf{Noun}             & \textbf{Verb}           & \textbf{Adjective}       &  \textbf{Adverb}    \\ % & All  \\ 
\midrule
\multirow{4}{*}{\rotatebox[origin=c]{90}{Dim=100}}  & SGE &  33.1            & 29.2            & 31.7            & 38.2 \\
 & SGE + context   &  37.2            & 31.6            & 37.1            & 42.2           \\
& HTLE &  \textbf{42.4} & \textbf{33.9} & \textbf{38.1} & \textbf{49.7}  \\ 
& STLE & 42.0 & 33.1 & 38.1 & 47.2 \\
\midrule 
\multirow{4}{*}{\rotatebox[origin=c]{90}{Dim=300}}  & SGE & 39.0 & 33.8 & 36.4 & 50.1 \\
 & SGE + context   &  39.2            & 35.0 & 39.0            & \textbf{55.4}   \\
& HTLE &  \textbf{44.9} & \textbf{37.0} & \textbf{41.0} & 50.9      \\ 
& STLE & 42.7 & 37.0 & 39.9 & 50.2 \\
\midrule
\multirow{4}{*}{\rotatebox[origin=c]{90}{Dim=600}}  & SGE & 39.1 & 34.3 & 36.9 & 52.8 \\
 & SGE + context   &   39.7            & 35.7            & 39.9            & \textbf{56.2}\\
& HTLE &  \textbf{45.2} & \textbf{37.2} & \textbf{42.1} & 51.9          \\ 
& STLE & 44.0 & 37.1 & 41.5 &  51.0 \\
\bottomrule
\end{tabular}
\end{table} 
\egroup

When increasing the dimension size of embeddings, the improvements hold up for all POS tags apart from adverbs. It can be inferred that larger dimension sizes capture semantic similarities for adverbs and context words better than other parts-of-speech categories. 
Additionally, we observe for both evaluation sets that the improvements are preserved when increasing the embedding size. 
It should also be noted that the distribution of POS tags in the test set is approximately uniform except for adverbs of which there are fewer instances.

\begin{table}  \setlength{\tabcolsep}{0.5em}
\begin{center}
\small
\caption{\label{lsexample} Examples of word substitution rankings and respective GAP scores. The gold rank includes substitution words and annotators' votes. The models are word embeddings with context (SGE), MSSG \citep{neelakantan2014efficient}, and our topic-sensitive model (HTLE). Target words in the contexts and correct words in the rankings are bold.}
\begin{tabularx}{\textwidth}{lXr}
\toprule
& \textbf{Substitution instance} & \textbf{GAP} \\
\midrule
   \textbf{ (a)}  & {During the siege, George Robertson had appointed Shuja-ul-Mulk, who was a \textit{\textbf{bright}} boy only 12 years old and the youngest surviving son of Aman-ul-Mulk , as the ruler of chitral.} &   \\
 \hdashline
\rule{0pt}{2.5ex} 
   Gold & intelligent (3), clever (3), smart (1) & \\
 \hdashline
\rule{0pt}{2.5ex} 
   SGE   & \textit{shining, luminous, vibrant, brilliant, vivid, colourful, gleam, light, sharp, \textbf{smart},} \ldots & 12.02 \\ % \multirow{3}{*}{\rotatebox[origin=c]{90}{ranks}}  
  MSSG  & \textit{vivid, shining, luminous, brilliant, colourful, vibrant, gleam, sharp, light, talented,} \ldots & 10.34 \\
   HTLE   & \textit{brilliant, gifted, talented, capable, sharp, \textbf{intelligent}, clear, vivid, colourful, shining,} \ldots & 16.00 \\ 
 \midrule
    \textbf{(b.1)}  & Trees on Anmyeondo used to be thick and lush to the extent of prompting a saying, ``you can become \textit{\textbf{rich}} with an axe'', but now only few trees are left due to reckless deforestation since the time of Korea's liberation from Japanese colonial rule.  &  \\
  \hdashline
\rule{0pt}{2.5ex} 
   Gold   & wealthy (5) &  \\
  \hdashline
\rule{0pt}{2.5ex} 
   SGE   & \textit{\textbf{wealthy}, abundant, vibrant, lush, abounding, lavish, ample, valuable,} \ldots & 100 \\
   MSSG  &\textit{ lush, abundant, \textbf{wealthy}, vibrant, abounding, valuable, lavish, ample,} \ldots & 33.33 \\
   HTLE   & \textit{abundant, valuable, lush, abounding, vibrant, ample, high, significant,} \ldots & 7.69 \\ 
 \midrule
\textbf{ (b.2)}  &Africas central problems in the WTO revolve around the imbalances and biases created by \textit{\textbf{rich}} countries in the Uruguay round agreement (URA) . &  \\
 \hdashline
\rule{0pt}{2.5ex} 
   Gold   & wealthy (5) &  \\
  \hdashline
\rule{0pt}{2.5ex} 
   SGE   & \textit{abundant, lush, \textbf{wealthy}, abounding, vibrant, valuable, ample, lavish,} \ldots &  33.33\\
   MSSG  & \textit{abundant, \textbf{wealthy}, vibrant, lush, abounding, lavish, ample, valuable,} \ldots & 50 \\
   HTLE   & \textit{\textbf{wealthy}, abundant, valuable, vibrant, significant, abounding, ample,} \ldots  & 100 \\ 
 \midrule
\textbf{ (c)}   & I feel I can get a lot more done as a selectman by being innovative and \textit{\textbf{fixing}} the problems we have with cash flows, because they occur every year. &  \\
  \hdashline
\rule{0pt}{2.5ex} 
   Gold   & resolve (2), solve (1), mend (1), repair (1) &  \\
  \hdashline
\rule{0pt}{2.5ex} 
   SGE  & \textit{\textbf{resolve}, \textbf{mend}, \textbf{repair}, heal, cure, improve, correct, stick, do, patch,} \ldots  &  79.45  \\
    MSSG  & \textit{do, \textbf{mend}, heal, cure, \textbf{resolve}, \textbf{repair}, correct, improve, stick, patch,} \ldots & 29.04 \\
   HTLE   & \textit{do, improve, heal, \textbf{repair}, cure, \textbf{resolve}, \textbf{mend}, determine, stick,} \ldots   & 21.72  \\ 
 \bottomrule
\end{tabularx}
\end{center}
\end{table}

Our findings confirm \citet{li-jurafsky:2015:EMNLP}'s observations to some extent: 
Higher dimension embeddings capture part of the information on semantic relations that models with multiple embeddings per word capture.
In our experiments, SGE with 600 dimensions performs better than HTLE with 100 dimensions. 
Given a relevant semantic task, this advantage of having multiple embeddings per word can be observed more appropriately regardless of dimension.
   
\section{Qualitative analysis} \label{embindepth}

In this section, we discuss some properties of our representations and provide an analysis of the semantic information captured by topic-sensitive embeddings in the lexical substitution evaluation. 
Table~\ref{lsexample} illustrates the performance of different models by providing examples from the lexical substitution task LS-SE07. 
In Table~\ref{lsexample} (a), for the word \textit{`bright'}, we observe that our model captures the meaning of the word in the context \textit{(`talented')} and provides a sensible substitution ranking, but the GAP scores are low. 
This is due to the gold substitution list being incomplete and the highest-ranking words of our model, despite matching the context, are not in the gold ranking. SGE and MSSG however, rank the candidates belonging to a different sense (\textit{`shiny'}) higher.

    \begin{figure}   
\centering
\includegraphics[scale=0.7]{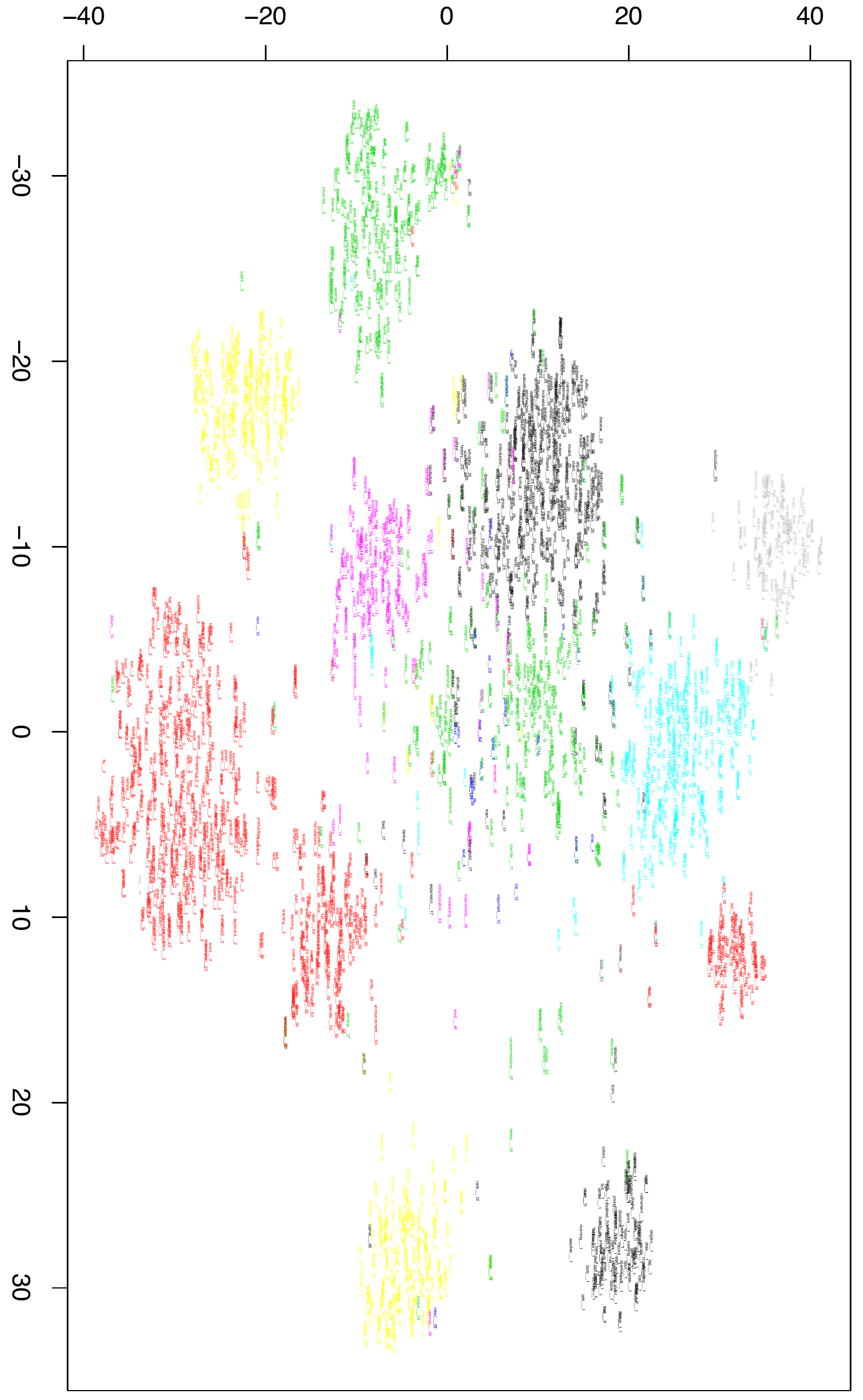} 
  \captionof{figure}{Visualizing a subset of word-topic pairs using t-SNE to showcase topic assignment separations. Colors distinguish topics. We observe that words that are labeled the same topics end up in the same clusters. }
  \label{vecvis}
   \end{figure}
   
Additionally, Table~\ref{lsexample} provides an example of two different contexts for the same word \textit{`rich'}. In Table~\ref{lsexample} (b.1), the topic of the sentence for the word \textit{`rich'} was learned correctly, but it is misleading for the substitution because the meaning of the word changes in the local context and SGE and MSSG perform better than our model and rank the correct substitution \textit{`wealthy'} higher. 
However, for the same word (\textit{`rich'}) in a different context, Table~\ref{lsexample} (b.2), the topic-sensitive model obtains a more accurate substitution ranking and SGE fails to identify the meaning of the word (\textit{`wealthy'}) in the context.
Example (c) in Table~\ref{lsexample} provides an instance for substituting the word \textit{`fixing'} in which we achieve a higher GAP score by using SGE. Here, both MSSG and our model detect an inaccurate sense for the context (\textit{`heal'}) and rank the words accordingly. 

These examples show that when we learn multiple embeddings per word, we can disambiguate words in context to a greater degree. 
Comparing the MSSG and the HTLE model, we see a slight difference in results.
One distinction between the MSSG model and the HTLE model is their definition of context.
The former uses a context window of length 10 \citep{neelakantan2014efficient} and the latter uses a context window of length 10 as well as the document-level (in this case the complete sentence) topical information. 
While both models perform similarly, HTLE can be more effective when requiring larger contexts for disambiguation.

Figure~\ref{vecvis} uses t-SNE \citep{7b54165e73a3424b8820136bcf61ca89} to visualize the embeddings of a subset of word-topic pairs in the vocabulary with colors distinguishing between topics. 
This figure gives us a basic understanding of the vector space and the distribution of topics.
We observe that in general, words are closer (more similar) to other words from the same topic, rather than the same words in different topics. 
Additionally, we observe that in some cases, the same word with different topic labels ends up with almost overlapping embeddings. 
This indicates that while we assign different topic labels to a word in different documents, as long as the sense of the word is the same, the embeddings we learn turn out very similar.

\section{Conclusion} \label{embconc}

Studying word embeddings is a good medium for getting an understanding of the impact of context in preserving word meaning.
%there is little work done on capturing the polysemic nature of language.
In this chapter, we have explored how document-level context can be useful to learn a more informed word representation.
We asked:  

\begin{enumerate}[label=\textbf{RQ1.\arabic* },wide = 0pt, leftmargin=2em]
\setlength\itemsep{1em}
 \setcounter{enumi}{0}
\item \acl{rq:topic1}

\medskip

\noindent We introduced a model that uses a hierarchical Dirichlet process to learn topic distributions over documents. 
We observed that these distributions distinguish between senses of words. 
 This method exploits the document-level context of words and does not require annotated data or linguistic resources. 
Using this information, we asked:

\item \acl{rq:topic2}

\medskip

\noindent We approximated the word senses with topics and further used this additional signal for training the embeddings of each topic-word pair separately. 
Our first model hard-labeled words with topics to learn representations. The second model jointly learned topic-labeled and generic representations for each word in order to share statistical information between different meanings of a particular word. 
The third model used topic distributions for each word following the notion that meanings of words are not mutually exclusive in a given context.

 \noindent  Lastly, we investigated the effectiveness of these embeddings by asking:

\item \acl{rq:topic3}

\medskip

\noindent 
When the evaluation tasks require less contextual information, the performance of our model was similar to the baselines.
 We evaluated word embeddings on the word similarity task and observed slight improvements under different settings. 
However, there was no clear indication that one model performs best across all data sets. 
Next, we evaluated the embeddings in a more context-aware setting.
Using the SCWS data set, where context is available for word pairs, we saw that two of our models obtained improvements over the baseline. 
However, with higher dimensions, the MSSG model \citep{neelakantan2014efficient} was the best performing system. 
Finally, we showed that in the lexical substitution ranking task \citep{mccarthy2007semeval} 
our models outperformed two competitive baselines and performed comparably to the best-performing methods even though ---unlike those methods---our approach did not use any syntactic information. 

\end{enumerate}

 \noindent  Taken together, these questions answered:

\paragraph{Research Question 1:} \acl{rq:topic} 

\medskip

 \noindent  Our experiments showed that we can use topic distribution over documents to improve the learning of word representations.
%We have introduced an approach to learn topic-sensitive word representations that exploits the document-level context of words and does not require annotated data or linguistic resources. 
With our proposed approach, we obtained improvements in the lexical substitution task without using any syntactic information. 
Our HTLE model which learns representations by hard-labeling topics to target words and learning individual embeddings achieved the best performance.
We observed that topic-sensitive representations capture different senses of the words to some extent and work best when context is available. 

\medskip

 \noindent It is worth mentioning that the methods proposed in this chapter predate more powerful neural models, such as transformers as well as complex language modeling objectives to learn dynamic contextual embeddings \citep{peters-etal-2018-deep,radford2019language,devlin-etal-2019-bert}. 
These models incorporate sentence-level context (and at times multiple sentences) in the computation of each word representation and have been shown to be very effective at capturing different meanings of words.

In this chapter, we studied how \textit{static embeddings} can benefit from larger contextual cues, namely the topic of the document. 
%While we used document topics as additional input to the model, the use of topical information can be further explored. 
As a byproduct of our models, we also learned representations for topics and our visualization of these embeddings suggested that words belonging to the same topic are indeed clustered together. 
Embeddings that integrate informative priors such as topics are more interpretable \citep{DBLP:journals/corr/abs-1807-07279} and can be used to advance our understanding of what word embeddings capture and represent \citep{hurtado-bodell-etal-2019-interpretable}.

% !TEX root = thesis-main.tex

\chapter{Data Augmentation for Rare Words}
\label{chapter:research-02}

\section{Introduction and research questions}

In the previous chapter, we observed the impact of context on learning word representations, in particular, modeling polysemy which is a challenging phenomenon in language. 
In the following chapters of this thesis, we investigate other challenges in learning word meaning. 
One medium to evaluate language understanding is machine translation. 
A machine translation system needs to understand the meaning in the source language and transfer it into the target language. 

The quality of a neural machine translation system depends substantially on the availability of sizable parallel corpora.
To train NMT models with reliable parameter estimations, these networks require numerous instances of sentence translation pairs with words occurring in diverse contexts, which is typically not available for low-resource language pairs.
As a result, NMT falls short of producing good-quality translations for these language pairs \citep{zoph-EtAl:2016:EMNLP2016,koehn2017six,gu-etal-2018-meta,ngo-etal-2019-overcoming}.
The solution is to either revise the learning models \citep{ostling2017neural}, or to provide more training data by manual annotation \citep{Melamed98manualannotation} or to perform automatic data augmentation \citep{sennrich-haddow-birch:2016:P16-11,wang-etal-2018-switchout}. 
Since manual annotation of data is time-consuming, data augmentation for low-resource language pairs is a more viable approach.

\begin{figure}
\centering
\includegraphics[width=0.75\linewidth]{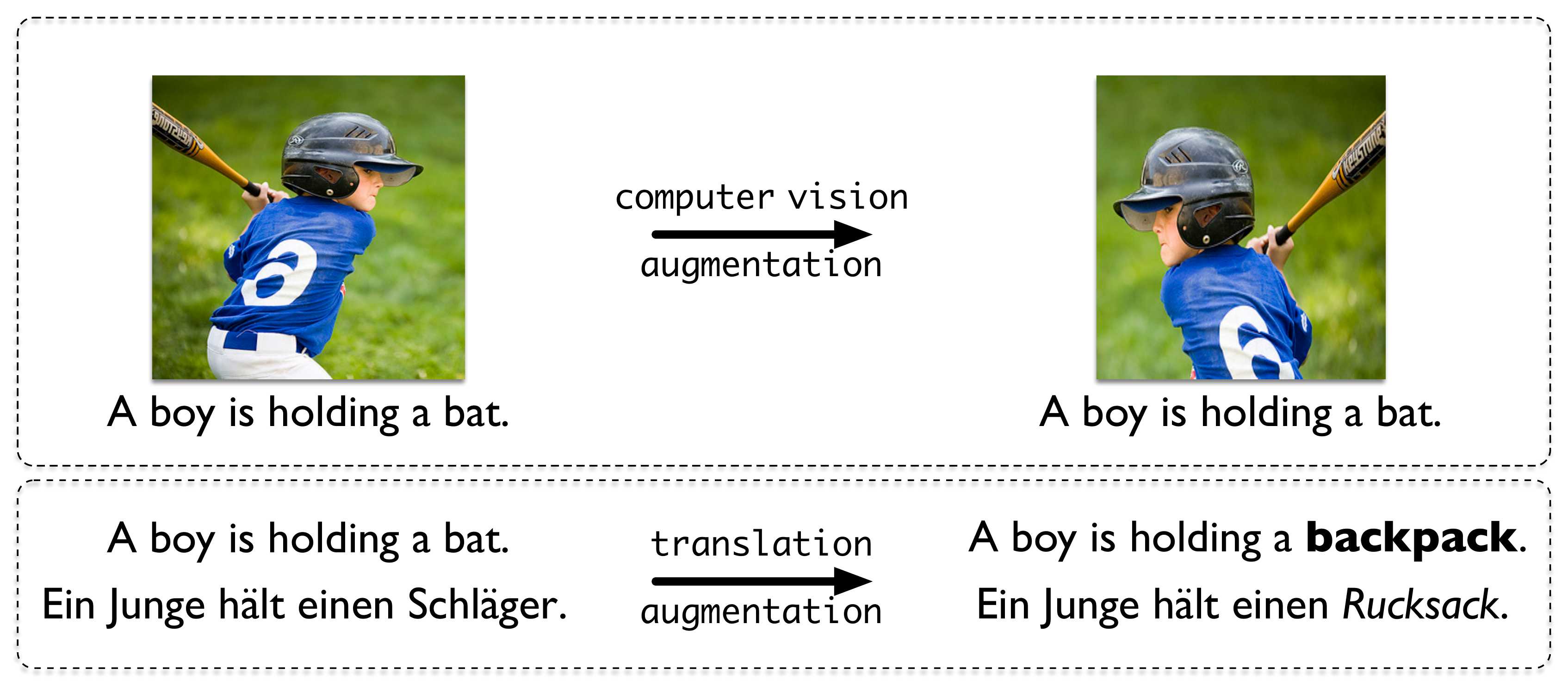}
\caption{Top: %A data augmentation technique to alter an image where the description remains the same. 
flip and crop, two label-preserving data augmentation techniques in computer vision.
Bottom: Altering one sentence in a parallel corpus requires changing its translation.}
\label{augfig}
\end{figure}

In computer vision, data augmentation techniques are widely used to increase robustness and improve the learning of objects with a limited number of training examples. 
In image processing, the training data is augmented  
by, for instance, horizontally flipping, random cropping, tilting, and altering the RGB channels of the original images \citep{NIPS2012_4824,DBLP:conf/bmvc/ChatfieldSVZ14}. Since the content of the new image is still the same, the label of the original image is preserved (see Figure~\ref{augfig}: top).
While data augmentation has become a standard technique to train deep networks for image processing, it is not common practice for training networks for NLP applications such as machine translation.

In this chapter, we address the challenge of translation of low-resource language pairs where the primary obstacle is the lack of sufficient training data.
Motivated by the success of data augmentation in computer vision, we investigate in this chapter whether NMT can benefit from data augmentation as well.
Concretely, we ask:

\paragraph{Research Question 2:} \acl{rq:tdabt} 

\medskip

 \noindent Research has shown that there is a strong correlation between the size of the training data and the quality of neural models \citep{Halevy:2009:UED:1525642.1525689}.
To investigate this relation in machine translation, we compare how the translation and generation of a word changes by adding diverse contexts to the training data.  
In this chapter, we focus on low-resource language pairs, simulating a low-resource setting as done in the literature \citep{Marton:2009:ISM:1699510.1699560,duong-EtAl:2015:EMNLP},  to examine the effects of the lack of data on translation quality.
In particular, we look into the translation and generation of rare words, thus asking: 
 
\begin{enumerate}[label=\textbf{RQ2.\arabic* },wide = 0pt, leftmargin=2em]
\setlength\itemsep{1em}
\item \acl{rq:tda1}

\medskip

\noindent The impact of training data scarcity on translation quality is especially noticeable for rare words \citep{sennrich-haddow-birch:2016:P16-12}. 
%Our quantitive analysis starts with examining rare words in a low-resource setting. 
We demonstrate that parameter estimation of rare words is challenging in NMT, and it is further exacerbated in a low-resource setting.
%Concretely, we measure the quality of translation and generation of these words.
We investigate the effects of additional context, generated automatically, on both \textit{translating} and \textit{generating} rare words. 
We achieve this by proposing a simple yet effective approach that augments the training data by altering existing sentences in the parallel corpus, similar in spirit to the data augmentation approaches in computer vision (see Figure~\ref{augfig}).
First, we propose a weaker notion of label preservation that allows altering both source and target sentences at the same time as long as they remain translations of each other. 
%We obtain substantial improvements for translating English$\rightarrow$German and German$\rightarrow$English.

Next, we examine augmentation during \textit{test time} by exploring a stronger notion of label preservation, and we ask: 

\item \acl{rq:tda2}

\medskip

\noindent For the augmentation process in this scenario to be possible, any change to a sentence in one language must preserve the meaning of the sentence.  
It is essential because we aim for not altering reference translations during evaluation. 
We hypothesize that it should be useful to alter the source sentence containing a rare or out-of-vocabulary word by paraphrasing it with a more common word.
In addition, we also investigate the performance of different paraphrasing resources.

\end{enumerate}

\paragraph{Organization.} This chapter is organized as follows: 
After reviewing the previous work (Section~\ref{tda:background}), we present our main data augmentation model in Section~\ref{tda:model}. 
In Section~\ref{tda:exp}, we introduce the general experimental setup, followed by a detailed description of the results of the translation experiments in Section~\ref{tda:results}. 
We analyze the effectiveness of the model further in Section~\ref{tda:analysis}.
Next, we discuss augmentation at inference and propose a meaning-preserving method in Section~\ref{tda:semantic}. 
Finally, we conclude in Section~\ref{tda:conc} with an outlook of future work.

\section{Previous work} \label{tda:background}

Relevant previous work for the work described in this chapter involves two research topics: 
First, we briefly review the literature on image data augmentation.
Second, we discuss researches studying challenges in translation of low-resource language pairs.
%We also briefly present previous work on neural language modeling. 

\subsection{Data augmentation in computer vision}

Neural models learn best when massive data is available \citep{Halevy:2009:UED:1525642.1525689}.
As a result, data augmentation has become one of the staple preprocessing steps in image classification \citep{NIPS2012_4824,huang2018gpipe, cubuk2019autoaugment}, image generation \citep{kynkaanniemi2019improved,karras2019style}, and object detection \citep{singh2018sniper,liu2016ssd}.
Data augmentation approaches address the overfitting problem in neural models from the perspective of the training data. 
Extensive research in computer vision has been done over the years on different techniques of data manipulation. 
There are various techniques to manipulate image data. For instance:

\begin{itemize}
 \item \textbf{Geometric transformations:} These changes are simple alterations that are applied to the images in the training data---for instance, flipping, rotation, and cropping. 
    
    \item \textbf{Mixing images:} These transformations are done by combining multiple images either from the same class or sampled from the entire training data---for instance, averaging pixel values of multiple images or randomly cropping and patching images. \citet{inoue2018data} show that by overlaying images and augmenting data, they achieve significant improvements in classification accuracy on CIFAR-100 data set.
\end{itemize}

Note that the augmentation techniques in computer vision have hardly any concerns about whether the image still \textit{remains} an equivalent image after the alteration. 
That is not the case for augmentation of sentences, where any alteration should be mindful of generating semantically and syntactically correct sentences. 

\begin{comment}
\begin{table}
\centering
\small
\begin{tabularx}{\linewidth}{lll}
 \toprule
\textbf{Technique} & \textbf{Augmentation strategy} &   \\ \midrule
\multirow{5}{*}{Data warping}  &  Geometric transformations & \\
& Kernel filters & \\
& Color space transformation & \\
& Random erasing & \\
& Mixing images & \\ \midrule
\multirow{3}{*}{Deep learning approaches} & Adversarial training & \\
& Neural style transfer & \\
& GAN data augmentation & \\
\bottomrule
\end{tabularx}
\caption{blah}
\label{backgroundcomputervision}
\end{table}
\end{comment}

\subsection{Low-Resource translation}

Parallel data, which is the primary source of learning for machine translation models, is constructed manually and is not available in abundance for every language pair (see Section~\ref{databg}). 
Neural translation models especially suffer from the lack of sufficient parallel data for training. 
\citet{koehn2017six} experiment with different corpora sizes and show that their NMT system only outperforms their phrase-based machine translation system when more than 100 million words of parallel data are available. 
However, \citet{sennrich-zhang-2019-revisiting} show that NMT models are highly sensitive to hyperparameters such as BPE vocabulary size. They observe strong improvements by adapting system parameters to low-resource settings.  

Additionally, in a low-resource setting, the problem of translating rare words is more pronounced. 
Both \citet{sutskever2014sequence} and \citet{DBLP:journals/corr/BahdanauCB14} observe that NMT models tend to translate sentences with many rare words more poorly than sentences containing mostly frequent words.
Several recent approaches have targeted the low-resource obstacle in machine translation in different ways. 
Based on their approach and viewpoint on addressing this problem, current research can be categorized into four main groups:

\paragraph{Leveraging monolingual data} 
We discussed several approaches to leverage monolingual data in translation in Section~\ref{bgbtref}.
These approaches address the problem by leveraging data resources other than the limited parallel corpora. 
\citet{sennrich-haddow-birch:2016:P16-11} propose a method to back-translate sentences from monolingual data and augment the bitext with the resulting pseudo-parallel corpora.  %This approach results in significant improvements in many language pairs.
While this approach is successful in improving the translation quality, it is not effective in very low-resource settings where the back-translation model cannot be trained to a sufficient level of quality \citep{abdulmumin2020using}.
This approach is further discussed in Chapter~\ref{chapter:research-03}.
\citet{currey2017copied} create a parallel corpus from monolingual data in the target language by copying it so that each source sentence is identical to its corresponding target sentence. With this simple technique, they observe improvements on relatively low-resource language pairs such as English$\leftrightarrow$Turkish and English$\leftrightarrow$Romanian.

\paragraph{Re-designing the model} 
These approaches target the model itself and propose changes to the standard neural translation models \citep{costa-jussa-fonollosa-2016-character,sennrich-haddow-birch:2016:P16-12}.
\citet{ostling2017neural} propose to learn sentence reordering during translation of low-resource language pairs by introducing more local dependencies.
They use word alignments to provide supervision to the reordering model.
\citet{lee-etal-2017-fully} introduce a fully character-level translation model that maps a character sequence in a source language to a character sequence in a target language. 
They observed significant improvements in the translation of morphologically rich languages where the word-level NMT models fail to translate rare and out-of-vocabulary words.
Previous approaches which propose different segmentations of the input sequence are effective; however, they present a different set of challenges:
With longer sequences, the model requires information to be retained over longer temporal spans. 
Moreover, since the meaning of a word is not a compositional function of its characters, the model must learn to memorize many character sequences as higher-level linguistic abstractions. \citep{cherry-etal-2018-revisiting}.

\paragraph{Cross-lingual transfer learning}
These strategies use models trained on high-resource language pairs to transfer various parameters and components to the low-resource language pair. 
\citet{zoph-EtAl:2016:EMNLP2016} propose to train a high-resource language pair first and then transfer some of the learned parameters to the low-resource pair to initialize and constrain training. 
\citet{gu2018universal} show that sharing lexical and sentence-level representations across multiple source languages aid in the translation of low-resource languages. They also use monolingual embeddings along with seed parallel data from all languages to build a universal representation.
Cross-lingual approaches are particularly valuable for multilingual translation learning, where a single NMT model learns to translate between multiple languages \citep{firat-etal-2016-multi,johnson-etal-2017-googles,blackwood-etal-2018-multilingual,aharoni-etal-2019-massively}.
While these approaches are very impressive in translating between language pairs not seen during training, this paradigm cannot outperform the individual models trained on bilingual corpus in many cases \citep{johnson-etal-2017-googles}.

\paragraph{Unsupervised learning} 
These studies focus on zero-resource learning, where there are no parallel corpora available for a language pair \citep{yang-etal-2018-unsupervised,artetxe-etal-2018-unsupervised,artetxe2018iclr,artetxe-etal-2019-effective}.
\citet{lample2018unsupervised} propose a model that takes sentences from monolingual data in two different languages and maps them into the same latent space.
The model learns to translate without using any parallel data by reconstructing both languages from the shared feature space. 
\citet{lample-etal-2018-phrase} address the challenge of only having access to monolingual corpora in each language. 
They use a smoothed n-gram language model (phrase-based model) as a data-driven prior to denoising sentences and automatically generate the parallel data by iterative back-translation (neural model).
%, they achieve impressive results. 
These approaches are effective; however, the pseudo sentences used for training are usually of low quality as translation mistakes accumulate during training. 
Additionally, while they perform well between languages that are from the same branch, they perform poorly between distant languages \citep{sun-etal-2020-knowledge}.

%\subsection{Recurrent neural language model} TODO: move to BG chapter

\section{Data augmentation for rare words} \label{tda:model}

In this section, we propose a novel approach for data augmentation of parallel corpora. Specifically, we use a Bidirectional RNN model trained on monolingual data to introduce completely new contexts for rare words in the bitext. In our approach we use sentences from the training data as starting points and use the probability distribution of the output layer of an RNN to insert words into new contexts: 
Given %a pair of 
a source and target sentence pair $(S,T)$, we want to alter it in a way that we obtain new contexts for these words while diversifying as much as possible the training examples.
A number of ways to do this can be envisaged, as for example paraphrasing (parts of) $S$ or $T$, or altering both and preserve the semantic equivalence between $S$ and $T$.
We explore both approaches in this chapter.

We choose to focus on a subset of the vocabulary that we know to be poorly modeled by our baseline NMT system, namely words that occur rarely in the parallel corpus.
Thus, the goal of our data augmentation technique is to provide novel contexts for rare words.
To achieve this we search for contexts where a common word can be replaced by a rare word and consequently replace its corresponding word in the other language by that rare word's translation:
\vspace{1mm}

\begin{center}\small
\setlength{\tabcolsep}{15pt}
\begin{tabular}{ll}
\textbf{original pair} & \textbf{augmented pair} \\\hline
$S: s_1, \ldots , s_i, \ldots, s_n$ & $S': s_1, \ldots , s_i', \ldots , s_n$ \\
$T: t_1, \ldots , t_j, \ldots , t_m$ & $T': t_1, \ldots , t_j', \ldots , t_m$ \\ %\textit{trans}(s_i')
\end{tabular}
\end{center}

\vspace{1mm}

\noindent%
where $t_j$ is a translation of $s_i$ and word-aligned to $s_i$, and $t_j'$ is the translation of $s_i'$.
Plausible substitutions are those that result in a fluent and grammatical sentence but do not necessarily maintain its semantic content.
As an example, the rare word \textit{motorbike} can be substituted in different contexts:

\smallskip

\begin{center}\small
\begin{tabular}{l | l}
Sentence [original \textbackslash \textbf{substituted}] &  \begin{tabular}[x]{@{}c@{}}Plausible\end{tabular}  \\
\hline
	My sister drives a  [car \textbackslash \textbf{motorbike}]  & yes \\
	My uncle sold his [house \textbackslash \textbf{motorbike}] & yes \\
	Alice waters the [plant \textbackslash  \textbf{motorbike}]  & no (semantics) \\
	John bought two [shirts \textbackslash\textbf{motorbike}] & no (syntax) \\
\end{tabular}
\end{center}

\smallskip

\noindent%
%where the third and fourth rows illustrate a semantically and a syntactically wrong context respectively.
%with highlighted [original / \textbf{substituted}] words. 
%Clearly, blindly substituting words can result in semantically or syntactically wrong sentences. %as shown in this table.  (third and fourth examples).
Implausible substitutions need to be ruled out during data augmentation.
To this end, rather than relying on linguistic resources %, like parsers or ontologies, 
which are not available for many languages, %we rule out inappropriate substitutions by using 
we rely on LSTM language models (LM) trained on large amounts of monolingual data in both forward and backward directions.

Our data augmentation method involves the following steps:
\paragraph{Targeted words selection:}
Following common practice, our NMT system limits its vocabulary $V$ to the $v$ most common words observed in the training corpus.
We select the words in $V$ that have fewer than $R$ occurrences and use this as our targeted rare word list $V_R$.
\paragraph{Rare word substitution:}
If the LM suggests a rare substitution in a particular context, we replace that word and add the new sentence to the training data.
Formally, given a sentence pair $(S,T)$ and a position $i$ in $S$, we compute the probability distribution over $V$ by the forward and backward LMs and select rare word substitutions $\mathcal{C}$ as follows:
\begin{align}
\overrightarrow{\mathcal{C}} & = \{ s'_i \in V_{R} : \topk  P_{\textrm{\textit{ForwardLM}}_S}(s'_i \given s_{1}^{i-1}) \} \\
\overleftarrow{\mathcal{C}} & = \{ s'_i \in V_{R} : \topk P_{\textrm{\textit{BackwardLM}}_S}(s'_i \given s_{n}^{i+1}) \} \\
\mathcal{C} \ & = \{s'_i \given s'_i \in \overrightarrow{\mathcal{C}} \land s'_i \in \overleftarrow{\mathcal{C}} \} 
\end{align}
where $s_{i}^{j}$ are context words from position $i$ to $j$ and $\topk$ returns the $K$ words with highest conditional probability according to the context. 
The selected substitutions $s'_i$, are used to replace the original word and generate a new sentence.
%
%\footnote{We accept as plausible substitutions the top $k$ words of the vocabulary ranked by the multiplication of forward and backward RNN LM probabilities.}

\paragraph{Translation selection:}
Using automatic word alignments\footnote{We use fast-align \citep{dyer-chahuneau-smith:2013:NAACL-HLT} to extract word alignments and a bilingual lexicon with lexical translation probabilities from the low-resource bitext.} trained over the bitext, we replace the translation of word $s_i$ in $T$ by the translation of its substitution $s_i'$. 
Following a common practice in statistical MT \citep{koehn-etal-2007-moses}, the optimal translation $t'_j$ is chosen by multiplying direct and inverse lexical translation probabilities with the LM probabilities of the translation in context: % the other language (?)}.
\begin{equation}
t'_j =  \mathop{\argmax}_{t \in \textit{trans}(s_i')} P(s'_i \given t) P(t \given s'_i) P_{\textrm{\textit{ForwardLM}}_T}(t \given t_{1}^{j-1}) P_{\textrm{\textit{BackwardLM}}_T}(t \given t_{n}^{j+1})
\end{equation}
If no translation candidate is found because the word is unaligned or because
%In case there is no substitution candidate due to no alignment, or 
the language model probability is less than a certain threshold, the augmented sentence is discarded.
This reduces the risk of generating sentence pairs that are semantically or syntactically incorrect.

\medskip

We use the described steps of targeting and substituting words in source and target sentences to augment the training data:

\paragraph{Sampling:}
We loop over the original parallel corpus multiple times, sampling substitution positions $i$ in each sentence and making sure that each rare word gets augmented at most $N$ times so that a large number of rare words can be affected.
We stop when no new sentences are generated in one pass over the training data.

\paragraph{Augmentation:}
Assuming the original parallel data is $\mathcal{D}$, our data augmentation process can be represented by the following mapping:
\begin{equation}
\phi:  \mathcal{D} \mapsto \mathcal{A}
\end{equation}
where $\mathcal{A}$ is the modified set with new contexts for rare words built from sentence pairs in $\mathcal{D}$. 
Note that some sentences in $\mathcal{D}$ may not be augmented because of the randomness of the sampling step and shortage of substitution suggestions from the language model.
As the final step, the training data is expanded as the union of the original data and the augmented data:
\begin{equation}
\mathcal{D'} = \mathcal{D} \cup \mathcal{A}
\end{equation}

\gap

\begin{figure}[htb!]
\centering
\includegraphics[width=0.88\linewidth]{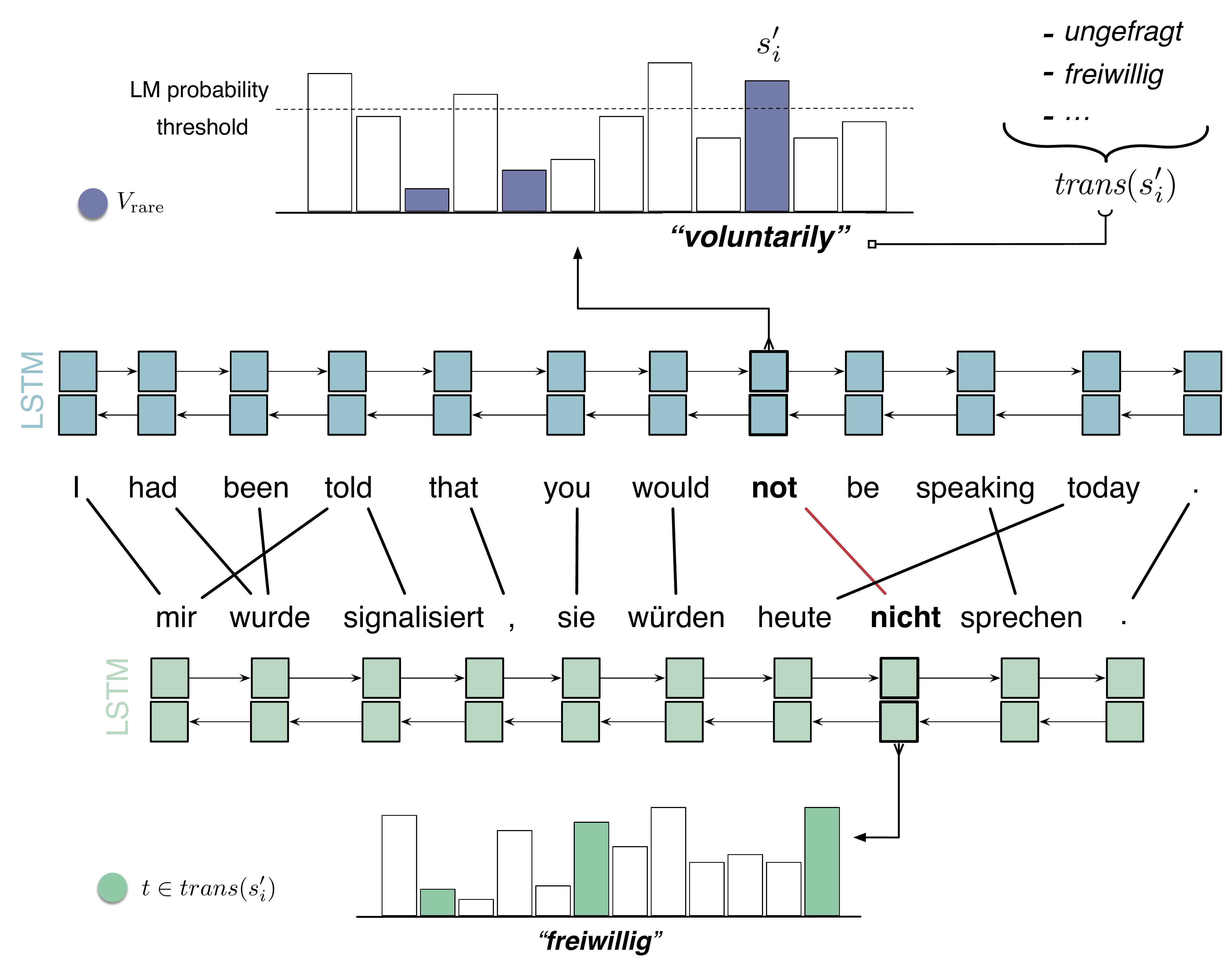}
\caption{A visual representation of our proposed mechanism for generating new sentence pairs.}
\label{tdaposter}
\end{figure}

Our proposed method is demonstrated in Figure~\ref{tdaposter} with an example:
Given the English sentence `\textit{I had been told that you would not be speaking today.}', we randomly sample the position of the word `\textit{not}' and explore the suggestions of the language model that fit the context. 
We select the word `\textit{voluntarily}' because it is a rare word in the low-resource setting of our experiments and the English language model has high confidence in substituting it in the sentence.
Correspondingly, we explore the translation candidates of the word `\textit{voluntarily}' to make comparable changes to the German sentence.
From the translation candidates, we choose the word that yields the most fluent sentence according to the German language model.
Therefore the word `\textit{freiwillig}' is selected to substitute `\textit{nicht}' in the German sentence.
The newly generated sentence pairs are then added to the training data. 

Table~\ref{examples} provides several examples resulting from our augmentation procedure. 
While using a large LM to substitute words with rare words mostly results in grammatical sentences, this does not mean that the meaning of the original sentence is always preserved. 
Note that meaning preservation is not an objective of the proposed approach in this section and we will explore this further in Section~\ref{tda:semantic}.  

\begin{table}[hbt!]
%\begin{minipage}[c]{0.48\linewidth}
\centering
\small
%\begin{tabularx}{\linewidth}{@{\ }l @{\ \ \ }X@{\ }}
\caption{Examples of augmented data with highlighted \underline{Original} words and \textbf{substituted} words. \label{examples}
% [$w_1$ / $w_2$] in the sentence indicates the replacement where $w_1$ is the original word and $w_2$ is substitution. 
%In the English sentence the substitution is suggested by the LM, and in the German sentence by the alignments to the English substitution. % WE SAID THIS BEFORE
}
\begin{tabularx}{\linewidth}{l@{\hskip 0.2in}X@{\hskip 0.2in}X}
 \toprule
 & \textbf{Original sentence pair} & \textbf{Synthetic sentence pair} \\
 \midrule
(a) & \textsc{src:} I had been told that you would \underline{not} be speaking today. &  \textsc{src:} I had been told that you would \textbf{voluntarily} be speaking today. \\
 & \textsc{tgt:} mir wurde signalisiert, sie w{\"u}rden heute \underline{nicht} sprechen. & \textsc{tgt:} mir wurde signalisiert, sie w{\"u}rden heute \textbf{freiwillig} sprechen.\\
 \midrule
(b) & \textsc{src:} the present situation is \underline{indefensible} and completely unacceptable to the commission. & \textsc{src:} the present situation is \textbf{confusing} and completely unacceptable to the commission.\\
  & \textsc{tgt:} die situation sei \underline{unhaltbar} und f{\"u}r die kommission g{\"a}nzlich unannehmbar. &  \textsc{tgt:} die situation sei \textbf{verwirrend} und f{\"u}r die kommission g{\"a}nzlich unannehmbar.\\
% EXAMPLE OK
\midrule
(c) &  \textsc{src:}  \ldots agree wholeheartedly with the institution of an ad hoc delegation of parliament on the turkish \underline{prison} system. &  \textsc{src:} \ldots agree wholeheartedly with the institution of an ad hoc delegation of parliament on the turkish \textbf{missile} system.\\
 &  \textsc{tgt:} \ldots ad-hoc delegation des parlaments f{\"u}r das regime in den t{\"u}rkischen \underline{gef{\"a}ngnissen} voll und ganz zustimmen. & \textsc{tgt:} \ldots ad-hoc delegation des parlaments f{\"u}r das regime in den t{\"u}rkischen \textbf{flugwaffen} voll und ganz zustimmen.\\
\midrule
(d) & \textsc{src:} cancellation fees are \underline{not} subject to \underline{judiciary} mitigation.  &  \textsc{src:} cancellation fees are \textbf{generally} subject to \textbf{western} mitigation. \\
 &  \textsc{tgt:} stornogeb{\"u}hren unterliegen \underline{nicht} dem \underline{richterlichen} m{\"a}{\ss}igungsrecht. &  \textsc{tgt:} stornogeb{\"u}hren unterliegen \textbf{allgemein} dem \textbf{westlichen} m{\"a}{\ss}igungsrecht. \\
\bottomrule
\end{tabularx}
\end{table}

In our experiments, two translation data augmentation (TDA) setups are considered: only one word per sentence can be replaced (TDA$_{r=1}$) or multiple words per sentence can be replaced, with the condition that any two replaced words are at least five positions apart (TDA$_{r\ge1}$).
The latter incurs a higher risk of introducing noisy sentences but has the potential to positively affect more rare words within the same amount of augmented data. 
We evaluate both setups in the following section.

\section{Data and experimental setup} \label{tda:exp}

In this section, we describe the experimental settings.
To simulate a low-resource setting we randomly sample 10\% of the English$\leftrightarrow$German WMT15 training data and report results on newstest 2014, 2015, and 2016 \citep{bojar-EtAl:2016:WMT1}. 
For reference, we also provide the result of our baseline system on the full data.
% sentence pairs from the corpora to train the baseline and use for augmentation experiments. 

As NMT system, we use a 4-layer attention-based encoder-decoder model as described in Section~\ref{RNN} trained with hidden dimension 1000, and batch size 80 
for 20 epochs.
NMT models often limit their vocabularies to be the top $K$ most frequent words in each language because of the computationally intensive nature of the softmax.
In all experiments, the NMT vocabulary is limited to the 30K most common words in both languages.
Note that our proposed data augmentation method does not introduce new words to the vocabulary.
In all experiments, we preprocess source and target language data with Byte Pair Encoding (BPE) \citep{sennrich-haddow-birch:2016:P16-12} using 30K merge operations. 
In the non-label-preserving augmentation experiments, BPE is performed after data augmentation. %, which also addresses to some extent the problem of rare words. 

For the LMs needed for data augmentation, we train 2-layer LSTM networks in forward and backward directions on the monolingual data provided for the same task 
(3.5B and 0.9B tokens in English and German, respectively) 
with embedding size 64 and hidden size 128.
We set the rare word threshold $R$ to 100, and top $K$ words to 1000. 
These values are determined heuristically from the training data.
 %and maximum number $N$ of augmentations per rare word to 500.
%
%and the German LM to choose the optimal word translation in context.
Another question we want to investigate is whether rare word substitution is more effective in the source or the target language. 
Therefore in the experiments, we augment the source side in English$\rightarrow$German, and target side in German$\rightarrow$English translation. 
In all experiments, we use the English LM for the rare word substitutions.
Since our first approach is not label-preserving, we only perform augmentation during training and do not alter source sentences during testing.
In Section~\ref{tda:semantic}, we also alter source sentences during testing while preserving labels.

\newcommand{\sigspace}{\rule{3ex}{0pt}}
\newcommand{\sigspacehalf}{\rule{1.6ex}{0pt}}
\begin{table}
\setlength{\tabcolsep}{5pt}
\centering
\caption{\label{bleuTBdeen} Translation performance (BLEU) on German-English WMT test sets (2014, 2015, and 2016) in a simulated low-resource setting. Back-translation refers to the work of~\citet{sennrich-haddow-birch:2016:P16-11}. 
Statistically significant improvements are marked $^\blacktriangleup$ at the $p < .01$ and $^\smalltriangleup$ at the $p < .05$ level, with the first superscript referring to baseline and the second to back-translation$_{1:1}$.}
\begin{tabular}{lcllllll}
\toprule
\multicolumn{2}{c}{}    & \multicolumn{6}{c}{\textbf{De-En}} \\\cline{3-8}
\rule{0pt}{2.5ex}    \textbf{Model}   & \textbf{Data} &  \multicolumn{2}{c}{\bf WMT14} &  \multicolumn{2}{c}{\bf WMT15} &  \multicolumn{2}{c}{\bf WMT16}\\ \midrule     
Full data (ceiling)  & 3.9M &  21.1 &  & 22.0 & & 26.9  &  \\\midrule
Baseline  & 371K & 10.6 & & 11.3 & & 13.1  & \\
Back-translation$_{1:1}$  & 731K & 11.4 & (+0.8)$^{\blacktriangleup}$\sigspacehalf & 12.2 & (+0.9)$^{\blacktriangleup}$\sigspacehalf & 14.6 & (+1.5)$^{\blacktriangleup}$  \\
Back-translation$_{1:3}$  & 1.5M & 11.2 & (+0.6)\sigspace & 11.2 & (--0.1)\sigspace & 13.3 & (+0.2)\sigspace \\\midrule
TDA$_{r=1}$ & 4.5M  &  11.9 & (+1.3)$^{\blacktriangleup,\mhyphen}$ & 13.4 & (+2.1)$^{\blacktriangleup,\blacktriangleup}$ & 15.2 & (+2.1)$^{\blacktriangleup,\blacktriangleup}$  \\
TDA$_{r\ge 1}$ &  6M    &  \textbf{12.6}  & (+2.0)$^{\blacktriangleup,\blacktriangleup}$ & \textbf{13.7} & (+2.4)$^{\blacktriangleup,\blacktriangleup}$ & \textbf{15.4} & (+2.3)$^{\blacktriangleup,\blacktriangleup}$ \\
Oversampling & 6M & 11.9 & (+1.3)$^{\blacktriangleup,\mhyphen}$ & 12.9 & (+1.6)$^{\blacktriangleup,\smalltriangleup}$ & 15.0 & (+1.9)$^{\blacktriangleup,\mhyphen}$  \\
\bottomrule
\end{tabular}
\end{table}

\begin{table}
\setlength{\tabcolsep}{5pt}
\centering
\caption{\label{bleuTBende} Translation performance (BLEU) on English-German WMT test sets (2014, 2015, and 2016) in a simulated low-resource setting. Back-translation refers to the work of~\citet{sennrich-haddow-birch:2016:P16-11}. 
Statistically significant improvements are marked $^\blacktriangleup$ at the $p < .01$ and $^\smalltriangleup$ at the $p < .05$ level, with the first superscript referring to baseline and the second to back-translation$_{1:1}$.}
\begin{tabular}{lcllllll}
\toprule
\multicolumn{2}{c}{}    & \multicolumn{6}{c}{\textbf{En-De}}  \\\cline{3-8}
\rule{0pt}{2.5ex} \textbf{Model}   & \textbf{Data} &  \multicolumn{2}{c}{\bf WMT14} &  \multicolumn{2}{c}{\bf WMT15} &  \multicolumn{2}{c}{\bf WMT16}\\ \midrule     
Full data (ceiling)  & 3.9M &   17.0 & & 18.5 & & 21.7 & \\\midrule
Baseline  & 371K   & 8.2 & & 9.2 & & 11.0 & \\
Back-translation$_{1:1}$  & 731K &  9.0 & (+0.8)$^{\blacktriangleup}$\sigspacehalf & 10.4 & (+1.2)$^{\blacktriangleup}$\sigspacehalf & 12.0 & (+1.0)$^{\blacktriangleup}\sigspacehalf$ \\
Back-translation$_{1:3}$  & 1.5M & 7.8 & (--0.4)\sigspace & 9.4 & (+0.2)\sigspace & 10.7 & (--0.3)\sigspace \\\midrule
TDA$_{r=1}$ & 4.5M  &  10.4 & (+2.2)$^{\blacktriangleup,\blacktriangleup}$  & 11.2 & (+2.0)$^{\blacktriangleup,\blacktriangleup}$ & 13.5 & (+2.5)$^{\blacktriangleup,\blacktriangleup}$  \\
TDA$_{r\ge 1}$ &  6M    & \textbf{10.7} & (+2.5)$^{\blacktriangleup,\blacktriangleup}$  & \textbf{11.5} & (+2.3)$^{\blacktriangleup,\blacktriangleup}$ & \textbf{13.9} & (+2.9)$^{\blacktriangleup,\blacktriangleup}$ \\
Oversampling & 6M & 9.7 & (+1.5)$^{\blacktriangleup,\smalltriangleup}$ & 10.7 & (+1.5)$^{\blacktriangleup,\mhyphen}$ & 12.6 & (+1.6)$^{\blacktriangleup,\mhyphen}$ \\
\bottomrule
\end{tabular}
\end{table}

We also compare our approach to~\citet{sennrich-haddow-birch:2016:P16-11} by back-translating monolingual data and adding it to the parallel training data.
Specifically, we back-translate sentences from the target side that are not included in our low-resource baseline with two settings: keeping a one-to-one ratio of back-translated %pseudo 
versus original data (\mbox{$1:1$}) following the authors' suggestion, or using three times more back-translated data (\mbox{$1:3$}).
% in their paper and kept the ratio of the back-translated and original parallel corpora one-to-one.
We measure translation quality by single-reference case-insensitive BLEU \citep{Papineni2001} computed with the \code{multi-bleu.perl} script from Moses.

\section{Results} \label{tda:results}

In this section, we discuss the results on the translation task and evaluate the effectiveness of our approach in a simulated low-resource NMT scenario.
We repeat the sampling and substitution step iteratively until we reach the desired corpus size for each experiment. 
In our various experiments, we successfully augment between $72\%$ to $81\%$ of targeted rare words. 
All translation results are displayed in Table~\ref{bleuTBdeen} and Table~\ref{bleuTBende} for German$\rightarrow$English and English$\rightarrow$German experiments, respectively. 
%shows the results for English$\leftrightarrow$German translation on different test sets. 

First, we observe that the low-resource baseline performs much worse than the full data system,
re-iterating the importance of sizable training data for NMT. %, but it is well performing proportional to the training size.
Next, we observe that both back-translation and our proposed TDA method significantly improve translation quality. However, TDA obtains the best results overall and significantly outperforms back-translation for all test sets. % (note that increasing the amount of back-translated data does not improve but actually worsen the results obtained)
This is an important finding considering that our method involves only minor modifications to the original training sentences and does not involve any costly translation process, while the back-translation approach augments with novel target sentences.
Improvements are consistent across both translation directions, %Similar trends appear in both translation directions, 
regardless of whether rare word substitutions are applied to the source or to the target side.
We also observe that altering multiple words in a sentence performs slightly better than altering only one word.
This indicates that addressing more rare words is preferable even though the augmented sentences are more likely to be noisy.  

To verify that the gains are actually due to the rare word substitutions and not just to the repetition of part of the training data, we perform a final experiment where each sentence pair selected for augmentation is added to the training data \textit{unchanged} (Oversampling row in Tables~\ref{bleuTBdeen} and~\ref{bleuTBende}). 
Surprisingly, we find that this simple form of sampled data replication outperforms both baseline and back-translation systems,\footnote{Note that this effect cannot be achieved by simply continuing the baseline training for up to 50 epochs.} while TDA\textsubscript{$_{r\ge 1}$} remains the best performing system overall.

\begin{table}[htb!]
\centering
\setlength{\tabcolsep}{4pt}
\caption{Average length of German$\rightarrow$English translation systems, along with the average length of human reference translations (bottom line). Predominantly, we favour longer translations that are close to human reference translations, i.e., models with higher \% Ref ratio.\label{sentlen1} }
\begin{tabular}{lcccc}
\toprule
& \multicolumn{3}{c}{ \textbf{De-En}}   \\
\cline{2-4}
\rule{0pt}{2.5ex}     & \bf WMT14 & \bf WMT15 & \bf WMT16 & \bf  \%Ref    \\
 \midrule
Baseline & 19.9 & 19.2 & 19.9 & 0.88  \\
TDA$_{r= 1}$ & 21.4 & 20.4 & 21.2 & \textbf{0.94} \\
%\rule{0pt}{2.5ex} 
TDA$_{r\ge 1}$ & 21.0 & 20.0 & 20.8 & 0.92  \\
\hdashline
%\rule{0pt}{2.5ex} 
Reference & 23.0 & 22.2 & 21.9 & 1.00  \\
\bottomrule
\end{tabular}
\end{table}
\begin{table}[htb!]
\centering
\setlength{\tabcolsep}{4pt}
\caption{Average length of English$\rightarrow$German translation systems, along with the average length of human reference translations (bottom line). Predominantly, we favour longer translations that are close to human reference translations, i.e., models with higher \% Ref ratio.\label{sentlen2} }
\begin{tabular}{lcccc}
\toprule
&  \multicolumn{3}{c}{ \textbf{En-De} }   \\
\cline{2-4} 
\rule{0pt}{2.5ex}     & \bf WMT14 & \bf WMT15 & \bf WMT16 & \bf  \%Ref    \\
 \midrule
Baseline  & 19.8 & 19.4 & 18.9 & 0.91 \\
TDA$_{r= 1}$ &  20.5 & 20.2 & 19.9 & \textbf{0.95} \\
%\rule{0pt}{2.5ex}  
TDA$_{r\ge 1}$  & 20.7 & 20.0 & 19.8 & \textbf{0.95} \\
\hdashline
%\rule{0pt}{2.5ex} 
Reference & 21.4 & 20.8 & 21.6 & 1.00 \\
\bottomrule
\end{tabular}
\end{table}
We also observe that the system trained on our augmented data tends to generate longer translations, which is favoured. % than the baseline. 
Tables~\ref{sentlen1} and~\ref{sentlen2} provide the average length of the translation outputs of different systems, along with the average length of human reference translations.
Averaging over all test sets and language pairs, the length of translations generated by the baseline is 0.89 of the average reference length, while for TDA\textsubscript{$_{r=1}$} and TDA\textsubscript{$_{r\ge 1}$} it is 0.94 and 0.93, respectively.
%during testing. 
We attribute this effect to the ability of the TDA-trained system to generate translations for rare words that were left untranslated by the baseline system.

\begin{figure}
\centering
\begin{tikzpicture}[shorten >=1pt, node distance=2.5cm,scale=1]
\pgfplotstableread{
X   Gp   Name        generated     rare         notrareanymoreGen   notrareanymoreNotGen 
8   baseline   14    604   3389  0   0
7   TDA test 2114   3389     2082  32   
5   baseline   15    480   2528    0    0
4   TDA   test    1522   2528    1422    100
2   baseline   16    627   3146   0    0
1   TDA   test  2072   3146    1998  70
}\datatable
% 2014 (33). 2015 (29). 2016 (19)
\begin{axis}[
    %axis lines*=left, ymajorgrids,
     width=8.5cm, height=6.5cm,
     xmin=0,
     %xbar stacked,
     xbar,
     xmax = 3500, 
     bar width=13pt,
     bar shift=0pt,
     ytick=data,
     reverse legend,
         xtick={1000, 2000, 3000},
     %axis y line*=right,
     yticklabels from table={\datatable}{Gp},
         ytick style={draw=none},
     %ylabel = test2014 \hspace{1em} test2015  \hspace{1em} test2016,
     legend style={at={(0.42,-0.15)},anchor=north, draw=none, /tikz/every even column/.append style={column sep=5pt}}, 
     legend image code/.code={%
             \draw[#1] (0cm,-0.1cm) rectangle (0.3cm,0.1cm);
        },
     legend cell align=left
]

\addplot [fill=blu, fill opacity=1] table [y=X, x=rare] {\datatable};  \addlegendentry{{Words in $V_R \cap V_{ref}$ not generated during translation}} %Rare words that are generated during translation}}
\addplot [fill=gr, fill opacity=1] table [y=X, x=generated] {\datatable}; \addlegendentry{{Words in $V_R \cap V_{ref}$ generated during translation }}
\end{axis}

\begin{axis}[
    %axis lines*=left, ymajorgrids,
    width=8.5cm, height=6.5cm,
    xmin=0,
    xbar stacked,
    xmax = 3000,
    bar width=13pt,
    ytick=data,
    axis y line*=right,
    y tick label style={rotate=90} ,
     yticklabels from table={\datatable}{Name},
    xticklabels=\empty,
    xtick=\empty,
    ytick style={draw=none},
     legend style={at={(0.35,-0.345)},anchor=north, draw=none, /tikz/every even column/.append style={column sep=5pt}}, 
     legend image code/.code={%
             \draw [pattern=north east lines](0cm,-0.1cm) rectangle (0.3cm,0.1cm);
        },
     legend cell align=left,
]
\addplot [fill=red, fill opacity=0,forget plot] table [y=X, x=notrareanymoreNotGen] {\datatable};
\addplot [pattern=north east lines] table [y=X, x=notrareanymoreGen] {\datatable}; \addlegendentry{Words in $V_R \cap V_{ref}$ affected by augmentation}

\end{axis}
\end{tikzpicture}
\caption{Effect of TDA on the number of unique rare words generated during De$\rightarrow$En translation. $V_R$ is the set of rare words targeted by TDA\textsubscript{$_{r\ge 1}$} and $V_{ref}$ the reference translation vocabulary.} % The hatched pattern indicates words that are not rare anymore i.e., were augmented in the training data.}
\label{rarewordfreqs}
\end{figure}
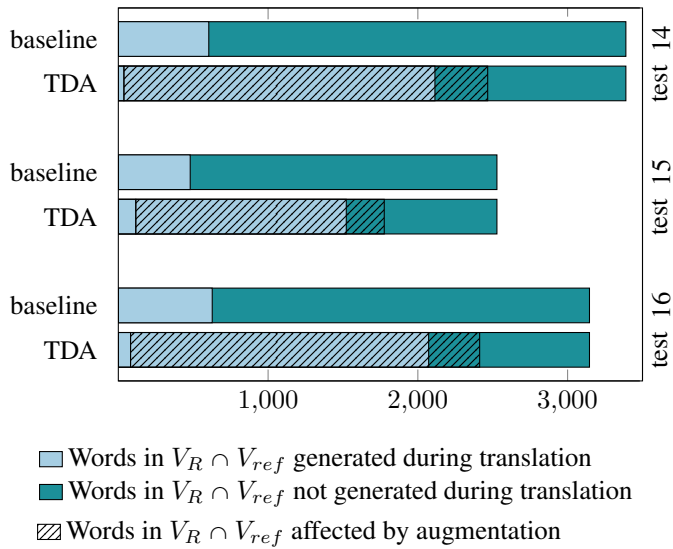

\section{Further analysis}  \label{tda:analysis} 

In this section, we further analyze our findings and discuss the results of our proposed models.
Our goal is to understand the impact of the introduced diverse contexts on the learning capabilities of the neural translation model. 

\begin{table}[p]
%\begin{minipage}[c]{0.48\linewidth}
\rotatebox{90}{
\begin{minipage}{\textheight}
\begin{center}
%\begin{tabularx}{\linewidth}{@{\ }l @{\ \ \ }X@{\ }}
\caption{An example from WMT14 illustrating the effect of augmenting rare words on generation at test time. 
The translation of the baseline does not include the rare word \textit{centimetres}, however, the translation of our TDA model generates the rare word and produces a more fluent sentence. Instances of the augmentation of the word \textit{centimetres} in training data are also provided. \label{transex}}
\begin{tabularx}{0.8\textheight}{l@{\hskip 0.02in}X@{\hskip 0.2in}X}
% \multirow{2}{*}{(a)} & \textsc{src:}  &  \textsc{src:}  \\
 %& \textsc{tgt:}  &  \textsc{tgt:}  \\
 %\multicolumn{3}{l}{ } \\
  \multicolumn{3}{l}{\textit{Example from WMT14}}  \\
 \toprule
\textsc{src} & \multicolumn{2}{l}{der tunnel hat einen querschnitt von 1,20 meter h{\"o}he und 90 \underline{zentimeter} breite.}\\ 
\textsc{ref}   & \multicolumn{2}{l}{the tunnel has a cross-section measuring 1.20 metres high and 90 \underline{centimetres} across.} \\
%\specialrule{.05em}{0em}{0em} 
Baseline  & \multicolumn{2}{l}{the wine consists of about 1,20 m and 90 of the canal.} \\
TDA\textsubscript{$_{r\ge 1}$}  & \multicolumn{2}{l}{the tunnel has a \texttt{unk} measuring meters 1.20 metres high and 90 \textbf{centimetres} wide.} \\
\bottomrule
% & $\bullet$ various shopping venues , countless bars and restaurants and a nightclub are to be found in the vicinity ( 100 [m / \textbf{centimetres}]  radius ) .\
\noalign{\vskip 5mm} 
 \multicolumn{3}{l}{ \textit{Examples from the training data displaying augmentations for the word \textbf{`centimetres'}}} \\
\toprule
& Original data & Augmented data \\ 
\midrule
(a) & the average speed of cars and buses is therefore around 20 \underline{kilometres}  per hour.  & the average speed of cars and buses is therefore around 20 \textbf{centimetres} per hour. \\
 \hdashline
(b) & grab crane in special terminals for handling capacities of up to 1,800 \underline{tonnes} per hour.  &  grab crane in special terminals for handling capacities of up to 1,800 \textbf{centimetres} per hour.\\ 
  \hdashline
(c) &  all suites and rooms are very spacious and measure between 50 and 70 \underline{m}. & all suites and rooms are very spacious and measure between 50 and 70 \textbf{centimetres}. \\ 
  \hdashline
(d) &   all we have to do is lower the speed limit everywhere to five \underline{kilometers} per hour. &  all we have to do is lower the speed limit everywhere to five \textbf{centimetres} per hour.\\ 
  \hdashline
(e) &  just 9.5 litres of water per minute flow through the innovative \texttt{unk} shower system, whereas the hansgrohe \texttt{unk} 85 green hand spray manages with only six \underline{litres} per minute. & just 9.5 litres of water per minute flow through the innovative \texttt{unk} shower system, whereas the hansgrohe \texttt{unk} 85 green hand spray manages with only six \textbf{centimetres} per minute.  \\ 
\bottomrule
\end{tabularx}
\end{center}
\end{minipage}
}
\end{table}

\subsection{Target words} 
A desired effect of our method is to increase the number of correct rare words generated by the NMT system at test time. 
To illustrate the impact of augmenting the training data by creating contexts for rare words on the \textit{target} side, Table~\ref{transex} provides an example for German$\rightarrow$English translation. 

We see that the baseline model is not able to generate the rare word \textit{`centimetres'} as a correct translation of the German word \textit{`zentimeter'}. However, this word is not rare in the training data of the TDA\textsubscript{$_{r\ge 1}$} model after our augmentation and is generated during translation. Table~\ref{transex} also provides several instances of augmented training sentences targeting the word \textit{`centimetres'}. Note that even though some augmented sentences are rather unusual (e.g., \textit{`the speed limit is five centimetres per hour'}), the NMT system still benefits from the new context for the rare word and is able to generate it during testing.

Figure~\ref{rarewordfreqs} demonstrates that this is indeed the case for many words: the number of rare words occurring in the reference translation ($V_R \cap V_{ref}$) is three times larger in the TDA system output than in the baseline output. 
%Figure~\ref{rarewordfreqs} shows the number of words generated during translations in different test sets for the baseline and TDA\textsubscript{multiple} system. 
One can also see that this increase is a direct effect of TDA
as most of the rare words are not `rare' anymore in the augmented data, i.e., they were augmented sufficiently often to occur more than 100 times (see the hatched pattern in Figure~\ref{rarewordfreqs}).
%We observe that the generation of rare words increases in our model in comparison with the baseline.
%This indicates that by having more instances of words in the training data, the model tends to use these words more during the translation process.
Note that in our experiments, we did not use any information from the evaluation sets.

\subsection{Source words}  

To gauge the impact of augmenting the contexts for rare words on the \textit{source} side, we examine normalized attention scores of these words before and after augmentation. When translating English$\rightarrow$German 
with our TDA model, the attention scores for rare words on the source side are on average 8.8\% higher than when translating with the baseline model. This suggests that having more accurate representations of rare words increases the model's confidence to attend to these words when encountered during test time.

\subsection{Negative examples}  

Table~\ref{tdaexampleswrong} provides examples of cases where augmentation results in incorrect sentences. 
In the first example, the sentence is ungrammatical after substitution (\textit{`of / yearly'}), which can be the result of choosing substitutions with low probabilities from the English LM $\topk$ suggestions.

\begin{table}[p]
\rotatebox{90}{
\begin{minipage}{\textheight}
\begin{center}
\caption{Examples of incorrectly augmented data with highlighted \underline{Original} words and \textbf{substituted} words. \label{tdaexampleswrong}}
\begin{tabularx}{0.8\linewidth}{l@{\hskip 0.07in}X@{\hskip 0.2in}X}
 \toprule
  & Original sentence pair & Synthetic sentence pair \\
\midrule
% \multirow{2}{*}{(a)} & \textsc{src:}  &  \textsc{src:}  \\
 %& \textsc{tgt:}  &  \textsc{tgt:}  \\
 %\multicolumn{3}{l}{ } \\
(a) & \textsc{src:} registered users will receive the \texttt{unk} newsletter free \underline{of} charge.&  \textsc{src:}  registered users will receive the \texttt{unk} newsletter free \textbf{yearly} charge. \\
 & \textsc{tgt:} registrierte user erhalten zudem regelm{\"a}{\ss}ig \underline{den} markenticker newsletter.  &  \textsc{tgt:}  registrierte user erhalten zudem regelm{\"a}{\ss}ig \textbf{j{\"a}hrlich} markenticker newsletter.  \\
\noalign{\vskip 2mm}   \hdashline
 \textsc{Problem:} & \multicolumn{2}{l}{ \textit{Substitution results in syntactically incorrect source and target sentences.} } \\
 \midrule
(b) & \textsc{src:}  the personal contact is \underline{essential} to us. &  \textsc{src:}  the personal contact is \textbf{entrusted} to us. \\
                      & \textsc{tgt:} pers{\"o}nliche kontakt ist uns sehr \underline{wichtig}.  &  \textsc{tgt:} pers{\"o}nliche kontakt ist uns sehr \textbf{betraut}. \\
\noalign{\vskip 2mm}   \hdashline
 \textsc{Problem:} & \multicolumn{2}{l}{ \textit{The German sentence is grammatically incorrect.} } \\
  \midrule
(c) & \textsc{src:} \texttt{unk}  \texttt{unk} wishes you very \underline{pleasant} holiday. &  \textsc{src:}  \texttt{unk}  \texttt{unk} wishes you very \textbf{crazy} holiday. \\
 & \textsc{tgt:}  \texttt{unk} \texttt{unk} w{\"u}nscht ihnen einen \underline{erholsamen} urlaub.  &  \textsc{tgt:} \texttt{unk}  \texttt{unk} w{\"u}nscht ihnen einen \textbf{verr{\"u}ckt} urlaub. \\
\noalign{\vskip 2mm}    \hdashline
 \textsc{Problem:} &  \multicolumn{2}{l}{\textit {German substitution has the wrong inflection.}} \\
 \midrule
(d) & \textsc{src:} \underline{consumers} are currently being \underline{deliberately} misled. &  \textsc{src:}  \textbf{schools} are currently being \textbf{widely} misled. \\
 & \textsc{tgt:}  \underline{die} verbraucher werden gegenw{\"a}rtig \underline{bewusst} get{\"a}uscht. &  \textsc{tgt:}  \textbf{schulen} verbraucher werden gegenw{\"a}rtig \textbf{weithin} get{\"a}uscht. \\
\noalign{\vskip 2mm}    \hdashline
 \textsc{Problem:} & \multicolumn{2}{l}{\textit{Substituted with the wrong German word because of wrong alignment.} } \\
\bottomrule
\end{tabularx}
\end{center}
\end{minipage}
}
\end{table}

Errors can also occur during translation selection, as in the second example where \textit{`betraut'} is an acceptable translation of \textit{`entrusted'} but would require a rephrasing of the German sentence to be grammatically correct.
Problems of this kind can be attributed to the German LM, but also to the lack of a more suitable translation in the lexicon extracted from the bitext.
Interestingly, this noise seems to affect NMT only to a limited extent.

\subsection{Word segmentation} 
BPE \citep{sennrich-haddow-birch:2016:P16-12} is an essential preprocessing step in NMT to address the problem of rare and unknown words in the training data and we use it in all experiments. 
Although crucial, it is not very effective in a low-resource setting \citep{ngo-etal-2019-overcoming}.
We suspect that this is caused by the scarcity of data, which results in inaccurate word splits with possibly rare subword units.
This can be observed in the example in Table~\ref{transex}. 
In the experiments, the English and German words \textit{`centi$\mid$metres'} and \textit{`zenti$\mid$meter'} are both split into two subword units. Still, the baseline model fails to translate it correctly.
This further stresses the importance of data augmentation with diverse contexts. 
Even though BPE segmentation is successful in rare word translations, our proposed approach yields additional improvements.

\section{Meaning-Preserving augmentation}  \label{tda:semantic}

In the previous section, we proposed a model with a weak notion of label preservation that allows modifying both source and target sentences at the same time as long as they remain translations of each other. 
As a result, we alter and augment the training data, and the test data remains unchanged. 
This approach improves translation quality by better translating rare words because of the additional contexts during training. 
However, it does not address the problem of translating out-of-vocabulary (OOV) words during testing.
To specifically target OOV words at test time, we propose a stronger notion of label preservation to only alter source sentences with paraphrases. 
Note that at test time, we only have access to source sentences and keep the reference sentences unchanged. 

In this section, we investigate how we can benefit from external lexical resources to address the problem of translating unknown words. 
We define OOV as words not listed in the 30k most common words in source and target vocabulary. 
While BPE is an effective approach in addressing the OOV translation problem, we do not use it in these experiments.
To benefit from external knowledge resources, we substitute OOVs with synonym words obtained from these resources that exist in our vocabulary. 
We experiment with three different resources to alter source sentences with paraphrases:

\subparagraph{PPDB} proposed by~\citet{ganitkevitch-etal-2013-ppdb}. This Paraphrase Database is an automatically extracted database from parallel corpora containing millions of paraphrases in multiple languages.

\subparagraph{Wordnet} proposed by~\citet{miller1995wordnet}. WordNet is a manually created large lexical database of English and includes relations between words and groups of cognitive synonyms. We use GermaNet \citep{hamp-feldweg-1997-germanet,HENRICH10.264} which is a lexical-semantic resource similar to Wordnet for the German language.

\subparagraph{CBOW} embeddings proposed by~\citet{mikolov2013distributed}. We use the Continuous Bag of Words (CBOW) model to identify the words most similar to each OOV word and interpret that as the synonym. %In contrast to the previous two resources, this method is trained using context. % and as a result provides context-aware synonyms. 
\subparagraph{HTLE} embeddings proposed in Chapter~\ref{chapter:research-01}. We use the multiple topic-sensitive representations per word to identify synonyms according to the context of the word. 
In contrast to the previous resources, this method provides context-sensitive synonyms which means the same word in different contexts has different synonym substitutions. 

\gap 

We substitute the OOV words in source sentences with synonyms that already exist in the NMT vocabulary. % acquired from the parallel training data.  
An example of the substitution is shown in Table~\ref{augwnexample} where the original word `\textit{fateful}' is OOV and is replaced by the \texttt{unk} symbol in the original training data.
%
%Here the word \textit{fateful} does not exist in the limited source vocabulary, and 
Each paraphrase resource suggests a substitution for the target word in the sentence. % that are different at times. 
%PPDB and Wordnet have fix suggestions for each word.
%However, the CBOW embedding uses the local context window and as a result, provides a list of most similar words specific to that context. 
We also experiment with targeting low-frequency words in the test data. 
Similar to OOVs, we substitute words that are rare in our training data (frequencies less than 100) with synonyms during test time. 

\begin{table}[htb!]
\small
\centering
\begin{tabular}{l l}
\toprule
 {original src} & {He said Lamb made the \textit{fateful} 911 call sometime after that.} \\ 
 {NMT input} & {He said Lamb made the \texttt{[unk]} 911 call sometime after that.} \\ \midrule
 +PPDB   & {He said Lamb made the \textbf{disastrous} 911 call sometime after that.}  \\
+Wordnet &  {He said Lamb made the \textbf{fatal} 911 call sometime after that.}   \\
+CBOW &  {He said Lamb made the \textbf{tragic} 911 call sometime after that.}  \\
+HTLE &  {He said Lamb made the \textbf{critical} 911 call sometime after that.}  \\
\bottomrule
\end{tabular}
 \caption{Examples of paraphrase modification. The out-of-vocabulary word \textit{fateful} in the source sentence is substituted with synonyms obtained from different lexicon resources. \label{augwnexample}}
\end{table}

Tables~\ref{semdeen} and~\ref{semende} provide the results for the translation of German$\rightarrow$English and English$\rightarrow$German, respectively.  
Note that the HTLE embeddings are available only in English and so we only use this approach in the English$\rightarrow$German translation experiments to augment English sentences on the source side.
Overall the results show improvements over the baseline. 
BLEU scores using PPDB and Wordnet are very similar for all experiments. %and are very close to the baseline results. 
Improvements using CBOW and HTLE are also similar, however, HTLE is the most effective method out of the four approaches.
This indicates, to a certain degree, the importance of context-aware substitutions in data augmentation.

\begin{table}[htb!]
\centering
\caption{Translation performance (BLEU) on German-English WMT news test sets (2014, 2015, and 2016). \texttt{OOV} signifies out-of-vocabulary words. \texttt{rare} words are selected with the frequency threshold of 100. \label{semdeen}}
\begin{tabular}{lllcccccc}
\toprule
   & & & \multicolumn{3}{c}{\textbf{De-En}}  \\\cline{4-6} 
\rule{0pt}{2.5ex} \textbf{Model} & \textbf{+lexical DB} & \textbf{Target}  & \bf WMT14 &  \bf WMT15 &  \bf WMT16 \\      
 \midrule
Baseline &  &  &    19.3 &   20.1 &  24.9   \\ \midrule
\multirow{4}{*}{Subs}  & {PPDB}  & \texttt{OOV} & 21.2 & 22.2 & 26.9 \\
& {GermaNet}   & \texttt{OOV}  & 20.3 & 21.9 & 25.2 \\
& {CBOW}   & \texttt{OOV}  & 21.2 & 22.4& 27.0 \\
\midrule
\multirow{4}{*}{Subs} & {PPDB}   & \texttt{OOV} + \texttt{rare}   & 21.3 & 22.3 & 26.9 \\
& {GermaNet}   & \texttt{OOV} + \texttt{rare}     & 20.5 & 22.0 & 25.4 \\
& {CBOW}   & \texttt{OOV} + \texttt{rare}  & \bf 21.4 & \bf 22.5& \bf 27.2 \\
\bottomrule
\end{tabular}
\end{table}
\begin{table}[htb!]
\centering
\caption{Translation performance (BLEU) on English-German WMT news test sets (2014, 2015, and 2016). \texttt{OOV} signifies out-of-vocabulary words. \texttt{rare} words are selected with the frequency threshold of 100. \label{semende}}
\begin{tabular}{lllccc}
\toprule
   & & & \multicolumn{3}{c}{\textbf{En-De}}   \\\cline{4-6} 
\rule{0pt}{2.5ex} \textbf{Model}  & \textbf{+lexical DB} & \textbf{Target} &  \bf WMT14 &  \bf WMT15 &  \bf WMT16 \\      
 \midrule
Baseline  & & & 15.9 &  17.6 &  20.0       \\ \midrule
\multirow{4}{*}{Subs} & {PPDB} & \texttt{OOV}     & 17.2 &  	18.5   	 & 21.8   \\
& {Wordnet}& \texttt{OOV}      & 17.2 &  	18.5   	& 21.7 \\
& {CBOW}& \texttt{OOV}      & 17.3 &  	18.6   	& 22.0 \\
& {HTLE}   & \texttt{OOV}  & 17.5 & 18.6 & 22.2 \\ \midrule
\multirow{4}{*}{Subs} & {PPDB}  & \texttt{OOV} + \texttt{rare}    & 17.2 &  	18.7   	 & 21.9  \\
& {Wordnet} & \texttt{OOV} + \texttt{rare}     & 17.1 &  	18.5   	& 21.8  \\
& {CBOW} & \texttt{OOV} + \texttt{rare}     & 17.4 &  	18.7   	& 22.2  \\
& {HTLE}   & \texttt{OOV} + \texttt{rare}  & \bf 17.9 &  	\bf 19.1   	& \bf 22.5  \\
\bottomrule
\end{tabular}
\end{table}

Finally, we look into the rate a neural model generates the \texttt{unk} symbol before and after augmentation.
When a source sentence contains several rare and OOV words, the translation model tends to use the \texttt{unk} symbol to represent these words. 
As a result, the model performs poorly and produces several \texttt{unk} symbols in the translation output.
We investigate the fluency of the translation outputs by observing the rate of the generation of the \texttt{unk} symbol.
We observe that in our experiments the number of \texttt{unk} symbols generated in the translation output drops. 
Table ~\ref{unkaveaug} provides statistics on the generation of  \texttt{unk} token. 
Surprisingly, the significant differences in the number of \texttt{unk} symbols in the translation outputs are not entirely reflected in the BLEU scores. 

\begin{table}
\centering
\caption{The impact of paraphrase augmentation on the generation of \texttt{unk} tokens in the translation output. Reductions are computed in comparison with the baseline model. Lower number of \texttt{unk}s is better.  \label{unkaveaug} }
\begin{tabular}{l|c|cccc}
%\toprule
& Baseline & PPDB & Wordnet & CBOW & HTLE \\ \midrule
Number of \texttt{unk}s & 4931 & 4851 & 4857 & 3018 & \bf 3003 \\
Reduction in number of \texttt{unk}s & - & 1.62$\%$ & 1.5$\%$ & 38.8$\%$ & \bf 39.1$\%$ \\
%\bottomrule
\end{tabular}
\end{table}

\section{Conclusion} \label{tda:conc}

In this chapter, we investigated the impact of diverse contexts on the translation of rare words.
The quality of an NMT system depends substantially on the availability of sizable parallel corpora, which is only available for a limited number of languages and domains.
%For low-resource language pairs, this is not the case, and it results in poor translation quality. 
The translation is particularly erroneous for low-frequency words; with only a few instances in the training data, the model has difficulties learning to translate these words. 
While the challenges of translating low-resource language pairs have been studied extensively, the impact of artificially generated contexts on the translation quality of words has hardly been studied. 
We addressed this issue in this chapter and investigated the effect of additional context in learning word translations, by asking:
 
\begin{enumerate}[label=\textbf{RQ2.\arabic* },wide = 0pt, leftmargin=2em]
\setlength\itemsep{1em}
\item \acl{rq:tda1}

\medskip

\noindent  Our experiments showed that by providing more diverse contexts for rare words, we improve the estimation of the model and subsequently increase the number of times the model generates these words correctly. 
We have proposed an effective approach to augment the training data of neural machine translation for low-resource language pairs. 
We generated new sentence pairs containing rare words in new contexts by leveraging language models trained on large amounts of monolingual data.
Our approach augments the data by diversifying the sentences of the parallel corpora, changing both source and target sentences.
We showed that this approach leads to generating rare words more often during translation and thus improves translation quality. 
%Our analysis further stresses that poor translation quality of rare words is a result of a lack of diverse training examples. 
We observed substantial improvements in simulated low-resource English$\rightarrow$German and German$\rightarrow$English settings.

 \noindent Having observed the impact of additional \textit{training} data on the translation of rare words, we looked into how we can perform augmentation during \textit{testing} and asked: 

\item \acl{rq:tda2}

\medskip

\noindent To answer this question, we first explained why our previously proposed method is not viable at test time.  
We do not have access to the reference translations during inference and as a result we only accept alterations to the source sentence that keep the meaning of the sentence unchanged.
We introduced a substitution method to replace both rare and out-of-vocabulary words in the source sentences with their paraphrases, using several knowledge resources.
We gained improvements in BLEU scores over the baselines.
However more interestingly, we significantly reduced the number of \texttt{unk} in the target output.

\end{enumerate}

 \noindent In summary, our extensive studies of the rare word translation challenge partially answered the following question:

\paragraph{Research Question 2:} \acl{rq:tdabt} 

\medskip

 \noindent To answer this question, we examined the effect of the availability of data, and rare words in particular, on translation quality.
We found that translating and generating rare words is a challenging task for NMT models. 
With the proposed data augmentation approach, we diversified and increased the contexts of rare words.
We improved the translation quality by augmenting the data with these new sentence pairs.

\medskip

 \noindent In this chapter, we looked into the long tail of words where statistical models have difficulties learning. 
We continue our investigation into this question in the next chapter by examining whether the trained model itself can identify words that will benefit from the addition of diverse contexts.

\chapter{Data Augmentation Based on Model Failure}

\label{chapter:research-03}

\section{Introduction and research questions}

In the previous chapter, we have observed that the availability of large-scale training data is essential for sequence-to-sequence neural models to achieve good translation quality.
We have shown that neural machine translation models benefit from data augmentation for rare words.
By using a combination of parallel and synthetic data, neural models learn to translate more effectively.
In this chapter, we study a more general approach to augmentation by identifying and targeting words that are most difficult to learn by the model.
%where the model identifies the words it has difficulty learning.
%We leverage target monolingual data and target these words to provide additional context for. 

Previous approaches have focused on leveraging monolingual data, which is available in much larger quantities than parallel data \citep{Lambert:2011:ITM:2132960.2132997}. 
%\citet{sennrich-haddow-birch:2016:P16-11} generate synthetic data by back-translating sentences randomly sampled from monolingual data using a reverse translation model.
\citet{sennrich-haddow-birch:2016:P16-11} proposed {back-translation} of monolingual target sentences to the source language and adding the synthetic sentences to the parallel data (discussed in Section~\ref{bgbtref}).
In this approach, a reverse model trained on parallel data is used to translate sentences randomly sampled from target-side monolingual data into the source language.
This synthetic parallel data is then used in combination with the actual parallel data to re-train the model.
This approach yields state-of-the-art results even when large amounts of parallel data are already available and has become common practice in NMT \citep{2017arXiv170800726S,2017arXiv170704499G,2017arXiv171107893H}.
Generally speaking, back-translation mitigates the problem of overfitting and fluency by exploiting additional data in the target language \citep{sennrich-haddow-birch:2016:P16-11}.

An important question for this technique is how to select the monolingual data in the target language that is to be back-translated into the source language in order to obtain the best possible translation quality. 
Earlier studies have explored to what extent data selection of parallel corpora can benefit translation quality \citep{D11-1033,vanderwees-bisazza-monz:2017:EMNLP2017}, but such selection techniques have not been investigated in the specific context of back-translation. 

Motivated by the success of back-translation in NMT, we investigate in this chapter whether back-translation can benefit from a more insightful data selection approach, i.e., \textit{targeted} sampling.
In particular, we explore what words benefit from the generation of additional context and how this information can help us develop more creative data selection methods and improve translation quality.
To this end, we ask: 

\paragraph{Research Question 2:} \acl{rq:tdabt} 

\medskip

 \noindent We partially examined this research question in the previous chapter. 
 In this chapter, we investigate whether model failures are good indicators for choosing new contexts.
 Methods similar to back-translation have a trained model at their disposal and use it to generate new contexts.
 We conduct a series of analyses on the learning process of neural translation models with synthetic data.
 Signals from a pre-trained model can show us where the model is struggling, which can be beneficial in selecting data for augmentation. 
 So we ask:
 %While back-translation has been shown to be very effective, it is not exactly clear why it helps. 
 
\begin{enumerate}[label=\textbf{RQ2.\arabic* },wide = 0pt, leftmargin=2em]
\setlength\itemsep{1em}
 \setcounter{enumi}{2}
\item \acl{rq:bt1}

\medskip

\noindent We review the influence of additional contexts generated by the back-translation approach on the learning process of NMT. 
Observing the loss function of the model during training, we study the changes in the prediction of every word in the vocabulary. 
Our findings show that it is mostly words that are difficult to predict in the target language that benefit from additional back-translated data. 
These low-confidence words have high prediction loss during training when the translation model converges.
Leveraging this information, we explore alternatives to random sampling to specifically target these words, thus asking:

\item \acl{rq:bt2}

\medskip

\noindent We propose alternatives to the random sampling approach with a focus on increasing occurrences of low-confidence words in the training data.
Our proposed approach is twofold: (i) identifying difficult \textit{words} and sampling to increase occurrences of these words, and (ii) identifying \textit{contexts} in which these words are difficult to predict and sample sentences similar to the difficult contexts.
We then analyze various ways of identifying difficult words and augmenting the training data.  
Our investigations show that targeted sampling of monolingual data improves the translation quality of NMT models compared to standard back-translation. 

\end{enumerate}
 
\paragraph{Organization.} This chapter is organized as follows: In Section~\ref{btrelated}, we provide an overview of existing work on data selection for machine translation. 
Next, in Section~\ref{btdata}, we describe our data and experimental setup.
We study different aspects of the back-translation method in Section~\ref{btbtanalysis}.
In Section~\ref{bttokpred}, we present a more in-depth analysis of the impact of the new contexts generated by back-translation on the prediction power of our NMT model. 
Next in Section~\ref{bttarget}, we propose a targeted sampling approach for selecting new contexts for the training data. 
We also provide experimental results and analyze the impact of different sampling methods. 
In Section~\ref{btcontextu}, we propose an alternative data selection approach with context-aware sampling and provide qualitative results in Section~\ref{btanalysisagain}. 
Finally, we discuss the conclusions and implications of this work in Section~\ref{btconc}.

\section{Related work} \label{btrelated}

In this section, we provide an overview of work related to this chapter on data selection methods in MT.

\subsection{Data selection in machine translation}

Before the emergence of neural models, several previous studies in PBMT have focused on choosing which portion of the parallel corpora to use for training \citep{moore-lewis-2010-intelligent,wang-etal-2013-edit}. 
For instance, \citet{D11-1033} computed cross-entropy scores for sentence pairs using a 4-gram language model trained on a pseudo in-domain corpus.
They sorted the sentence pairs based on this criterion and augmented the training data with the $\topk$ sentence pairs that are most relevant to the target domain.

With the development of neural models in machine translation, most works greedily use all available training data for a given language pair. 
However, it is unlikely that all data is equally useful for creating the best-performing system.
Additionally, when the domain of the training and testing data is different, it is essential to carefully select the portion of the data that is most helpful for training. 
Various data selection methods have been proposed to address the problem of domain adaptation in MT \citep{silva-etal-2018-extracting,wang-etal-2019-dynamically}.  
These methods aim to reduce the model size and result in shorter training times by using a subset of the available data while maintaining high performance.
For instance, \citet{vanderwees-bisazza-monz:2017:EMNLP2017} introduced dynamic data selection, where they vary the selected data subsets during each training epoch. The ranking criteria are based on bilingual cross-entropy differences similar to \citet{D11-1033}.
They significantly reduce the training data size by only using parts of the data which are most relevant to the translation task at hand. 

\citet{2019arXiv190607808P} use two transductive data selection methods, infrequent n-gram recovery and feature decay algorithms, to select a subset of sentence pairs from synthetic training data. 
This approach ensures that the selected sentence pairs share n-grams with the test set.
\citet{DBLP:journals/corr/abs-1909-03750} investigate whether combining the two paradigms of NMT and PBMT in generating synthetic data contributes to translation quality. 
In their proposed approach, they randomly select source sentences from the PBMT synthetic data and the NMT synthetic data with a one-to-one ratio.
They show that mixing PBMT and NMT back-translated data further improves over using each type of data alone.

\section{Data and experimental setup} \label{btdata}

In this chapter, we conduct several experiments to evaluate the impact of synthetic context on translation quality. 
For the translation experiments, we use the English$\leftrightarrow$German WMT17 training data and report results on WMT news test sets 2014, 2015, 2016, and 2017 \citep{bojar-EtAl:2017:WMT1}. 
As NMT system, we use a 2-layer attention-based encoder-decoder LSTM model described in Section~\ref{RNN} implemented in OpenNMT \citep{2017opennmt}.
We train this model with embedding size 512, hidden dimension size 1024, and batch size of 64 sentences.
We preprocess the training data with joint Byte Pair Encoding (BPE) using 32K merge operations \citep{sennrich-haddow-birch:2016:P16-12}. 

We compare the results to~\citet{sennrich-haddow-birch:2016:P16-11} by back-translating random sentences from the monolingual data and combine them with the parallel training data.
To lessen the arbitrary effect of random sampling, we perform random selection and re-training three times and report the averaged outcomes for the three models.
In all experiments, the sentence pairs are shuffled before each epoch.
We measure translation quality by single-reference case-sensitive {BLEU} \citep{Papineni2001} computed with the \code{multi-bleu.perl} script from Moses.

\begin{figure}[htb!]
\begin{center}
\includegraphics[width=\textwidth]{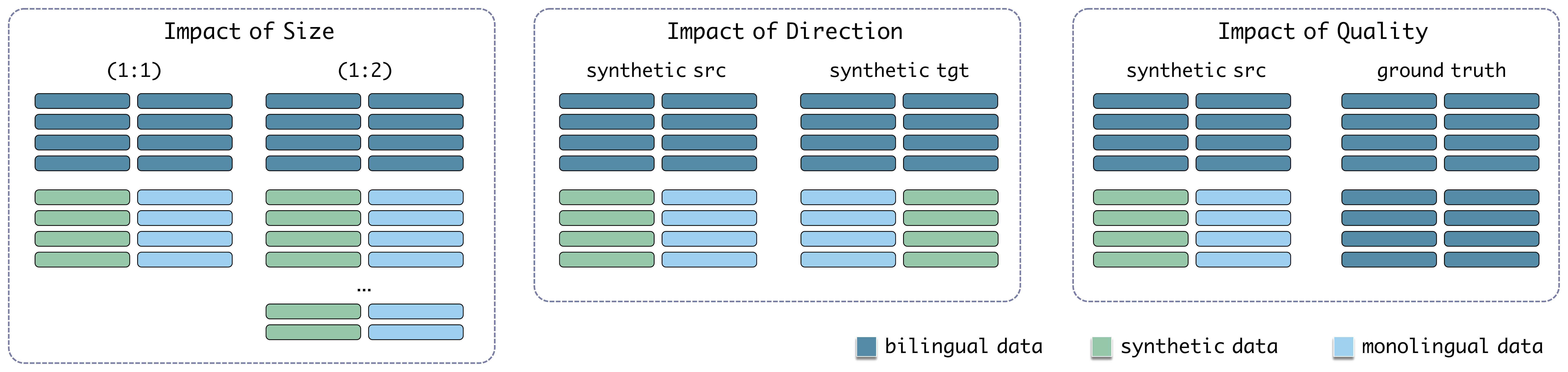}
\caption{Illustration of three set of experiments analyzing different impacts of synthetic data as additional training data: \textit{size} (left), \textit{direction} (middle), and \textit{quality} (right). %Green indicates synthetic data and blue indicates real data generated manually.
}
\label{btmets}
\end{center}
\end{figure}

\section{Analyzing back-translation with random sampling} \label{btbtanalysis}

In this section, we investigate different aspects and modeling challenges of integrating the back-translation method into the NMT pipeline.
We are interested in investigating the impact of synthetic data on translation quality with random sampling augmentation (see Figure~\ref{btmets}).

\subsection{Impact of synthetic data size}

\begin{figure}[htb!]
\begin{center}
\includegraphics[width=0.8\textwidth]{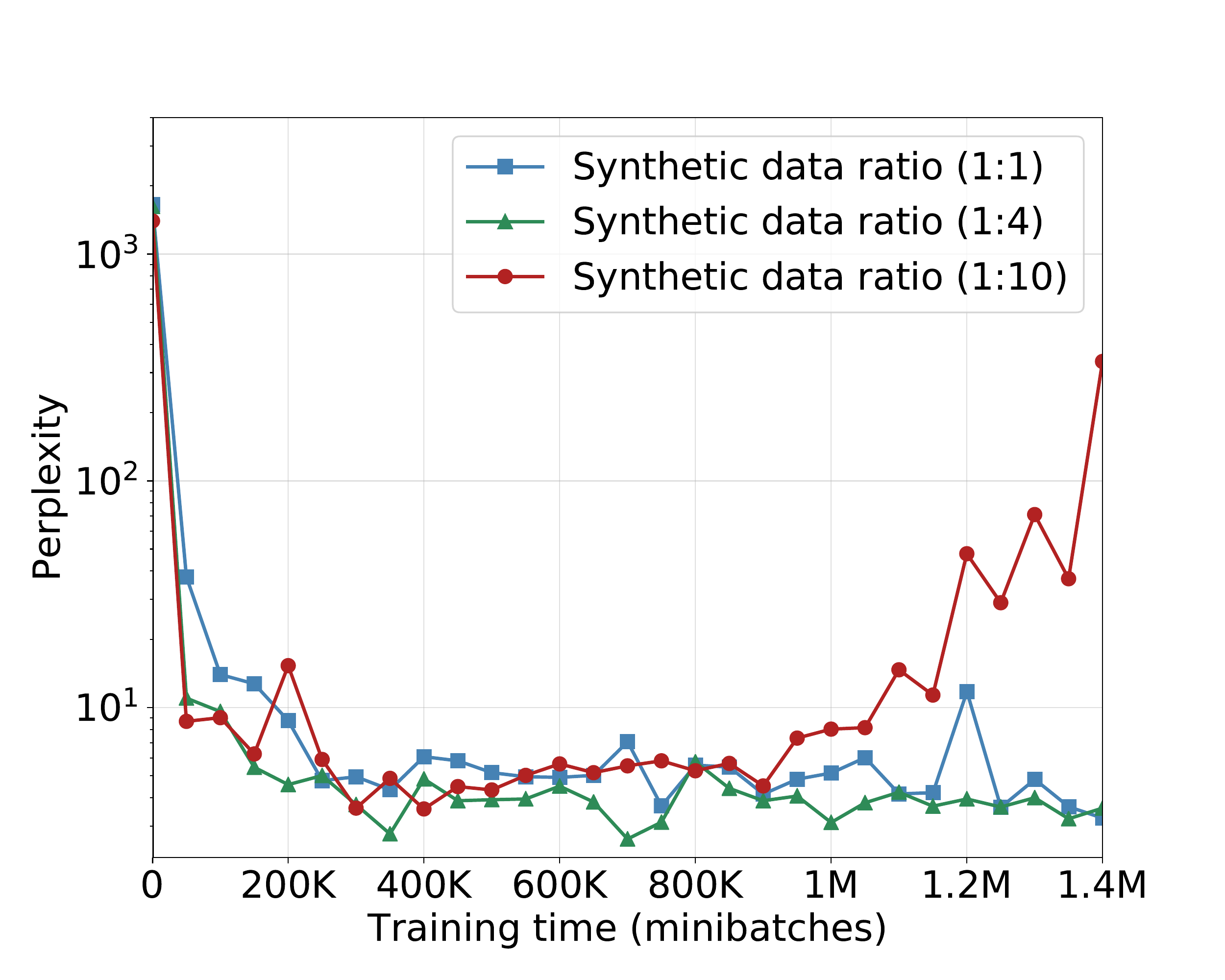}
\caption{Training plots for systems with different ratios of  (\mbox{$real:syn$}) training data, showing perplexity on development set.}
\label{ppl}
\end{center}
\end{figure}

One selection criterion for using back-translation is the ratio of real to synthetic data.
\citet{sennrich-haddow-birch:2016:P16-11} showed that higher ratios of synthetic data lead to decreases in translation performance. 
In order to investigate whether improvements in translation quality increase with higher ratios of synthetic data, we perform three experiments with different sizes of synthetic data (see Figure~\ref{btmets}: left).
Figure~\ref{ppl} shows the perplexity as a function of training time for different sizes of synthetic data.

We find that all systems perform similarly in the beginning and converge after observing increasingly more training instances.
However, the model with the ratio of (\mbox{1:10}) synthetic data becomes increasingly biased towards the noisy data after 1M instances.
Decreases in performance with more synthetic than real data is also in line with findings of \citet{2018arXiv180406189P}.
Comparing the systems using ratios of \mbox{1:1} and \mbox{1:4}, we see that the latter achieves lower perplexity during training.
Table~\ref{bigbigger} shows the performance of these systems on the German$\rightarrow$English translation task. 
\setlength{\tabcolsep}{4pt}
\begin{table}[htb!]
\begin{center}
\caption{\label{bigbigger} German$\rightarrow$English translation quality ({BLEU}) of systems with different ratios of \textit{\mbox{real:syn}} data.}
\begin{tabular}{lccccc}
 \toprule \bf  & \bf 	Size &  \bf WMT14 &  \bf	WMT15 &  \bf	WMT16 &  \bf	WMT17 \\ \midrule%
 Baseline	& 4.5M	& 26.7	&27.6	&32.5	&28.1\\
%Parallel + synthetic src (random)	& 8.9M	&	28.8	&30.0	&36.5	&31.1\\
%Parallel + synthetic src (random 2)	& 9M	& 28.9	&29.6&	36.2&	30.7\\
%Parallel + synthetic src (random 3)	& 9M	&	28.5	&29.6	&36.2&	30.6\\
\hspace{.15cm} + synthetic (\mbox{1:1})	& 9M	&	28.7&	29.7	&36.3	&30.8\\
\hspace{.15cm} + synthetic (\mbox{1:4}) 	&23M	&	29.1&	30.0&	36.9	&31.1\\
\hspace{.15cm} + synthetic (\mbox{1:10})  & 50M	&22.8&23.6&	29.2	&		 23.9\\
\bottomrule
\end{tabular}
\end{center}
\end{table}
The {BLEU} scores show that the translation quality does not improve linearly with the size of the synthetic data.
The model trained on \mbox{1:4} real to synthetic ratio of training data achieves the best results, however, the performance is close to the model trained on a ratio of \mbox{1:1}.

Similar to the study in this section, \citet{edunov-etal-2018-understanding} showed that it is possible to achieve the best results with a large-scale model trained on a \mbox{1:50} ratio of real to synthetic data. 
However, it is crucial to upsample the real data during training so that training batches contain on average an equal amount of real and synthetic data.

\subsection{Impact of translation direction}

Adding monolingual data in the target language to the training data has the potential benefit of introducing new contexts and improving the fluency of the translation model.
The automatically generated translations in the source language introduce new contexts for the source words and, despite not being perfect, improve the quality of the final re-trained model.

Monolingual data is available in large quantities for many languages. 
The decision of the direction of back-translation is subsequently not based on the monolingual data available, but on the advantage of having more fluent source or target sentences.

\begin{table}[htb!]
\begin{center}
\caption{\label{dir} English$\rightarrow$German translation quality ({BLEU}) of systems using forward and reverse models for generating synthetic data.}
\begin{tabular}{lccccc}
 \toprule \bf  & \bf 	Size &  \bf WMT14 &  \bf	WMT15 &  \bf	WMT16 &  \bf	WMT17 \\ \midrule%
Baseline	& 4.5M	&	21.2&	23.3&	28.0&	22.4\\
\hspace{.15cm} + synthetic tgt &	9M		&22.4	&25.3	&29.8	&23.7\\
\hspace{.15cm} + synthetic src &	9M		&24.0&	26.0	&30.7	&24.8\\
\bottomrule
\end{tabular}
\end{center}
\end{table}

\citet{Lambert:2011:ITM:2132960.2132997} showed that adding synthetic source and real target data achieves improvements in traditional phrase-based machine translation. Similarly, in previous works in NMT, back-translation is applied to monolingual data in the target language.
\citet{zhang-zong-2016-exploiting} proposed a self-learning algorithm to generate synthetic data from monolingual source sentences. 
During re-training, they distinguish between real and synthetic data by freezing the parameters of the decoder for the synthetic data.

We perform a small experiment to compare the impact of translation direction and where to incorporate monolingual data (see Figure~\ref{btmets}: middle).
Table~\ref{dir} shows that in both directions, the performance of the translation system improves over the baseline. 
This is in contrast to the findings of~\citet{Lambert:2011:ITM:2132960.2132997} for PBMT systems where they show that using synthetic target data does not lead to improvements in translation quality.

Still, when back-translating target monolingual data, BLEU scores in the target language are higher than when translating monolingual data in the source language. 
This indicates the importance of having fluent sentences in the target language.

\subsection{Impact of quality of the synthetic data}

One selection criterion for back-translation is the quality of the synthetic data.
\citet{W18-2709} studied the effects of noise in the training data on a translation model and discovered that NMT models are less robust to many types of noise than PBMT models.
In order for the NMT model to learn from the parallel data, the data should be fluent and close to the manually generated translations.
However, automatically generating sentences using back-translation is not as accurate as manual translations.

To investigate the \textit{oracle gap} between the performance of manually created and back-translated sentences, we perform a simple experiment using the existing parallel training data (see Figure~\ref{btmets}: right).
In this experiment, we divide the parallel data into two parts, train the reverse model on the first half of the data, and use this model to back-translate the second half.
The manually translated sentences of the second half are considered as ground truth for the synthetic data.

Table~\ref{groundt} shows the {BLEU} scores of the experiments.
As to be expected, re-training with additional parallel data yields higher performance than re-training with additional synthetic data.
However, the differences between the {BLEU} scores of these two models are surprisingly small.
\begin{table}[htb!]
\begin{center}
\caption{\label{groundt} German$\rightarrow$English translation quality ({BLEU}) of systems using synthetic source and human generated source data.}
\begin{tabular}{lccccc}
 \toprule \bf  & \bf  Size  &  \bf WMT14 &  \bf	WMT15 &  \bf	WMT16 &  \bf	WMT17 \\ \midrule%
Baseline  &	2.25M	&24.3	&24.9	&29.5	&25.6\\
\hspace{.15cm} + synthetic src &	4.5M	 &26.0	&26.9	&32.2	&27.5\\
\hspace{.15cm} + ground truth	& 4.5M	& 26.7	&27.6	&32.5	&28.1\\
\bottomrule
\end{tabular}
\end{center}
\end{table}
This indicates that performing back-translation with a reasonably good reverse model already achieves results that are close to a system that uses additional manually translated data.
This is in line with findings of \citet{sennrich-haddow-birch:2016:P16-11} who observed that the same monolingual data translated with three translation systems of different quality and used in re-training the translation model yields similar results.

\section{Back-Translation and token prediction loss} \label{bttokpred}

In the previous section, we observed that using back-translated data yields almost the same improvements as gold parallel data with the same target side.
However, there is a limit in learning from synthetic data, and with higher ratios of synthetic data the model biases too much towards the synthetic data.

In this section, we investigate the influence of the sampled sentences on the model.
In Chapter~\ref{chapter:research-02}, we showed that targeting specific words during data augmentation improves the generation of these words in the right context. 
Specifically, adding synthetic sentences containing those words to the training data has an impact on the prediction probabilities of individual words.
In this chapter, we further examine the effects of the back-translated synthetic data on the prediction of target tokens.

As mentioned in Section~\ref{nmtsec}, the objective function of training an NMT system is to minimize $\mathcal{L}$ by minimizing the prediction loss, $-\log p(y_t \mid \vt{y}_{<t}, \vt{s}_n)$, for each target token in the training data, where:
\begin{align}
p(y_t \mid y_{<t}, \vt{s}_n) =  \softmax \,(\vt{W}_o\widetilde{\vt{h}}_t)
\end{align}

\noindent Here, $\widetilde{\vt{h}}_t$ is the top hidden layer of the decoder and $\vt{W}_o$ is the output weight matrix. 
The addition of monolingual data in the target language improves the estimation of the probability $p({Y})$ and consequently, the model generates more fluent sentences.

\begin{figure}[htb!]
\centering
\includegraphics[width=\textwidth]{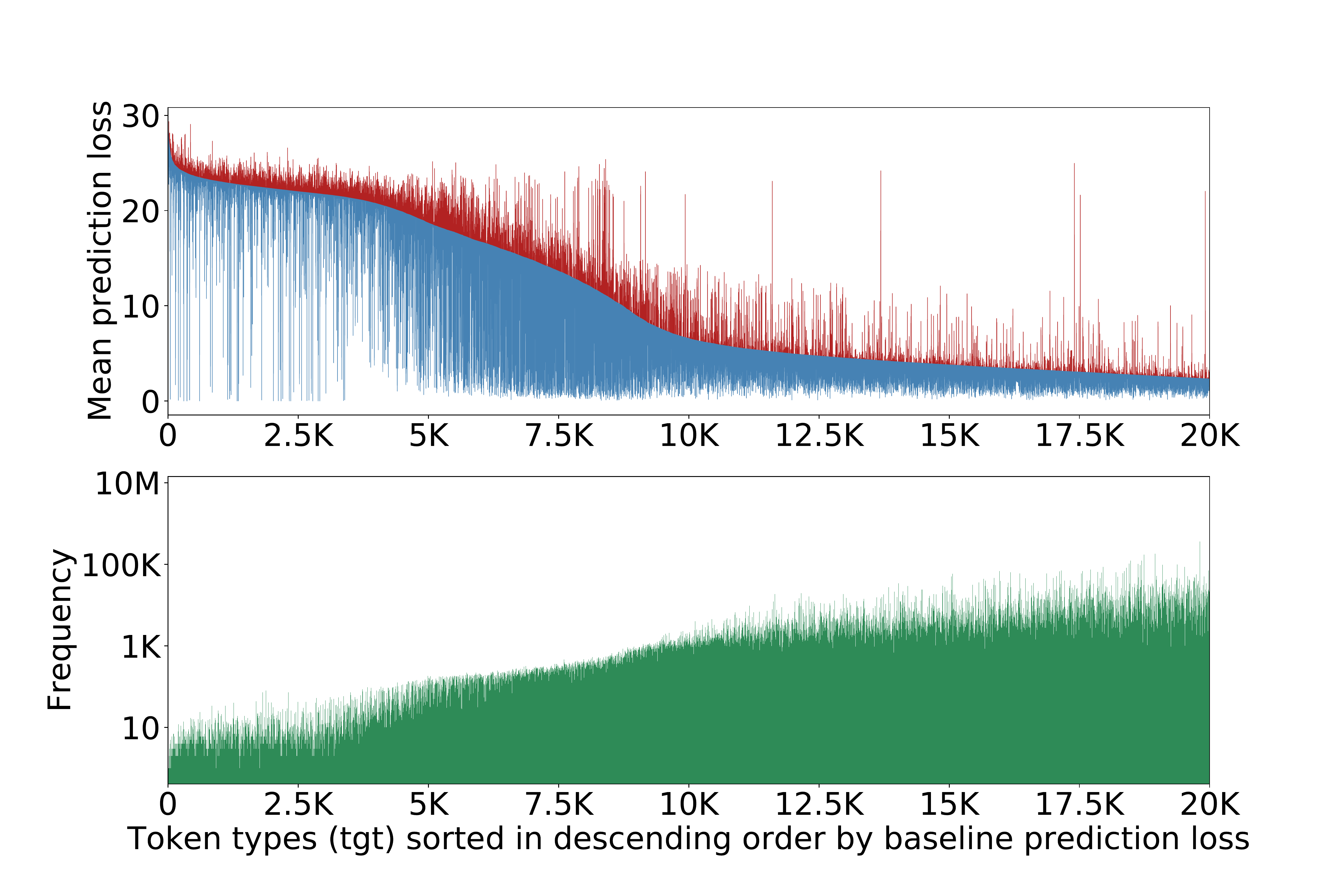}
\caption{Top: Changes in mean prediction loss after re-training with synthetic data sorted by mean prediction loss of the baseline system (x-axis). Decreases and increases in values are marked blue and red, respectively. Bottom: Frequencies (log) of target tokens in the baseline training data. Note that data points in both plots (x-axis) represent \textit{token} types in the vocabulary.}
\label{loss}
\end{figure}

\citet{sennrich-haddow-birch:2016:P16-11} show that by using back-translation, the system with target-side monolingual data reaches a lower perplexity on the development set.
This is expected since the domain of the monolingual data is news and therefore similar to the domain of the development set.
To investigate the model's accuracy independently from the domains of the data, we collect statistics of the target token prediction loss during training.

Figure~\ref{loss} shows the changes of token prediction loss % $-\log p(y_t | y_{<t}, \vt{s}_n)$ 
when training is close to converging and the weights are verging on being stable.
The values are sorted in decreasing order by the tokens' mean prediction losses of the system trained on real parallel data (before augmentation).
We observe an effect similar to distributional smoothing \citep{chen1996empirical}: 
First, we observe that the prediction loss increases slightly for most tokens (red).
Next, we spot an irregular pattern in \textit{decrease} of prediction loss (blue).
The largest decrease in loss occurs for tokens with high prediction loss values when trained on the parallel data only. 
This indicates that by randomly sampling sentences for back-translation, the model improves its estimation of tokens that were originally more difficult to predict, i.e., tokens that had a high prediction loss. 

Note that we compute the token prediction loss, without updating the weights, in just one pass over the training corpus with the final model and as a result, the loss is not biased towards the order of presentation of the training sentences.
% it is not biased towards the order of the data.

This finding motivates us to further explore sampling criteria for back-translation that contribute considerably to the parameter estimation of the translation model. 
We propose that by oversampling sentences containing difficult-to-predict tokens, we can maximize the impact of using the monolingual data.
After translating sentences containing such tokens and including them in the training data, the model becomes more robust in predicting these tokens.
In the next two sections, we propose several methods of using the target token prediction loss to identify the most beneficial sentences for back-translation and re-training the translation model.

\section{Targeted sampling based on model failure} \label{bttarget}

One of the main benefits of using synthetic data is getting a better estimation of words that were originally difficult to predict as measured by their high prediction losses during training.
In this section, we propose four variations of how to identify these words and perform sampling to target these words.
The first three variations are described in Algorithm~\ref{alg1} where the list of difficult tokens is defined in two different ways.
The third variation is described in Algorithm~\ref{alg2}. 
The following subsections provide details of these model variants. 

 {\centering
\begin{minipage}{.85\linewidth}
\begin{algorithm}[H]
\caption{Sampling for difficult words}\label{alg1}
 \hspace*{\algorithmicindent} {\textbf{Input:}} Difficult tokens \textit{$\mathfrak{D}=\{y_i\}_{i=1}^{D}$}, monolingual 
corpus \textit{$\mathbb{M}$}, number \\
\hspace*{\algorithmicindent} of required samples \textit{N}   \\
\hspace*{\algorithmicindent} {\textbf{Output:}} Sampled sentences $S=\{S_i\}_{i=1}^{N}$ where each
sentence $S_i$ is \\
\hspace*{\algorithmicindent} sampled from $\mathbb{M}$
 \begin{algorithmic}[1]
\Procedure{\textsc{DiffSampling} ($\mathfrak{D}, \mathbb{M}, N$):}{}
%\State Initialize $D = \{\}$ \Comment{list of difficult words}
\State Initialize $S=\{\}$ %\Comment{sampled sentences set}
\Repeat
\State Sample $S_c$ from $\mathbb{M}$
\ForAll{tokens $y$ in $S_c$}
\If{$y \in \mathfrak{D}$} 
	\State Add $S_c$ to $S$
\EndFor
\Until{$|S| = N$}
\State \textbf{return} $S$ %\Comment{The gcd is b}
\EndProcedure
\end{algorithmic}
\end{algorithm}
\end{minipage}
\par
}
\vspace{\baselineskip}% Insert a blank line

\subsection{Token frequency as a feature of difficulty} 

Figure~\ref{loss} shows that the majority of tokens with high mean prediction losses have low frequencies in the training data.
Additionally, the majority of decreases in prediction loss after adding synthetic sentence pairs to the training data occurs with less frequent tokens.
Note that these tokens are not necessarily \textit{rare} and some of them may have up to 1000 different occurrences in the training data.
We observe in Figure~\ref{loss} that approximately half of the tokens in the target vocabulary benefit from back-translated data.

We propose a sampling criterion based on token frequencies. 
Sampling new contexts from monolingual data provides context diversity proportional to the token frequencies and less frequent tokens benefit most from new contexts.
Algorithm~\ref{alg1} presents this approach where the list of difficult tokens is defined as: 
\begin{align}
\mathfrak{D} = \{\forall y_i \in V_t \colon freq(y_i) < \eta \}
\end{align}
\noindent where $V_t$ is the target vocabulary and $\eta$ is the frequency threshold for deciding on the difficulty of the token.

\subsection{Tokens with high mean prediction losses}

In this approach, we use the mean losses to identify difficult-to-predict tokens. 
The mean prediction loss $\hat{\ell}(y)$ of token $y$ during training is defined as follows:
\begin{align}\label{ave}
\hat{\ell}(y) = \frac{1}{n_y}\sum_{n=1}^{N}\sum_{t=1}^{|Y^{n}|} -\log p(y^{n}_t \mid {y}^{n}_{<t}, \vt{s}_n) \delta(y^n_t,y)
\end{align}
where $n_y$ is the number of times token $y$ is observed during training, i.e., the token frequency of $y$, $N$ is the
number of sentences in the training data, $|\vt{Y}^{n}|$ is the length of target sentence $n$, and $\delta(y^n_t,y)$ is the Kronecker delta function, which is 1 if $y^n_t = y$ and 0 otherwise.
By specifically providing more sentences for difficult words, we improve the model's estimation and decrease its prediction uncertainty.

%During sampling from the monolingual data, we select sentences that contain difficult words.

Algorithm~\ref{alg1} presents this approach where the list of difficult tokens is defined as: 
\begin{align}
\mathfrak{D} = \{\forall y_i \in V_t \colon \hat{\ell}(y_i) > \mu \}
\end{align}

\noindent
where $V_t$ is the vocabulary of the target language and $\mu$ is the threshold for the difficulty of the token.

\subsection{Tokens with skewed prediction losses}

By using the mean loss for target tokens as defined above, we do not discriminate between differences in prediction loss for occurrences in different contexts. This lack of discrimination can be problematic for tokens with high loss variations. For instance, there can be a token with ten occurrences, out of which two have high and eight have low prediction loss values. 

We hypothesize that if a particular token is easier to predict in some contexts and harder in others, the sampling strategy should be context-sensitive, allowing to target specific contexts in which a token has a high prediction loss. 
In order to distinguish between tokens with a skewed and tokens with a more uniform prediction loss distribution, we use both the mean and standard deviation of the token prediction losses to identify difficult tokens.
Hence, we target tokens that have both high mean prediction loss and high amount of variation in different contexts.
Algorithm~\ref{alg1} formalizes this approach where the list of the difficult tokens is defined as: 
\begin{align}
\mathfrak{D} = \{\forall y_i \in V_t \colon \hat{\ell}(y_i) > \mu \land \sigma(\ell(y_i)) > \rho \}
\end{align}

\noindent where $\hat{\ell}(y_i)$ is the mean and $\sigma(\ell(y_i))$ is the standard deviation of prediction loss of token $y_i$, $V_t$ is the vocabulary list of the target language, and $\mu$ and $\rho$ are the thresholds for deciding the difficulty of the token.

 \subsection{Preserving sampling ratio of difficult occurrences}

Above, we examined the mean of prediction loss for each token over all occurrences, in order to identify difficult-to-predict tokens.
However, the uncertainty of the model in predicting a difficult token varies for different occurrences of the token: one token can be easy to predict in one context, and hard in another.
While the sampling step in the previous approaches targets these tokens, it does not ensure that the distribution of sampled sentences is similar to the distribution of problematic tokens in difficult contexts.

{\centering
\begin{minipage}{.85\linewidth}
\begin{algorithm}[H]
\caption{Sampling with ratio preservation}\label{alg2}
 \hspace*{\algorithmicindent} \textbf{Input:} Difficult tokens and the corresponding sentences in the bitext \\
  \hspace*{\algorithmicindent}  \textit{$\mathfrak{D}=\{y_t, Y_{y_t}=[y_1, \ldots, y_t, \ldots, y_m]\}$}, monolingual corpus \textit{$\mathbb{M}$}, \\
  \hspace*{\algorithmicindent} number of required samples \textit{N} \\
\hspace*{\algorithmicindent} \textbf{Output:} Sampled sentences $S=\{S_i\}_{i=1}^{N}$ where each sentence $S_i$ is \\
\hspace*{\algorithmicindent}  sampled from $\mathbb{M}$
 \begin{algorithmic}[1]
\Procedure{\textsc{PredLossRatioSampling}($\mathfrak{D}, \mathbb{M}, N$): }{}
\State Initialize $S=\{\}$
\State $H(y_t) = \frac{ N \times \mid(y_t, \boldsymbol{\cdot})\in \mathfrak{D}\mid}{\mid(y_{\boldsymbol{\cdot}}, \boldsymbol{\cdot})\in \mathfrak{D}\mid}$
%\State Initialize $H = []$ \Comment{Hash of difficult words and intended ratio in the sampled set}
\Repeat
\State Sample $S_c$ from $\mathbb{M}$
\ForAll{tokens $y$ in $S_c$}
\If{$|y \in S| < H(y)$} 
	\State Add $S_c$ to $S$
\EndFor
\Until{$|S| = N$}
\State \textbf{return} $S$ %\Comment{The gcd is b}
\EndProcedure
\end{algorithmic}
\end{algorithm}
\end{minipage}
\par
}
\vspace{\baselineskip}% Insert a blank line

To address this issue, we propose an approach where we consider the number of times a token occurs in difficult-to-predict contexts and sample sentences accordingly, thereby ensuring the same ratio as the distribution of difficult contexts.
If token $y_1$ is difficult to predict in two contexts and token $y_2$ is difficult to predict in four contexts, the number of sampled sentences containing $y_2$ is double the number of sampled sentences containing $y_1$.
Algorithm~\ref{alg2} formalizes this approach.

 \subsection{Results}
 
We measure the translation quality of various models for German$\rightarrow$English and English $\rightarrow$ German translation tasks.
As baseline we compare our approach to \citet{sennrich-haddow-birch:2016:P16-11}. 
For all experiments we sample and back-translate sentences from WMT monolingual data, keeping a one-to-one ratio of back-translated versus original data (\mbox{1:1}).
We set the hyperparameters $\mu$, $\rho$, and $\eta$ to 5, 10, and 5000, respectively.
The values of the hyperparameters are chosen on a small sample of the parallel data based on the token loss distribution. 

\begin{figure}[htb!]
\begin{center}
\includegraphics[width=\textwidth]{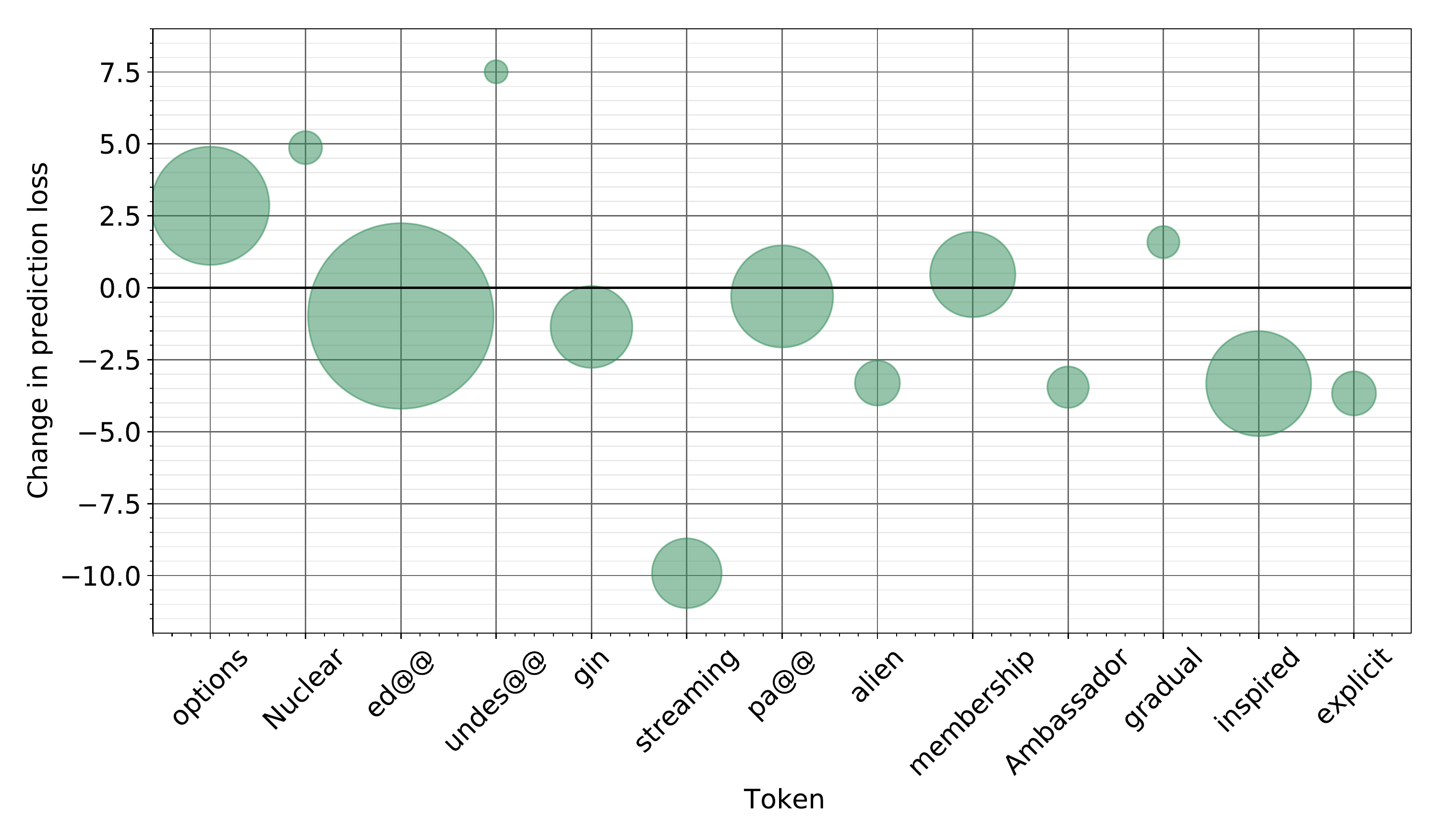}
\caption{Examples of changes in average prediction loss after augmentation. Lower is better. The sizes of dots are proportional to the increase in the number of contexts for each word in the training data. Subword unit boundaries are marked with `@@'. \label{predlossexample}}
\end{center}
\end{figure}

Our proposed augmentation method increases instances of targeted words in the training data which leads to an overall decrease in average prediction loss per token. Figure~\ref{predlossexample} provides examples of tokens in the training data and their changes after data augmentation.  
The results of the translation experiments are presented in Tables~\ref{meanresultsende} and \ref{meanresultsdeen}.

\begin{table*}
\begin{center}
%\small%
\tabcolsep=2.5pt%
\caption{\label{meanresultsende}  English$\rightarrow$German translation quality ({BLEU}). Experiments marked $^\dagger$ are averaged over 3 runs. \textsc{Random} is the standard back-translation approach with random sampling. \textsc{MPL} and \textsc{Freq} are difficulty criteria based on mean prediction loss and token frequency, respectively. \textsc{MPL + sPL} is experiments with upsampling tokens with skewed prediction losses. \textsc{PPLR} preserves the  ratio of the distribution of difficult contexts.	}
\begin{tabular}{@{ }lccccc@{}}
\toprule
   & \multicolumn{4}{c}{\textbf{En-De}}  \\ \cmidrule{2-6} 
 \textbf{System}    &  \bf WMT14 &  \bf WMT15 &  \bf WMT16 &  \bf WMT17 \\ 
\midrule     
 \textsc{Baseline}$^\dagger$		& 21.2	&23.3&	28.0&	22.4\\
%Parallel + synthetic src (random)	& 8.9M	&	28.8	&30.0	&36.5	&31.1\\
%Parallel + synthetic src (random 2)	& 9M	& 28.9	&29.6&	36.2&	30.7\\
%Parallel + synthetic src (random 3)	& 9M	&	28.5	&29.6	&36.2&	30.6\\
\textsc{Random}$^\dagger$  &	 24.0&	26.0&	30.7	&24.8\\ \midrule
 \textbf{Difficulty criterion} & &&      \\ \midrule
%\textsc{PredLoss}, $n>=3$	& 	28.6	&	&29.3	&&	35.9	&&30.5&	&&&& &\\
\textsc{Freq} 	&24.2&27.0&31.7 &25.2\\			
\textsc{MPL}$^\dagger$	 &\textbf{24.7}&26.8&31.5&\textbf{25.5}\\ %$n>=1$	
\textsc{MPL + sPL} 	&24.1&26.9  &31.0&25.3\\
\textsc{PPLR}  &24.5&\textbf{27.2}&\textbf{31.8} &25.5\\
\bottomrule
\end{tabular}
\end{center}
\end{table*}

\begin{table*}
\begin{center}
\tabcolsep=2.5pt%
\caption{ \label{meanresultsdeen}  German$\rightarrow$English translation quality ({BLEU}). Experiments marked $^\dagger$ are averaged over 3 runs. \textsc{Random} is the standard back-translation approach with random sampling. \textsc{MPL} and \textsc{Freq} are difficulty criteria based on mean prediction loss and token frequency, respectively. \textsc{MPL + sPL} is experiments with upsampling tokens with skewed prediction losses. \textsc{PPLR} preserves the  ratio of the distribution of difficult contexts.	}
%\vspace{10mm}
\begin{tabular}{@{ }lccccc@{}}
\toprule
   & \multicolumn{4}{c}{\textbf{De-En}}  \\ \cmidrule{2-6}
 \textbf{System}  & \bf WMT14 & \bf  WMT15 &   \bf WMT16 &   \bf WMT17 \\ 
\midrule     
 \textsc{Baseline}$^\dagger$		& 26.7 &	27.6 &	32.5& 	28.1\\
%Parallel + synthetic src (random)	& 8.9M	&	28.8	&30.0	&36.5	&31.1\\
%Parallel + synthetic src (random 2)	& 9M	& 28.9	&29.6&	36.2&	30.7\\
%Parallel + synthetic src (random 3)	& 9M	&	28.5	&29.6	&36.2&	30.6\\
\textsc{Random}$^\dagger$ & 	28.7& 		29.7	&	36.3		&30.8 \\ \midrule
 \textbf{Difficulty criterion} & &&      \\ \midrule
%\textsc{PredLoss}, $>=3$	& 	28.6	&	&29.3	&&	35.9	&&30.5&	&&&& &\\
\textsc{Freq} 	& 29.7&30.5	&37.5&31.4 &\\		 	
\textsc{MPL}$^\dagger$ &	29.9 &		\textbf{30.9} 		&\textbf{37.8}	&\textbf{32.1}	\\ %$n>=1$	
%\textsc{MPL}, $\nless 3$ & 	28.6	&	29.3	&	35.9	&30.5 \\
\textsc{MPL + sPL} 	& 	\textbf{30.0}&		30.9		&37.7	&31.9  \\
\textsc{PPLR}	&		29.8	&30.9&37.4& 31.6  \\
\bottomrule
\end{tabular}
\end{center}
\end{table*}

As expected using random sampling for back-translation improves the translation quality over the baseline.
However, each of the proposed targeted sampling techniques outperforms random sampling. 
Specifically, the best performing model for German$\rightarrow$English uses the mean of prediction loss (\textsc{MPL}) for the target vocabulary to frequently sample sentences including these tokens.

For the English$\rightarrow$German experiments we obtain the best translation performance when we preserve the prediction loss ratio during sampling.
We also observe that even though the model targeting tokens with skewed prediction loss distributions ($\textsc{MPL + sPL}$) improves over random selection of sentences, it does not outperform the model using only mean prediction losses. 
Note that frequency-based sampling, the simplest method proposed in this chapter, is very effective.
We observe that the gains of other proposed approaches over frequency-based sampling are quite small. 
Therefore using frequency-based sampling remains a good strategy to improve translation quality over random sampling. 

\section{Context-Aware targeted sampling} \label{btcontextu}

In the previous section, we proposed methods for identifying difficult-to-predict tokens and performed targeted sampling from monolingual data.
While the objective was to increase the occurrences of difficult tokens, we ignored the context of these tokens in the sampled sentences. 

Arguably, if a word is difficult to predict in a given context, providing more examples of the same or similar context can aid the learning process.
In this section, we focus on the context of difficult-to-predict words and aim to sample sentences that are similar to the corresponding difficult context.
We first identify difficult-to-predict words and the local \textit{context} where the prediction loss is high. 
Next, we sample sentences from the monolingual data that contain a difficult-to-predict word. 
We then compare the context of the difficult word in the sampled sentence and the initial difficult context and select the sampled sentence if the contexts are similar. 
Finally, we back-translate the selected sentences and augment the training data.

{\centering
\begin{minipage}{.85\linewidth}
\begin{algorithm}[H]
\caption{Sampling with context}\label{alg3}
  \hspace*{\algorithmicindent} \textbf{Input:} Difficult tokens and the corresponding sentences  in the bitext \\
    \hspace*{\algorithmicindent}  \textit{$\mathfrak{D}=\{y_t, Y_{y_t}=[y_1, \ldots, y_t, \ldots, y_m]\}$},  monolingual corpus \textit{$\mathbb{M}$},  \\
      \hspace*{\algorithmicindent} context function \textit{$context$}, number of required samples \textit{N}, similarity \\
     \hspace*{\algorithmicindent} threshold $s$ \\
\hspace*{\algorithmicindent} \textbf{Output:} Sampled sentences $S=\{S_i\}_{i=1}^{N}$ where each sentence $S_i$ is \\
  \hspace*{\algorithmicindent} sampled from $\mathbb{M}$
 \begin{algorithmic}[1]
\Procedure{\textsc{contextSampling}($\mathfrak{D}, \mathbb{M}, context, N, s$): }{}
\State Initialize $S=\{\}$
\Repeat
\State Sample $S_c$ from $\mathbb{M}$
\ForAll{tokens $y_t$ in $S_c$}
	\If{$y_t \in \mathfrak{D}$} 
%\State $C_m \leftarrow substr(S_c,i-w, i+w)$
	\State $C_m \leftarrow context(S_c,$ index\_of$(S_c, y_t))$ \label{cont1}
	\ForAll{$Y_{y_t}$}
	%\State $C_p \leftarrow substr(Y_{y_t}, t-w, t+w)$
		\State $C_p \leftarrow context(Y_{y_t},$  index\_of$(Y_{y_t}, y_t)) $ \label{cont2}
                         \State \If{Sim$(C_m, C_p) > s$}  \label{simss}:  Add $S_c$ to $S$
	\EndFor
\EndFor
\Until{$|S| = N$}
\State \textbf{return} $S$ %\Comment{The gcd is b}
\EndProcedure
\end{algorithmic}
\end{algorithm}
\end{minipage}
\par
}
\vspace{\baselineskip}% Insert a blank line

The general algorithm is described in Algorithm~\ref{alg3}.
In the following sections, we discuss different definitions of the local context ($context$ function in line~\ref{cont1} and line~\ref{cont2}) and similarity measures ($\text{Sim}$ function in line~\ref{simss}) in this algorithm and report the results.

\subsection{Definition of local context}

Prediction loss is a function of the source sentence and the target context.
We hypothesize that one of the reasons that a token has a high prediction loss in only some contexts is because of the complexity of those contexts.
This complexity can be caused by an infrequent event such as a rare sense of the word, a domain that is different from other occurrences of the word, or an idiomatic expression. 

We identify \textit{pairs} of tokens and sentences from parallel data where in each pair, the NMT model suffers a high prediction loss for the token in the given context. 
Note that a token can occur several times in this list since it can be considered as difficult-to-predict in different sentences.

We propose two approaches to define the local context of a difficult token:

\paragraph{Neighboring tokens}

A straightforward way is to use positional context: tokens that precede and follow the target token, typically in a window of $w$ tokens to each side.
For sentence $S$ containing a difficult token at index $i$, the \textit{context} function in Algorithm~\ref{alg3} is:
\begin{align}
context(S, i) = [S^{i-w}, \ldots, S^{i-1}, S^{i+1}, \ldots, S^{i+w}]
\end{align}
\noindent
where $S^{j}$ is the token at index $j$ in sentence $S$.
Note that in this approach, we look at a window of fixed size and as a result, not all subwords from the same word may end up in this context window. For instance for the sentence \textit{`a professor and a \underline{colleague} at Stan$\mid$ford'}, with target word \textit{`colleague'}, and $w=2$, the context is [\textit{`and', `a', `at', `Stan'}]. Here, the subword \textit{`ford'} as part of the word \textit{`Stanford'} is not included in the context window. The symbol `$\mid$' signifies subword unit boundary. 

\paragraph{Sibling tokens}

In our analysis of prediction loss during training, we observe that several tokens that are difficult to predict are indeed subword units.
Current state-of-the-art NMT systems apply BPE to the training data to address large vocabulary challenges \citep{sennrich-haddow-birch:2016:P16-12}.
By using BPE, the model generalizes common subword units towards what is more frequent in the training data.
This is inherently useful since it allows for better learning of less frequent words.
However, a side effect of this approach is that at times the model generates subword units that are not linked to any words in the source sentence.
As an example, in Table~\ref{exbpe1}, the German source and the English reference translation highlight this problem.
The word \textit{`B$\mid$ahr'} consisting of two subword units is incorrectly translated into \textit{`B$\mid$risk'} because of an unintended side-effect of both sharing the subword unit \textit{`B'}.

We address the insufficiency of the context for subword units with high prediction losses by targeting these tokens in sentence sampling.
Algorithm~\ref{alg3} formalizes this approach in sampling sentences from the monolingual data.
For a sentence $S$ containing a difficult subword at index $i$, the context function is defined as:
\begin{align}
context(S, i) = [ S^{n},\ldots, S^{i-1}, S^{i+1}, \ldots, S^{m} ]
\end{align}
where every token $S^{j}$ in the local context is a subword unit and part of the same word as $S^i$.
Table~\ref{exbpe2} presents examples of sampled sentences for the difficult subword unit \textit{`Stan'}.
In this case, the difficult context for this token is \textit{`Stan$\mid$ford'} and we use it for computation of similarity.
This suggests that the subword unit \textit{`Stan'} is difficult to predict when the context is for the word \textit{`Stan$\mid$ford'}.
This excludes other contexts where the subword unit \textit{`Stan'} is part of another word, such as \textit{`Stan$\mid$dard'}.
\begin{table}[thb!]
\begin{center}\small
\caption{\label{exbpe1} An example from the synthetic data where the word \textit{B$\mid$ahr} is incorrectly translated to \textit{B$\mid$risk}. Subword unit boundaries are marked with `$\mid$'.}
\begin{tabularx}{0.7\columnwidth}{lX}
\toprule
\textit{source} & wer glaube, dass das Ende, sobald sie in Deutschland ank$\mid${\"a}$\mid$men, ir$\mid$re, erz{\"a}hlt \textbf{B$\mid$ahr}. \\
\textit{reference } & if you think that this stops as soon as they arrive in Germany, you'd be wrong, says \textbf{B$\mid$ahr}.\\
\textit{NMT output} & who believe that the end, as soon as they go to Germany, tells \textbf{B$\mid$risk}.\\
\bottomrule
\end{tabularx}
\end{center}
\end{table}
\begin{table}[htb!]
\begin{center}\small
\caption{\label{exbpe2} Results of context-aware targeted sampling with sibling tokens for the difficult subword unit \textit{`Stan'}. In this example, the difficult context in which the subword \textit{`Stan'} has a high prediction loss is the complete word \textit{`Stanford'} and we sample sentences containing this word.}
\begin{tabularx}{0.8\columnwidth}{Xr}
\toprule
 \textit{Sentence from bitext containing difficult token \textbf{`Stan'}} & \\ \midrule
\multicolumn{2}{X}{He attended \textit{\textbf{Stan}$\mid$ford} University, where he double maj$\mid$ored in Spanish and History.}  \\  
  \midrule
  \textit{Sampled sentences from monolingual data} & \\% \textit{Similarity} \\ 
  \midrule
   %$-$ 
  \rowcolor{tablegray}  The group is headed by Aar$\mid$on K$\mid$ush$\mid$ner, a \textit{\textbf{Stan}$\mid$ford} University gradu$\mid$ate who formerly headed a gre$\mid$eting card company. & \\% 0.74 \\ %0.74 
   Ford just opened a new R\&D center near \textit{\textbf{Stan}$\mid$ford} University, a hot$\mid$bed of such technological research. & \\%0.73 \\%0.73
   \rowcolor{tablegray} Joe Grund$\mid$fest, a professor and a colleague at \textit{\textbf{Stan}$\mid$ford} Law School, outlines four reasons why the path to the IP$\mid$O has become so steep for asp$\mid$iring companies. & \\% 0.73 \\%0.73
\bottomrule
\end{tabularx}
\end{center}
\end{table}

\subsection{Similarity of the local contexts}

In context-aware targeted sampling, we compare the context of a sentence candidate and the difficult context in the parallel data and select the sentence if they are \textit{similar}.
In the following, we propose two approaches for measuring the similarities. 

\paragraph{Matching the local context (Exct)}

In this approach, we aim to sample sentences containing the difficult token matching the exact context to the problematic context.
By sampling sentences that match in a local window with the problematic context and differ in the rest of the sentence, we have more instances of the difficult token for training.
Algorithm~\ref{alg3} formalizes this approach where the similarity function is defined as:
\begin{align}
\text{Sim}(C_m, C_p) = \frac{1}{c} \sum_{i=1}^{c}\delta(C_m^i, C_p^i)
\end{align}

\noindent $C_m$ and $C_p$ are the contexts of the sentences from monolingual and parallel data, respectively, and $c$ is the number of tokens in the contexts. The $\delta$ function returns 1 when $C_m^i$ and $C_p^i$ are the same token, and 0 otherwise. 

\paragraph{Word representations (Sem)}

Another approach to sampling sentences that are similar to the problematic context is to weaken the matching assumption.
Allowing sentences that are similar in subject and not match the exact context words allows for lexical diversity in the training data.
We use embeddings obtained by training the Skipgram model \citep{mikolov2013efficient} on monolingual data to calculate the similarity of the two contexts.
For this approach we define the similarity function in Algorithm~\ref{alg3} as:
\begin{align}
\text{Sim}(C_m, C_p)  = \cos(\vt{v}(C_m), \vt{v}(C_p))
\end{align}
where $\vt{v}(C_m)$ and $\vt{v}(C_p)$ are the averaged embeddings of the tokens in the contexts.
Table~\ref{context} gives examples of sampled sentences for the difficult word \textit{Rock}.
In this example, the context where the word \textit{`Rock'} has high prediction loss is about the \textit{music genre} and not the most prominent sense of the word, \textit{stone}.
Sampling sentences that contain this word in this particular context provides an additional signal for the translation model to improve parameter estimation. 

\begin{table}[htb!]
\begin{center}\small
\caption{\label{context} Results of context-aware targeted sampling for the difficult token \textit{`Rock'} }
\begin{tabularx}{0.89\columnwidth}{Xcc}
\toprule
 \textit{Sentence from bitext containing difficult word} & & \\ \midrule
  Bud$\mid$dy Hol$\mid$ly was part of the first group induc$\mid$ted into the \textbf{Rock} and R$\mid$oll Hall of F$\mid$ame on its formation in 1986. &  \\  \midrule
  \textit{Sentences from monolingual data} & \textit{Similarity} & \textit{Sampled} \\ \midrule
  \rowcolor{tablegray} A 2008 \textbf{Rock} and R$\mid$oll Hall of F$\mid$ame induc$\mid$t$\mid$ee, Mad$\mid$onna is ran$\mid$ked by the Gu$\mid$inn$\mid$ess Book of World Rec$\mid$ords as the top-selling recording artist of all time. & 0.86 & \cmark \\% (\textit{Sim=0.86}) \\
   The winners were chosen by 500 voters, mostly musicians and other music industry veter$\mid$ans, who belong to the \textbf{Rock} and R$\mid$oll Hall of F$\mid$ame Foundation.  & 0.81 & \cmark \\
  \rowcolor{tablegray} The \textbf{Rock} and R$\mid$oll Hall of Fam$\mid$ers gave birth to the California rock sound. & 0.79& \cmark  \\%(\textit{Sim=0.79}) \\
 After an ice cold San Miguel beer at the H$\mid$ard \textbf{Rock} Caf\'e (Ay$\mid$ala Center) just enter the Bur$\mid$gos Street and enjoy the different clubs. & 0.42 & \xmark \\
  \rowcolor{tablegray}  See a play on Broad$\mid$way, enjoy stunning views from the Top of the \textbf{Rock}, or spend the day at the Museum of Mo$\mid$dern Art, all situated nearby. & 0.34 & \xmark  \\
The Library received the donations and endo$\mid$w$\mid$ments of prominent individuals such as John D. \textbf{Rock}$\mid$ef$\mid$eller and James B. Wil$\mid$b$\mid$ur. & 0.29 & \xmark  \\
\bottomrule
\end{tabularx}
\end{center}
\end{table}

 \subsection{Results}

\begin{table*}[p]
	\centering
	\setlength{\tabcolsep}{6pt}
\rotatebox{90}{
\begin{minipage}{\textheight}
\begin{center}
\caption{\label{contextresultsdeen} German$\rightarrow$English translation quality ({BLEU}). Experiments marked $^\dagger$ are averaged over 3 runs. \textsc{Random} is the standard back-translation approach with random sampling. \textsc{PredLoss} is the contextual prediction loss and \textsc{MPL} is the average loss. \textit{token} and \textit{SubUnit} are context selection definitions from neighboring tokens and subword units, respectively. Note that token includes both subword units and full words. \textit{Sent} regards the entire sentence as the context. \textit{Sem} is computing context similarities with token embeddings and \textit{Exct} is comparing the context tokens.}
\begin{tabular}{lcccccccccc}
\toprule
 &&&&&& \multicolumn{4}{c}{\textbf{De-En}}   \\  \cmidrule{7-11} 
 \textbf{System}   && &&&&  \bf WMT14 & \bf WMT15 & \bf WMT16 & \bf WMT17   \\ 
 \midrule    
 \textsc{Baseline} $^\dagger$ & &&&&& 26.7 &	27.6 &	32.5	&28.1\\
%Parallel + synthetic src (random)	& 8.9M	&	28.8	&30.0	&36.5	&31.1\\
%Parallel + synthetic src (random 2)	& 9M	& 28.9	&29.6&	36.2&	30.7\\
%Parallel + synthetic src (random 3)	& 9M	&	28.5	&29.6	&36.2&	30.6\\
\textsc{Random} $^\dagger$ & &&&&&  28.7& 		29.7	&	36.3	&	30.8\\ \midrule
\multirow{2}{*}{ \textbf{Difficulty criterion}} & \multicolumn{3}{c}{\textbf{Context}} &  \multicolumn{2}{c}{\textbf{Similarity}}       \\   \cmidrule(lr){2-4}   \cmidrule(lr){5-6}   %\midrule
   & Neighbor & Sibling& Sentence & Exct & Sem   &    \\ \midrule
\textsc{Freq} & \cmark && & & \cmark  &	 30.0&30.8&37.6&31.7  \\	
\textsc{PredLoss} & &\cmark&  & \cmark &	& 29.1&30.1&	36.9&31.0\\			
%\textsc{PredLoss} & \textsc{Swords}  & \textsc{emb }	&29.6 &30.5&37.5	&31.7&&24.4&\textbf{27.5} &29.8&23.7\\			
\textsc{PredLoss}  & \cmark && &\cmark& 	&	29.7 & 	30.6	&37.6&	31.8  \\ %trigram-bi-uni
\textsc{PredLoss}  & \cmark && && \cmark	 &29.9&30.8	&37.7&31.9  \\
\textsc{PredLoss} & &  &\cmark& & \cmark 	& 24.9&	25.5	&30.1&	26.2   \\	
%\textsc{PredLoss},	&&  {semantic } & 		30.0&	30.7&	37.7	&31.6\\
%\textsc{PredLoss},  .fixed ratio?	&&	 {semantic } & 	29.0&	30.2&	36.6&	31.1\\
%\textsc{PredLoss},   noStop $+$ src	&&	 {semantic } &	30.0	&30.8	&37.7&	31.7\\		
%\midrule	
\textsc{MPL } & \cmark &&  & &\cmark &   \textbf{30.2}	&\textbf{31.4}	&\textbf{37.9}	&\textbf{32.2}\\
\bottomrule
\end{tabular}
\end{center}
\end{minipage}
}
\end{table*}

\begin{table*} [p]
	\centering
		\setlength{\tabcolsep}{6pt}
\rotatebox{90}{
\begin{minipage}{\textheight}
\begin{center}
\caption{\label{contextresultsende} English$\rightarrow$German translation quality ({BLEU}). Experiments marked $^\dagger$ are averaged over 3 runs. \textsc{Random} is the standard back-translation approach with random sampling. \textsc{PredLoss} is the contextual prediction loss and \textsc{MPL} is the average loss. \textit{token} and \textit{SubUnit} are context selection definitions from neighboring tokens and subword units, respectively. Note that token includes both subword units and full words. \textit{Sent} denotes the sentence as the context. \textit{Sem} is computing context similarities with token embeddings and \textit{Exct} is comparing the context tokens.}
\begin{tabular}{lcccccccccc}
\toprule
 &&&&&& \multicolumn{4}{c}{\textbf{En-De}}   \\ \cmidrule{7-11} 
 \textbf{System}   && &&&&  \bf WMT14 & \bf WMT15 & \bf WMT16 & \bf WMT17   \\ 
 \midrule    
 \textsc{Baseline} $^\dagger$& &&&&&  21.2	&23.3&	28.0&	22.4\\
%Parallel + synthetic src (random)	& 8.9M	&	28.8	&30.0	&36.5	&31.1\\
%Parallel + synthetic src (random 2)	& 9M	& 28.9	&29.6&	36.2&	30.7\\
%Parallel + synthetic src (random 3)	& 9M	&	28.5	&29.6	&36.2&	30.6\\
\textsc{Random} $^\dagger$ & &&&&&	 24.0&	26.0&	30.7	&24.8\\ \midrule
\multirow{2}{*}{ \textbf{Difficulty criterion}} & \multicolumn{3}{c}{\textbf{Context}} &  \multicolumn{2}{c}{\textbf{Similarity}}       \\   \cmidrule(lr){2-4}   \cmidrule(lr){5-6}   %\midrule
   & Neighbor & Sibling& Sentence & Exct & Sem   &    \\ \midrule
\textsc{Freq} & \cmark && &  & \cmark  &	 24.4 &   26.3 &31.5&25.6 \\	
\textsc{PredLoss} & & \cmark & & \cmark &		&23.8&26.2 &28.8&23.2\\			
%\textsc{PredLoss} & \textsc{Swords}  & \textsc{emb }	&29.6 &30.5&37.5	&31.7&&24.4&\textbf{27.5} &29.8&23.7\\			
\textsc{PredLoss} & \cmark && & \cmark & 	&	 24.3  &27.4&31.6 & 25.5\\ %trigram-bi-uni
\textsc{PredLoss} &\cmark && & & \cmark & 24.5  &\textbf{27.5}&31.7 & 25.6\\
\textsc{PredLoss} &&& \cmark  & & \cmark		 &22.0&	24.6&	27.9	&22.5\\					
\textsc{MPL } & \cmark &&  & &\cmark  &  \textbf{24.4}& 27.2&\textbf{31.8}&\textbf{25.6}\\
\bottomrule
\end{tabular}
\end{center}
\end{minipage}
}
\end{table*}

The results of the translation experiments are given in Tables~\ref{contextresultsdeen} and \ref{contextresultsende} for German$\rightarrow$ English and English$\rightarrow$German, respectively.
In these experiments, we set the hyperparameters $s$ and $w$ to 0.75 and 4, respectively.
Comparing the experiments with different similarity measures, \textit{Exct} and \textit{Sem}, we observe that in all test sets we achieve the best results when using word embeddings.
This indicates that for targeted sampling it is more beneficial to have diversity in the context of difficult words as opposed to having the exact n-grams. 
When using embeddings as the similarity measure, it is worth noting that with a context of size 4 the model performs very well but fails when we increase the window size to include the whole sentence. 
The experiments focusing on tokens from the same words ({sibling} tokens) achieve improvements over the baselines, however, they perform slightly worse than the experiments using {neighboring} tokens as context.

The best BLEU scores are obtained with the mean of prediction loss as difficulty criterion (\textsc{MPL}) and using word representations to identify the most similar contexts.
We observe that summarizing the distribution of the prediction losses by its mean is more beneficial than using individual losses.
Our results motivate further explorations of using context for targeted sampling sentences for back-translation.

\section{Qualitative results} \label{btanalysisagain}

Finally, we review our proposed approach and further investigate individual token losses.
We observed that the \textit{individual} token loss, even after training converges, has a degree of instability and for the same word, it varies from context to context. 
However, in our experiments in the previous section, using local context to identify these difficult words was not very successful. 

\begin{figure}[htb!]
\begin{minipage}{\textwidth} 
\begin{center}
\includegraphics[width=\textwidth]{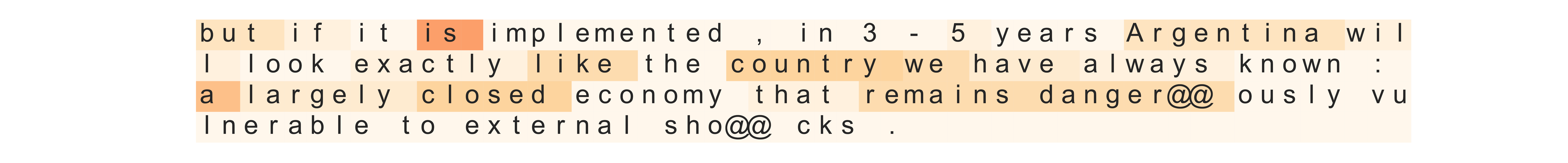}    
\end{center}
\end{minipage}   
\par\smallskip % force a bit of vertical whitespace
\hspace{\fill}  %% no blank line before of after this instruction
%\begin{minipage}{\textwidth} 
%\includegraphics[width=\textwidth]{05-research-03/figs/sent2}    %extra
%\end{minipage}    
%\hspace{\fill}  %% no blank line before of after this instruction
%\vspace{0.75cm}
\begin{minipage}{\textwidth} 
\includegraphics[width=\textwidth]{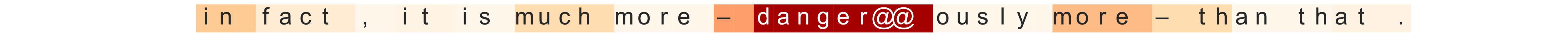}    
\end{minipage}   
\par\smallskip % force a bit of vertical whitespace
\begin{minipage}{\textwidth} 
\includegraphics[width=\textwidth]{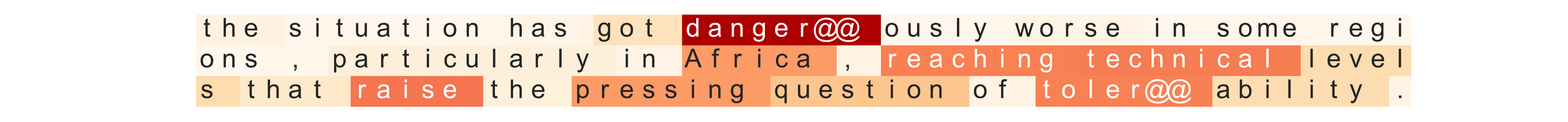}    
\end{minipage}  
%\begin{minipage}{\textwidth} 
%\includegraphics[width=\textwidth]{05-research-03/figs/sent5}     %extra
%\end{minipage}  
\caption{Visualization of token prediction loss (final training epoch) for the subword \textit{\textbf{danger@@}} in three different sentences. Darker means the model has less confidence predicting the word. Subword unit boundaries are marked with `@@' \label{loosexamples}}  
\end{figure}

We look at some examples from the training data where individual token loss is unstable in different contexts.
Figure~\ref{loosexamples} illustrates several sentences from the training data containing the subword \textit{`danger@@'} and the respective prediction losses of the trained model. The symbol `@@' signifies subword unit boundary. 
In all instances, this subword unit is part of the word \textit{`dangerously'}. 
All of the source sentences of these examples include the same German translation, \textit{`gef{\"a}hrlich'}, that corresponds to the translation of this word.
In this particular example, we see no clear indication in the context of why the model's confidence for the token \textit{`danger'} is considerably different for different contexts. 

Finally, we study the importance of the position of the token in the confidence of the model.
\begin{figure}[htb!]
\begin{center}
\includegraphics[width=\textwidth]{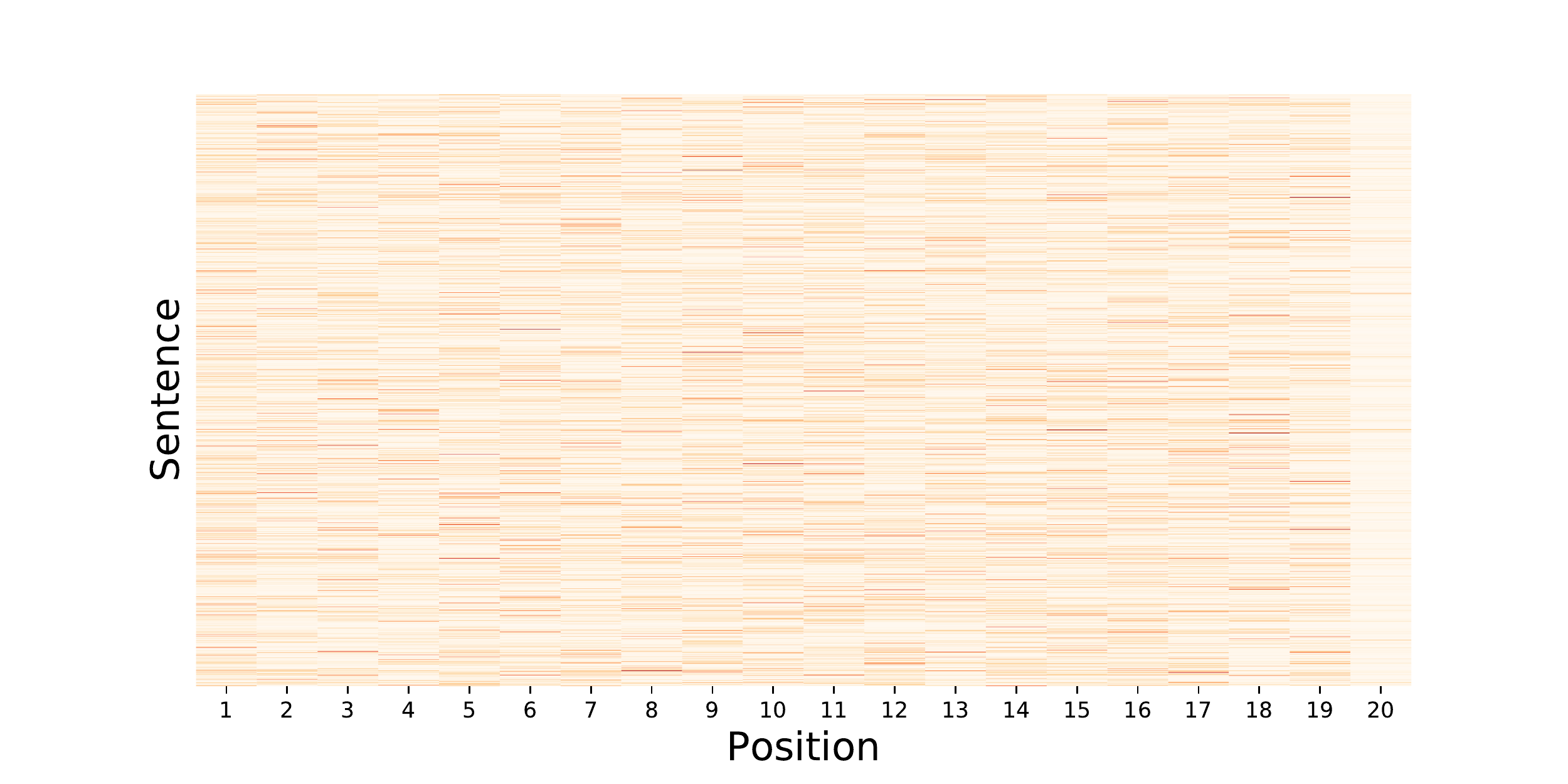}
\caption{Prediction losses of 1000 randomly sampled sentences of the same length (20 tokens) from the training data. Darker means the model has less confidence predicting the word. \label{losepos}}
\end{center}
\end{figure}
The prediction loss of the token $i$ in the target sentence is conditioned on the source sentence and the target tokens generated up to the token in position $i$.
As a result, words that occur later in the sentence have more contextual evidence than words appearing earlier and this could lead to better prediction. 
We examine whether the position of the token in the sentence is a notable factor in prediction loss values.

We randomly sample 1000 sentences of the same length (20 tokens) from the training data and observe the confidence of the model in the prediction. 
Figure~\ref{losepos} shows the spread of prediction loss values in each position in the sentence. 
The only distinct pattern we observe is that the last position has consistently low prediction loss.
This is expected since the end-of-sentence symbol always follows the end of sentence markers, such as `.', `!', or `?'.
We observe that the average loss values are slightly higher for the first position ($12\%$ higher). 
However, other positions have very similar average losses.  
We conclude that the position in the sentence is not a significant factor in individual prediction losses.

\section{Conclusion} \label{btconc}

Motivated by our observations in Chapter~\ref{chapter:research-02} that synthesizing new context is useful for translation of rare words, we further explored in this chapter the impact of additional contexts on the translation of words that are difficult to predict by the baseline model. 
We asked:
 
\begin{enumerate}[label=\textbf{RQ2.\arabic* },wide = 0pt, leftmargin=2em]
\setlength\itemsep{1em}
 \setcounter{enumi}{2}
\item \acl{rq:bt1}

\medskip

\noindent In this chapter, we explored different aspects of the back-translation method to gain a better understanding of its performance.
Our analyses showed that the quality of the synthetic data has a small impact on the effectiveness of back-translation once there is sufficient training data available.
However, the ratio of synthetic to real training data plays a more critical role.
When the ratio of synthetic to real data is high, the model becomes biased towards noise in the synthetic data and the quality decreases.

Next, we examined the NMT model and found that words with high prediction losses after training benefit the most from additional back-translated data.
%We observed that words with high prediction losses in the original model undergo the most changes after training with synthetic data. 
While individual prediction losses are not a distinctive factor in identifying difficult words and same words in very similar contexts have high variance of prediction loss, by averaging these values we can successfully spot difficult words.
Our findings showed that, when original model has a low confidence in predicting words, the addition of contexts for those words to the training data increases the overall accuracy of the model on the unseen test set. %conc

\noindent 
Equipped with this information, we asked:

\item \acl{rq:bt2}

\medskip

\noindent As an alternative to random sampling target sentences for back-translation, we proposed targeted sampling and specifically targeted words that are difficult to predict.
%We proposed several variants of using the prediction loss in identifying relevant sentences to back-translate.
%We also used the contexts of difficult words as a feature to sample sentences for back-translation.
We found that data augmentation with the goal of increasing contexts for difficult-to-predict words improved the translation quality in German$\leftrightarrow$English. % by up to 1.7 {BLEU} points.
Interestingly, the proposed frequency-based sampling approach is a simple, yet effective strategy that is hard to outperform.
This indicates that signals from the data distribution are on par with signals from the failures of the model.

\end{enumerate}

 \noindent  This allows us to answer our more general question: 

\paragraph{Research Question 2:} \acl{rq:tdabt} 

\medskip

 \noindent In this chapter, we continued our study on the influence of having diverse contexts on translation quality. 
 We found that translation quality improves when we diversify the context of difficult words.
%In particular, we focused on the back-translation method and the distinct synthetic contexts that are generated with a reverse NMT model. 
%Despite the fact that back-translation has become a standard method in current NMT models, there is little research on understanding how models benefit from synthetic data.  
We investigated the effective method of back-translation for NMT and explored alternatives to the typically used random selection of target sentences that are to be back-translated into the source language.

\medskip

 \noindent  In Chapters~\ref{chapter:research-02} and~\ref{chapter:research-03}, we studied the impact of the availability of data and why models suffer from {a lack of} diverse contexts during training.
We proposed two main data augmentation approaches with multiple variants targeting different problems in translation.  
Both approaches proposed in Chapters~\ref{chapter:research-02} and~\ref{chapter:research-03} lead to improvements in translation quality.

%We investigate the ability of NMT models to translate fairly% !TEX root = thesis-main.tex

\chapter{Translating Idiomatic Expressions}
\label{chapter:research-04}

\section{Introduction and research questions}

In Chapter~\ref{chapter:research-01}, we experimented with changing the scope of the context from sentence-level to document-level to capture the meaning of ambiguous words. 
In Chapters~\ref{chapter:research-02} and \ref{chapter:research-03}, we demonstrated that added context significantly help translating rare and difficult words.
The next question we are interested in is which phenomena we still \textit{fail} to capture with current approaches to using contexts.
Neural machine translation has achieved substantial improvements in translation of different linguistic phenomena over traditional rule-based and phrase-based models. 
For instance, reordering, subject-verb agreement, double-object verbs, and overlapping subcategorization are various areas where neural models successfully overcome the limitations of phrase-based models \citep{isabelle2017challenge,bentivogli-EtAl:2016:EMNLP2016}.

In this chapter, we examine which phenomena are not fully captured by current NMT models.
NMT models use both source and target sentences as contexts to generate a target word.
We are interested in cases where this scope is not sufficient for the NMT model. 
To shed light on this vulnerability of current NMT models, we ask:

\paragraph{Research Question 3:} \acl{rq:vol} 

\medskip

 \noindent We study the ability of NMT models to translate fairly complex linguistic phenomena. 
To examine this question, in this chapter, we focus on the translation of units that possibly require cues beyond the literal context, namely idiomatic expressions. 
Idioms, a category of multiword expressions, are an interesting language phenomenon where the overall meaning of the expression cannot be inferred from the meanings of its parts.
For the most part, idiom acquisition for humans requires additional resources such as explicit definitions of the expressions.  
NMT models, however, only have access to the nearby and local context of the idiomatic expression. 

To further investigate why neural translation models struggle in this area, we ask:

\begin{enumerate}[label=\textbf{RQ3.\arabic* },wide = 0pt, leftmargin=2em]
\setlength\itemsep{1em}
\item \acl{rq:id1}

\medskip

\noindent The first challenge for learning and evaluating idiom translation is the lack of dedicated data sets.
In this chapter, we address this problem by creating the first large-scale data set for idiom translation. % for German$\leftrightarrow$English . 
Building a hand-crafted data set for idiom translation is costly and time-consuming.
In this chapter, we automatically build a new bilingual data set for idiom translation 
extracted from an existing general-purpose German$\leftrightarrow$English parallel corpus.
The first part of our data set consists of 1,500 parallel sentences where the German side contains an idiom, while the second part consists of 1,500 parallel sentences where the English side contains an idiom.
Additionally, we provide the corresponding training data sets for German$\rightarrow$English and English$\rightarrow$German translation where source sentences including an idiom phrase are identified. 

\end{enumerate}

\noindent We then study how this data set can aid in assessing the translation quality of idiomatic expressions, thus asking: 

\begin{enumerate}[label=\textbf{RQ3.2 },wide = 0pt, leftmargin=2em]

\item \acl{rq:id2}

\medskip

\noindent Having prepared the idiom translation training and test data, we investigate how to assess the translation quality of idiomatic expressions. 
The labels in our data are indicators of the existence of idioms in a sentence. 
We use these labels as an additional signal during training of the NMT model and examine whether this flag is sufficient in identifying a phrase as idiomatic and translating it correctly. 
Finally, we introduce several metrics to evaluate the translation quality of idiom phrases in a sentence. 

\end{enumerate}

\paragraph{Organization.} This chapter is organized as follows: 
In Section~\ref{idrel}, we provide an overview of existing work on idiom identification and translation. 
Next, in Section~\ref{idiomdata} we introduce our data collection procedure and details on the extracted training and test data. 
Section~\ref{idexperiments} describes the design of the experiments for translating idioms.
Section~\ref{idevalus} proposes various metrics to locally evaluate idiom translation and provides experimental results on the translation task and analyzes the performance. 
Finally, we discuss the conclusions and implications of this work in Section~\ref{idconc}.

\section{Idiomatic expressions} \label{idrel}

Non-compositional multiword expressions, or idioms, are lexical semantic units where the meaning is often not merely a function of the meaning of its constituent parts \citep{10.2307/416483,doi:10.1093/applin/17.3.326}.

The non-compositionality characteristic of idiomatic expressions exists to different degrees in a language \citep{10.2307/416483}.
In English for example, for the idiom \textit{"spill the beans"}, the word \textit{`spill'} symbolizes \textit{`reveal'} and \textit{`beans'} symbolizes the \textit{`secrets'}. 
For the idiomatic expression \textit{"kick the bucket"}, on the other hand, no such analysis is possible.
Automatically identifying these idiomatic expressions in a sentence is challenging. 
In the following section, we discuss previous works in this area.

\subsection{Idiom identification} 

Expressions that potentially have idiomatic meanings can be recognized using various lexical association measures \citep{evert-krenn-2001-methods,evert-kermes-2003-experiments}.
However, other methods are necessary to decide whether a particular multiword expression (MWE) has an idiomatic use in a particular context.
\citet{katz-giesbrecht-2006-automatic} use distributional semantics as a model of context similarity to examine whether the local context of an MWE can distinguish its idiomatic use from its literal use.
\citet{salehi-cook-2013-predicting} use the translation of the components of the MWE in multiple languages to compute similarities between strings. This {compositionality score} illustrates the relative degree of compositionality of the MWE.
\citet{salehi-etal-2015-word} implement a similar approach but uses word embeddings to compute the compositionality score of an MWE.

 \citet{salton-etal-2016-idiom} use skip-thought vectors, sent2vec, first introduced by \citet{NIPS2015_5950} for idiom classification. 
 In this approach, they define the classes as to whether an MWE is used literally or idiomatically. 
More recently, \citet{klyueva-etal-2017-neural} propose to use an RNN that predicts the possible tags of an MWE. 
The system scored better in more `syntactic' MWEs like inherently reflexive verbs, light verb constructions, and verb-particle constructions.
However, they were not able to detect idioms with reasonable accuracy.

\subsection{Idiom translation}

Automatic translation of idiomatic phrases is a long-established problem in NLP \citep{Schenk:1986:IRM:991365.991458}.
As we illustrated in previous chapters, NMT models battle with translating rare words.
In a way, idioms are similar to this problem. 
While the occurrence of the expression might not be rare, the idiomatic meaning of the expression in a particular context is often uncommon \citep{salton-etal-2014-evaluation,isabelle2017challenge,agrawal-etal-2018-beating}. 
\begin{table}[htb!]
\centering
\small
\caption{Example of an idiomatic phrase in German and its translation. We compare the output of state-of-the-art commercial models (DeepL and GoogleNMT), as well as our trained model (based on OpenNMT). In translating a sentence containing this idiomatic phrase, we notice that none capture the idiom translation correctly. \label{examplesinc}}
\begin{tabularx}{0.8\linewidth}{p{2.7cm} X}
 \toprule
German phrase & \textit{eine wei{\ss}e Weste haben}  \\
 Literal translation & to have a white vest \\
 Idiomatic translation  &  to have clean slate  \\
	\midrule
 Sentence & Coca-Cola und Nestl{\'e} geh{\"o}ren zu den Unterzeichnern. Beide \textbf{haben} nicht gerade \textbf{eine wei{\ss}e Weste}. \\
  Reference translation & Coca Cola and Nestl{\'e} are two signatories with \textit{"spotty" track records}.\\
  \midrule
 DeepL   & Coca-Cola and Nestl{\'e} are among the signatories. Neither of them is \textbf{exactly the same}.  \\
GoogleNMT  & Coca-Cola and Nestl{\'e} are among the signatories. Both do not \textbf{have just a white vest}.  \\
OpenNMT & Coca-Cola and Nestl{\'e} are among the signatories. Both don't \textbf{have a white essence}.  \\
\bottomrule
\end{tabularx}
\end{table}
The challenge of translating idiomatic phrases in NMT is partly due
to the underlying complexity of identifying a phrase as idiomatic and generating its correct non-literal translation, and partly due
to the fact that idioms are rarely encountered in the standard data sets used for training NMT systems. 

As an example, in Table~\ref{examplesinc}, we provide an idiomatic expression in German and the literal and idiomatic translations in English.
We note that the literal translation of an idiom is not the correct translation; neither does it capture part of the meaning. 
To illustrate the problem of idiom translation we also provide the output of three
NMT systems for this sentence: GoogleNMT \citep{wu2016google}, DeepL\footnote{\url{www.deepl.com/translator}}, and the OpenNMT implementation \citep{2017opennmt} based on \citet{DBLP:journals/corr/BahdanauCB14} and \citet{luong:2015:EMNLP} trained on WMT17 parallel corpora. 
All systems fail to generate the proper translation of the idiomatic expression. 
This problem is particularly pronounced when the source idiom is very different from its equivalent in the target language, as is the case here.

Although there are a number of monolingual data sets available for identifying idiomatic expressions \citep{muzny2013automatic,markantonatou2017proceedings}, there is limited work on building a parallel corpus annotated with idioms, which is necessary to investigate this problem more systematically.  
\citet{salton-ross-kelleher:2014:HyTra} selected a small subset of 17 English idioms, collected 10 sentence examples for each idiom from the Internet, and manually translated them into Brazilian-Portuguese to use for translation.
\citet{isabelle2017challenge} built a challenge set of 108 short sentences that each focus on one difficult phenomenon of the language.
Their manual assessment of the eight sentences containing an idiomatic phrase showed that NMT systems struggle with the translation of these phrases.

\citet{shao-etal-2018-evaluating} introduced a new evaluation metric for detecting literal translation errors in Chinese$\rightarrow$English translation, and concluded that idiom translation remains an open problem in MT.
\citet{moussallem-etal-2018-lidioms} released a multilingual resource on idioms currently containing five languages: English, German, Italian, Portuguese, and Russian. In this work, the authors built the data set by crawling various sources and then have them manually evaluated by native speakers.

While these approaches are valuable for studying the problem of idiom translation, they each require manual efforts to identify and label idioms.
To further research in idiom translation, we still need \textit{large-scale} training and testing resources which are hard to obtain with manual labeling.  

\section{Data collection} \label{idiomdata}

In this section, we introduce our proposed data collection procedure for building a training and test set.
We focus on German$\leftrightarrow$English translation of idioms. 
This is an established language pair commonly used in machine translation literature.
Automatically identifying idiomatic phrases in a parallel corpus requires a gold standard data set annotated manually by linguists. 
We use an online dictionary containing idiomatic and colloquial phrases\footnote{\url{www.dict.cc}}, which is built manually, as our gold standard for extracting idiom phrase pairs. 

Examining the WMT German$\leftrightarrow$English test sets from 2008 to 2016 \citep{bojar-EtAl:2017:WMT1}, we observe very few sentence pairs containing an idiomatic expression. 
The standard parallel corpora available for training however contain a sizeable number of such sentence pairs.
Therefore, we automatically select sentence pairs from the training corpora where the source sentence contains an idiom phrase to build the new test set.
Note that we only focus on idioms on the source side and we have two separate lists of idioms for German and English.
Hence, we independently build two test sets (for German idiom translation and English idiom translation) with different sentence pairs selected from the parallel corpora.

 \begin{table}[htb!]
\centering
\small
\caption{Two examples displaying different constraints of matching an idiom phrase with occurrences in the sentence. \label{two}}
\begin{tabularx}{0.85\linewidth}{@{\ }l @{\ \ \ }X@{\ }}
 \toprule
German idiom & \textit{alles {\"u}ber einen kamm scheren}  \\
English equivalent & to measure everything by the same yardstick \\
Matching German sentence & Aber man kann eben nicht \textbf{alle} Inseln \textbf{\"{u}ber einen Kamm scheren}.\\
English translation &  But we cannot measure all islands by the same standards.\\ 
 \midrule
German idiom & \textit{in den kinderschuhen stecken}  \\
English equivalent & to be in the fledgling stage \\
Matching German sentence & Es \textbf{steckt} immer noch \textbf{in den Kinderschuhen}. \\
English translation &  It is still in its infancy. \\
\bottomrule
\end{tabularx}
\end{table}

Depending on the language, the words making up an idiomatic phrase are not always contiguous in a sentence. 
For instance, in German, the subject can appear between the verb and the prepositional phrase making up the idiom. 
German also allows for several re-orderings of the phrase.

In order to generalize the process of identifying idiom occurrences, we lemmatize the phrases and consider different re-orderings of the words in the phrase as an acceptable match. 
%Table~\ref{exmp} showcases an examples of this match in the corpora.
We also allow for a fixed maximum number of words to occur in between the words of an idiomatic phrase.
%In table~\ref{exmp}, the second example showcases a sentence pair matched with this constraint.
Table~\ref{two} shows two examples of idiom occurrences that match these criteria.
Following this set of rules, we extract sentence pairs containing idiomatic phrases and create a set of sentence pairs for each unique idiom phrase.

There are various ways of combining regular and idiomatic sentences and building training and test data. 
We know that the NMT model is capable of translating a word correctly at test time if it has observed it at training time.
In the previous chapters, we showed that the frequency of occurrences in the training data and the quality of the contexts are important factors in helping the model learn to translate. 
Motivated by this, we distribute sentences with idiomatic phrases between training and test sets so that there are no idioms in the test set that we have not {seen} during training.
%It is preferable if there are a sufficient number of occurrences during training. 
We also make sure that there is no overlap between the training and test sets. 

\begin{figure*}[hbt!]
\centering
\includegraphics[width=0.95\linewidth]{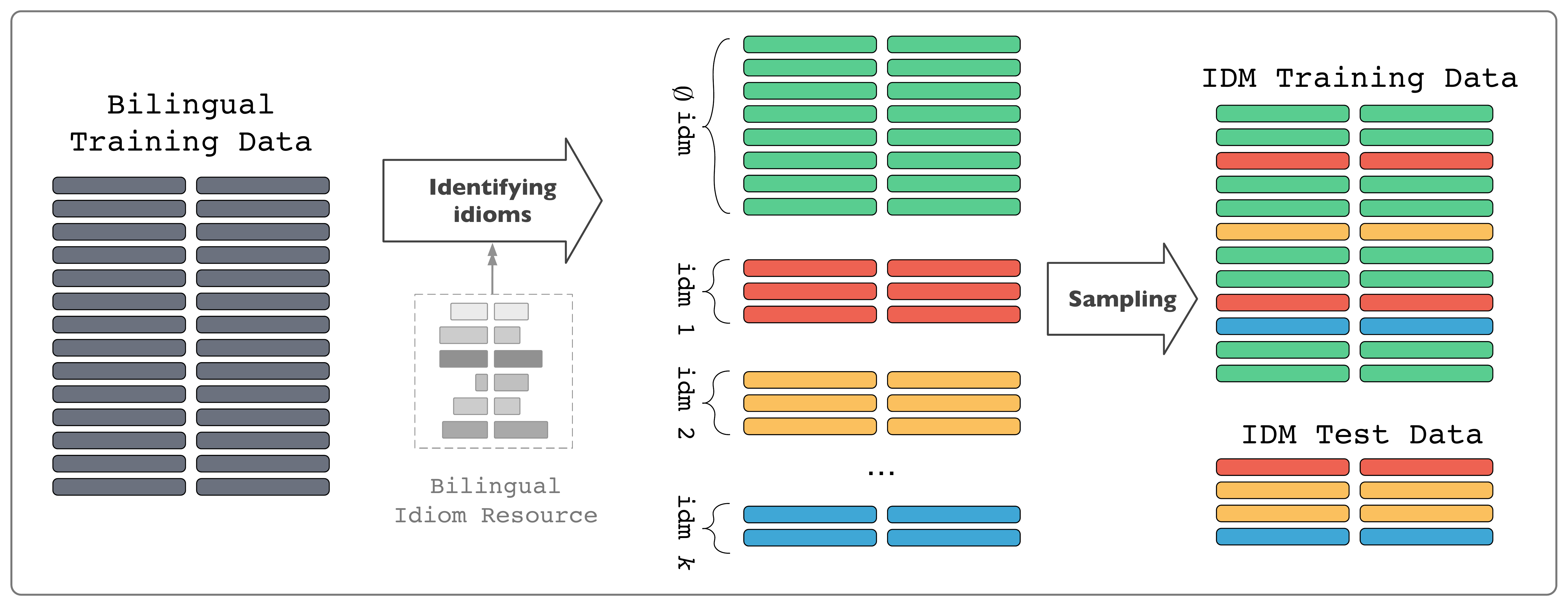}
\caption{The process of data collection and construction of the test set containing only sentence pairs with idiomatic phrases.}
\label{augfig2}
\end{figure*}

 \begin{table}[htb!]
\centering
\small
\caption{Statistics of the constructed German and English idiom translation data sets. %Sentence pairs are based on the training and test sets. 
\label{stats}}
\begin{tabular}{l r}
 \toprule
  German idiom translation data set & \\
  \midrule
Number of unique idioms & 103   \\
  Training size & 4.5M \\   
  Idiomatic sentences in training data & 1848 \\
 Test size & 1500  \\
\toprule
  English idiom translation data set & \\
  \midrule
Number of unique idioms &  132  \\
  Training size & 4.5M \\
  Idiomatic sentences in training data & 1998  \\
 Test size & 1500  \\
\bottomrule
\end{tabular}
\end{table}

Considering these principles, we build the training and test data as follows:
First, we sample without replacement from WMT data sets and select individual sentence pairs to build the idiom test set.
To build the new training data, we use the remaining sentence pairs in each idiom set as well as the sentence pairs from the original parallel corpora that did not include any idiomatic phrases.
In this process, we ensure that for each idiomatic expression there is at least one occurrence in both training and test data and that no sentence pair is included in both training and test data. 

  \begin{table*}[htbp]
  \rotatebox{90}{
\begin{minipage}{\textheight}
\begin{center}
% \begin{center}
 \caption{Examples from the German idiom translation test set.  \label{biggest}}
 \begin{tabularx}{0.85\linewidth}{lX}
       \toprule
       German idiom & \textit{in den kinderschuhen stecken} \\
       English equivalent &  to be in the fledgling stage   \\
       German sentence &     Eine Bemerkung, Gentoo/FreeBSD \textbf{steckt} noch \textbf{in den Kinderschuhen} und ist kein auf Sicherheit achtendes System. \\
       English sentence &     Note that Gentoo/FreeBSD is still \textbf{in its infancy} and is not a security supported platform. \\
       \midrule
       German idiom & \textit{den kreis schlie{\ss}en}   \\
       English equivalent  &  to bring sth. full circle      \\
       German sentence  & Die europ{\"a}ische Krise \textbf{schlie{\ss}t den Kreis}.\\
       English sentence  & The European crisis is \textbf{coming full circle}. \\
\midrule
       German idiom &    \textit{auf biegen und brechen} \\
       English equivalent   & by hook or crook    \\
       German sentence  & Nehmen wir zum Beispiel die W{\"a}hrungsunion: Sie soll \textbf{auf Biegen und Brechen} eingef{\"u}hrt werden.\\
       English sentence  & Take, for example, the introduction -\textbf{come what may}- of the single currency.\\
\midrule
       German idiom &  \textit{sie haben das wort} \\
       English equivalent  &   the floor is yours    \\
       German sentence  & Berichterstatterin. - (FR) Herr Pr{\"a}sident! Danke, dass \textbf{Sie} mir \textbf{das Wort} erteilt \textbf{haben}.\\
       English sentence  & rapporteur. - (FR) Mr President, thank you for \textbf{giving} me \textbf{the floor}.\\
\bottomrule
 \end{tabularx}
 \end{center}
\end{minipage}
}
 %\end{center}
 \end{table*}
 
Figure~\ref{augfig2} visualizes the process of constructing the new training and test sets.
As a result of this construction, for each language direction, we obtain a targeted test set for idiom translation and the corresponding training corpus representing a natural distribution of sentences with and without idioms.
We annotate each sentence pair with the canonical form of its source-side idiom phrase and its equivalent in the target language. 

Table~\ref{stats} provides some statistics of the two data sets. 
For each unique idiom in the test set, we also provide the frequency of the respective idiom in the training data. 
Note that this is based on the lemmatized idiom phrase under the constraints mentioned in Section~\ref{idiomdata} and is not necessarily an exact match of the phrase.
Table~\ref{biggest} shows several examples from the data set for German idiom translation. 
We observe that for some idioms the literal translation in the target language is close to the actual meaning, while for others it is not the case.

Note that multiword expressions that at times have an idiomatic meaning can also be translated literally depending on the context (e.g., \textit{"spill the beans"} to literally describe the act of \textit{spilling the beans}). 
This data set represents this additional difficulty: Models cannot just memorize fixed translation of idioms but also have to consider the specific context in which they are used.

\section{Translation experiments} \label{idexperiments}
    
While the main focus of this chapter is to generate data sets for training and evaluating idiom translation, we also perform a few NMT experiments using our data set to measure the problem of idiom translation on large-scale data. 

In the first experiment, following the conventional settings, we do not use any labels indicating whether a particular phrase is used idiomatically or not in the training data.
In the second experiment, we use the labels in the training data as an additional feature to investigate the effect of informing the model of the existence of an idiomatic phrase in a sentence during training.
We perform %two 
a German$\rightarrow$English experiment by providing the model with additional input flags.
This approached is similar to the work by \citet{sennrich-etal-2016-controlling}, where they control the honorifics produced at test time by adding a side constraint to the source side. 

The additional flag indicates whether a source sentence contains an idiom
and are implemented as a special extra token \texttt{<idm>} that is prepended to each source sentence containing an idiom both in the training and test data.
This a simple approach that can be applied to any sequence-to-sequence architecture. 

%Most NMT systems have a sequence-to-sequence architecture where an encoder builds up a representation of the source sentence and a decoder, using the previous LSTM hidden states and an attention mechanism, generates the target translation \citep{DBLP:journals/corr/BahdanauCB14,sutskever2014sequence,cho2014properties}.
We use a 4-layer attention-based encoder-decoder model as described in Section~\ref{RNN} trained with hidden dimension size of 1000, and batch size of 80 for 20 epochs. 
In all experiments, the NMT vocabulary is limited to the most common 30K words in both languages and we preprocess source and target language data with Byte Pair Encoding (BPE) \citep{sennrich-haddow-birch:2016:P16-12} using 30K merge operations. 
We also use a phrase-based translation system similar to Moses \citep{koehn-etal-2007-moses} as baseline to measure PBMT performance for idiom translation.
Several examples of our idiom translation test set and the output translations of the PBMT and NMT models are illustrated in Table~\ref{smtnmt}.

  \begin{table*}[htbp]
%  \rotatebox{90}{
%\begin{minipage}{\textheight}
\begin{center}
% \begin{center}
 \caption{Examples from the resulting test set of sentence pairs containing idiomatic expressions. NMT and PBMT translations of sentences are provided, highlighting the challenge of idiom translation.  \label{smtnmt}}
 \begin{tabularx}{0.95\linewidth}{lX}
       \toprule
%       \textsc{src} & Alles, was ich brauche, finde ich \textbf{vor Ort}. \\
   %    \textsc{ref} & I can find everything I need \textbf{locally}. \\
      % PBMT  & everything I do, I think, \underline{on the ground}. \\
       %NMT  &  All I need is \underline{on site}. \\
       %\midrule
 \textsc{src} &   Seitdem aber begannen sich zwischen Pol\'{i}vka und Harabi\v{s} die Streitigkeiten zu h\"{a}ufen, die in der Absetzung ``K\"{o}nig Boleslavs I. gew\"{a}hlt \textbf{bis zum Sankt-Nimmerleins-Tag}'' gipfelten. \\
       \textsc{ref} &  From then on , quarrels begin to accumulate between Pol\'{i}vka and Harabi\v{s} , which culminated in the dethronement of ``the king Boleslav I elected \textbf{forever and ever}''.   \\
       PBMT  &  Since then, however, the disputes between Pol\'{i}vka and Harabi\v{s} began to accumulate, culminating in the departure of King Boleslavs I.   \\
       NMT  &   Since then, but began between Pol\'{i}vka and Harabi\v{s} disputes to accumulate in the removal ``king Boleslavs I. elected by the \underline{Sankt-Nimmerleins-Tag} culminated''.    \\
              \midrule
 \textsc{src} &   Sie wurde \textbf{vor Ort} not\"{a}rztlich behandelt und von Rettungskr\"{a}ften in ein Krankenhaus gebracht. \\
       \textsc{ref} &   She was treated \textbf{at the site} by an emergency doctor and taken to hospital by ambulance.  \\
       PBMT  &  It was treated \underline{on site in the field}, and it was brought to a hospital from the rescue forces.   \\
       NMT  &   she was \underline{on the ground} and not\"{a}rztlich treated by rescue workers in a hospital.    \\
              \midrule
 \textsc{src} & Janson ist selbst ein \textbf{alter Hase} in seinem Metier, der Schauspielkunst.   \\
       \textsc{ref} &   Janson is an \textbf{old hand} himself when it comes to his profession, the art of acting.  \\
       PBMT  &   Janon himself is an \underline{old hase} in his painter, the artistic art.  \\
       NMT  &    Janson itself is an \underline{old hand} in his subjects, the schauspielkunst.   \\
       \midrule
     \textsc{src} &   Mit unserer Mitteilung vom letzten Sommer haben wir \textbf{den Stein ins Rollen} gebracht und demonstriert, dass Europa an der Erarbeitung eines internationalen Instruments beteiligt ist. \\
       \textsc{ref} & Our communication of last summer enabled us to \textbf{get things up and running} and to demonstrate that Europe was participating in the drawing up of an international instrument. \\
       PBMT  &    With our communication of last summer we \underline{have the ball rolling} and demonstrated that Europe in the drafting of an international instrument is involved. \\
       NMT  &   With our communication last summer, we \underline{launched the stone} and demonstrated that Europe is involved in the development of an international instrument. \\
\bottomrule
 \end{tabularx}
 \end{center}
 \end{table*}
 
\section{Idiom translation evaluation} \label{idevalus}

Ideally, idiom translation should be evaluated manually, but this is a very costly process.
%one should do manual evaluation of idiom translations, however since it is costly, in this work we consider using automatic metrics. 
Automatic metrics, on the other hand, can be used on large data sets at no cost and have the advantage of replicability (Section~\ref{bgexp}).
We use three metrics to evaluate the translation quality with a specific focus on idiom translation accuracy: BLEU, Modified Unigram Precision, and Word-level Idiom Accuracy. We describe each metric below.

\subsection{BLEU} The traditional BLEU score \citep{Papineni2001}, discussed in Section~\ref{bgexp}, is a good measure to determine the overall quality of the translations. 
However, this measure considers the precision of \textit{all} $n$-grams  in a sentence and by itself does not focus on the translation quality of the idiomatic expressions.

\subsection{Modified unigram precision} 
To specifically concentrate on the quality of the translation of idiomatic expressions, we also look at the \textit{localized} precision. 
In this approach, we translate the idiomatic expression in the context of a sentence and only evaluate the translation quality of the idiom phrase. 

To isolate the idiom translation in the sentence, we look at the word-level alignments between the idiomatic expression in the source sentence and the generated translation in the target sentence. 
We use \texttt{fast-align} \citep{dyer-chahuneau-smith:2013:NAACL-HLT} to extract word alignments.
Since idiomatic phrases and the respective translations are not contiguous in many cases, we only compare the unigrams of the two phrases.
We compute unigram matches between the reference translation and the translation output, the $candidates$ set, as follows:
\begin{equation} 
UniPrec = \frac{ \sum\limits_{\vt{C} \in \{Candidates\}}^{} \sum\limits_{{\unigram} \in \vt{C}}^{} \countop_{clip}({\unigram}) }{  \sum\limits_{\vt{C'} \in \{candidates\}}^{} \sum\limits_{{\unigram }' \in \vt{C'}}^{} \countop({\unigram}')  }
\end{equation}

\noindent where $\countop_{clip} = \min(\countop, \textit{max\_ref\_count})$.
By computing the clipped count, we truncate each word's count so that it does not exceed the largest count observed in a reference for that word.
Note that for this metric %and \textit{Word-level Idiom Accuracy} 
we have two references: The idiom translation as an independent expression, and the human-generated idiom translation in the target sentence. 

\subsection{Word-Level idiom accuracy} We also use another metric to evaluate the word-level translation accuracy of the idiom phrase. 
We use word alignments between source and target sentences to determine the number of correctly translated words.
We use the following equation to compute the accuracy:
\begin{align} 
WAcc = \frac{H-I}{N}
\end{align} 

\noindent where $H$ is the number of correctly translated words, $I$ is the number of extra words in the idiom translation, and $N$ is the number of words in the gold idiomatic expression.

\smallskip

\begin{table*}[htb!]
\begin{center}
\caption{Translation performance on the German idiom translation test set. \textit{Word-level Idiom Accuracy} and \textit{Unigram Precision} are computed only on the idiom phrase and its corresponding translation in the sentence.   \label{numbers}}
\begin{tabularx}{0.72\textwidth}{@{\extracolsep{4pt}} l cccc}
\toprule
 & \multicolumn{1}{c}{\bf WMT08-16}  & \multicolumn{3}{c}{\bf Idiom test set}\\ \cline{2-2} \cline{3-5}
    \bf Model &  BLEU  &   BLEU  &    UniPrec &  WAcc  \\ 
     \toprule
PBMT baseline & 20.2  & 19.7  &  57.7 & 71.6 \\
NMT baseline   &  26.9  &    24.8       &  53.2 &   67.8  \\
NMT  \textsc{src} flag  & 25.2  &  22.5       & 64.1 & 73.2     \\
NMT \textsc{tgt} flag  & 17.8  & 16.2           &   54.3 &  64.0  \\
      \bottomrule
\end{tabularx}
 \end{center}
\end{table*}

\subsection{Evaluation results}
Table~\ref{numbers} presents the results for the translation task using different metrics. 
Looking at the overall BLEU scores, we observe that baseline performance on the idiom-specific test set is lower than on the union of the standard test sets (WMT 2008-2016). 
While the scores on these two data sets are not directly comparable, this result is in line with previous findings that sentences containing idiomatic expressions are harder to translate \citep{isabelle2017challenge}.
We can also see that the performance gap is %However, this is 
not as pronounced for a PBMT system, suggesting that phrase-based models are capable of \textit{memorizing} the idiomatic phrases to some extent.

The NMT model using a special input flag to indicate the presence of an idiom in the source sentence performs better than PBMT but slightly worse than the NMT baseline in terms of BLEU.
Despite this drop in BLEU performance, by examining the \textit{unigram precision} and \textit{word-level idiom accuracy} scores, we observe that this model generates more accurate idiom translations.
When comparing this to having the idiom flag on the source or target side, we observe a significant difference:
The experiment with the target flag performs the worst, partially because during inference, only the source sentence is available and hence there is no contextual signal to aid the model.
 
 \begin{figure*}[ht!]
\centering
\begin{subfigure}[h]{0.495\textwidth}
\includegraphics[width=\linewidth]{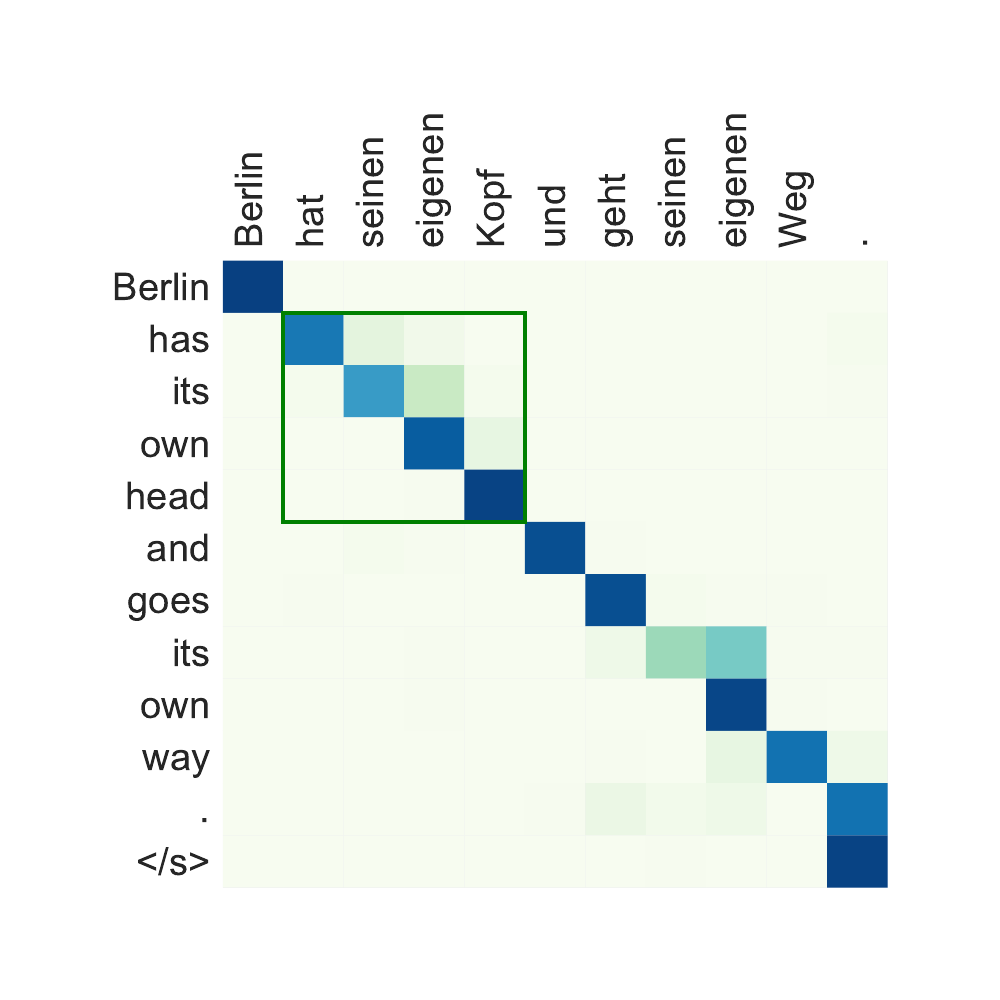}
\end{subfigure}
\begin{subfigure}[h]{0.495\textwidth}
\includegraphics[width=\linewidth]{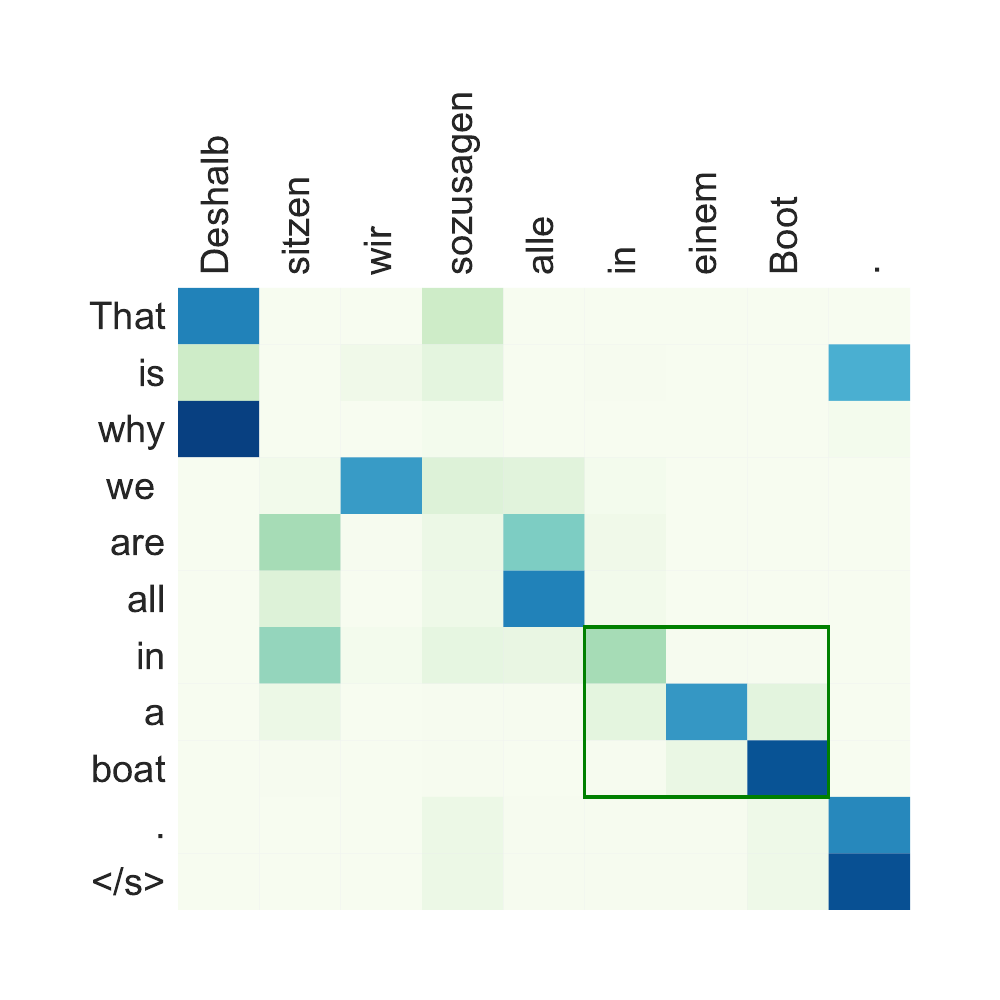}
%\caption{Attention matrix for translation of a German sentence. The blocked area marks the idiomatic expression and its generated translation. The reference translation is: \textit{``x''}\label{att2}}
\end{subfigure}
\caption{Attention visualization of the translation of two sampled German sentences. Darker color means higher weight. The blocked area marks the idiomatic expression and its generated translation. The reference translations are: (left) \textit{``Berlin \underline{has a mind of its own} and is doing its own thing.''} and (right) \textit{``We are therefore all \underline{in the same boat}, so to speak.''} \label{att2}}
\end{figure*}

Figure~\ref{att2} illustrates the attention distribution of the NMT model during translation of an example German sentence. 
We expected that in order to translate each word in the idiomatic expression correctly, the model would pay a noticeable degree of attention to the other words in the expression. 
However, we see that it does not happen and the model essentially translates the sentence word by word, i.e., literally. 
%We suspect that for idiom translation to be successful, NMT models need a stronger signal from context.
%
These preliminary experiments reiterate the problem of idiom translation with neural models, and in addition show that with a labeled data set, we can devise simple models to address this problem to some extent.

\section{Conclusion} \label{idconc}

Motivated by our observations in Chapters~\ref{chapter:research-02} and \ref{chapter:research-03} that illustrate the importance of local context on learning difficult words, we further explored in this chapter some shortcomings of current models.
Concretely, we were interested in finding cases where NMT models are unsuccessful.
One case of a complex language phenomenon is idiom translation where it is very difficult to infer the meaning of the phrase without explanation and only from the observed context.

To investigate the behaviour of NMT models in translating idiomatic expressions, we asked:

\begin{enumerate}[label=\textbf{RQ3.\arabic* },wide = 0pt, leftmargin=2em]
\setlength\itemsep{1em}
\item \acl{rq:id1}

\medskip

\noindent We identified two main challenges when translating idiom phrases, namely lack of dedicated data sets and lack of targeted evaluation metrics.
%As an essential step towards assessing idiom translation quality in neural models, we require an explicitly tailored test data.
To address this problem, we have extracted a parallel data set for training and testing idiom translation for German$\rightarrow$English and English$\rightarrow$German.
In the test sets, we included sentences with at least one idiom on the source side.
In the training set, we included a mixture of idiomatic and non-idiomatic sentences with labels to distinguish between the two.
We release our new data sets which can be used to further investigate and improve NMT performance of idiom translation.
Using our new resources, we performed preliminary translation experiments to evaluate the quality of idiom translation.
Experiments on this test set showed PBMT models scored higher than NMT models on our metrics which explicitly measure idiom translation quality. 

\item \acl{rq:id2}

\medskip

\noindent We observed that even though the NMT model achieved a higher overall BLEU score, it performed worse on idiom translation metrics in comparison with PBMT model.
%This is in agreement with previous work on investigating the weak points of neural models.
Next, we studied whether a flag in the training data can help to distinguish between when a phrase is to be translated literally and when it should be translated idiomatically.
Our experiments showed that adding a side flag during training improves the quality of idiom translation.
However, we found that the BLEU score on standard test sets declined.
Our experiments suggest that there is no correlation between overall BLEU scores and the localized precision of idiomatic phrase translations.

\end{enumerate}

 \noindent It allows us to return to our more general research question:

\paragraph{Research Question 3:} \acl{rq:vol} 

\medskip

 \noindent To answer this question, we specifically examined non-compositional multiword expressions.
%Idiom translation is one of the more difficult challenges of machine translation. 
Since the literal meaning of the components is different from the idiomatic meaning of the entire expression, the model needs to know in which context to translate it literally and in which idiomatically. 
We showed that NMT models perform poorly on idiom translation despite their overall strong advantage over previous MT paradigms. 
We conclude that further research on idiom translation can benefit from having a dedicated data set. 

\noindent In the next chapter, we continue investigating this question by examining cases where there are no complex linguistic phenomena, such as non-compositional phrases, in the observed context.

\chapter{Volatilities of Neural Models}
\label{chapter:research-05}

\section{Introduction and research questions}

In the previous chapters, we first investigated how to enhance the use of context to address some of the shortcomings of neural translation models. 
%Challenges for neural translation models, when sufficient data is not available, include understanding rare words \citep{W18-2712} and domain mismatch between training and testing \citep{koehn2017six,W18-2709}.
Then in Chapter~\ref{chapter:research-04}, we showed that the translation of idiomatic phrases is challenging for the current NMT models. 
We saw that the scope of the observed context was not sufficient to infer the meaning of idioms.
Based on these findings, in this chapter, we continue examining the following question:

\paragraph{Research Question 3:} \acl{rq:vol} 

\medskip

 \noindent 
 While the lack of suitable context exposes shortcomings in current models, we extend our research in this chapter to situations where appropriate data \textit{is} available. 
We first look into the robustness of current translation models.
Namely, we investigate what is the effect of small perturbations of the source sentence on the translation.
Observing that in some cases translations change unexpectedly with these small perturbations, 
we study whether and to what extent it can be replicated and quantified with automatically modified test data. 
Concretely we ask:
 
\begin{enumerate}[label=\textbf{RQ3.\arabic* },wide = 0pt, leftmargin=2em]
\setlength\itemsep{1em}
 \setcounter{enumi}{2}
\item \acl{rq:vol1}

\medskip

\noindent To answer this question, we locally modify sentence pairs in the test set and identify examples where a trivial modification in the source sentence causes an `unexpected change' in the translation.
These modifications are generated conservatively to avoid insertion of any noise or rare words in the data (Section~\ref{secsentvar}). 
Our goal is not to \textit{fool} the NMT models, but instead, to identify common cases where the models exhibit unexpected behaviour and in the worst cases result in incorrect translations. 
We identify these unexpected and erroneous changes in the translation output as a sign of an underlying \textit{volatility} of NMT models. 

\item \acl{rq:vol2}

\medskip

\noindent We investigate to what extent two current state-of-the-art NMT models are robust against changes in the input during inference.
We observe that our modifications expose volatilities of both RNN and Transformer translation models in $26\%$ and $19\%$ of sentence variations, respectively.
Our findings show how vulnerable current NMT models are to trivial linguistic variations, putting into question the generalization abilities of these models. 
%This volatile behavior of translating extremely similar sentences in surprisingly different ways highlights the underlying generalization problem of current NMT models. 

\end{enumerate}

\paragraph{Organization.} The chapter is organized as follows: 
Section~\ref{isitanothernoise} discusses prior works on the impact of noise on the performance of machine translation.
In Section~\ref{secvol}, we provide an example of unexpected behaviour of NMT models and discuss how it is different from the unexpected behaviour when encountering noise in the input text. 
In Section~\ref{secsentvar}, we introduce our sentence variation generation approach and provide details on the experimental settings.
Section~\ref{secvolassess} proposes various metrics to identify and quantify these unexpected changes and provides experimental results on a translation task.
Finally, we discuss the conclusions and implications of this work in Section~\ref{secvolconc}. 

\section{Noisy text translation} \label{isitanothernoise}

Recently, several approaches investigated NMT models when encountering noisy input and how \textit{worst-case examples} of noisy input can `break' state-of-the-art NMT models \citep{D18-1050}. 
Noisy input text can cause mistranslations in most translation systems, and there has been growing research interest in studying the behaviour of translation systems when encountering noisy input \citep{li-EtAl:2019:WMT1}.

\citet{DBLP:journals/corr/abs-1711-02173} show that character-level noise in the input %, for instance swapping random letters of a word, 
leads to poor translation performance.
They propose to swap or randomize letters in a word in the input sentence. For instance, they change the word `\textit{noise}' in the source sentence into `\textit{iones}'.
\citet{halluc} randomly insert words in different positions in the source sentence and observe that in some cases the translations are completely unrelated to the input. 
\citet{D18-1050} propose a benchmark data set for translation of noisy input sentences, consisting of noisy, user-generated comments on Reddit.
The types of noisy input text they observe include spelling or typographical errors, word omission/insertion/repetition, and grammatical errors.

\section{Volatility in machine translation} \label{secvol}

In the discussed works in Section~\ref{isitanothernoise}, the focus of the research is on studying how the translation systems are not robust when handling noisy input text. In these approaches, the input sentences are semantically or syntactically incorrect which leads to mistranslations. 
However, in this chapter, our focus is on input text that does \textit{not} contain any types of noise. We modify input sentences in a way that the outcomes are still syntactically and semantically correct. 
We investigate how translation systems exhibit volatile behaviour in translating sentences that are extremely similar and only differ in one word without any noise injection.
While it is to some extent expected that the performance of NMT models that are trained on predominantly clean but tested on noisy data deteriorates, other changes are more unexpected.

\begin{table}[ht]
\centering
\caption{Insertion of the German word \textit{`sehr'} (English: \textit{`very'}) in different positions in the source sentence results in substantially different translations. Note that all source sentences are syntactically correct and semantically plausible. We use a Transformer model trained on WMT data with 6 encoder and decoder layers and 8 attention heads. $^\dagger$ indicates the original sentence from WMT 2017. \label{first}}
\begin{tabularx}{0.7\textwidth}{lll}
%Source & Translation \\ 
\multicolumn{3}{l}{\textcolor{mygray}{Source:} \textit{Ich bin \underline{\ \ \ \ \ \ }\ \raisebox{-1.5ex}{\tiny{\fbox{$1$}}} \textbf{erleichtert} und \underline{\ \ \ \ \ \ }\ \raisebox{-1.5ex}{\tiny{\fbox{$2$}}} bescheiden.}}\\
%\multicolumn{3}{l}{\textcolor{mygray}{Reference:} \textit{I am relieved and humble.}} \\
\hline
\\[-0.5ex]
\fbox{$1$} &  \fbox{$2$} & \textcolor{mygray}{NMT output} \\
\hline
%\multicolumn{4}{l}{} \\ % & I am relieved and very modest. \\
%\multicolumn{2}{l}{ I am relieved and \textbf{very} modest.} \\
  $\phi$ & $\phi$   & I am \underline{easier} and modest.  \\ 
  $\phi$ & \textit{sehr} $ ^{\dagger}$&  I am \textbf{relieved} and very modest. \\ 
%Ich bin erleichtert und bescheiden. & I am easier and modest. \\
  \textit{sehr} & $\phi$ & I am {very} much \underline{easier} and modest.  \\ 
%Ich bin sehr erleichtert und bescheiden. & I am very much easier and modest. \\
  \textit{sehr} & \textit{sehr} & I am {very} \underline{easy} and {very} modest.   \\ 
%Ich bin sehr erleichtert und sehr bescheiden. & I am very easy and very modest. \\
\\
&& \textcolor{mygray}{Reference}\\
\hline
 $\phi$ & $\phi$   & \textit{I am relieved and humble.}\\
  \textit{sehr} & \textit{sehr} & \textit{I am very relieved and very humble.}   \\ 
\end{tabularx}
\end{table}

In this chapter, we explore unexpected and erroneous changes in the output of NMT models.
Consider the simple example in Table~\ref{first} where the Transformer model 
is used to translate very similar sentences.
Surprisingly, we observe that by simply altering one word in the source sentence---inserting the German word \textit{`sehr'} (English: \textit{`very'})---an unrelated change occurs in the translation.
In principle, an NMT model that generates the translation of the word \textit{`erleichtert'} (English: \textit{`relieved'}) in one context, should also be able to generalize and translate it correctly in a very similar context.
Note that there are no infrequent words in the source sentence and after each modification, the input is still syntactically correct and semantically plausible.
We call a model \textit{volatile} if it displays inconsistent behaviour across similar input sentences during inference.

\section{Variation generation}   \label{secsentvar}

While there are various ways to automatically modify sentences, we are interested in simple semantic and syntactic modifications. % in changing the meaning of the sentence.
These trivial linguistic variations should have almost no effect on the translation of the rest of the sentence.

%\subsection{Classes of Modifications}

%The modifications with the least variations in meaning result in very similar sentences to the original sentence. 
%We are interested in trivial linguistic variations with almost no effect on the translation of the rest of the sentence.
We define a set of rules to slightly modify the source and target sentences in the test data and keep the sentences syntactically correct and semantically plausible. 

%The rules take the form $r = (w_i \rightarrow w_j)$, where the word $w_i$ in the original source sentence is replaced by the word $w_j$.
%Depending on the rule, we also modify the corresponding target sentence. 

\begin{itemize}%[noitemsep,nolistsep]
\item \textbf{Delete}: A conservative approach to modifying a sentence automatically without breaking its grammaticality is to remove adverbs. 
We identify a list of the 50 most frequent adverbs in English and their translations in German.\footnote{Here, we use the \url{dict.cc} online dictionary.}
For every sentence in the WMT test sets, if we find a sentence pair containing both a word and its translation from this list, we remove both words and create a new sentence pair. 
\item \textbf{Insert}: Randomly inserting words in a sentence has a high chance of producing a syntactically incorrect sentence.
To ensure that sentences remain grammatical and semantically plausible after modification, we define a bidirectional n-gram probability for inserting new words as follows:
\begin{equation}
P(w_3 \mid w_1 w_2 w_4 w_5) = \frac{C(w_1 w_2 w_3 w_4 w_5)}{\sum_j C(w_1 w_2 w_j w_4 w_5)}
\end{equation}
$w_3$ is inserted in the middle of the phrase $w_1 w_2 w_4 w_5$, if the conditional probability is greater than a predefined threshold.  
The probabilities are computed on the WMT data.
This simple approach, instead of using a more complex language model, serves our purposes since we are interested in inserting very common words that are already captured by the n-grams in the training data.

\item \textbf{Substitute number}: Another simple yet effective approach to safely modifying sentences % ensuring that we do not inject noise into the test set 
is to substitute numbers with other numbers.
In this approach, we select every sentence pair from the test sets that contain a number and substitute the number $i$ in both source and target sentences with $i+k$ where $1 \leq k \leq 5$.
We choose a small range for change so that the sentences are still semantically correct for the most part and result in a few implausible sentences.
%The new test set is augmented with all of the new sentence pairs.
\item \textbf{Substitute gender}: Finally, a local modification  
is to change the gender of the pronoun in the sentences. 
The goal of this modification is to investigate the existence and severity of gender bias in our models.
This is inspired by recent approaches that have shown that NMT models learn social stereotypes such as gender bias from training data \citep{escude-font-costa-jussa-2019-equalizing,stanovsky-etal-2019-evaluating}.  
\end{itemize}

Note that in a minority of cases, these procedures can lead to semantically incorrect sentences.
For instance, by substituting numbers we can potentially generate sentences such as \textit{''She was born on October 34th``}. 
While this can cause problems for a reasoning task, it barely affects the translation task, as long as the modifications are consistent on the source and target side.

Table~\ref{examplesofpert} shows examples of generated variations. %from each class of modification.
We emphasize that only modifications with local consequences have been selected and we intentionally ignore cases such as \textit{negation} which can result in wider structural changes in the translation of the sentence.

\begin{table}[ht]
\begin{center}\small
\setlength\tabcolsep{4pt} % default value: 6pt
\caption{\label{examplesofpert} Examples of different variations from WMT. [$w_i$\textbackslash$w_j$] indicates that $w_i$ in the original sentence is replaced by $w_j$. ${\phi}$ is the empty string.}
\begin{tabularx}{0.9\columnwidth}{lX}
\toprule
\bf Modification & \bf Sentence variations\\
\midrule
\textit{Delete} &  Some 500 years after the Reformation, Rome  \textbf{[now\textbackslash$\pmb{\phi}$]} has a Martin Luther Square. \\
\textit{Insert} & I loved Amy and she is  \textbf{[$\pmb{\phi}$\textbackslash also]} the only person who ever loved me.  \\
%according to press reports , the party have threatened Trump with removing financial support for his election campaign if he is unable to do any better in the \textbf{opinion} polls . \\
 %\midrule
%\texttt{del} & But despite the smiles for the cameras, few here are convinced-especially [\textbf{now} / $\pmb{\phi}$], just before parliamentary elections.\\
\textit{Subs number} & I'm very pleased for it to have happened at Newmarket because this is where I landed \textbf{[30\textbackslash31]} years ago. \\
\textit{Subs gender} & \textbf{[He\textbackslash She]} received considerable appreciation and praise for this. \\
%but despite the smiles for the cameras , few here are convinced - especially , just before parliamentary elections .\\
\bottomrule
\end{tabularx}
\end{center}
\end{table}

%In the next section we discuss the results and analyze volatilities of the NMT models.
We generate 10K sentence variations by applying these modifications to all sentence pairs in WMT test sets 2013--2018 \citep{bojar-EtAl:2018:WMT1}. %to generate variations of the same sentences. %and for each class of modification different number of new sentences are generated.% (presented in Table~\ref{counts}).
%Note that we generate different number of new sentences for different modification classes due to the nature of the modifications and the available words in the sentences.
% from WMT test sets.
We use RNN and Transformer models to translate sentences and their variations.

\subsection{Experimental setup}

In the translation experiments, we use the standard English$\leftrightarrow$German WMT-2017 training data % and report results on newstest 2013-2018 
\citep{bojar-EtAl:2018:WMT1}.
We perform NMT experiments with two different architectures as described in Sections~\ref{RNN} and~\ref{TRNN}: RNN \citep{luong:2015:EMNLP} and Transformer \citep{vaswani2017attention}.
We preprocess the training data with Byte-Pair Encoding (BPE) using 32K merge operations \citep{sennrich-haddow-birch:2016:P16-12}. %to segment words into subword units for both languages.
Table~\ref{bleus} shows the case-sensitive BLEU scores as calculated by \texttt{multi-bleu.perl}. %on WMT test sets 2016 and 2017.

\begin{table}[ht]
\center
\setlength\tabcolsep{3pt} % default value: 6pt
\caption{\label{bleus} BLEU scores of different baseline models on the WMT news data for translation of German$\leftrightarrow$English.}
\begin{tabular}{@{\extracolsep{4pt}}lcccccc@{}}
\toprule
 & \multicolumn{3}{c}{\bf De-En} &  \multicolumn{3}{c}{\bf En-De} \\ \cline{2-4}  \cline{5-7}   
 &   WMT16 &  WMT17 &  WMT18  &  WMT16 &  WMT17  &  WMT18  \\ \midrule
RNN & 32.5 & 28.2 &35.2 & 28.1  & 22.4 & 34.6 \\ 
Transformer & 36.2 & 32.1 & 40.1 & 33.4 & 27.9& 39.8  \\
\bottomrule
\end{tabular}
\end{table}

\paragraph{RNN} We use a 2-layer bidirectional attention-based LSTM model implemented in OpenNMT \citep{2017opennmt} trained with an embedding size of 512, hidden dimension size of 1024, and batch size of 64 sentences. We use Adam \citep{kingma2014adam} for optimization.

\paragraph{Transformer} We also experiment with the Transformer model \citep{vaswani2017attention} implemented in OpenNMT. We train a model with 6 layers, the hidden size is set to 512, and the filter size set to 2048. The multi-head attention has 8 attention heads. We use Adam \citep{kingma2014adam} for optimization. All parameters are set based on the suggestions by \citet{2017opennmt} to replicate the results of the original paper.

\section{Unexpected and erroneous changes} \label{secvolassess}

The modifications described above generate sentences that are extremely similar %to the original sentences 
and hence are expected to have a very similar difficulty of translation. 
First, we evaluate the NMT models on how robust and consistent they are in translating these sentence variations rather than their absolute quality.
Next, we perform manual evaluation on the translation outputs to assess the impact of unexpected changes on translation quality.  
%To this end, we provide a more detailed analysis on unexpected changes in translations here.

\subsection{Deviations from original translation}\label{ctsec}

The variations are aimed to have minimal effect on changing the meaning of the sentences. 
Hence, major changes in the translations of these variations can be an indication of volatility in the model.
To assess whether the proposed sentence variations result in major changes in the translations, we measure changes in the translations of sentence variations with Levenshtein distance \citep{levenshtein1966binary}. %and \textit{span of change}.
%Specifically, Levenshtein distance measures the edit distance between the two translations. %while span of change indicates the spread of changes in the sentence.
We also use the first and last positions of change in the translations, which represents the span of changes.

\begin{figure*}[tbh!]
\centering
\includegraphics[width=\textwidth]{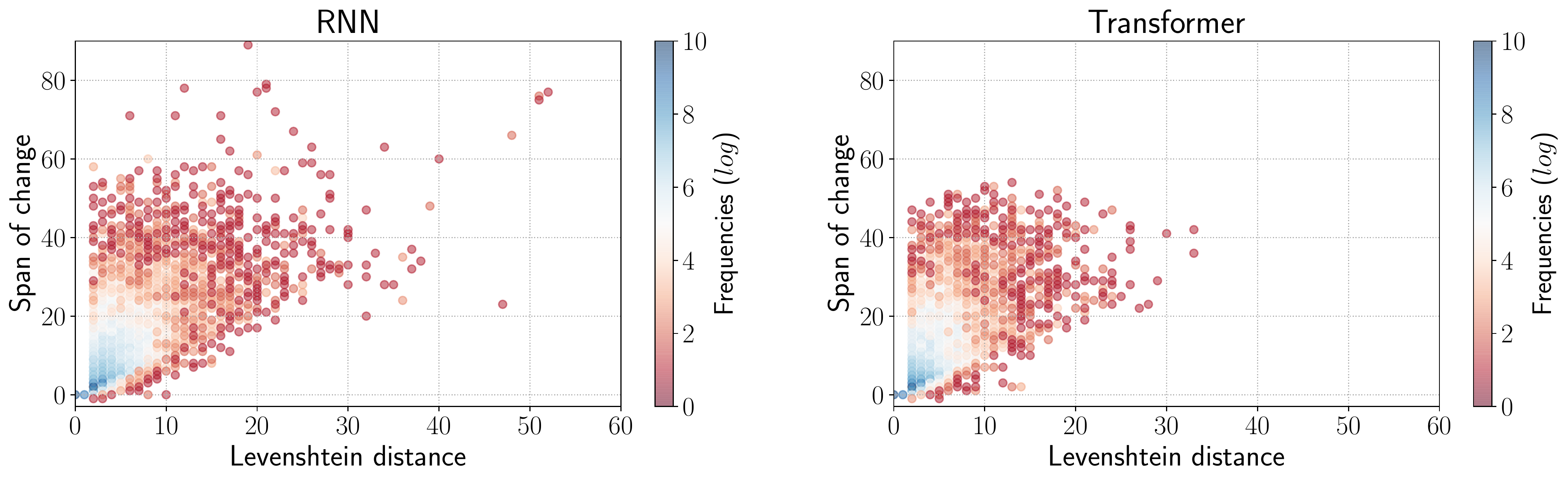}
\caption{\textit{Levenshtein distance} and \textit{span of change} between translations of sentence variations for RNN and Transformer. %With extremely similar sentence variations, we observe major differences between the translations for a surprising number of sentences.
The majority of sentence variations fall into the category of \textit{minor} changes between translations (blue area). However, a surprising number of cases have significant changes (red area). RNN exhibits a slightly more unstable pattern i.e., sentence variations with large edit differences and large spans of change.
}
\label{fig1}
\end{figure*}

Ideally, with our simple modifications, we expect a value of zero for the span of change and a value of at most 2 for the Levenshtein distance for a translation pair.
This indicates that there is only one token difference between the translation of the original sentence and the modified sentence. 

We define two types of changes based on these measures: \textit{minor} and \textit{major}.
We choose the threshold to distinguish between minor and major changes conservatively to allow for more variations in the translations.
The change in translations is empirically considered \textit{major} if both metrics are greater than 10, and \textit{minor} if both are less than 10. 
Note that edit distances and spans are based on BPE subword units.

With two very similar source sentences, we expect the Levenshtein distance and span of change between translations of these sentences to be small.
%\todo{This means that unexpected change is just the same as major??? So why bother with the extra terminolog?} 
Figure~\ref{fig1} shows the results for the RNN and Transformer model. 
%Surprisingly, we observe that our local modifications results in significant changes in a number of sentences. 
While the majority of sentence variations have minor changes, %after the modification, % i.e., small Levenshtein distance and span of change, 
a substantial number of sentences, $18\%$ of RNN and $13\%$ of Transformer translations, result in translations with major differences. % major changes in the translations % in the sentence variations.
This is a surprising indication of volatility since these trivial modifications, in principle, should only result in minor and local changes in the translations.

\begin{table}[htb!]
\centering
\caption{An example of the generated variations of an English sentence and different sentence-level metrics for the translation of each variation. We compute the oscillation range for each sentence in the test data. \label{oscmetexams}}
\begin{tabularx}{.97\columnwidth}{ll}
%Source & Translation \\ 
\toprule
Source: & \textit{Mr Ivanov took up the post in December {$\spadesuit$}.} \\
Reference: & \textit{Ivanov nahm den Posten im Dezember {$\spadesuit$} an.} \\
\hline
\\[-0.5ex]
{$\spadesuit$}  & \textcolor{mygray}{NMT output} \\
\hline
  \textit{2012} & Herr Ivanov hat den Beitrag im Dezember 2012 {\"u}bernommen.  \\
  &  \textcolor{mygray}{\textit{bleu=22.78 meteor=60.54 ter=36.36 LengthRatio=109.09}} \\
  \textit{2013} &  Herr Ivanov nahm den Beitrag im Dezember 2013 auf.  \\
  &  \textcolor{mygray}{\textit{bleu=43.67, meteor=66.89, ter=27.27, LengthRatio=109.09}} \\
  \textit{2014} &  Herr Ivanov nahm den Beitrag im Dezember 2014 auf. \\
  &  \textcolor{mygray}{\textit{bleu=43.67, meteor=66.89, ter=27.27, LengthRatio=109.09}} \\
  \hline % \\[-2ex]
{Oscillation range:} & \textit{bleu=20.9, meteor=6.4, ter=9.1, LengthRatio=0} \\
\bottomrule
\end{tabularx}
\end{table}

\subsection{Oscillations of variation in translations}

In this section, we look into various sentence-level metrics to further analyze the observed behaviour. 
In particular, we focus on the \textit{substitute numbers} modification since with this modification, we can easily generate numerous variations of the same sentence. % is easy and noise-proof. 
Having a high number of variations for each sentence gives us the opportunity of observing oscillations of various string matching metrics. 

We use sentence-level BLEU, METEOR \citep{denkowski-lavie-2011-meteor}, TER \citep{Snover06astudy}, and LengthRatio to quantify changes in the translations.
LengthRatio represents the translation length over reference length as a percentage.
For a given source sentence, we define the \textit{oscillation range} as changes in the sentence-level metric for
 the translations of all variations of the sentence (see Table~\ref{oscmetexams} for an example).

\begin{figure*}[htb!]
\centering
\includegraphics[width=1\textwidth]{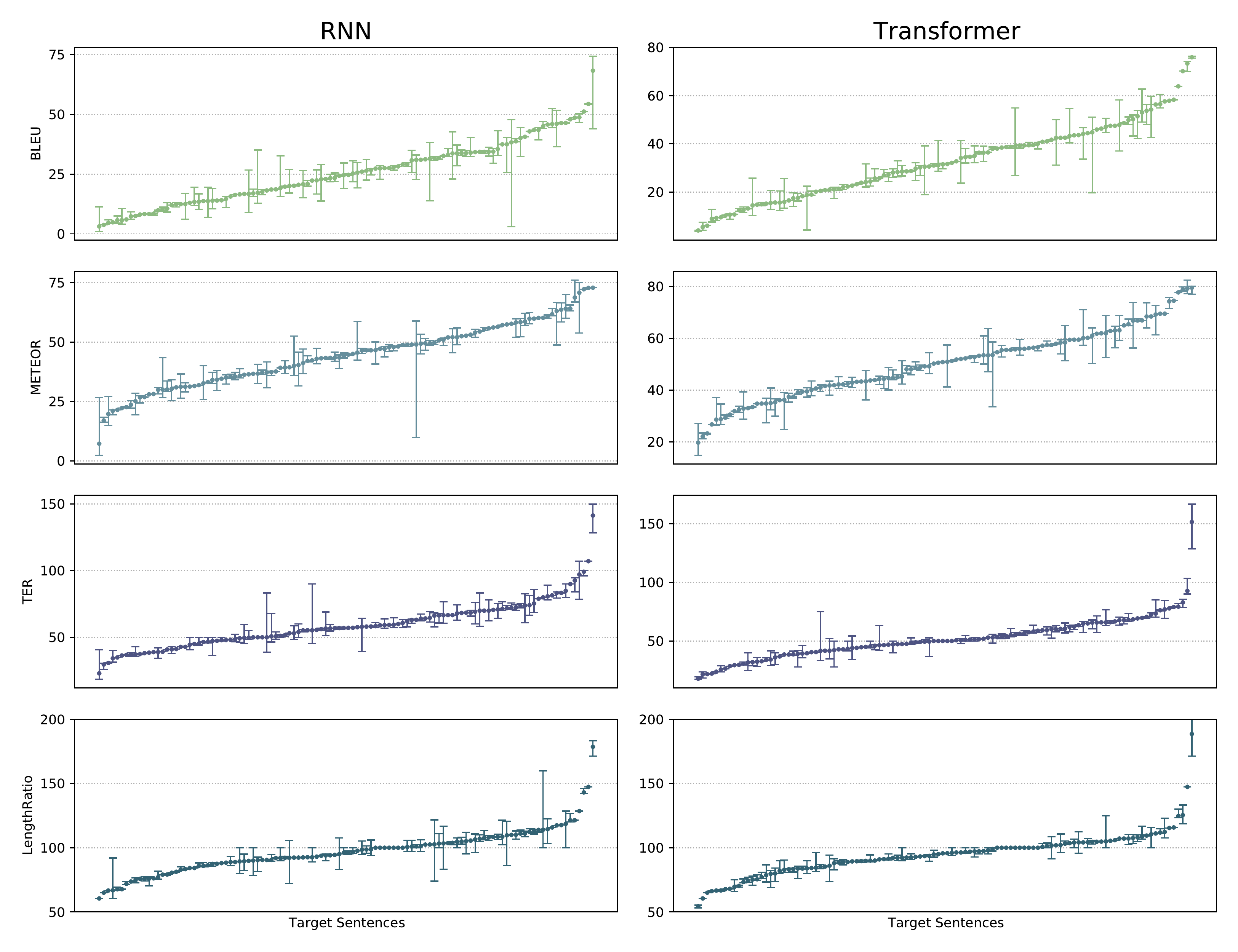}
\caption{Oscillations of various sentence-level attributes for randomly sampled sentences from our test data and their \textit{substitute number} variations. 
The data points are the mean values for all variations of each sentence, and the error bars indicate the range of oscillation of the metrics. The x-axis represents test sentence instances, sorted based on the corresponding metric. 
Ideally, each data point should have zero oscillation. }
\label{figosc}
\end{figure*}

\begin{table}[ht]
\center
\small
\caption{\label{oscstats} Mean oscillations for \textit{substitute number} variations. In theory, the variations should result in zero oscillations for every metric.}
\begin{tabularx}{0.63\columnwidth}{lcccc}
\toprule
% & \multicolumn{3}{c}{\bf DE-EN} &  \multicolumn{3}{|c}{\bf EN-DE} \\
& \bf BLEU & \bf METEOR & \bf TER & \bf LengthRatio \\
\midrule
RNN & 4.0 & 3.8& 5.2 & 5.3 \\ 
Transformer & 3.8 & 3.3 & 4.2 & 3.4  \\
\bottomrule
\end{tabularx}
\end{table}

While sentence-level metrics are not reliable indicators of translation quality, they do capture fluctuations in translations.  
With the variations we introduced, in theory, there should be no fluctuations in the translations. 
Figure~\ref{figosc} and Table~\ref{oscstats} provide the results. 
We observe that even though these sentence variations differ by only one number, there are many cases where an insignificant change in the sentence results in unexpectedly large oscillations. Both RNN and Transformer exhibit this behaviour to a certain extent.

\subsection{The effect of volatility on translation quality}

While edit distances, spans of change, and oscillation in variations provide some indication of volatility, they do not capture all aspects of this unexpected behaviour. 
It is also not entirely clear what effect these unexpected changes have on translation quality.
To further investigate this, we also perform two manual evaluations by eight fellow PhD students working on information and language processing systems.
The native language of the annotators consists of English, Dutch, and Chinese. 
All non-native annotators use English as a second language.
Our manual evaluation does not require familiarity with the German language.

In the first evaluation, we provide annotators with a pair of sentence variations and their corresponding translations
%the translation pairs of sentence variations 
and ask them to identify the differences between the two sentence pairs.
In the second evaluation, we additionally provide the source sentences and reference translations and ask the annotators to rank the sentence variations based on the translation quality similar to \citet{bojar-EtAl:2016:WMT1}.
In total, the annotators evaluated 400 randomly selected sentence quadruplets. 
 
 \begin{table*}[htb!]
\centering \small
\caption{Definitions of different labels of changes and examples for each category. The annotators identified these differences between the translations. \label{guidelinestab} }
\begin{tabularx}{0.95\columnwidth}{lXX}
\bf Label & \bf Definition & \bf Example  \\ \toprule
\texttt{Word form} & One or more words are different in form but belong to the same lexeme.   & \textit{observe} $\rightarrow$ \textit{observation} \\
\texttt{Reordered} & One or more words are reordered in the translation sentence.   & \textit{he said go}  $\rightarrow$ \textit{go, he said} \\
\texttt{Paraphrased} & One word is replaced with a synonym or a section of the sentence is paraphrased. & \textit{first six months} $\rightarrow$ \textit{first half of the year} \\
\texttt{Add/Remove} & One or more words are added or dropped from the translation sentence. & \textit{will participate} $\rightarrow$  \textit{will also participate}   \\
\texttt{Other}  & Other changes in the translation sentence.  & \textit{were torn through} $\rightarrow$  \textit{have been bypassed}   \\
\end{tabularx}
\end{table*}
\begin{table*}[htbp!]
 \begin{small}
\begin{center}
\setlength\tabcolsep{4pt} % default value: 6pt
\caption{\label{man}  A random sample of sentences from the WMT test sets and our proposed variations shown with `unexpected change' annotations ($\Delta Translation$). The cases where the unexpected change leads to a change in translation quality are marked in column $\Delta Quality$. [$w_i$\textbackslash$w_j$] indicates that $w_i$ in the original sentence is replaced by $w_j$. $S$ is the original and modified source sentence, $R$ is the original and modified reference translation, $T$ is the translation of the original sentence, and $T_m$ is the translation of the modified sentence. Differences in translations related to annotations are underlined.}
\begin{tabularx}{\textwidth}{lX}
\toprule
${S}$ &   Coes letztes Buch "Chop Suey" handelte von der chinesischen K{\"u}che in den USA, w{\"a}hrend Ziegelman in ihrem Buch "\textbf{[97\textbackslash 101]} Orchard" {\"u}ber das Leben in einem Wohnhaus an der Lower East Side aus der Lebensmittelperspektive erz{\"a}hlt.  \\
${R}$ & Mr. Coe's last book, "Chop Suey," was about Chinese cuisine in America, while Ms. Ziegelman told the story of life in a Lower East Side tenement through food in her book "\textbf{[97\textbackslash 101]} Orchard."  \\ 
%\midrule
${T}$ &   Coes\underline{'s} last book, "Chop Suey," was about Chinese cuisine in the \underline{US}, while Ziegelman, \underline{in her book "97 Orchard" talks about} living in a lower East Side.  \\
${T_m}$ &  Coes last book "Chop Suey" was about Chinese cuisine in the \underline{United States}, while Ziegelman \underline{writes in her book "101 Orchard" about} living in a lower East Side. \\ %&&\cmark&\cmark&&& \xmark \\
%\hdashline
  \rowcolor{tablegray}   \multicolumn {2}{l}{\textcolor{mygray}{\texttt{$\Delta Translation$: [reordered] [paraphrased] |  $\Delta Quality$: No }}}  \\
  \midrule
 ${S}$ &  Man h{\"a}lt \textbf{[bereits\textbackslash$\pmb{\phi}$]} Ausschau nach Parkbank, Hund und Fu{\ss}ball spielenden Jungs und M{\"a}dels. \\
$R$ & You are \textbf{[already\textbackslash$\pmb{\phi}$]} on the lookout for a park bench, a dog, and boys and girls playing football. \\
%\midrule
${T}$ & \underline{We are} already \underline{looking} for Parkbank, dog and football playing boys and girls.  \\
${T_m}$& \underline{Look} for Parkbank, dog and football playing boys and girls. \\ %&\cmark&&&\cmark& & \cmark \\
%\hdashline
  \rowcolor{tablegray}   \multicolumn {2}{l}{\textcolor{mygray}{\texttt{$\Delta Translation$: [word form] [add/remove] | $\Delta Quality$: Yes }}}  \\
  \midrule
%$\Delta$ & \multicolumn {1}{l}{\textcolor{mygray}{\texttt{[word form] [add/remove]}}}\\
%\midrule
${S}$ 	& Bei einem Unfall eines Reisebusses mit \textbf{[43\textbackslash 45]} Senioren als Fahrg{\"a}sten sind am Donnerstag in Krummh{\"o}rn (Landkreis Aurich) acht Menschen verletzt worden.  \\
$R$ & On Thursday, an accident involving a coach carrying \textbf{[43\textbackslash 45]} elderly people in Krummh{\"o}rn (district of Aurich) led to eight people being injured. \\
%\midrule
${T}$  &	In the event of an accident involving \underline{a coach with 43 senior citizens} as \underline{passengers}, eight people were injured on Thursday in \underline{Krummaudin (County Aurich)}.  \\
${T_m}$   &	In the event of an accident involving \underline{a 45-year-old coach} as a \underline{passenger}, eight people were injured on Thursday in  \underline{the district of Aurich}. \\% &\cmark&&&\cmark& \cmark &\cmark\\
%\hdashline
%$\Delta$ & \multicolumn {1}{l}{ \textcolor{mygray}{\texttt{[word form] [add/remove] [other]}}}\\
  \rowcolor{tablegray} \multicolumn {2}{l}{\textcolor{mygray}{\texttt{$\Delta Translation$: [word form] [add/remove] [other] | $\Delta Quality$: Yes }}}  \\
\midrule
${S}$  &  Es ist ein anstrengendes Pensum, aber die Dorfmusiker helfen \textbf{[normalerweise\textbackslash$\pmb{\phi}$]}, das Team motiviert zu halten.  \\	
 $R$ & It's a backbreaking pace, but village musicians \textbf{[usually\textbackslash$\pmb{\phi}$]} help keep the team motivated.    \\
% \midrule
${T}$  & It\underline{'s} a \underline{demanding child}, but the village \underline{musicians} usually \underline{help} keep the team motivated.  \\
${T_m}$  & It \underline{is} a \underline{hard-to-use}, but the village \underline{musician} \underline{helps to} keep the team motivated.	\\ %	&\cmark&&& & \cmark&\cmark \\	
%\hdashline
%$\Delta$ & \multicolumn {1}{l}{ \textcolor{mygray}{\texttt{[word form] [other]}}}\\
  \rowcolor{tablegray}   \multicolumn {2}{l}{\textcolor{mygray}{\texttt{$\Delta Translation$: [word form] [other] | $\Delta Quality$: Yes }}}  \\
\bottomrule
\end{tabularx}
 \end{center}
%}
 %\end{center}
 \end{small}
\end{table*}
  \begin{figure*}[htb!]
\centering
\includegraphics[width=1\textwidth]{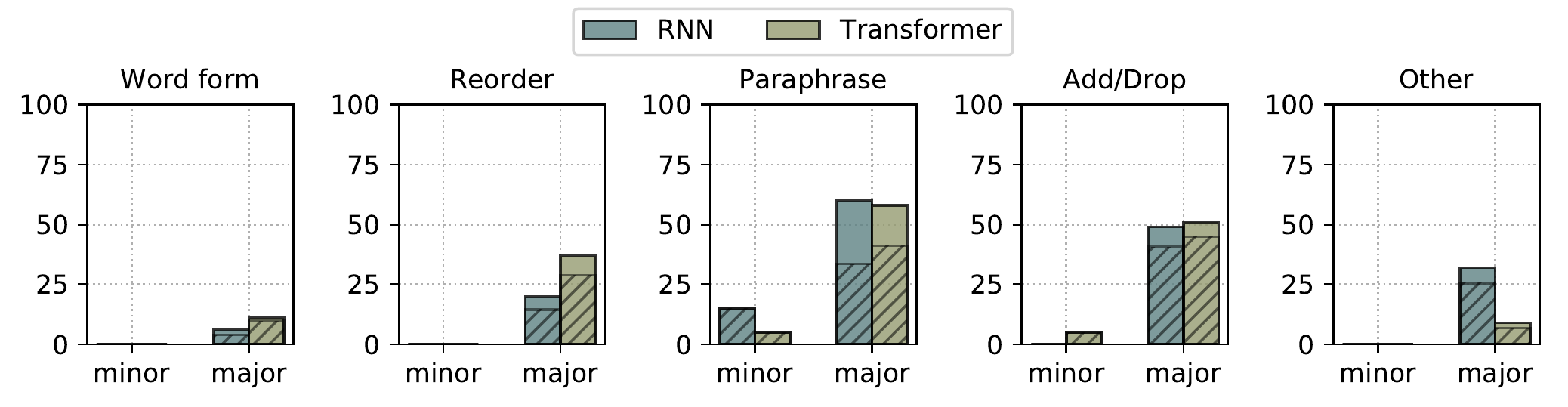}
\caption{Categories of unexpected changes in the translation of sentence variations as provided by annotators. The percentage of sentence variations with \textit{minor} and \textit{major} edit differences, as defined in~\ref{ctsec}, are shown separately. The hatched pattern indicates the ratio of sentence variations for which the translation quality changes.} %for different sentence variations.}
%Levenshtein distance between translations of sentences before and after modification grouped in 3 buckets. Hatched pattern is the ratio of sentence pairs in each bucket with more than 10 unigram differences.} %en-de.
\label{fig2}
\end{figure*}

Table~\ref{guidelinestab} shows the identified categories of changes from annotators' labels.  
The main types of unexpected changes identified by the annotators are a `change of word form', e.g., verb tense, `reordering of phrases', `paraphrasing' parts of the sentence, and an `other' category, e.g., preposition.  
A sentence pair can have multiple labels based on the types of changes. 
Table~\ref{man} provides examples from the test data. %of the types of changes. % in each sentence variation. 
%Note that we are interested in \textit{unexpected} changes and ignore the changes that are a direct consequence of the modifications.

Statistics for each category of unexpected change are shown in Figure~\ref{fig2}.
Our first observation is that, as to be expected, there are very few `unexpected changes' when two variations lead to translations with \textit{minor} differences.
Interestingly, the vast majority of changes are due to paraphrasing and adding or dropping of words.
Comparing the performance of the RNN and Transformer model, we see that both RNN and Transformer display inconsistent translation behaviour. 
%While Transformer has slightly fewer sentences with major changes, it has a higher number of sentence variations in the major category that result in a change in translation quality. % both RNN and Transformer have roughly similar 
From the annotators' assessments, we find that in $26\%$ and $19\%$ of sentence variations, the modification results in a change in translation quality for the RNN and Transformer model, respectively.
From the manual evaluations, we conclude that the oscillations in translation outputs captured by our metrics indeed point to harmful changes in translation quality. 
This behaviour is not exposed by standard test sets and evaluation metrics. 

\subsection{Generalization and compositionality}

Because of their ability to generalize beyond their training data, deep learning models achieve exceptional performances in numerous tasks.
The generalization ability allows translation systems to generate long sentences not seen before. 
Recently there has been some interest in understanding whether this performance depends on recognizing
shallow patterns, or whether the networks are indeed capturing and generalizing linguistic rules \citep{linzen-etal-2016-assessing,chowdhury-zamparelli-2018-rnn}.

The capability of generalization of current deep learning models can be interpreted as whether compositionality arises in learning problems where the compositional structure has not been explicitly declared.
The principle of compositionality \citep{Frege1892} has been extremely influential throughout the history of formal semantics and cognitive science with many arguments for and against it \citep{montague1974d,Pelletier94theprinciple,DBLP:journals/jolli/Janssen01}. 

In simple terms, compositionality can be defined as the ability to construct larger linguistic expressions by combining simpler parts. For instance, if a model applies the correct compositional rules to understand \textit{`John loves Mary'}, it must also understand \textit{`Mary loves John'} \citep{fodor2002compositionality}.

Investigating the compositional behaviour of neural networks in real-world natural language problems is a challenging task. 
Recently, several approaches have studied deep learning models' understanding of compositionality in natural language by using synthetic and simplified languages \citep{DBLP:journals/corr/abs-1902-07181,babyai_iclr19}.
%He raises the question that whether making neural networks more compositional will make them more adaptive.
\citet{ijcai2020-708} designed theoretically grounded tests based on different interpretations of compositionality. 
Their experiments showed that the current state-of-the-arts neural network architectures struggle to capture different aspects of compositionality in language and there is a need for a more extensive set of evaluation criteria to evaluate these models.
\citet{886a37b5fc2f43449e4bca3b5557e3ae} introduced the SCAN data set consisting of simplified natural language commands and their translations into sequences of actions.
They showed that, when there is a systematic difference between training and test data, neural models fail to generalize because they lack the ability to extract systematic rules from the training data. 
\citet{DBLP:journals/corr/abs-1904-00157} observed that current models seem to be able to generalize without any compositional rules.
He argued that to a certain extent neural networks can be productive without being compositional.
 %and further studies can lead to new insights into the less systematic generalization patterns.}

Although we do not specifically look into the compositional potential of translation systems, we are inspired by compositionality in defining our modifications.
We argue that the observed volatile behaviour of the translation systems in this chapter is a side effect of current models not being compositional.  
If a translation system has a good understanding of the underlying structures of the sentences \textit{`Mary is 10 years old'} and \textit{`Mary is 11 years old'}, it must also translate them very similarly regardless of the accuracy of the translation. 
While current evaluation metrics capture the accuracy of the NMT models, these volatilities go unnoticed.  

Current neural models are successful in generalizing without learning any explicit compositional rules, however, our findings show that they still lack robustness.
We highlighted this lack of robustness in this chapter and suspect that it is associated with these models' lack of understanding of the compositional nature of language.

\section{Conclusion} \label{secvolconc}

Motivated by our findings in the previous chapters, we continued investigating the circumstances where current models do not perform as expected. 
Specifically, we are interested in cases where the semantic and syntactic complexity of the sentence remains unchanged with minor modifications to the observed context. 
Hence we investigated:

\begin{enumerate}[label=\textbf{RQ3.\arabic* },wide = 0pt, leftmargin=2em]
\setlength\itemsep{1em}
 \setcounter{enumi}{2}
\item \acl{rq:vol1}

\medskip

\noindent We studied an unexpected and erroneous behaviour in current NMT models by examining various metrics to quantify oscillations in translations of very similar sentences.   
We show that even with minor modifications preserving the grammaticality and plausibility of the sentence, we can effectively identify a surprising number of cases where the translations of extremely similar sentences are unexpectedly different. 
%By further manual inspection, we observed that these differences included `changes of word forms', `reorderings of phrases', `changes by paraphrasing', `adding or dropping words from the translations', and `semantically different translations'. 
Our experiments on our test set showed that current NMT models are not completely robust and they expose their weaknesses when probed with specific test cases.

\item \acl{rq:vol2}

\medskip

\noindent Models that have compositional understanding of the world are capable of generalizing to unseen composite cases \citep{886a37b5fc2f43449e4bca3b5557e3ae,cogswell2019emergence}.
We proposed an approach to examine the compositional understanding and measure the generalization capability of NMT models. 
We did so by introducing a specific test set and various evaluation metrics. 
If a model is capable of compositional understanding, it should have hardly any oscillations in the translation outputs on this test set. 
However, once we probed these models with extensive test cases, we observed that they do in fact exhibit unexpected changes in the translation outputs.
Our analyses showed that both RNN and Transformer models exhibit volatile behaviour with changes in translation quality for $26\%$ and $19\%$ of sentence variations, respectively.  
Our experiments highlighted the need for a comprehensive evaluation setup for deeper analyses of current neural models.

\end{enumerate}

\noindent 
This concludes the main research question of this chapter as follows:

\paragraph{Research Question 3:} \acl{rq:vol} 

\medskip

 \noindent  To answer this question, we examined the vulnerability of current NMT models to small changes to the observed context. 
Primarily, we focused on modifications that do {not} introduce new linguistic complexities for the translation model. 
We proposed a simple approach to modifying standard test sentences without introducing noise. 
By creating this data set, we can automatically measure if NMT models lack robustness and exhibit volatile behaviour. 
We observed that neural models, even with high performances on standard test sets, struggle in showing compositional understanding and suffer from a lack generalization.

\bookmarksetup{startatroot} % Lift from parts
\addtocontents{toc}{\bigskip} % skip

% !TEX root = thesis-main.tex

\chapter{Conclusions}
\label{chapter:conclusions}

In this thesis, we explored the role of context in machine translation as well as lexical modeling, using deep learning frameworks.
With the increase of computational power of machines and the availability of data, progress in deep learning has focused on developing more advanced models. 
In fact, with the increasing amount of training data, the performance of many computer vision and language understanding tasks has increased significantly. 
In this thesis, we are motivated by \textit{what} these models learn from the available data and \textit{how} we can use this information to resolve linguistic challenges that arise from statistical learning from data. 
Specifically, we investigated the influence of contextual cues in understanding different words and proposed several approaches that improve how these models learn from context.

Firstly, we looked into ambiguous words and studied how document-level context assists in distinguishing different meanings of a word. 
Next, we focused on the influence of context in the bilingual setting of machine translation. 
We examined how neural translation models use context to learn and transfer meaning and showed that by diversifying data for difficult words, we can improve translation quality.
In order to identify difficult words, we first looked at the data distribution and specifically targeted rare words.
Since there is a lack of diverse contexts in the training data for rare words, the translation of them is challenging.
Next, we investigated the failures of a trained model to identify difficult words.
These difficult words are words that the model has low confidence in predicting after training on a sizable amount of data.
We identified difficult words for an NMT model and performed data augmentation targeting these words. 
By creating new contexts for difficult words, we improved the generation capability of the NMT model and the translation quality.

Next, we addressed the shortfalls of relying only on the contexts observed in the training data to learn the meanings of words. 
We examined under which conditions context is not enough for capturing various linguistic phenomena. 
In particular, we studied the interesting case of idiom translation and showed that current NMT models often fail to capture such nuances.
Neural networks optimize the learning process on the available data and the lack of data for complex linguistic phenomena such as idiom translation is an obstacle for developing stronger models. 

Finally, we raised more general questions about the learning capabilities of current state-of-the-art translation models.
We analyzed how these models fail unexpectedly even in cases where there are no evident complex linguistic phenomena. 
By introducing simple contextual modifications, we identified an underlying generalization problem of state-of-the-art translation models.

In the following section, we revisit our research questions and summarize the main findings of this thesis.
We then propose a number of questions that remain open for further exploration. 

\section{Main findings}

%Concretely, we answered the following research questions in this thesis:

In Chapter~\ref{chapter:research-01}, we started with a preparatory question on the importance of context when learning word representations for difficult words.
Going past local neighboring words, we looked at document-level contexts by asking:

\begin{enumerate}[label=\textbf{RQ1.\arabic* },wide = 0pt, leftmargin=2em]
\setlength\itemsep{1em}
\item \acl{rq:topic1}  \label{rq:topic1}
%\medskip

\noindent To answer this question, we investigated whether document-level information is an adequate contextual cue to help distinguish between different senses of a word.
We experimented with a hierarchical Dirichlet process for modeling document topics which generated two sets of distributions that we used in our methods: distributions over topics for words in the vocabulary and distributions over topics for documents in the corpus. 
We observed that these distributions distinguish between senses of words. 
Next, we examined how we can leverage this information to learn word representations by asking:

\item \acl{rq:topic2}  \label{rq:topic2}

\medskip

\noindent We found that the distribution over topics is different for different senses of an ambiguous word. 
This motivated us to combine this distribution with the Skipgram model to provide information on word senses to the embeddings. 
To achieve this, we devised three model variations that learned multiple representations per word based on the assigned topic in different contexts. 
We then evaluated these embeddings by asking: 

\item \acl{rq:topic3}  \label{rq:topic3}

\medskip

\noindent We evaluated word embeddings on the word similarity task and observed slight improvements under different settings. 
However, there was no clear winner across all data sets. 
Since word similarity data sets consider individual words in isolation and do not provide any contexts, we then evaluated the embeddings in a more context-aware setting.
Our evaluation on the lexical substitution task showed that topic distributions capture word senses to a large extent. 
Moreover, we obtained statistically significant improvements in a lexical substitution task without using any syntactic information.  
The best results were achieved by our HTLE model which learns topic-sensitive representations by hard-labeling topics to target words and not using generic representations. 
%We observed that topic-sensitive representations captured senses of the words to some extent.

\end{enumerate}

These three sub-questions together allowed us to answer our first main research question: 

\begin{enumerate}[label=\textbf{Research Question \arabic*:},ref={RQ\arabic*},wide = 0pt]
\setlength\itemsep{1em}

\item \acl{rq:topic} 

\medskip

 \noindent 
 Our experiments showed that we can use document-level topic distribution to improve word representation learning.
To summarize, we introduced an approach in Chapter~\ref{chapter:research-01} to learn topic-sensitive word representations that exploits the document-level context of words and does not require annotated data or linguistic resources. 
Additionally, we also learned representations for topics and our qualitative analyses showed that words belonging to the same topics also tend to be clustered together.

Having observed the effectiveness of wider context in capturing polysemy in word embeddings, we investigated in Chapter~\ref{chapter:research-02} the impact of context on the translation of difficult words. 
We first asked:

\begin{enumerate}[label=\textbf{RQ2.\arabic* },wide = 0pt, leftmargin=2em]
\setlength\itemsep{1em}
\item \acl{rq:tda1} \label{rq:tda1}

\medskip

\noindent 
By leveraging language models trained on large amounts of monolingual data, we generated new sentence pairs containing rare words in new contexts. 
We first confirmed that the translation performance is primarily affected by low-frequency and out-of-vocabulary words. 
Our analysis in Section~\ref{tda:analysis} further showed that the poor translation quality of rare words is a result of a lack of diverse training examples. 
To address this problem, we proposed a method to automatically generate new contexts for these words.
Next, we used this data to augment the parallel corpus used to train the translation model and re-trained the entire system. 
Our results showed that this approach improves the representations of rare words learned by the model and consequently increases the number of times the model generates these words correctly. 

We observed substantial improvements in simulated low-resource English$\rightarrow$German and German$\rightarrow$English settings.
%This is a relatively simple, yet effective approach to augment parallel training data for low-resource language pairs. 

%In answering this question in Chapter~\ref{chapter:research-02}, we observed the impact of additional \textit{training} data on the translation of rare words.
\medskip

A natural follow-up question is whether we can perform augmentation during test time as well. So we asked:

\item \acl{rq:tda2} \label{rq:tda2}

\medskip

\noindent 
%We first demonstrated why our previously proposed method is not viable at test time: 
Our experiments showed that augmentation at test time reduces the number of \texttt{unk}s in the output and results in more fluent sentences. 
We cannot modify a source sentence resulting in a change of meaning without modifying the reference translation as well. 
Since we do not have access to the reference translations during inference, any alteration we made to the source sentence must keep the meaning of the sentence unchanged. 
In Section~\ref{tda:semantic}, we introduced a substitution via paraphrasing method to replace rare and out-of-vocabulary words in source sentences. We used different paraphrase knowledge resources to do this: WordNet, PPDB, GermaNet, CBOW, and our embedding approach proposed in Chapter~\ref{chapter:research-01}.
We gained improvements in BLEU scores while significantly reducing the number of \texttt{unk} generated in the target output. 

\medskip

In Chapter~\ref{chapter:research-03}, we continued addressing our second research question by further analyzing the effectiveness of additional context for learning the meaning of a word.
Rather than looking at the distribution of the training data, we investigated the behaviour of a neural MT system during training by asking:

\item \acl{rq:bt1} \label{rq:bt1}

\medskip

\noindent 
We found that signals from failures of the model can be used to identify where the model is not learning satisfactorily.
To investigate this question we first explored different aspects of other influential augmentation methods, in particular back-translation, in Section~\ref{btbtanalysis}. 
Our analyses showed that the quality of the synthetic data generated with a reasonably good model has a small impact on the effectiveness of back-translation, but that the ratio of synthetic to real training data plays a more important role.
With a higher ratio, the model becomes biased towards noise in the synthetic data and unlearns the parameters.
Next, we looked into which words benefit most from additional back-translated data.
We observed that words with high prediction losses in the original model undergo the most changes after training with synthetic data. 
Our findings showed that with the addition of contexts for words with high prediction loss, we can increase the overall accuracy of the model.

Equipped with this information, we addressed the following question:

\item \acl{rq:bt2} \label{rq:bt2}

\medskip

\noindent In Section~\ref{bttarget}, we proposed our sampling approach targeting words that are difficult to predict.
These words benefit the most from a more diverse context after augmentation.
Our approach included several variants of using the prediction loss for identifying relevant sentences to back-translate.
We also used the contexts of difficult words by incorporating context similarities as a feature to sample sentences for back-translation.
We discovered that using the prediction loss to identify weaknesses of the translation model and providing additional synthetic data targeting these shortcomings improved the translation quality of German$\rightarrow$English and English$\rightarrow$German translations. % by up to 1.7 {BLEU} points.
%Finally, our proposed frequency-based sampling approach remains a simple, yet effective strategy that is hard to outperform.

\end{enumerate}

Having discussed our specific sub-questions, we return to our more general question:

\item \acl{rq:tdabt}  \label{rq:tdabt}

\medskip

 \noindent In Chapters~\ref{chapter:research-02} and~\ref{chapter:research-03}, we investigated the effect of the availability of data, where \textit{the lack of} diverse contexts during training causes difficulties.
Rare words, by definition, suffer from a lack of diverse context. 
Our studies showed that both translating and generating rare words is a challenging task with NMT models. 
We then continued with the impact of diverse contexts on translation quality in NMT. 
In particular, we focus on the back-translation method and the synthetic contexts that are generated with a reverse trained NMT model. 
We investigated this method and explored alternatives to select the monolingual data in the target language that is to be back-translated into the source language to improve translation quality.
Both data augmentation approaches proposed in Chapters~\ref{chapter:research-02} and~\ref{chapter:research-03} lead to improvement of translation quality by generating diverse contexts for training.

\medskip

Continuing our research on the impact of context on the quality of NMT models, we subsequently investigated some of its limitations. 
In Chapter~\ref{chapter:research-04}, we looked into idiom translation with neural models.
Since the literal meaning of the components is different than the idiomatic meaning of the entire expression, the model needs to know in which context to translate it literally and in which idiomatically. 
Neural MT, in particular, has been shown to perform poorly on idiom translation despite its overall strong advantage over previous MT paradigms \citep{isabelle2017challenge}. 

We began by asking:

\begin{enumerate}[label=\textbf{RQ3.\arabic* },wide = 0pt, leftmargin=2em]
\setlength\itemsep{1em}
\item \acl{rq:id1} \label{rq:id1}

\medskip

\noindent One of the main challenges of studying idiom translation is the lack of dedicated and labeled data for evaluation and analysis.
As an essential step towards answering this question, we required a test set explicitly tailored to evaluating idiom translation quality.
To this end, we harvested a parallel data set for training and testing idiom translation for German$\rightarrow$English and English$\rightarrow$German.
The test sets included sentences with at least one idiom on the source side while the training data is a mixture of idiomatic and non-idiomatic sentences with labels to distinguish between the two.
Using our new resources, 
we performed preliminary translation experiments and proposed different metrics to evaluate the quality of idiom translation.
%Experiments on this test set showed that although PBMT achieved a lower overall BLEU score in comparison with NMT, it scored higher on our metrics which explicitly measure idiom translation quality. 

\medskip

We then evaluated the translation quality of neural models on our idiom translation test set:

\item \acl{rq:id2} \label{rq:id2}

\medskip

\noindent We observed that the NMT model achieved a higher overall BLEU score but scored lower in idiom translation metrics.
This is in agreement with previous works on investigating idioms as one of the weak points of neural models \citep{shao-etal-2018-evaluating}.
Since there are no explicit signals in the sentence to identify when a phrase is to be translated literally and when it is to be translated idiomatically, we examined whether such a signal would help. 
Our experiments in Section~\ref{idevalus} showed that adding a flag during training to indicate idiomatic use improves the quality of idiom translation in general.
However, the overall BLEU score declined slightly.
We concluded that there is little correlation between overall BLEU scores and the localized precision of idiomatic phrase translation.
Our experiments showed that idiom translation can benefit from having a tailored development and test set and more specific metrics for evaluation.

\noindent Our next research question focused on other cases where NMT models fail to generate a correct translation given the observed context. 
In Chapter~\ref{chapter:research-05}, we first examined how to expose this shortcoming in translation models by asking:  

\item \acl{rq:vol1} \label{rq:vol1}

\medskip

\noindent We studied the behaviour of NMT models and observed an unexpected but recurring pattern: A model that translates a given phrase correctly in a sentence fails to translate it correctly in another sentence which is very similar to the first.
To explain these observations, we introduced new quantitative metrics measuring such unexpected changes. 
These metrics measured oscillations in translations of very similar sentences.  
Our experiments further showed that even with minor modifications preserving the grammaticality and plausibility of the sentence, we can effectively identify a surprising number of cases where the translations of extremely similar sentences are very different. 
By further manual inspection, we observed that these differences included `changes of word forms', `reorderings of phrases', `changes by paraphrasing', `adding or dropping words from the translations', and `semantically different translations'. 
We concluded that with contextual modifications during testing, we can reveal a lack of robustness of translation models.

\medskip

Knowing this shortcoming of NMT models, we then asked:

\item \acl{rq:vol2} \label{rq:vol2}

\medskip

\noindent We created a test set from our the sentence variation method proposed in Section~\ref{secsentvar}. 
Next, we examined the robustness of current NMT models with this new test data and various evaluation metrics. 
Our analyses showed that both RNN and Transformer models exhibit volatile behaviour with changes in translation quality.
We observed that these models fall short of capturing the compositional nature of the language, which confirms previous findings on the lack of compositional behaviour of NMT systems \citep{886a37b5fc2f43449e4bca3b5557e3ae}.
Additionally, we found that current evaluation sets do not spot the unexpected patterns we identified with our test data.

\end{enumerate}

\medskip

These answers allow us to return to our more general question:

\item \acl{rq:vol} \label{rq:vol}

\medskip

\noindent 
To study the influence of the observed context, we investigated how NMT models handle translating \textit{non-compositional} and \textit{compositional} events.
Our findings in Chapters~\ref{chapter:research-04} and~\ref{chapter:research-05} showed that even well-performing models with high translation quality still suffer from a number of problems in both cases. 

First, we looked into non-compositional multiword expressions or idioms.
Idiom translation is one of the more difficult challenges of machine translation. 
To study this, in Section~\ref{idiomdata}, we created a data set for training and testing idiom translation.
We found that with the observed context, current NMT models struggle with translating non-compositional expressions. PBMT models on the other hand, despite underperforming on general-purpose data sets, achieve better idiom translation quality.

Next, we investigated the impact of observed context on translating compositional expressions.
To achieve this, we defined a test set and evaluation metrics and investigated the NMT model's behaviour in this particular setting. 
In creating this test set, we focused on modifications that do \textit{not} explicitly introduce new challenges for the translation model. 
Large oscillations in translations in the test set are an indication that the models do not capture composition in a systematic way, but often rely on memorized patterns to translate new sentences.
In Section~\ref{secsentvar}, we proposed a simple approach to modify standard test sentences without introducing noise and hence generating semantically and syntactically correct variations.
Our findings showed that even well-performing models with high translation quality are prone to this problem and more extensive evaluations are necessary for assessing a system. 
We believe that our insights will be useful for developing more robust NMT models.

\end{enumerate}

\section{Future work}

In this thesis, we studied how context is used by neural models to learn and generate words in a language and proposed methods to improve it.
%We chose machine translation for the experiments as a good representative for evaluating language understanding.
While this work highlighted the potentials of neural translation models in learning from data and studied some cases where they fall short, there are still many questions left to explore.
Here, we discuss a few of these questions:

 \paragraph{Are evaluation metrics for generation tasks still adequate?}

 Since manual evaluation is very expensive, several automatic metrics have been designed to evaluate generation tasks as described in Section~\ref{bgexp}, including BLEU \citep{Papineni2001}, ROUGE \citep{lin-2004-rouge}, METEOR \citep{banerjee-lavie-2005-meteor}, and TER \citep{Snover06astudy}.

These metrics compare an automatically generated candidate with a reference that is created manually. 
Matching the words in the candidate and reference gives us useful and reproducible results on the performance of a system and has helped advance tasks such as machine translation.
However, with the improvements in generation models, the errors in the candidates are becoming increasingly subtle and idiosyncratic and existing metrics are not fully capable of highlighting them.
While this issue has recently been addressed by \citet{chen-etal-2019-evaluating} and \citet{ribeiro-etal-2020-beyond}, further research on approaches and metrics that highlight deeper problems in generation models and go
beyond $n$-gram based matching are necessary.

\paragraph{How can we learn complex nuances and structures of language?}

Many neural models still struggle with complex language structures, such as idiomatic expressions, in their respective tasks.
One reason is that many interesting phenomena in language do not occur frequently. 
As a result, exclusively data-driven models fail to capture these nuances. 
One way of addressing this issue requires constructing data sets that are both adequately large and of high quality. 
With the availability of data sets that target specific linguistic phenomena, the process of learning them will be measurable and developing models targeting these challenges will be more accessible.  

\paragraph{How compositional are sequence-to-sequence models?}

Compositionality is the ability to construct larger linguistic expressions by combining simpler parts \citep{Frege1892,fodor1992holism}.
Investigating the compositional behaviour of neural networks in real-world natural language problems is a challenging task. 

Current NMT models deliver high average translation quality provided enough training data and a good training-test domain match. It is not entirely clear, though, how much of this success stems from learning the underlying compositional structure of the sentence.
In general, many traits of neural models are still a black box which hinders advancements to some extent.  
Recently, a few studies have focused on studying the level of compositionality in neural sequence-to-sequence models using toy data sets \citep{DBLP:journals/corr/abs-1711-00350,ijcai2020-708}.
Creating evaluation paradigms to further analyze this aspect can potentially lead to a better understanding of the inner workings of these influential models.

% Start the back matter
% Numbering stays the same
% Add 1) bibliography, 2) abstract (Dutch/English)
\backmatter
% !TEX root = thesis-main.tex

% Include list of notation
%\chapter*{List of Notation}
%\addcontentsline{toc}{chapter}{List of Notation}
%\markboth{List of Notation}{List of Notation}

% Include the bibliography

% Do some formatting of the chapter title and page headers
% Set the item separation to 0 (saves a few pages)
\renewcommand{\bibsection}{\chapter{Bibliography}}
\renewcommand{\bibname}{Bibliography}
\markboth{Bibliography}{Bibliography}
\renewcommand{\bibfont}{\footnotesize}
\setlength{\bibsep}{0pt}

% Include the actual bib file
% Add the chapter to the ToC (it doesn't happen automatically if you loose the chapter number)
\bibliographystyle{abbrvnat}
\bibliography{thesis_clean}

% Include abstract(s)

\chapter{Summary}

Neural networks learn patterns from data to solve complex problems.
To understand and infer meaning in language, neural models have to learn complicated nuances.
Discovering distinctive linguistic phenomena from data is not an easy task. 
For instance, lexical ambiguity is a fundamental feature of language which is challenging to learn. 
Even more prominently, inferring the meaning of rare and unseen lexical units is difficult with neural networks. 
Meaning is often determined from \textit{context}.
With context, languages allow meaning to be conveyed even when the specific words used are not known by the reader. 
To model this learning process, a system has to learn from a few instances in context and be able to generalize well to unseen cases.
Neural models use a sizable amount of data that often consists of contextual instances to learn patterns.
The learning process is hindered when training data is scarce for a task.
Even with sufficient data, learning patterns for the long tail of the lexical distribution is challenging. 

In this thesis, we focus on understanding certain potentials of contexts in neural models and design augmentation models to benefit from them. 
We focus on machine translation as an important instance of the more general language understanding problem. 
To translate from a source language to a target language, a neural model has to understand the meaning of constituents in the provided context and generate constituents with the same meanings in the target language.
This task accentuates the value of capturing nuances of language and the necessity of generalization from few observations.
The main problem we study in this thesis is what neural machine translation models learn from data and how we can devise more focused contexts to enhance this learning.
First, we study how document-level contexts aid in distinguishing different meanings of a word. 
Second, we investigate how translation models exploit context to learn and transfer meaning and show that different and diverse contexts resolve various obstacles of translation.
Third, we examine under which conditions the observed context in the data is not enough for inferring meaning and capturing various linguistic phenomena. 

Looking more in-depth into the role of context and the impact of data on learning models is essential to advance the Natural Language Processing (NLP) field. 
Understanding the importance of data in the learning process and how neural network models interact with and benefit from data can help develop more accurate NLP systems. 
Moreover, it helps highlight the vulnerabilities of current neural networks and provides insights into designing more robust models.

% Add at least a Dutch one, the English one could also go on the back cover
% Again, add the chapter to the ToC
\chapter{Samenvatting}

Neurale netwerken zijn computermodellen die patronen leren uit data, om zo complexe problemen op te lossen.
Om betekenis in taal te begrijpen en af te leiden, moeten neurale netwerken ingewikkelde nuances leren.
Voorbeelden van taalkundige fenomenen die niet gemakkelijk zijn voor een neuraal netwerk, zijn 
\textit{lexicale ambigu\"{i}teit} (woorden met meerdere betekenissen) en zeldzame of nieuwe woorden die het neurale model niet of slechts enkele keren heeft gezien in de trainingsdata.
De betekenis van zeldzame of ambigue woorden hangt af van de context waarin ze voorkomen. Door middel van context kunnen talen betekenis overbrengen, zelfs als de lezer bepaalde gebruikte woorden niet kent.
Om dit leerproces te modelleren, moet een systeem leren van slechts enkele voorbeelden in een bepaalde context en moet het goed kunnen generaliseren naar ongeziene gevallen.
Neurale modellen gebruiken een aanzienlijke hoeveelheid data, vaak bestaande uit voorbeelden in context die gebruikt worden om patronen te leren.
Als er maar weinig trainingsdata voor een bepaalde taak is, wordt het leerproces gehinderd, maar ook als er genoeg data beschikbaar is, blijft het leren van zeldzame woorden een uitdaging.

De focus van dit proefschrift ligt op het begrijpen van de mogelijkheden die context biedt voor neurale modellen om hier vervolgens van te kunnen profiteren.
We richten ons op machinaal vertalen als een belangrijk voorbeeld van het meer algemene probleem van taalbegrip.
Om te vertalen van een brontaal naar een doeltaal, moet een neuraal model de betekenis van woorden of zinnen in de gegeven context begrijpen en woorden of zinnen met dezelfde betekenis genereren in de doeltaal.
Deze taak benadrukt het belang van het vastleggen van nuances in taal en de noodzaak om te kunnen generaliseren vanuit slechts een beperkt aantal observaties.
Het belangrijkste probleem dat we in dit proefschrift bestuderen is wat neurale machinale vertaalmodellen leren van data en hoe we meer gefocuste contexten kunnen bedenken om dit leerproces te verbeteren.
Allereerst bestuderen we hoe context op documentniveau kan helpen bij het onderscheiden van verschillende betekenissen van een woord.
Daarnaast onderzoeken we hoe vertaalmodellen context gebruiken om betekenis te leren en over te dragen en laten we zien dat het gebruik van verschillende contexten een aantal obstakels tijdens het vertalen kan overwinnen.
Ten slotte onderzoeken we onder welke voorwaarden de waargenomen context in de data onvoldoende is om betekenis af te kunnen leiden en bepaalde taalkundige fenomenen te vatten.

Door dieper in te gaan op op de rol die context speelt en de impact van data op het leren van modellen, draagt het werk in dit proefschrift bij aan de vooruitgang van Natural Language Processing (NLP).
Het is belangrijk dat we goed begrijpen hoe neurale modellen interageren met en profiteren van data, om zo accuratere NLP-systemen te ontwikkelen.
Bovendien helpt het onderzoek in dit proefschrift om de kwetsbaarheden van huidige neurale netwerken te benadrukken en geeft het inzicht in hoe robuustere modellen ontworpen kunnen worden.

%% Close
\end{document}